\DocumentMetadata{pdfversion=2.0, lang=en-US}

\documentclass[12pt,          
               phd,           
               onehalfspacing 
               ]{ncsuthesis}

\usepackage{ncsu-accessibility}
\usepackage{iftex}
\usepackage[T1]{fontenc}
\usepackage[utf8]{inputenc}
\usepackage{amsmath}
\usepackage{amssymb}
\usepackage{libertinus}
\usepackage{libertinust1math}
\usepackage{amsthm}       
\usepackage{textcomp}
\usepackage{lipsum}       
\usepackage{longtable}    
\usepackage{natbib}       

\bibpunct{(}{)}{;}{a}{}{,}

\dispositionformat{\bfseries}            

\headingformat{\normalsize\MakeUppercase}

\newcommand{\NCSUStudentFirstMiddle}{Md Mirajul}
\newcommand{\NCSUStudentLast}{Islam}
\newcommand{\NCSUStudentFullName}{\NCSUStudentFirstMiddle\space \NCSUStudentLast}

\newcommand{\NCSUThesisTitle}{Offline Apprenticeship Learning for Human-Centric Environments}

\newcommand{\NCSUProgramName}{Computer Science}
\newcommand{\NCSUDegreeYear}{2026}

\newcommand{\NCSUPDFSubject}{PhD Dissertation, Department of Computer Science, North Carolina State University}
\newcommand{\NCSUPDFKeywords}{apprenticeship learning, imitation learning, inverse reinforcement learning, intelligent tutoring systems, offline reinforcement learning, reward modeling, energy-based models, healthcare decision making}

\committeesize{4}
\chair{Dr. Min Chi}
\memberI{Dr. James Lester}
\memberII{Dr. David Roberts}
\memberIII{Dr. Maria Mayorga}

\student{\NCSUStudentFirstMiddle}{\NCSUStudentLast}
\program{\NCSUProgramName}
\thesistitle{\NCSUThesisTitle}
\degreeyear{\NCSUDegreeYear}

\usepackage{xcolor}

\definecolor{NCSULinkBlue}{HTML}{005389}
\definecolor{NCSUReynoldsRed}{HTML}{990000}

\definecolor{NCSUSoftBlue}{HTML}{EEF6FB}
\definecolor{NCSUSoftBlueTitle}{HTML}{D9ECF7}
\definecolor{NCSUBlueBorder}{HTML}{2F5D7C}

\definecolor{NCSUSoftGreen}{HTML}{F0F7F2}
\definecolor{NCSUSoftGreenTitle}{HTML}{DCEFE2}
\definecolor{NCSUGreenBorder}{HTML}{3F6F4F}

\definecolor{NCSUBoxGray}{HTML}{F6F6F6}
\definecolor{NCSUBoxGrayTitle}{HTML}{EDEDED}

\usepackage[
  unicode=true,
  bookmarksnumbered=true,
  bookmarksopen=true,
  linktoc=all,
  colorlinks=true,
  allcolors=black,
  pdfdisplaydoctitle=true,
  pdftitle={\NCSUThesisTitle},
  pdfauthor={\NCSUStudentFullName},
  pdfsubject={\NCSUPDFSubject},
  pdfkeywords={\NCSUPDFKeywords},
  pdfcreator={pdfLaTeX}
]{hyperref}

\usepackage{bookmark}

\usepackage{microtype}

\sectionfont{\normalsize\bfseries}
\subsectionfont{\normalsize\bfseries}
\subsubsectionfont{\normalsize\bfseries}
\paragraphfont{\normalsize\bfseries}
\subparagraphfont{\normalsize\bfseries}

\newcounter{ncsuStatement}[chapter]
\renewcommand{\thencsuStatement}{\thechapter.\arabic{ncsuStatement}}

\newcommand{\NCSUStatementTitle}[3]{%
  \textbf{#1~#2%
  \if\relax\detokenize{#3}\relax
  \else
    : #3%
  \fi.}%
}

\newcommand{\NCSUStatementBegin}[4]{%
  \refstepcounter{ncsuStatement}%
  \par\addvspace{12pt}%
  \noindent\NCSUStatementTitle{#1}{\thencsuStatement}{#2}\quad
  \begingroup
  \ignorespaces
}

\newcommand{\NCSUStatementEnd}{%
  \par
  \endgroup
  \par\addvspace{12pt}%
}

\DeclareDocumentEnvironment{definition}{O{}}{%
  \NCSUStatementBegin{Definition}{#1}{NCSUSoftGreenTitle}{NCSUGreenBorder}%
}{%
  \NCSUStatementEnd
}

\DeclareDocumentEnvironment{theorem}{O{}}{%
  \NCSUStatementBegin{Theorem}{#1}{NCSUSoftBlueTitle}{NCSUBlueBorder}%
  \itshape
}{%
  \NCSUStatementEnd
}

\DeclareDocumentEnvironment{lemma}{O{}}{%
  \NCSUStatementBegin{Lemma}{#1}{NCSUSoftBlueTitle}{NCSUBlueBorder}%
  \itshape
}{%
  \NCSUStatementEnd
}

\DeclareDocumentEnvironment{proposition}{O{}}{%
  \NCSUStatementBegin{Proposition}{#1}{NCSUSoftBlueTitle}{NCSUBlueBorder}%
  \itshape
}{%
  \NCSUStatementEnd
}

\DeclareDocumentEnvironment{corollary}{O{}}{%
  \NCSUStatementBegin{Corollary}{#1}{NCSUSoftBlueTitle}{NCSUBlueBorder}%
  \itshape
}{%
  \NCSUStatementEnd
}

\DeclareDocumentEnvironment{example}{O{}}{%
  \NCSUStatementBegin{Example}{#1}{NCSUBoxGrayTitle}{black!55}%
}{%
  \NCSUStatementEnd
}

\DeclareDocumentEnvironment{remark}{O{}}{%
  \NCSUStatementBegin{Remark}{#1}{NCSUBoxGrayTitle}{black!45}%
}{%
  \NCSUStatementEnd
}

\IfFileExists{student-macros.tex}{

\usepackage{multirow}
\usepackage{booktabs}     
\usepackage{algorithm}
\usepackage{algorithmic}
\usepackage{subcaption}
\usepackage{dsfont}
\usepackage{mathrsfs}
\usepackage{xspace}

\providecommand{\thead}[1]{\textbf{#1}}

\usepackage{newunicodechar}

\usepackage{tikz}
\usepackage{pgfplots}
\pgfplotsset{compat=1.18}
\usetikzlibrary{
  arrows,
  arrows.meta,
  backgrounds,
  calc,
  decorations.pathmorphing,
  decorations.pathreplacing,
  fit,
  positioning,
  shapes.geometric,
  shapes.misc,
  shadows.blur
}

\definecolor{lightblue}{RGB}{173, 216, 230}
\definecolor{gpblue}{HTML}{2166AC}
\definecolor{critred}{HTML}{D6604D}
\definecolor{jointgreen}{HTML}{1A9641}
\definecolor{edmpur}{HTML}{762A83}
\definecolor{irloran}{HTML}{B35806}
\definecolor{arrowgray}{HTML}{555555}
\definecolor{boxframe}{HTML}{333333}
\colorlet{ticcgray}{black!40}

\tikzset{
  mainbox/.style={
    draw=#1!70!black, fill=#1!12, rounded corners=4pt,
    minimum width=3.8cm, minimum height=1.1cm,
    align=center, font=\small\sffamily, line width=0.7pt,
  },
  tinybox/.style={
    draw=#1!60!black, fill=#1!10, rounded corners=3pt,
    minimum width=2.3cm, minimum height=0.75cm,
    align=center, font=\footnotesize\sffamily, line width=0.6pt,
  },
  arrow/.style={
    -{Stealth[length=5pt,width=4pt]}, color=arrowgray, line width=0.9pt,
  },
  thinarrow/.style={
    -{Stealth[length=4pt,width=3pt]}, color=arrowgray!70, line width=0.6pt, dashed,
  },
  seclabel/.style={font=\small\bfseries\sffamily, color=boxframe},
}

\DeclareMathOperator*{\argmin}{argmin}
\DeclareMathOperator*{\argmax}{argmax}

\newcommand{\THEMES}{THEMES\xspace}
\newcommand{\themes}{THEMES\xspace}

\newcommand{\halide}{\textsc{Halide}}

\newcommand{\cchal}{\textsc{Cchal}}

\newcommand{\eg}{e.g.,\xspace}

\newcommand{\crithi}[1]{\mathcal{C}_{\mathrm{hi}}(#1)}
\newcommand{\critlo}[1]{\mathcal{C}_{\mathrm{lo}}(#1)}
\newcommand{\gpcredit}[1]{w^{\mathrm{GP}}_{#1}}
\newcommand{\jointw}[1]{w_{#1}}
\newcommand{\JSD}{\mathrm{JSD}}

\providecommand{\alt}[1]{}
}{}

\IfFileExists{optional.tex}{

\newcommand{\NCSUUseLibertinusFonts}{%
  \setmainfont{LibertinusSerif}[
    Extension=.otf, UprightFont=*-Regular, ItalicFont=*-Italic,
    BoldFont=*-Bold, BoldItalicFont=*-BoldItalic,
    Ligatures=TeX
  ]%
  \setsansfont{LibertinusSans}[
    Extension=.otf, UprightFont=*-Regular, ItalicFont=*-Italic,
    BoldFont=*-Bold,
    Ligatures=TeX,
    Scale=MatchLowercase
  ]%
  \setmonofont{LibertinusMono-Regular.otf}[
    Scale=MatchLowercase
  ]%
  \setmathfont{LibertinusMath-Regular.otf}%
}

\ifPDFTeX\else\NCSUUseLibertinusFonts\fi 

\newif\ifNCSUUseOptionalChapterHeadings

\NCSUUseOptionalChapterHeadingstrue

\makeatletter

\newcommand{\NCSUUseAccessibleChapterHeadings}{%
  \renewcommand{\@makechapterhead}[1]{%
    \vspace*{1.0in}%
    {\parindent \z@ \raggedright \normalfont
      \ifnum \c@secnumdepth >\m@ne
        {\normalsize\bfseries\color{black}%
        \MakeUppercase{\@chapapp}\space \thechapter\par}%
        \vspace{0.25in}%
      \fi
      {\Large\bfseries ##1\par}%
      \vspace{0.18in}%
      {\color{black}\rule{0.45\textwidth}{0.8pt}\par}%
      \vspace{0.75in}%
    }%
  }%

  \renewcommand{\@makeschapterhead}[1]{%
    \vspace*{1.0in}%
    {\parindent \z@ \raggedright \normalfont
      {\Large\bfseries ##1\par}%
      \vspace{0.18in}%
      {\color{black}\rule{0.45\textwidth}{0.8pt}\par}%
      \vspace{0.75in}%
    }%
  }%
}

\makeatother

\newif\ifNCSUPrettyTheorems

\NCSUPrettyTheoremsfalse

\makeatletter

\newcommand{\NCSUApplyOptionalVisualChoices}{%
  \ifNCSUUseOptionalChapterHeadings
    \NCSUUseAccessibleChapterHeadings
  \fi

  \ifNCSUPrettyTheorems
    \@ifundefined{NCSUStatementBegin}{%
      \PackageWarningNoLine{ncsu-template}{%
        Pretty theorem cards were requested, but the base theorem macros were not found.
        Make sure ncsu-required.tex is loaded before optional visual choices are applied.
      }%
    }{%
      \renewcommand{\NCSUStatementBegin}[4]{%
        \refstepcounter{ncsuStatement}%
        \def\NCSUCurrentStatementRuleColor{##4}%
        \par\addvspace{12pt}%
        \noindent
        \begingroup
          \setlength{\fboxsep}{6pt}%
          \colorbox{##3}{%
            \makebox[\dimexpr\linewidth-2\fboxsep\relax][l]{%
              \strut\NCSUStatementTitle{##1}{\thencsuStatement}{##2}%
            }%
          }%
        \endgroup
        \par\nobreak\vspace{6pt}%
        \begingroup
          \leftskip=1em
          \rightskip=1em
          \noindent\ignorespaces
      }%

      \renewcommand{\NCSUStatementEnd}{%
        \par
        \endgroup
        \vspace{4pt}%
        \noindent{\color{\NCSUCurrentStatementRuleColor}\rule{\linewidth}{0.7pt}}%
        \par\addvspace{12pt}%
      }%
    }%
  \fi
}

\makeatother}{}

\providecommand{\NCSUApplyOptionalVisualChoices}{}
\NCSUApplyOptionalVisualChoices

\begin{document}

\frontmatter


\begin{abstract}

Effective decision-making in human-centric environments such as Intelligent Tutoring Systems (ITSs) and clinical healthcare requires policies that are adaptive, data-efficient, and ethically deployable. Apprenticeship Learning (AL) offers a principled path: rather than engineering reward functions, AL infers policies directly from expert demonstrations by recovering the underlying reward structure, enabling strictly \emph{offline} learning without environment interaction. This dissertation develops a progression of offline AL frameworks addressing increasingly realistic complexities of human demonstration data, organized around two central challenges: \textbf{reward modeling} and \textbf{demonstration quality}.

On the \textbf{reward modeling} front, we first relax the standard assumption that all demonstrations share a single, homogeneous reward function. We apply \textbf{EM-EDM}, an Expectation-Maximization framework built upon Energy-based Distribution Matching (EDM), which simultaneously clusters demonstrations by their latent reward functions and induces cluster-specific policies over large continuous state spaces. We show that EM-EDM outperforms AL baselines and two DRL induced policies on student pedagogical decision prediction, demonstrating that heterogeneous reward modeling is both feasible and effective from a small number of expert demonstrations. We then address the further assumption that reward functions remain \emph{fixed within} a trajectory. We leverage \textbf{THEMES} (\textbf{T}ime-aware \textbf{H}ierarchical \textbf{EM} \textbf{E}nergy-based \textbf{S}ubtrajectory AL), which partitions trajectories into temporally coherent sub-trajectories and induces policies over each partition guided by a hierarchically learned reward regulator. Validated across one educational and three real-world clinical healthcare datasets, THEMES consistently outperforms all flat and hierarchical AL baselines, establishing that modeling temporally evolving reward functions substantially improves policy induction in human-centric domains.

On the \textbf{demonstration quality} front, we first challenge the assumption that demonstrations are uniformly optimal. We introduce \textbf{HALIDE} (\textbf{H}ierarchical \textbf{A}pprenticeship \textbf{L}earning from \textbf{I}mperfect \textbf{D}emonstrations with \textbf{E}volving Rewards), which integrates continuous demonstration quality signals directly into the hierarchical policy induction process. HALIDE establishes that incorporating suboptimal trajectories \emph{without} quality-aware ranking degrades performance, whereas ranked hierarchical learning enables principled reuse of mixed-quality data, yielding consistent gains under cross-semester distribution shift. We then go further and jointly attribute demonstration quality to individual decisions via \textbf{CCHAL} (\textbf{C}ritical \textbf{C}ontrast \textbf{H}ierarchical \textbf{A}pprenticeship \textbf{L}earning). Rather than weighting entire trajectories uniformly, CCHAL distributes trajectory-level demonstration ranks to per-decision weights via Bayesian credit assignment, separately scores each decision's criticality by contrasting expert against non-expert behavior in a dual-arm design, and sharpens reward learning with a Bradley-Terry pairwise contrastive IRL objective over observed failure trajectories. Evaluated across three clinical EHR datasets and a student ITS dataset, CCHAL outperforms flat and hierarchical AL baselines, including its direct predecessor HALIDE, establishing that decision-level outcome attribution and behavioral criticality are both necessary to fully exploit demonstration quality signals.

Finally, the dissertation is grounded by a real-world classroom evaluation of \textbf{StuEDM}, an EDM-induced policy that lets students make their own decisions about how to engage with each problem and intervenes only when it disagrees with the student's choice, trained on a substantially smaller and shorter-span demonstration corpus than a comparable Conservative-Q-Learning-induced policy. StuEDM significantly improves posttest performance and normalized learning gain over a pretest-equivalent Expert baseline that assigns among the same domain-expert-designed presentation modes with no adaptive guidance, and significantly improves the pretest-adjusted posttest score over the CQL-induced policy trained on a much larger historical corpus, while overriding students' own decisions less than half as often; together these results provide empirical evidence that offline AL from a substantially smaller, agency-preserving demonstration corpus can rival reward-optimizing DRL trained on far larger corpora in live educational deployments.

Taken together, this dissertation establishes offline AL as a practical, data-efficient, and ethically viable paradigm for human-centric environments, tracing a principled path from reward modeling to quality-aware demonstration learning, and from retrospective evaluation to real-world classroom deployment.

\end{abstract}

\makecopyrightpage

\maketitlepage

\begin{dedication}
\centering To my parents, Mohammad Rofiqul Islam and Rehena Parveen, to my brother, Md Raisul Islam, and to my wife, Tasfia Tafannum, for supporting and encouraging me during this PhD journey.
\end{dedication}

\begin{biography}
Md Mirajul Islam was born in Bangladesh. He attended Bangladesh University of Engineering and Technology (BUET), where he earned his Bachelor of Science degree in Computer Science and Engineering. After graduating, he spent two years working in academia and industry, an experience that sharpened his interest in applying machine learning to real, human-centered problems rather than purely theoretical ones. In 2021, he joined the Department of Computer Science at North Carolina State University to pursue his Ph.D. under the supervision of Professor Min Chi.

During his doctoral studies, Mirajul worked at the NC State Center for Educational Informatics, where he developed data-efficient, human-centered machine learning frameworks for modeling and optimizing decision-making in education and healthcare. Through several projects with real-world stakes, he grew increasingly drawn to Offline Apprenticeship Learning (AL) and Imitation Learning (IL): learning from mixed-quality human demonstrations in settings, such as classrooms and intensive care units, where online trial-and-error is either infeasible or unsafe. Building on this foundation, he developed a series of frameworks that progressively tackled heterogeneous, evolving, and unsupervised reward structures within these human-centric domains, work that was published at leading venues in artificial intelligence and education research.

Beyond his academic work, Mirajul sought out industry experience to see these ideas tested outside the lab. In the summers of 2022, 2024, and 2025, he interned at MarginEdge, Under Armour, and Amazon, respectively. At MarginEdge, he built automated data-processing pipelines using Amazon Textract to extract and structure handwriting data from invoices. At Under Armour, he applied machine learning to athlete performance modeling, combining biomechanics and perception analytics to guide footwear design. At Amazon, as an Applied Scientist Intern, he developed a hierarchical multi-agent framework leveraging large language models to enhance content generation for Amazon Learn. These experiences deepened his interest in reinforcement learning, generative AI, and human--AI collaboration, and his Amazon internship led to a return offer.

In May 2026, Mirajul successfully defended his Ph.D. dissertation and began working as an Applied Scientist at Amazon Web Services, Inc.
\end{biography}

\begin{acknowledgements}

I am deeply grateful to my advisor, Dr.\ Min Chi, for her guidance, patience, and unwavering support throughout every stage of this dissertation. Her mentorship shaped not only the research presented here but also how I think about research itself, and I could not have asked for a better advisor to guide me through this journey.

I would also like to thank my committee members, Dr.\ James Lester, Dr.\ David Roberts, and Dr.\ Maria Mayorga, for their thoughtful feedback, probing questions, and generous time throughout my proposal and defense; their perspectives meaningfully strengthened this work.

My sincere thanks to my labmates at the Center for Educational Informatics. To those who graduated before me and helped show me the ropes---Xi Yang, Mark Abdelshiheed, John Hostetter, Mitchell Plyer, and Ge Gao---thank you for your early guidance and collaboration. To those who were still in the lab through my own defense---Louis Rabinowitz, Adittya Soukarjya Saha, Rajesh Debnath, and Seongsoo Kim---thank you for the countless discussions and day-to-day camaraderie. And to Shengran Guo and Yong Cho, who joined the lab just as I was on my way out, I wish you both the very best in the work ahead. I am also grateful to Yeo Jin Kim, who, though our time in the lab did not overlap, generously shared resources and insight from her own experience as Dr.\ Chi's student.

I would also like to thank Dr.\ Tiffany Barnes for being a wonderful mentor throughout our collaboration between the Pyrenees and Deepthought labs. Through this collaboration I had the pleasure of working with the Deepthought team, including Nazia Alam and Sutapa Dey Tithi, and I thank them both for their collaboration.

I am also thankful to my friends---Mohammad Mahinur Rahman, Naimur Rahman Chowdhury, Chowdhury Amir Abdullah, Imranur Rahman, Rashik Intisar Siddiquee, Mohaiminul Islam, Mohammad Mahmudul Hassan, Md.\ Sazzad Hossain, Md.\ Enayet Chowdhury, Md Nahiyan Uddin, Md Moinul Hossain, Md Kamrul Hasan Shovon, Taharina Tasnim, and many more---for their encouragement and for the many needed breaks from research over the years.

Finally, none of this would have been possible without my family. To my parents, Mohammad Rofiqul Islam and Rehena Parveen, my brother, Md Raisul Islam, and my wife, Tasfia Tafannum: thank you for your unconditional love, patience, and belief in me throughout this journey.

\end{acknowledgements}

\thesistableofcontents

\thesislistoftables

\thesislistoffigures

\chapter*{AUTHORSHIP STATEMENT}
\addcontentsline{toc}{chapter}{Authorship Statement}

\noindent Dr. Min Chi, dissertation advisor and committee chair, provided conceptualization, research guidance, supervision, and manuscript revision across every chapter of this dissertation.

\vskip 0.5cm

\begin{description}
  \item[Chapter 1.] \emph{Introduction.} Md Mirajul Islam: sole author.
  \item[Chapter 2.] \emph{Background.} Md Mirajul Islam: sole author.
  \item[Chapter 3.] \emph{Modeling Heterogeneous Student Pedagogical Strategies.} Published as Islam, Md Mirajul; Yang, Xi; Hostetter, John; Saha, Adittya Soukarjya; and Chi, Min, \emph{EDM 2024}. Md Mirajul Islam: primary research, experiments, and writing. Xi Yang: methodology. John Hostetter: data collection. Adittya Soukarjya Saha: data collection.
  \item[Chapter 4.] \emph{Modeling Evolving Reward Functions.} Published as (i) Islam, Md Mirajul; Yang, Xi; Debnath, Rajesh; Saha, Adittya Soukarjya; and Chi, Min, \emph{AIED 2025}, and (ii) Yang, Xi; Islam, Md Mirajul; Gao, Ge; and Chi, Min, \emph{KDD 2025}. Md Mirajul Islam: primary research, experiments, and writing for (i); experiments for (ii). Xi Yang: methodology for (i); primary research, methodology, and writing for (ii). Rajesh Debnath and Adittya Soukarjya Saha: data collection for (i). Ge Gao: writing help for (ii).
  \item[Chapter 5.] \emph{Hierarchical Apprenticeship Learning from Imperfect Demonstrations with Evolving Rewards.} Published as Islam, Md Mirajul; Debnath, Rajesh; Saha, Adittya Soukarjya; and Chi, Min, \emph{AIED 2026}. Md Mirajul Islam: primary research, methodology, experiments, and writing. Rajesh Debnath and Adittya Soukarjya Saha: data collection.
  \item[Chapter 6.] \emph{Critical Contrast Hierarchical Apprenticeship Learning.} Md Mirajul Islam: primary research, methodology, experiments, and writing. Shengran Guo: additional experiments, run using code and experimental setup provided by Md Mirajul Islam.
  \item[Chapter 7.] \emph{Empirical Evaluation of Offline AL in Classroom Study.} Md Mirajul Islam: primary research, methodology, experiments, and writing.
  \item[Chapter 8.] \emph{Conclusions and Future Work.} Md Mirajul Islam: sole author.
\end{description}

\vskip 1cm

\noindent\textbf{Use of generative artificial intelligence:} Generative AI (Claude, Anthropic) was used only to assist with \LaTeX{} formatting, template migration, and document accessibility remediation (e.g., margin corrections, caption formatting, table layout). It was not used to generate, analyze, or write any research content, findings, or substantive text of this dissertation.

\mainmatter

\chapter{Introduction}
\label{chap-one}

\section{Overview}
\label{sec:overview}

Many human-centric environments---such as education, healthcare, and personalized support systems---can be formulated as sequential decision-making problems, where an intelligent agent must determine \emph{what} action to take and \emph{when} to take it over time. These settings are naturally modeled using Reinforcement Learning (RL), which optimizes long-term outcomes through trial-and-error interaction with an environment \citep{sutton2018reinforcement}. In an Intelligent Tutoring System (ITS), for example, a pedagogical agent must decide whether to present a student with a worked example, prompt independent problem solving, or engage in collaborative construction---adapting to each student’s evolving knowledge state across dozens of steps. In clinical settings, a treatment agent must select interventions for a patient whose condition changes over time. Despite the considerable success of RL and Deep RL (DRL) in simulated domains, their adoption in real-world, human-facing applications remains limited by two longstanding challenges.

The first challenge is \textbf{sample inefficiency}. Classic DRL algorithms such as Deep Q-Networks (DQN) require millions of environment interactions to learn stable policies and generalize to unseen states. Yet interactions with humans---patients, students, users---are scarce, costly, and ethically constrained. Many prior studies on RL in educational settings attempt to derive pedagogical policies from fewer than 3,000 student--ITS interaction trajectories, and the resulting policies can prove ineffective \citep{shen2016reinforcement}. The second challenge is \textbf{reward function design}. Designing an accurate reward function is a prerequisite for RL, yet it is notoriously difficult in human-centric domains where desired behaviors are complex, latent, and context-dependent \citep{kaelbling1996reinforcement}. Reward functions are typically hand-crafted, dependent on domain expertise, and prone to blind spots \citep{abbeel2004apprenticeship, goyal2019using}. Worse, misspecified rewards can produce policies that are inconsistent with intended goals \citep{amodei2016concrete}. Moreover, in sensitive human-centric domains---classrooms, clinics---live policy experimentation raises ethical concerns, making it infeasible to collect the exploratory data that RL methods require.

\subsection{Apprenticeship Learning as an Alternative}

Apprenticeship Learning (AL) offers a principled alternative by inferring the underlying reward function directly from expert demonstrations, enabling policy induction without explicit reward specification \citep{abbeel2004apprenticeship}. By assuming expert agents make optimal or near-optimal decisions, AL can efficiently approximate expert behavior from a limited number of demonstrated trajectories \citep{abbeel2004apprenticeship, ziebart2008maximum}. Classical AL approaches, however, are predominantly \emph{online}: they require iterative interaction with the environment to collect new data and update the model \citep{ziebart2008maximum, ho2016generative, finn2016guided}. This precludes their use in fully offline settings. \emph{Strictly offline} AL methods---those that learn solely from pre-collected logs without any environment interaction---are therefore essential in human-centric domains. A recent advance, Energy-Based Distribution Matching (EDM) \citep{jarrett2020strictly}, achieves strong performance in offline AL by matching state-action occupancy distributions between expert demonstrations and a learned policy using energy-based models, without requiring knowledge of model transitions or off-policy evaluations.

Despite EDM’s success, it---like most offline AL methods---rests on a sequence of idealized assumptions that are routinely violated in practice. This dissertation progressively relaxes these assumptions, developing a series of frameworks that address the realistic complexities of human demonstration data across educational and clinical settings.

\subsection{Reward Modeling: Heterogeneous and Evolving Reward Functions}

A foundational assumption of most offline AL methods is that all demonstrations share a single, homogeneous reward function that remains fixed throughout each trajectory. Both parts of this assumption are routinely violated in human-centric settings, and this dissertation addresses them in turn.

The first violation is \emph{cross-trajectory heterogeneity}: different demonstrators pursue fundamentally different objectives. In an ITS, some students are learning-oriented while others prioritize efficiency; their pedagogical decisions reflect different underlying reward structures. Prior work addressing this, such as EM-IRL \citep{yang2020student}, is limited to small discrete state spaces (17 features). The first study of this dissertation adapts \textbf{EM-EDM}---an Expectation-Maximization framework built upon EDM---which simultaneously clusters student trajectories by their latent reward functions and induces cluster-specific policies, scaling to continuous state spaces with over 140 features. Even with as few as 24 expert demonstrations, EM-EDM outperforms four state-of-the-art AL baselines and two DRL baselines on predicting student pedagogical decisions.

The second violation is \emph{intra-trajectory evolution}: even a single demonstrator’s goals shift across phases of a task. A student who begins a tutoring session cautiously requesting help may shift toward confident independent problem-solving as the session progresses; a clinician’s treatment priorities evolve as a patient’s condition changes. Such temporal dynamics cannot be captured by any static reward function. The second study leverages \textbf{THEMES} \citep{yang2025themes}---a \textbf{T}ime-aware \textbf{H}ierarchical \textbf{EM} \textbf{E}nergy-based \textbf{S}ubtrajectory AL framework---which automatically partitions expert trajectories into temporally coherent sub-trajectories, applies EM-EDM over each partition, and refines the segmentation through a hierarchically learned reward regulator. Applied across one educational and three clinical healthcare datasets, THEMES consistently outperforms all flat and hierarchical AL baselines, establishing that capturing temporally evolving reward functions substantially improves policy induction in human-centric domains.

\subsection{Demonstration Quality: Imperfect and Negative Demonstrations}

A second foundational assumption of AL is that all provided demonstrations are at minimum acceptable in quality---that imperfect trajectories should be discarded and failed ones carry no learnable signal. This dissertation challenges both parts of this assumption.

The first challenge concerns \emph{imperfect demonstrations}: in practice, students explore, make errors, and revise strategies, producing trajectories of varying quality. Discarding all suboptimal data wastes the majority of available demonstrations; yet treating all as equally informative introduces noise that corrupts learned policies. The key insight is that imperfect demonstrations constitute structured signals---provided their relative quality is explicitly ranked. The third study introduces \textbf{HALIDE} (\textbf{H}ierarchical \textbf{A}pprenticeship \textbf{L}earning from \textbf{I}mperfect \textbf{D}emonstrations with \textbf{E}volving Rewards), which integrates continuous quality signals---derived from Normalized Learning Gain---directly into the hierarchical policy induction process. HALIDE establishes a critical finding: incorporating suboptimal data \emph{without} quality-aware ranking degrades performance, whereas ranked hierarchical learning enables principled reuse of mixed-quality trajectories and yields consistent gains in cross-semester generalization. Crucially, ranking in AL serves a fundamentally different role from reward functions in RL: rather than defining what is optimal, it modulates how much influence each demonstrated behavior exerts during learning.

The second challenge concerns \emph{decision-level importance}: existing quality-weighting methods, including HALIDE, apply a single trajectory-level rank uniformly across every decision, even though most decisions occur at routine states while a minority occur at critical junctures where the choice made is pivotal to the outcome---and failed trajectories are typically discarded entirely, even though they encode informative contrast about which decisions distinguish success from failure. The fourth study introduces \textbf{CCHAL} (\textbf{C}ritical \textbf{C}ontrast \textbf{H}ierarchical \textbf{A}pprenticeship \textbf{L}earning), which jointly attributes trajectory-level demonstration quality to individual decisions via Bayesian credit assignment in feature-action space, separately scores each decision's criticality by contrasting expert against non-expert behavior within the same initial risk stratum, and combines both signals to sharpen a Bradley-Terry pairwise contrastive IRL objective over observed failure trajectories. Evaluated across three clinical EHR datasets and a student ITS dataset, CCHAL outperforms flat and hierarchical AL baselines, including its direct predecessor HALIDE, establishing that decision-level outcome attribution and behavioral criticality are both necessary to fully exploit demonstration quality signals.

\subsection{Real-World Classroom Validation}

The progression from EM-EDM through THEMES, HALIDE, and CCHAL addresses increasingly realistic data conditions in retrospective evaluation. The fifth study closes the loop with a real-world classroom deployment. \textbf{StuEDM} is a pedagogical policy in which students make their own decisions about how to engage with each training problem---choosing to solve it independently, study a worked example, or collaborate with the tutor---and an EDM-derived policy, trained on only a limited set of expert demonstrations, intervenes only when it disagrees with the student’s decision. In a controlled classroom study with 104 undergraduate students, StuEDM---trained on 244 students' own pedagogical decisions across five semesters---is compared against StuCQL, an analogous policy whose override logic is provided by Conservative Q-Learning trained on a much larger, longer-span historical interaction corpus (2,716 students, 13 semesters), and against a pretest-equivalent Expert baseline that assigns among the same domain-expert-designed presentation modes with no adaptive guidance. StuEDM significantly improves posttest performance and normalized learning gain over the Expert baseline, and significantly improves the pretest-adjusted posttest score over StuCQL; a decision-level analysis further shows StuEDM overrides students' own decisions less than half as often as StuCQL. StuEDM students also engaged with the tutor significantly longer than StuCQL students, a further engagement-related outcome favoring StuEDM, and the underlying training-corpus efficiency remains a genuine advantage, providing direct empirical evidence that offline AL from a substantially smaller, agency-driven demonstration corpus can rival reward-optimizing DRL trained on far larger datasets in real educational deployments.

\section{Research Questions}
\label{sec:research_questions}

\textbf{RQ1: How can offline apprenticeship learning be used to model heterogeneous student pedagogical strategies?}
Students exhibit diverse learning approaches and latent objectives, meaning a single monolithic pedagogical policy may not suit all learners. This question addresses the challenge of capturing such heterogeneity by adapting offline AL to simultaneously discover multiple latent reward functions and induce cluster-specific policies from continuous state spaces. By inducing policies directly from varied demonstration data, the aim is to uncover the distinct strategy types that different students follow without requiring reward labels or online experimentation.

\textbf{RQ2: How can we capture the evolving nature of reward functions across different phases of sequential decision-making in education and healthcare?}
Complex decision-making tasks unfold in distinct phases during which the decision-maker’s goals and trade-offs shift over time. Traditional RL and imitation learning assume a fixed reward function, which struggles to reflect these temporal shifts. RQ2 investigates how to automatically segment demonstration trajectories into sub-trajectories and infer phase-specific reward regimes from data, without requiring manually defined task boundaries---enabling policies that adapt to different decision-making stages in both educational and clinical domains.

\textbf{RQ3: How can apprenticeship learning leverage imperfect human demonstrations while accounting for evolving reward structure and demonstration quality?}
In real-world settings, purely expert demonstrations are rare. Available data consist largely of mixed-quality trajectories---partially successful behaviors, exploratory decisions, and non-optimal strategies. This question explores how to rank demonstrations by pedagogical quality and integrate that ranking into a hierarchical learning framework that preserves temporal structure and distinguishes transient errors from meaningful progress, enabling robust offline learning without reward labels or online interaction.

\textbf{RQ4: Can decision-level outcome attribution and behavioral criticality, combined with contrastive learning from failed demonstrations, improve hierarchical apprenticeship learning beyond trajectory-level quality weighting?}
Existing approaches to mixed-quality demonstration learning apply a single quality rank uniformly across all decisions within a trajectory, and typically discard failed demonstrations entirely. This question asks whether redistributing trajectory-level quality to individual decisions via Bayesian credit assignment, combined with a criticality score that contrasts expert against non-expert behavior at each decision, and a pairwise contrastive objective that treats observed failures as informative negatives, improves policy induction over trajectory-level quality weighting alone.

\textbf{RQ5: Can offline apprenticeship learning from limited expert demonstrations outperform reward-optimizing deep RL trained on large data corpora in a real classroom deployment?}
Advances in offline AL are typically evaluated retrospectively. This question asks whether the practical advantages of AL---no reward engineering, data efficiency, alignment with expert behavior---translate to superior student learning outcomes in a live classroom study, compared to a conservative deep RL policy trained on a large-scale interaction corpus.

\section{Contributions}
\label{sec:contributions}

This dissertation contributes to the field of offline apprenticeship learning by advancing modeling capabilities across five interconnected dimensions:

\begin{enumerate}
    \item \textbf{Heterogeneous Reward Modeling (Chapter~\ref{chap:edm24}):} We adapt the EM-EDM framework to educational data, discovering latent learner types and inducing cluster-specific pedagogical policies. This enables accurate modeling of diverse student strategies in large-scale tutoring logs with continuous state spaces, outperforming four AL baselines and two DRL baselines with as few as 24 expert demonstrations.

    \item \textbf{Evolving Reward Modeling (Chapter~\ref{chap:aied25}):} We leverage THEMES, a time-aware hierarchical framework that automatically segments long-horizon trajectories into sub-trajectories governed by different latent objectives. Applied across one educational and three real-world clinical healthcare environments, THEMES captures temporal shifts in human strategies and substantially outperforms flat and hierarchical AL baselines.

    \item \textbf{Quality-Aware Hierarchical Learning from Imperfect Demonstrations (Chapter~\ref{chap:aied26}):} We introduce HALIDE, a ranked hierarchical apprenticeship learning framework that integrates continuous demonstration quality signals into the policy induction process. HALIDE enables principled reuse of mixed-quality student data while preserving the temporal and strategic structure induced by hierarchical modeling, consistently improving over unweighted hierarchical baselines in cross-semester generalization.

    \item \textbf{Critical Contrast Hierarchical Learning from Imperfect Demonstrations (Chapter~\ref{chap:ccal}):} We introduce CCHAL, which extends hierarchical AL by jointly attributing outcome quality to individual decisions via Bayesian credit assignment and scoring each decision's behavioral criticality through a dual-arm contrastive comparison of expert against non-expert behavior, then sharpens reward learning with a Bradley-Terry pairwise contrastive IRL objective over observed failure trajectories. Evaluated on three clinical EHR datasets and one intelligent tutoring system dataset, CCHAL outperforms flat and hierarchical AL baselines, including its direct predecessor HALIDE, across educational and clinical domains.

    \item \textbf{Real-World Empirical Validation (Chapter~\ref{chap:stuedm}):} We deploy StuEDM---an EDM-induced policy that lets students make their own pedagogical decisions and intervenes only on disagreement, trained on an order-of-magnitude smaller demonstration corpus---in a controlled classroom study with 104 students. StuEDM significantly improves posttest performance and normalized learning gain over an Expert baseline that assigns among the same domain-expert-designed presentation modes with no adaptive guidance, and significantly improves the pretest-adjusted posttest score over CQL trained on a much larger historical corpus; StuEDM also overrides students' own decisions less than half as often as CQL, providing direct empirical evidence that offline AL from a substantially smaller, agency-preserving demonstration corpus can rival reward-optimizing DRL in real educational settings.
\end{enumerate}

Together, these contributions define a progression from modeling \emph{heterogeneous}, to \emph{evolving}, to \emph{imperfect}, to \emph{contrastive} reward structures, culminating in real-world empirical validation. They establish a unified framework for understanding and learning from complex human decision-making, with demonstrated applications in both education and healthcare.

The rest of the dissertation is organized as follows. Chapter~\ref{chap:background} surveys relevant background and prior work. Chapter~\ref{chap:edm24} addresses RQ1 through heterogeneous reward modeling via EM-EDM. Chapter~\ref{chap:aied25} addresses RQ2 by leveraging THEMES for evolving reward functions across educational and clinical tasks. Chapter~\ref{chap:aied26} addresses RQ3 through HALIDE, a quality-aware hierarchical framework for imperfect demonstrations. Chapter~\ref{chap:ccal} addresses RQ4 through CCHAL, extending hierarchical AL with outcome-attributed decision weighting, dual-arm contrastive criticality scoring, and Bradley-Terry pairwise contrastive IRL. Chapter~\ref{chap:stuedm} addresses RQ5 through a live classroom deployment of StuEDM. Chapter~\ref{chap:conc} concludes with a summary of findings and future directions.

\chapter{Background}
\label{chap:background}

This chapter reviews the literature that motivates and contextualizes the contributions of this dissertation. We begin by outlining the foundations of apprenticeship learning (AL) and its relationship to reinforcement learning (RL), with particular emphasis on settings where reward functions are difficult to specify and online exploration is infeasible. We then survey advances in inverse reinforcement learning, adversarial imitation learning, and energy-based offline AL that seek to recover robust policies from fixed datasets. Building on this, we review methods addressing heterogeneous and evolving reward functions, imperfect and mixed-quality demonstrations, and contrastive supervision from negative trajectories. Finally, we discuss prior applications of RL and AL to intelligent tutoring systems, which serve as the primary testbed throughout this dissertation. Together, these threads highlight the core methodological challenges---reward ambiguity, distributional shift, and variation in demonstration quality---that this dissertation addresses through a progression of offline apprenticeship learning frameworks.

\section{Sequential Decision-Making under Uncertainty}

In many human-centric decision-making environments---such as education and healthcare---an agent must decide which action to take and when to take it as interactions unfold over time. These settings naturally form sequential decision-making problems under uncertainty, where actions influence both immediate and long-term outcomes. Reinforcement Learning (RL) provides a principled framework for modeling such processes by seeking policies that maximize cumulative reward without requiring explicit knowledge of the optimal action at each time step \citep{sutton2018reinforcement}.

Recent advances have demonstrated the effectiveness of RL and Deep RL (DRL) for learning data-driven policies in personalized education, clinical decision support, and adaptive behavioral interventions \citep{abdelshiheed2023bridging,ju2020aiedcritical,mandel2014offline,rowe2015improving}. However, several limitations constrain their practical use in real-world human-centric domains. A major challenge is \textbf{sample inefficiency}: DRL algorithms such as Deep Q-Networks (DQN) typically require millions of interactions to learn effective policies, whereas real-world environments often provide only a few thousand trajectories \citep{DBLP:conf/aied/AusinMBC20,shen2016reinforcement}. Another challenge arises from \textbf{reward function design}: in domains involving human behavior---such as learning progress or patient well-being---reward signals are difficult to define, non-stationary, and often partially observable \citep{kaelbling1996reinforcement}. Hand-crafted reward functions are time-consuming, prone to subjective biases, and may misalign with desired outcomes \citep{goyal2019using,abbeel2004apprenticeship,amodei2016concrete}. Moreover, in sensitive human-centric domains, live policy experimentation raises ethical concerns, making it infeasible to collect the exploratory data that RL methods require. These limitations motivate the use of Apprenticeship Learning as an alternative.

\section{Apprenticeship Learning (AL)}

Apprenticeship Learning (AL) offers an alternative to reinforcement learning in settings where reward functions are difficult to specify and environment interaction is limited. Rather than relying on hand-crafted rewards or extensive exploration, AL seeks to learn effective policies directly from \emph{expert demonstrations} by inferring the latent reward structure that best explains expert behavior \citep{abbeel2004apprenticeship}. Under the assumption that expert trajectories reflect optimal or near-optimal decision-making, AL methods aim to approximate expert-level performance with far fewer samples than traditional RL \citep{abbeel2004apprenticeship,ziebart2008maximum}.

Early AL approaches include behavior cloning (BC) \citep{pomerleau1991efficient,syed2012imitation,raza2012teaching}, which directly maps states to actions by imitating expert behavior. While simple and suitable for purely offline settings, BC suffers from compounding error and poor generalization when demonstrations do not cover the full state space \citep{pmlr-v15-ross11a}. More principled approaches derive from inverse reinforcement learning (IRL) \citep{abbeel2004apprenticeship,ziebart2008maximum}, which iteratively: (1) infers a reward function from expert data, (2) learns a policy via RL, (3) rolls out the learned policy, and (4) updates reward parameters based on discrepancies from expert trajectories. Although IRL methods are more robust than BC, their reliance on online rollouts limits applicability in human-centric domains where repeated interactions are infeasible or unsafe.

Adversarial imitation learning, exemplified by GAIL \citep{ho2016generative} and AIRL \citep{fu2018learning}, advances AL by training a discriminator to distinguish expert from learned behavior and a generator (policy) to minimize this distinction. Off-policy variants \citep{kostrikov2018discriminator,kostrikov2019imitation} reduce the need for online rollouts but inherit adversarial training instability and reliance on off-policy critics, which can introduce bias in offline settings.

These limitations motivated the development of \emph{offline} AL methods designed to operate on fixed datasets without additional environment interaction. Offline AL must address challenges absent in online settings: distributional shift, the inability to validate policies through rollouts, and reward identifiability without on-policy data. Batch IRL methods \citep{klein2011batch,lee2019truly,chan2021scalable,raghu2017continuous} remove online rollouts but often rely on restrictive reward parameterizations or off-policy evaluation, leading to unstable or suboptimal solutions.

\section{AL with Heterogeneous Reward Functions}

To address heterogeneity \emph{across} demonstrations, several AL methods explicitly model multiple reward functions. Bayesian multi-task IRL \citep{dimitrakakis2011bayesian} learns separate reward functions for each trajectory using a shared prior distribution. Choi and Kim \citep{choi2012nonparametric} extended this using a Dirichlet process mixture to cluster demonstrations without predefining the number of reward types. Arora \textit{et al.} \citep{arora2021min} combined Dirichlet processes with maximum entropy IRL to jointly infer clusters and reward functions.

Babes \textit{et al.} \citep{babes2011apprenticeship} proposed an EM-based IRL method that alternates between clustering trajectories and optimizing cluster-specific rewards via maximum-likelihood IRL---a foundational formulation that this dissertation builds upon. Yang \textit{et al.} applied EM-IRL to student behavior modeling in ITSs \citep{yang2020student}, discovering latent learner subtypes from pedagogical decision data. However, like many multi-intent methods, EM-IRL is constrained to discrete state representations with limited feature sets, and does not generalize to continuous state spaces---a critical limitation in data-rich ITS and healthcare settings with hundreds of continuous features. This dissertation addresses this gap by adapting an EM framework to EDM, enabling heterogeneous reward modeling at scale in continuous spaces.

\section{AL with Evolving Reward Functions}

Beyond cross-trajectory heterogeneity, reward functions may also \emph{evolve over time within a single trajectory}. In tutoring, a student's goals shift across problem-solving phases; in clinical care, treatment priorities change as a patient's condition evolves. Single-reward AL and even multi-intent methods fail to capture such intra-trajectory dynamics.

Krishnan \textit{et al.} introduced Hierarchical IRL (HIRL) \citep{krishnan2016hirl}, which segments demonstrations into locally linear sub-tasks using Bayesian nonparametric methods and learns sub-task-specific reward functions via maximum entropy IRL. Hausman \textit{et al.} \citep{hausman2017multi} proposed Multi-modal Imitation Learning (MIL), which jointly discovers sub-tasks and learns corresponding sub-policies using a generative adversarial framework augmented with a latent intention variable. Wang \textit{et al.} \citep{wang2020multi,wang2022hierarchical} developed hierarchical imitation learning models that simultaneously learn high-level policies and sub-policies for individual sub-tasks. However, these methods commonly rely on online policy rollouts or adversarial training, making them ill-suited for fully offline settings. Meta-imitation learning \citep{finn2016guided} has also been explored for few-shot generalization, but typically requires manually defined sub-task boundaries---impractical in complex, uncertain human-centric environments.

To address these limitations, Yang \textit{et al.} proposed \textbf{THEMES} \citep{yang2025themes}---a Time-aware Hierarchical EM Energy-based Subtrajectory AL framework---which automatically partitions trajectories into temporally coherent sub-trajectories using a data-driven method, applies EM-EDM to induce cluster-specific policies over each partition, and refines segmentation through a hierarchically learned reward regulator, all in a strictly offline manner. This dissertation leverages THEMES to model evolving pedagogical and clinical decision-making strategies across both educational and healthcare datasets.

\section{AL from Imperfect and Mixed-Quality Demonstrations}

Most AL methods assume access to expert-labeled or at least high-quality demonstrations. In practice, human demonstration corpora are heterogeneous: carefully curated expert trajectories are scarce, while large quantities of suboptimal or partially successful behaviors are readily available. Learning robust policies from mixed-quality data---without inheriting suboptimal biases---is a central challenge.

Early offline approaches based on behavioral cloning perform well with sufficient expert coverage but suffer from compounding error and distributional shift when expert coverage is limited \citep{pmlr-v15-ross11a}. A growing body of work studies learning from imperfect or mixed-quality demonstrations. Confidence-weighted imitation learning \citep{wu2019imperfectil} incorporates reliability signals to reweight state-action pairs. Discriminator-based reweighting \citep{xu2021discriminatorweighted} learns to down-weight suboptimal data. Positive-unlabeled formulations \citep{liu2025pngail} cope with the absence of explicit expert labels. Value- and reward-based approaches such as ORIL \citep{zolna2020oril} and SPRINQL \citep{hoang2024sprinql} integrate expert and suboptimal experience using offline RL techniques. DemoDICE \citep{kim2022demodice} and NoisyBC \citep{sasaki2021noisybc} filter or reweight demonstrations using noise assumptions. ILEED \citep{Beliaev2022ImitationLB} estimates per-demonstrator expertise scores but requires known demonstrator identities and assumes consistent skill levels within trajectories.

Despite this progress, most mixed-quality AL methods operate in flat policy spaces and apply quality signals at coarse granularity---typically trajectory-level weighting---which can still allow suboptimal decisions to propagate into learned policies. Critically, they do not account for hierarchical or temporally evolving decision-making, where a single trajectory may contain high-quality decisions in some phases and low-quality decisions in others. This dissertation addresses this gap with HALIDE, which integrates continuous demonstration quality signals---derived from Normalized Learning Gain---within a hierarchical policy induction framework, enabling ranked, quality-aware learning while preserving temporal and strategic structure.

\section{Contrastive AL from Negative Demonstrations}

A foundational assumption across the imitation learning literature is that failed or negative demonstrations---sequences where outcomes are poor or strategies fail---carry no learnable signal beyond their exclusion. In practice, failed trajectories encode structured information about \emph{which strategies fail and why}. Contrastive methods aim to exploit this structure.

Prior contrastive approaches to imitation learning use negative demonstrations primarily as contrast signals to improve the discriminability of expert behavior \citep{tangkaratt2021robust}. These methods typically apply anti-imitation uniformly across failed trajectories, treating all steps as equally undesirable. This is problematic in hierarchical settings: a failed trajectory often contains segments of strategy-consistent, correct behavior interrupted by a localized divergence that precipitates failure. Applying uniform anti-imitation across such trajectories discards informative correct sub-trajectories and introduces noisy gradient signals.

This dissertation introduces CHALIDE, which addresses this limitation by localizing failure-inducing divergence points within negative trajectories through a novel cluster-assignment divergence statistic derived from learned hierarchical strategy clusters. By applying temporally-weighted anti-imitation only at detected divergence points and enforcing cluster-level reward ordering constraints during hierarchical reward inference, CHALIDE exploits the full informative content of negative trajectories---both the strategy-consistent segments and the failure-inducing transitions---without the distortions of uniform anti-imitation.

\section{RL and AL in Intelligent Tutoring Systems}

Intelligent Tutoring Systems (ITSs) provide a primary testbed for sequential decision-making methods, as they require adaptive pedagogical policies that respond to individual student learning states over extended interactions. A large body of prior work has applied RL and DRL to pedagogical policy induction \citep{mandel2014offline,rowe2015improving,shen2016reinforcement}. Shen \textit{et al.} \citep{shen2016reinforcement} applied value iteration to induce policies targeting student performance improvement. Zhou \textit{et al.} \citep{zhou2019hierarchical} employed offline hierarchical RL to derive pedagogical policies, achieving superior results over flat RL baselines in classroom studies. Abdelshiheed \textit{et al.} \citep{abdelshiheed2023leveraging} demonstrated that deep RL-based metacognitive interventions can close skill gaps between student subgroups and support preparation for future learning.

These RL approaches typically require large-scale training data and rely on predefined reward functions such as normalized learning gain, limiting adaptability to diverse student populations \citep{shen2016reinforcement,zhou2019hierarchical}. Conservative Q-Learning (CQL) \citep{kumar2020conservative} has been applied to ITSs to address overestimation bias in offline settings, most notably as a student-choice-with-override policy \citep{hostetter2023leveraging} in which the system intervenes only when its recommendation disagrees with the student's autonomous choice.

AL has been explored as an alternative to reward-based RL in ITS settings. Rafferty \textit{et al.} \citep{rafferty2015inferring} applied IRL to infer learners' beliefs in an educational game, demonstrating IRL's potential for interpreting interactive learning data. Yang \textit{et al.} \citep{yang2020student} applied EM-IRL to cluster students by pedagogical behavior, though limited to discrete state representations. Islam \textit{et al.} \citep{islam2024generalized} extended this with EM-EDM, enabling policy induction from heterogeneous expert demonstrations in large continuous state spaces and demonstrating that AL outperforms two DRL and four AL baselines on pedagogical action prediction. This dissertation extends this line of work, culminating in the first live classroom deployment of an EDM-induced student-choice-with-override policy (StuEDM), which outperforms CQL trained on a large historical corpus across multiple normalized learning gain measures without additional study time.
\chapter{Modeling Heterogeneous Student Pedagogical Strategies}
\label{chap:edm24}

* \textit{This chapter has been published as:}
\begin{quote}
\textbf{Md Mirajul Islam}, Xi Yang, John Hostetter, Adittya Soukarjya Saha, and Min Chi. \textbf{“A Generalized Apprenticeship Learning Framework for Modeling Heterogeneous Student Pedagogical Strategies.”} \textit{Proceedings of the 17th International Conference on Educational Data Mining (EDM 2024).}
\end{quote}

A key challenge in e-learning environments like Intelligent Tutoring Systems (ITSs) is to induce effective pedagogical policies efficiently. While Deep Reinforcement Learning (DRL) often suffers from \textbf{\emph{sample inefficiency}} and \textbf{\emph{reward function}} design difficulty, Apprenticeship Learning (AL) algorithms can overcome them. However, most AL algorithms can not handle heterogeneity as they assume all demonstrations are generated with a homogeneous policy driven by a single reward function. Still, some AL algorithms which consider heterogeneity, often can not generalize to large continuous state space and only work with discrete states. In this paper, we propose an expectation-maximization(EM)-EDM, a general AL framework to induce effective pedagogical policies from given optimal or near-optimal demonstrations, which are assumed to be driven by heterogeneous reward functions. We compare the effectiveness of the policies induced by our proposed EM-EDM against four AL-based baselines and two policies induced by DRL on two different but related tasks that involve pedagogical action prediction. Our overall results showed that, for both tasks, EM-EDM outperforms the four AL baselines across all performance metrics and the two DRL baselines. This suggests that EM-EDM can effectively model complex student pedagogical decision-making processes through the ability to manage a large, continuous state space and adapt to handle diverse and heterogeneous reward functions with very few given demonstrations.

\section{Overview}
Reinforcement Learning (RL) and Deep RL (DRL) have experienced significant development in recent years and have been successfully applied in e-learning systems such as intelligent tutoring systems (ITSs) \citep{abdelshiheed2023bridging, chi2011empirically, chi2011evaluation, hostetter2023self, iglesias2009learning, shen2016reinforcement}. More specifically, in an ITS, the pedagogical agent leverages RL to learn a decision-making policy to maximize the expected cumulative \emph{rewards} over an extended period, with the overarching aim of benefiting students in the long run \citep{abdelshiheed2023leveraging, hostetter2023leveraging, ju2022student}. Despite DRL's great success, several obstacles prevent the broader application of DRL to educational systems in practice. One is \textbf{\emph{sample inefficiency}}. For example, it takes Deep Q-Networks (DQN), one type of classic DRL algorithm, hundreds of millions of interactions with the environment to learn a good policy and generalize to unseen states. 
In many previous studies on RL and DRL in ITSs, a prevalent practice involves generating an exploratory corpus. This is typically done by training a cohort of students on an ITS that produces random, yet \emph{reasonable}, decisions. Subsequently, RL is applied to derive pedagogical policies from this training corpus. However, it is noteworthy that despite this approach, a significant portion of prior research aims to derive pedagogical policies from fewer than 3000 logs of student-tutor interactions; the resultant RL policies may sometimes be proven ineffective \citep{DBLP:conf/aied/AusinMBC20, shen2016reinforcement}.

 Another major obstacle is  \textbf{\emph{reward function}} design.  As a fundamental element in RL, the \emph{reward function} defines the goals to achieve when the agent interacts with the environment, which provides the agent incentive to adopt better decision-making behaviors \citep{kaelbling1996reinforcement}. Specifically, the reward function maps each perceived state (or state-action pair) of the environment to a scalar value and gives it as a credit or punishment to the RL agent, and the agent will learn a decision-making policy accordingly to maximize the expected cumulative rewards. Just as supervised models depend heavily on accurate \emph{labels} for the training dataset, the effectiveness and robustness of RL approaches depend heavily on an accurate \emph{reward} function. 
Despite the importance of the reward function in RL, it is usually difficult to design, especially for \emph{human-centric} tasks like education, where multiple factors need to be covered and traded off. Generally, the reward function is hand-crafted beforehand, separate from the policy induction. However, manually designing an appropriate reward function is always labor-intensive and time-consuming \citep{goyal2019using}. It commonly depends on the domain knowledge, and it is hard to avoid the expertise blind spots \citep{abbeel2004apprenticeship}. Moreover, the manually specified reward function will likely be miss-specified, which can be inconsistent with the expected policy \citep{amodei2016concrete}.

To overcome the two obstacles, \textbf{Apprenticeship Learning (AL)} algorithms have been proposed. Instead of inducing the policy under the guidance of a pre-defined reward function and using a large amount of training data, AL aims at learning via observing and imitating a few demonstrations provided by expert agents, who make decisions \emph{optimally} or \emph{near-optimally} with respect to an unknown underlying reward function \citep{abbeel2004apprenticeship}. By learning from demonstrations, the learner agent, \textit{i.e.}, the apprentice, aims to learn a decision-making policy to behave as well as the experts. Getting rid of the difficulties in reward function design in RL, AL has been widely applied in various applications \citep{asoh2013application, hostetter2023xai, pan2019dissecting,rafferty2015inferring,wang2020inferring}. 

AL approaches have been categorized as being either online or offline. In the former category, the agent learns while interacting with the environment; the latter learns the policy from pre-collected data.
The existing AL are commonly \emph{online}, requiring interacting with the environment iteratively for collecting new data and then updating the model accordingly \citep{abbeel2004apprenticeship, ziebart2008maximum, finn2016guided}.  Online AL algorithms are generally appropriate for domains where interacting with simulations and actual environments is computationally cheap and feasible. On the other hand, for domains such as e-learning, building accurate simulations or simulated students is incredibly challenging because human learning is a rather complex, poorly understood process; moreover, learning policies while interacting with students can not only be unethical but not allowed. Therefore, we focus on offline AL approaches. A recently proposed energy-based distribution matching (EDM) approach \citep{jarrett2020strictly} has advanced the state-of-the-art in \emph{offline} AL. However, EDM assumes all demonstrations are generated using a homogeneous policy with a \emph{single} reward function. In real-world scenarios, the reward function can be \textbf{heterogeneous} across the demonstrations. For example, in the education domain, when students make decisions during learning, their reward functions can be learning-oriented, efficient-oriented, or not learning \citep{yang2020student}. Therefore, in this work, to handle the heterogeneous reward functions, we propose an expectation-maximization(EM)-EDM, a general AL framework \emph{to induce pedagogical policies from given optimal or near-optimal demonstrations, which are assumed to be driven by multiple heterogeneous reward functions}.

While EM-Inverse Reinforcement Learning (EM-IRL) was introduced in \citep{yang2020student}, it is limited because it can only handle relatively small discrete state representation (17 discrete features involved in \citep{yang2020student}). In contrast, our proposed EM-EDM can handle over 140 continuous features.
In this study, we focused on the decisions on whether to present the next problem as a Worked Example (WE), a Problem Solving (PS), or a Faded Worked Example (FWE). In WE, students were given a detailed example showing how the tutor solves a problem; in PS, by contrast, students were tasked with solving the same problem on their own on the ITS; in FWEs, the students and the tutor \emph{co-construct} in that their solutions are intertwined.

In AL, it is usually assumed that the demonstrations are optimally or near optimally executed by experts \citep{abbeel2004apprenticeship}. Thus, the quality of the demonstrations matters in order to induce more effective policies. In this study, rather than gathering exploratory data in DRL, our EM-EDM will derive pedagogical policies directly from the pedagogical decisions and behaviors demonstrated by \textbf{\emph{students}} while interacting with the ITS. In this study, we involve 53 ``optimal'' or ``near-optimal'' demonstrations, which empower students to make their own pedagogical decisions during learning, leading to positive learning outcomes (see details in Section~\ref{TrajSelection-ch3}).

We compare the effectiveness of the policies induced by our proposed EM-EDM against four AL-based baselines using the 53 demonstrations and two policies induced by DRL using 2,716 and 1,819 student-ITSs interactive trajectories, respectively. This evaluation is conducted on two tasks that are related but distinct from each other. For Task 1, we adhere to standard AL research practices by conducting training and evaluation of the methods for pedagogical action prediction. This is accomplished through 5-fold cross-validation, utilizing all selected demonstrations.
Task 2 holds greater relevance and practical significance in the educational data mining field. It involves using trajectories from a previous semester as training data and trajectories from a subsequent semester as testing data. The objective is to assess whether the clustering results obtained from the training data of the previous semester can effectively predict students' actions in the later semester. 
Our overall results showed that, for both tasks, EM-EDM outperforms the four AL baselines across all performance metrics as well as the two DRL baselines. This suggests that EM-EDM is more data efficient and effective in learning students' heterogeneous strategies in ITSs. Beyond this educational setting, we additionally ask whether the heterogeneous-reward principle behind EM-EDM generalizes to a second, higher-stakes human-centric domain: we validate EM-EDM against its two core baselines---behavior cloning (BC) and EDM's own single-reward assumption---on three clinical EHR datasets for septic shock prevention.

In summary, our contributions can be outlined as follows:
\begin{enumerate}
\item  The key contribution lies in the capability of EM-EDM to effectively model complex student pedagogical decision-making processes through the ability to manage a large, continuous state space and its adaptability in handling diverse and heterogeneous reward functions.
\item  The empirical evaluation of EM-EDM involves comparing four state-of-the-art AL baselines and two DRL baselines on two significant tasks.
\item  The results obtained from EM-EDM showed that students' demonstrations exhibit heterogeneous reward functions. Notably, EM-EDM can induce distinct pedagogical policies for different reward functions even with as few as 24 demonstrations.
\item  We validate EM-EDM beyond education on three clinical EHR datasets for septic shock prevention, showing that modeling heterogeneous reward functions is not specific to student pedagogical decisions but generalizes to clinical decision-making as well.
\end{enumerate}

\section{Related Work}
\subsection{RL and DRL for Pedagogical Policy Induction}

In general, there have been two types of prior research on using Reinforcement Learning (RL) to educational policy induction: classic RL and DRL approaches. With the ultimate goal of helping students in the long run, the pedagogical agent in an ITS uses RL to learn a decision-making policy that maximizes the predicted cumulative benefits over a prolonged period \citep{ju2022student}. Despite DRL's enormous success, several challenges are preventing its wider implementation in actual educational institutions; sample inefficiency is one. For instance, learning a suitable policy and generalizing to unknown states requires hundreds of millions of interactions with the environment for Deep Q-Networks (DQN), one kind of classic DRL method. 

The design of reward functions is another significant barrier. They are a key component of reinforcement learning and establish objectives for the agent to meet when interacting with the environment. This motivates the agent to make better decisions \citep{kaelbling1996reinforcement}. In particular, the reward function assigns a credit or punishment to each perceived state (or state-action combination) of the environment. Based on this, the RL agent learns a decision-making policy that maximizes the expected cumulative rewards. An accurate reward function is critical to the efficacy and robustness of reinforcement learning techniques, much as supervised models rely significantly on accurate labels for the training dataset. 

Even though the reward function is crucial to reinforcement learning, designing one is typically challenging, particularly for tasks that focus on people, like teaching, and require balancing various considerations. Usually, the policy induction is done separately from the reward function, which is manually created previously. However, creating a suitable reward function by hand is usually time-consuming and labor-intensive \citep{goyal2019using}. 

To summarize, previous research indicates that RL and DRL-induced pedagogical policies can be sample inefficient and require handling the challenge of reward function design.

\subsection{Apprenticeship Learning (AL)}
Behavior cloning \citep{syed2012imitation,raza2012teaching} is a traditional offline AL technique that learns a mapping from states to actions by avariciously copying the best practices of experts who have been shown \citep{pomerleau1991efficient}. Several inverse reinforcement learning (IRL)-based  \citep{abbeel2004apprenticeship,ziebart2008maximum} and adversarial imitation learning-based \citep{ho2016generative} techniques have been developed to better capture the data distribution in experts' presentations.

In general, iterative loops are used in IRL-based approaches to 1) infer a reward function, 2) induce a policy using traditional reinforcement learning, 3) roll out the learned policy, and 4) update the reward parameters in response to discrepancies between the roll-out behaviors and expert demonstrations. The technical and theoretical drawbacks of the original IRL approaches that adhere to the aforementioned online model will be retained when applied to an offline environment. Furthermore, the reward is typically modeled by IRL-based approaches using a certain tractable format, such as a linear function that maps states and state-action pairings to reward values \citep{abbeel2004apprenticeship,ziebart2008maximum,babes2011apprenticeship}. Moreover, certain batch-IRL have been proposed in order to prevent the learned policy from being implemented.

Typically, adversarial imitation learning-based techniques involve learning a discriminator to discern learned behaviors from expert demonstrations and a generator to implement the policy iteratively. Some off-policy learning techniques based on off-policy actor-critic have been developed under the batch setting in order to avoid rolling out the policy \citep{kostrikov2018discriminator,kostrikov2019imitation}. However, according to \citep{ho2016generative}, these techniques inherit the complicated alternating max-min optimization from general adversarial imitation learning.

Jarrett \textit{et al.} proposed and evaluated the \textbf{EDM} \citep{jarrett2020strictly} on a wide range of benchmarks, online environments like \textit{e.g.}, Acrobot, LunarLander, and BeamRider, as well as offline environment MIMIC-III. It was shown that EDM can outperform both \emph{IRL-based} and \emph{adversarial imitation learning-based} methods. In this work, EDM serves as one of our baselines.

\subsection{AL with Multiple Intentions}
A few AL methods have been proposed to consider multiple reward functions. A Bayesian multi-task IRL was proposed, which models the heterogeneity of reward functions by formalizing it as a statistical preference elicitation via a joint reward-policy prior \citep{dimitrakakis2011bayesian}. Choi and Kim integrated a Dirichlet process mixture model into Bayesian IRL to cluster the demonstrations \citep{choi2012nonparametric}. Using a Bayesian model, they incorporated the domain knowledge of multiple reward functions. Similarly, Arora \textit{et al.} combined the Dirichlet process with a maximum entropy IRL to learn the clusters of demonstration with different reward functions \citep{arora2021min}. 

Babes \textit{et al.} derived an EM-based IRL approach that clusters trajectories based on their different reward functions \citep{babes2011apprenticeship}. As a component of the EM-based IRL, a maximum-likelihood IRL uses a gradient ascent method to optimize the reward parameters, which has successfully identified unknown reward functions. Xi et al. used EM-Inverse Reinforcement Learning (EM-IRL) to subtype students from their pedagogical behavior data \citep{yang2020student}. Still, as most other multiple intentions works, it is limited because it can only handle discrete state representation (17 discrete features involved in \citep{yang2020student}) and can not generalize to large continuous state space.

\subsection{WE, PS, and FWE}

Many studies have examined the effectiveness of WE, PS, and FWE, as well as their different combinations \citep{van2011effects,renkl2002,najar2014}. 
Renkl et al. \citep{renkl2002} compared WE-FWE-PS with WE-PS pairs, and the results showed that the WE-FWE-PS condition significantly outperformed the WE-PS condition on posttest scores. Similarly, Najar et al. \citep{najar2014} compared adaptive WE/FWE/PS with WE-PS pairs and found that the former is significantly more effective than the latter on improving student learning. 
Overall, it is demonstrated that adaptively alternating amongst WE, PS, and FWE is more effective than hand-coded expert rules in improving student learning. 
However, when students make decisions among WE, PS, and FWE, there's no significant difference with tutors making decisions on students' learning performance \citep{guojing2017tutorstudent}. As far as we know, no prior research has explored how to combine students' decision-making with RL-induced policy decision-making to facilitate learning.

\section{Methodology}
AL generally follows the procedure of 1) taking the demonstrations as input to infer the latent reward function, based on which a policy can be induced, then 2) rolling out the learned policy to update the reward function by minimizing the divergence between the rolled-out behaviors versus the demonstrations. This procedure will be conducted iteratively until convergence. 

\begin{figure}
    \centering
    \includegraphics[scale=0.4,alt={Block diagram of the EM-EDM framework: expert trajectories enter the EM loop where the E-step assigns each trajectory a probability of belonging to each cluster, and the M-step uses EDM to induce a separate policy per cluster. The loop repeats until convergence, outputting K cluster-specific policies.}]{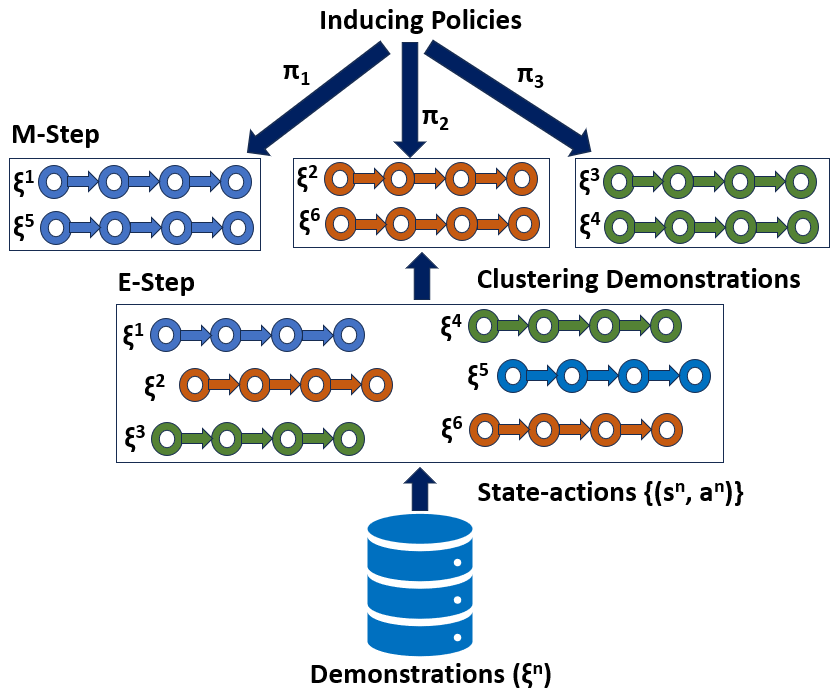}
    \caption{EM-EDM framework.}
    \label{fig:emedm}
\end{figure}

The input of EM-EDM contains $N$ demonstrated trajectories $\{\xi^{n}\} = \{(\mathbf{s}_t^n, a_{t}^{n})|t=1, ..., T^{n}; n = 1,..., N\}$ provided by experts, where $\mathbf{s}_t^n\in \mathbb{R}^m$ is the $t$-th multivariate state with $m$ features, and $a_{t}^{n}$ is the corresponding action in the $n$-th trajectory with the length of $T^{n}$. In this work, we assume that $\{\xi^{n}\}$ are carried out with \emph{multiple} policies. EM-EDM would cluster the trajectories first and then induce policies for each cluster to induce effective policies. Taking our ITSs as an example, EM-EDM will cluster student trajectories and induce different policies or strategies over learned clusters (as shown in Figure \ref{fig:emedm}).

\begin{algorithm}
    \caption{\textbf{EM-EDM}.}
    \label{alg:emedm}
    \renewcommand{\thealgorithm}{}
    \begin{algorithmic}[1]
           \STATE \textbf{Input}: Expert trajectories $\{\xi^{n}\}$, number of clusters $\{K\}$\\ 
           \STATE \textbf{Initialize}: Prior probability $\nu_{j}$ and policy parameter $\theta_{j}$, $j = 1,...K$ randomly
            \REPEAT
                \STATE \textbf{E Step}: Compute the $u_{ij}$, $i = 1,...N$, $j = 1,...K$
                \STATE \textbf{M Step}: Update $\nu_{j}$ and learn $\theta_{j}$ via EDM, $j = 1,...K$
            \UNTIL{\textit{stop\_criteria} is True}
            \STATE \textit{Output}: Cluster-wise policies $\{\Pi^{\theta}_j|j=1,...K\}$ 
   \end{algorithmic}
\end{algorithm}

\subsection{EDM}

As a strictly \emph{offline} AL, energy-based distribution matching (EDM) \citep{jarrett2020strictly} can learn the policy merely based on the experts' demonstrations, not requiring any knowledge of model transitions or off-policy evaluations. It assumes that the demonstrations $\{\xi^{n}\}$ are carried out with a policy $\Pi^{\theta}$ parameterized by $\theta$, driven by a \emph{single} reward function. 

We denote the state-action pair as $(\mathbf{s}, a)$. The occupancy measures for the demonstrations and the learned policy are denoted as $\rho_{\xi}$ and $\rho_{\Pi^{\theta}}$, respectively. 
The probability density for each state-action pair can be measured as:  $\rho_{\Pi^{\theta}}(\mathbf{s},a) =  \mathbb{E}_{\Pi^{\theta}}[\sum_{t=0}^{\infty} \gamma^{t} \mathds{1}{\{\mathbf{s}_{t}=\mathbf{s}, a_{t}=a\}}]$, where $\gamma$ is a discount factor, then the probability density for each state can be measured by: $\rho_{\Pi^{\theta}}(\mathbf{s}) = \sum_{a} \rho_{\Pi^{\theta}}(\mathbf{s},a)$. To induce the policy $\Pi^{\theta}$, our goal is minimizing the KL divergence between $\rho_{\xi}$ and $\rho_{\theta}$: 

\vspace{-0.2cm}
\begin{equation} 
    \argmin_{\theta} D_{KL}(\rho_{\xi}||\rho_{\theta}) = \argmin_{\theta} -\mathbb{E}_{\mathbf{s},a \sim \rho_{\xi}} \log \rho_{\theta}(\mathbf{s},a)
\end{equation}

Since $\Pi^{\theta}(a|\mathbf{s}) = \rho_{\Pi^{\theta}}(\mathbf{s},a)/\rho_{\Pi^{\theta}}(\mathbf{s})$, we can formulate the objective function as: 

\vspace{-0.2cm}
\begin{equation}
\label{eq:6}
    \argmin_{\theta} - \mathbb{E}_{\mathbf{s} \sim \rho_{\xi}} \log \rho_{\Pi^{\theta}}(\mathbf{s}) - \mathbb{E}_{\mathbf{s},a \sim \rho_{\xi}} \log \Pi^{\theta}(a|\mathbf{s}) 
\end{equation} 

\noindent When there is no access to roll out the policy $\Pi^{\theta}$ in an \emph{online} manner, $\rho_{\Pi^{\theta}}(\mathbf{s})$ in the first term of Eq.\eqref{eq:6} would be challenging to estimate. EDM can handle this issue by utilizing an energy-based model \citep{grathwohl2019your}.

According to energy-based model, the probability density $\rho_{\Pi^{\theta}}(\mathbf{s}) \propto e^{-E(\mathbf{s})}$, with $E(\mathbf{s})$ being an energy function. 
Then the occupancy measure for state-action pairs can be represented as:  $\rho_{\Pi^{\theta}}(\mathbf{s},a) = e^{f_{\Pi^{\theta}}(\mathbf{s}) [a]} / Z_{\Pi^{\theta}}$, and the occupancy measure for states can be obtained by marginalizing out the actions: $\rho_{\Pi^{\theta}}(\mathbf{s}) = \sum_{a} e^{f_{\Pi^{\theta}}(\mathbf{s}) [a]} / Z_{\Pi^{\theta}}$. Herein, $Z_{\Pi^{\theta}}$ is a partition function, and $f_{\Pi^{\theta}}: \mathbb{R}^{|S|} \rightarrow \mathbb{R}^{\mathbb{|A|}}$ is a parametric function that mapping each state to $A$ real-valued numbers. 


The parameterization of $\Pi^{\theta}$ implicitly defines an energy-based model over the state's distribution, where the energy function can be defined as: $E_{\Pi^{\theta}}(\mathbf{s}) = -\log \sum_{a} e^{f_{\Pi\theta}(\mathbf{s})[a]}$. 
Under the scope of the energy-based model, the first term in Eq.\eqref{eq:6} can be reformulated as an ``occupancy” loss: 

\vspace{-0.2cm}
\begin{equation}
\label{eq:8}
    \mathcal{L}_{\rho}(\theta) = \mathbb{E}_{\mathbf{s} \sim \rho_{\xi}} E_\theta (\mathbf{s}) - \mathbb{E}_{\mathbf{s} \sim \rho_{\Pi^{\theta}}} E_\theta (\mathbf{s}) 
\end{equation}

\noindent where $\nabla_\theta \mathcal{L}_\rho (\theta) = - \mathbb{E}_{\mathbf{s} \sim \rho_{\xi}} \nabla_\theta \log \rho_{\Pi^{\theta}}(\mathbf{s})$ can be solved by existing optimizers, \textit{e.g.}, stochastic gradient Langevin dynamics \citep{welling2011bayesian}. Therefore, by substituting the first term in Eq.\eqref{eq:6} as Eq.\eqref{eq:8} via energy-based model, we can derive a \emph{surrogate objective function} to get the optimal solution without the need of \emph{online} rolling out the policy. 


\subsection{EM-EDM}

To deal with \emph{multiple} reward functions varying across the demonstrations, Babes-Vroman \textit{et al.} proposed an EM-based inverse reinforcement learning \citep{babes2011apprenticeship} by iteratively clustering the demonstrations in \textit{E-step} and inducing policies for each cluster by IRL in \textit{M-step}. Specifically, in the \textit{M-step}, they explored several IRL methods based on the \emph{discrete} states, which is not scalable for large continuous state spaces, \textit{e.g.}, ITSs logs. 
Enlightened by this EM framework and the success of EDM, we proposed an EM-EDM (Algorithm \ref{alg:emedm}) \citep{xi2023themes}. 

Taking trajectories $\{\xi^{i} | i=1,..., N\}$ as input, with $N$ being the number of trajectories, EM-EDM aims to cluster these trajectories and learn the cluster-wise policies $\{\Pi_j|j=1,...K\}$, where $K$ is the number of clusters. $\nu_{j}$ and $\theta_{j}$, denote the prior probability and the policy parameter for each cluster, and both are randomly initialized.  
The objective function of EM-EDM is to maximize the log-likelihood defined in Eq.\eqref{eq:10}:


\begin{equation}
\label{eq:10}
    \argmax_{\theta_j} \underset{j=1}{\overset{K}{\sum}} \underset{i=1}{\overset{N}{\sum}} log(u_{ij})
\end{equation}

\noindent where $u_{ij}$ denotes the probability that trajectory $\xi^{i}$ follows the policy of the $j$-th cluster. It is defined in Eq.\eqref{eq:5}, with $U$ being a normalization factor.

\begin{equation}
\label{eq:5}
     u_{ij} = Pr(\xi^{i}|\theta_{j}) = \underset{(\mathbf{s},a) \in \xi^{i}}{\prod} \frac{\Pi_{\theta_{j}}(\mathbf{s},a)\nu_{j}}{U}
\end{equation}

During the EM process, in the \textit{E-step}, the probability that trajectory $\xi^{i}$ belonging to cluster $j$ is calculated by Eq.\eqref{eq:5}. 
Then in the \textit{M-step}, the prior probabilities are updated via $\nu_{j} = \sum_{i} u_{ij}/N$, and the policy parameters $\theta_{j}$ is learned by EDM. The \textit{E-step} and \textit{M-step} are iteratively executed until converged. Finally, the output of EM-EDM is the clustered trajectories with their respective policies.

\subsection{Apprenticeship Learning Baselines}

\subsubsection{Behavior cloning (BC)}
Behavior cloning \citep{syed2012imitation,raza2012teaching} is a traditional offline AL technique that learns a mapping from states to actions by avariciously copying the best practices of experts who have been shown \citep{pomerleau1991efficient}. It is a supervised learning approach used in AL where an agent learns to imitate an expert by directly mapping observations to actions based on the expert's demonstrations. The objective is to mimic the expert's behavior. The learning process involves minimizing the difference between the agent's actions and those demonstrated by the expert \citep{bain1999bc}.

In a simplified form, let's consider a dataset of expert demonstrations where each example consists of an observation-action pair: $(s_i, a_i)$, where $s_i$ is an observation and $a_i$ is the corresponding action. The behavior cloning objective is to learn a policy $\pi_{\theta}(a|s)$ parameterized by $\theta$ that approximates the expert's behavior.

The loss function for behavior cloning can be defined as the standard supervised learning loss:

\begin{equation}
    L(\theta) = \frac{1}{N} \sum_{i=1}^{N} \lVert \pi_{\theta}(a_i|s_i) - a_i \rVert^2    
\end{equation}

Where $N$ is the length of trajectories. The goal is to minimize this loss, and the optimal $\theta$ corresponds to a policy that imitates the expert's actions.

It's important to note that while BC is straightforward, it can be sensitive to distribution shifts and compounding errors, especially when the learned policy deviates from the expert's demonstrations.

\subsubsection{Generative Adversarial Imitation Learning (GAIL)}
Generative Adversarial Imitation Learning (GAIL) is another AL approach that uses a generative adversarial network (GAN) \citep{goodfellow2014generative} to model the reward function, making the learned policy less sensitive to differences between the demonstrator and the learner \citep{ho2016gail}. GAIL aims to generate a reward function that encourages the learner's policy to match the demonstrated behavior.

Given a set of expert trajectories $\tau_E \sim \pi_E$, and initial policy and discriminator parameter to be $\theta_0, w_0$, GAIL iteratively updates the discriminator parameters from $w_i$ to $w_{i+1}$ with the gradient as per Equation \ref{eqn:gail1}.

\begin{equation}
    \hat{E}_{\tau_i} \left[ \nabla_w \log(D_w(s, a)) \right] + \hat{E}_{\tau_E} \left[ \nabla_w \log(1 - D_w(s, a)) \right]
    \label{eqn:gail1}
\end{equation}

And then, take a KL-constrained natural gradient step with Equation \ref{eqn:gail2}.

\begin{equation}
\begin{split}
   \hat{E}_{\tau_i} \left[ \nabla_{\theta} \log \pi_{\theta}(a | s) Q(s, a) \right] - \lambda \nabla_{\theta} H(\pi_{\theta}),\\
   \text{ where } Q(\bar{s}, \bar{a}) = \hat{E}_{\tau_i} \left[ \log(D_{w_{i+1}}(s, a)) \,|\, s_0 = \bar{s}, a_0 = \bar{a} \right]
\end{split}
\label{eqn:gail2}
\end{equation}



Where $\pi_{\theta}$ is the policy of the learner, and $\pi_{\text{E}}$ is the expert policy. The discriminator aims to maximize this objective, while the learner's policy minimizes it. The reward function is derived from the discriminator's output.

GAIL addresses distribution shift issues in AL and is more robust than behavior cloning. However, it may face challenges such as mode collapse during GAN training.

\subsubsection{Adversarial Inverse Reinforcement Learning (AIRL)}
Adversarial Inverse Reinforcement Learning (AIRL) is an adversarial approach in AL similar to GAIL that adversarially trains a policy against a discriminator that aims to distinguish the expert demonstrations from the learned policy. Unlike GAIL, AIRL recovers a reward function more generalizable to changes in environment dynamics \citep{fu2018learning}. 

Given a set of expert demonstrations: $\tau_i^E$, initial policy $\pi$ and discriminator $D_{\theta,\phi}$, AIRL iteratively collect trajectories $\tau_i$ by executing $\pi$, train the discriminator via binary logistic regression to classify expert data from samples. The reward is updated in each step following the Equation \ref{eqn:airl1}. Later, the policy is updated with respect to reward using any policy optimization method.

\begin{equation}
r_{\theta,\phi}(s, a, s') = \log D_{\theta,\phi}(s, a, s') - \log(1 - D_{\theta,\phi}(s, a, s'))
\label{eqn:airl1}
\end{equation}

where $\pi_{\theta}$ is the policy of the learner, and $\pi_{\text{expert}}$ is the expert policy. Like GAIL, the discriminator aims to maximize this objective while the learner's policy minimizes it. The reward function is derived from the discriminator's output.

AIRL offers advantages such as handling distribution shifts and being less sensitive to suboptimal demonstrations. However, it may be computationally expensive.

\subsection{Two Deep RL Baselines : CQL \& CQL-T}

\subsubsection{CQL}
A good number of prior research has explored the Deep RL (DRL) (e.g., \citep{ju2020aiedcritical,mandel2014offline,rowe2015improving}), and it has shown that they can be used to induce effective pedagogical policies for ITSs \citep{mandel2014offline,wang2017interactive}. As a DRL baseline, we choose Conservative Q-Learning (CQL) \citep{kumar2020conservative}, known for addressing overestimation issues in comparison to various existing RL methods, including Deep Q-learning, which is extensively explored in prior DRL studies within ITSs.


CQL represents a Q-learning or actor-critic algorithm designed to learn Q-functions, ensuring that the anticipated policy value under the learned Q-function provides a conservative estimate of the actual policy value. To achieve Q-values with lower bounds, CQL goes beyond standard minimization by concurrently minimizing the Q-function under a selected distribution and maximizing it under the data distribution \citep{kumar2020conservative}. The training objective for the Q-function is accordingly structured to attain this goal:

\begin{equation}
\begin{aligned}
\hat{Q}_{\mathrm{CQL}}^\pi:=\arg \min _Q \alpha \cdot(\underbrace{E_{\mathbf{s} \sim \mathcal{D}, \mathbf{a} \sim \mu(\mathbf{a} \mid \mathbf{s})}[Q(\mathbf{s}, \mathbf{a})]}_{\text {minimize Q-values }}\\
-\underbrace{E_{\mathbf{s} \sim \mathcal{D}, \mathbf{a} \sim \hat{\pi}_\beta(\mathbf{a} \mid \mathbf{s})}[Q(\mathbf{s}, \mathbf{a})]}_{\text {maximize Q-values under data }}) \\
+\frac{1}{2} \underbrace{E_{\mathbf{s}, \mathbf{a}, \mathbf{s}^{\prime} \sim \mathcal{D}}\left[\left(Q(s, a)-\left(\mathcal{R}(\mathbf{s}, a)+\gamma\left[\max _{a \in \mathcal{A}} Q\left(\mathbf{s}^{\prime}, a\right)\right]\right)^2\right]\right.}_{\text {standard Bellman error }}
\end{aligned}    
\end{equation}

\vspace{2.5pt}
Then, the optimization of the policy is done w.r.t: \\~\\
\begin{equation}
    \begin{aligned}
\hat{Q}_{\mathrm{CQL}}^\pi: \pi \leftarrow \arg \max _\pi E_\pi\left[\hat{Q}_{\mathrm{CQL}}^\pi\right]
\end{aligned}
\end{equation}

The offline training process continues until it hits convergence, a predetermined error threshold, or the maximum number of iterations. As the amount of accessible data increases, the magnitude of $\alpha$ may decrease, so over here, $\alpha$ > 0 represents a trade-off factor \citep{kumar2020conservative}. Here, $\hat{\pi}_{\beta}$ denotes the behavior policy,
$\gamma \in [0, 1]$ denotes the discount factor, $\mathbf{s}$ is the current state, $\mathbf{s}'$ is the next state and $\mathscr{R}$ corresponds to the reward function.

\subsubsection{CQL-T}
CQL-T refers to the same methodology for policy induction as CQL except that it utilizes a combined reward function of both student training time and Normalized Learning Gain (NLG) motivated by prior research that response time reveals student proficiency \citep{schnipke2002exploring}. In particular, there was a significant negative correlation between student average response time and student final exam score taken at the end of the semester \citep{gonzalez2007innovative}.

\section{Experiments}
\begin{figure}
    \centering
    \includegraphics[scale=0.25,alt={Screenshot of the Pyrenees probability ITS web interface showing a multi-step probability problem with step-by-step input fields, hint buttons, and three presentation mode buttons at the top labeled PS (Problem Solving), WE (Worked Example), and FWE (Faded Worked Example).}]{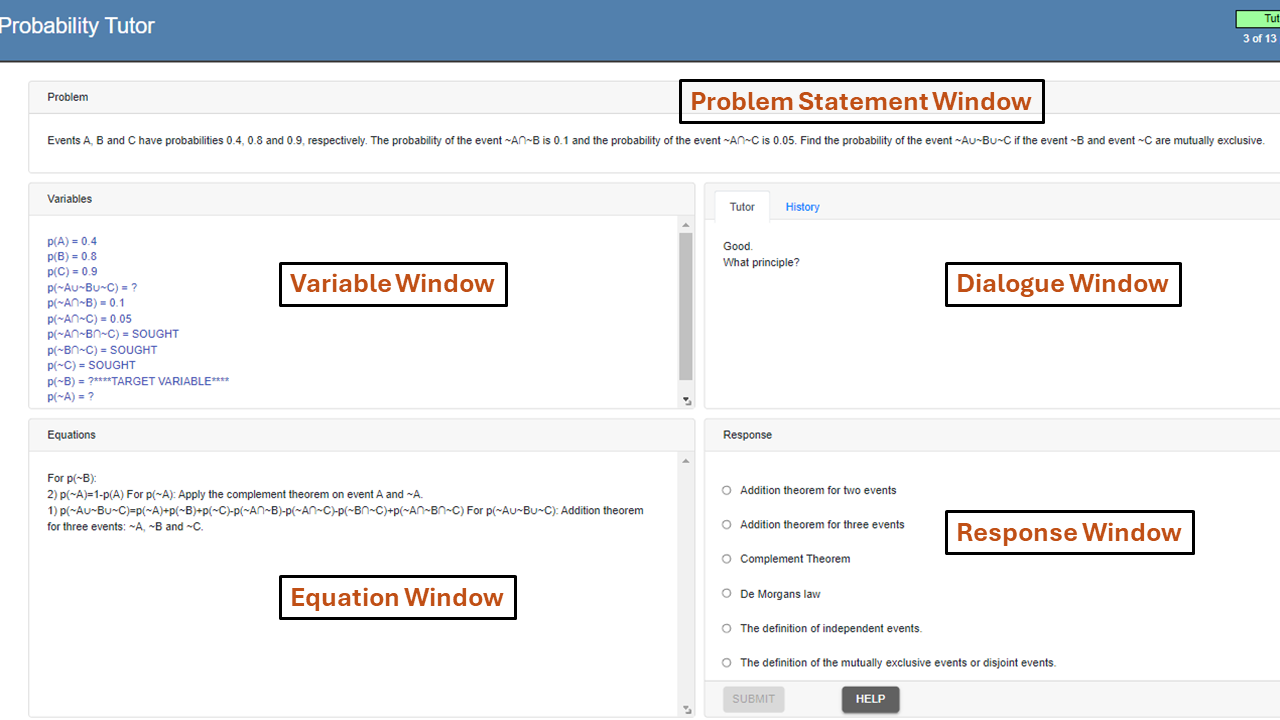}
    \caption{The interface of our probability tutor.}
    \label{fig:pyrenees-interface-ch3}
\end{figure}

We gathered our data by letting undergraduate students from the same major and same year complete an online intelligent tutoring system (ITS) (Figure \ref{fig:pyrenees-interface-ch3}) that introduced them to probability concepts like Bayes' Theorem and Addition Theorem. Using training problems, the students were guided through the instruction. The tutor gave detailed instructions, quick responses, and on-demand assistance for every problem. The assistance was given in the form of progressively more detailed hints. The bottom-out hint, the final one in the sequence, gave the students precise instructions. Students could decide throughout training how to solve the next stage pedagogically by either working through it alone, watching the tutor work through it, or doing it collaboratively. The tutor will ask questions to get the answer if they want to solve it independently; if not, the instructor will show or tell.

\subsection{Data Collection}

Our experiment involves 128 students from two semesters, Spring 21 (67 students) and Spring 22 (61 students). The textbook, pre-test, training, and post-test were the four stages every student taking part in our data collection process went through. All students studied the domain fundamentals from a textbook on probability during the textbook. They studied examples of each principle, read a general explanation, and worked through various single- and multi-principle problems. After that, the students completed a pretest with 8 problems. They would not receive feedback on their responses during this phase and would not be permitted to revisit previous questions (this also applied to the post-test). During training, twelve problems are shown in a fixed sequence, allowing students to practice by choosing to solve it alone (PS), solving it \emph{collaboratively} with the ITS (FWE), or seeing the solution as a worked-out example (WE). Students work through the problem step-by-step, define variables, type equations, etc. The fewest steps required to complete each training problem varied between 20 and 50. While making decisions on how to solve the next problem in training, if PS is chosen, the ITS asks questions to elicit the next step's answer from the student; if FWE is chosen, the ITS chooses to elicit or tell the answer; if WE is chosen, the ITS tells the answer. There were between three and eleven domain principles needed to address each problem. At last, every student completed the post-test, which comprised 12 questions. Four problems were more complex, non-isomorphic multiple-principle problems, and the remaining 8 were isomorphic to the problems provided in the pre-test phase.

Students had to write and solve one or more equations to arrive at an answer for the pre and post-tests. Three scoring criteria were applied: binary, partial credit, and one point for each premise. A solution received one point under the binary rubric for being entirely correct and zero points for being incorrect. The percentage of correctly applied principles seen in the solution determined each problem's score under the partial credit rubric. For example, learners would receive a 0.8 score if they correctly implemented 4 of the 5 potential principles. A point was awarded under the one-point-per-principle rubric for every correctly applied principle. Each student's answers to all the pre-test and post-test problems were double-blind, graded by 2 experienced graders, and then aggregated into one final grade by resolving disagreements if there were any. All test scores were normalized to the range of [0 - 100] for comparison purposes.

Moreover, all students' data were obtained anonymously through an exempt IRB-approved protocol, and we scored them using double-blinded grading rubrics. All data were de-identified, and no demographic data or grades were collected. This research seeks to remove societal harms from lower engagement and retention of students who need more personalized interventions for introductory Computer Science Courses.

\subsection{States \& Actions}

\subsubsection{Continuous State Space}
142 continuous state features were extracted from the student system interaction log data. Here is a brief description of the features:

\noindent $\bullet$ \textbf{Autonomy(10 features)}: the amount of work done by a student, such as the number of elicits since the last tell;\\
\noindent $\bullet$ \textbf{Temporal (29 features)}: time-related information about the student's behavior, such as the average time per step;\\
\noindent $\bullet$ \textbf{Problem Solving (35 features)}: information about the current problem-solving context, such as problem difficulty;\\
\noindent $\bullet$ \textbf{Performance (57 features)}: information about the student's performance so far, such as the percentage of correct entries;\\
\noindent $\bullet$ \textbf{Hints (11 features)}: information about the student's hint usage, such as the total number of hints requested.

\subsubsection{Pedagogical Action Space}
The students can take three decisions which can be interpreted as actions which are elicit(PS), tell(WE) or collaboration(FWE), i.e., to elicit the solution by themselves through asking questions, to let the tutor tell them the solution directly or to work with the tutor collaboratively towards the solution respectively.

\subsection{Selecting Demonstrations (A total of 53)}\label{TrajSelection-ch3} 
To enhance the selection of high-quality demonstrations for inducing more accurate yet effective AL policies, it is typically believed that the experts are performing the demonstrations in an optimal or near-optimal manner \citep{abbeel2004apprenticeship}. 
Our original dataset contains 128 students from two semesters, Spring 21 (67 students) and Spring 22 (61 students). Each student spent \~ 2 hours on the probability ITS and completed around 400 steps.
To select higher-quality trajectories from all students' interaction with the ITS, we use a qualitative measurement called Quantized Learning Gain (QLG) \citep{lin2017comparisons}, which is a binary qualitative measurement of students' learning gains from the pretest to the posttest to determine whether a student has benefited from a learning environment \citep{mao2018tracing}.

\begin{figure}
    \centering
    \includegraphics[scale=0.3,alt={Flowchart of the Quantized Learning Gain (QLG) metric. Students are split into Low, Medium, and High performance groups based on pre-test and post-test scores. Solid arrows show High QLG paths: remaining in High or moving up to High. Dashed arrows show Low QLG paths: staying in Low or Medium, or dropping to a lower group.}]{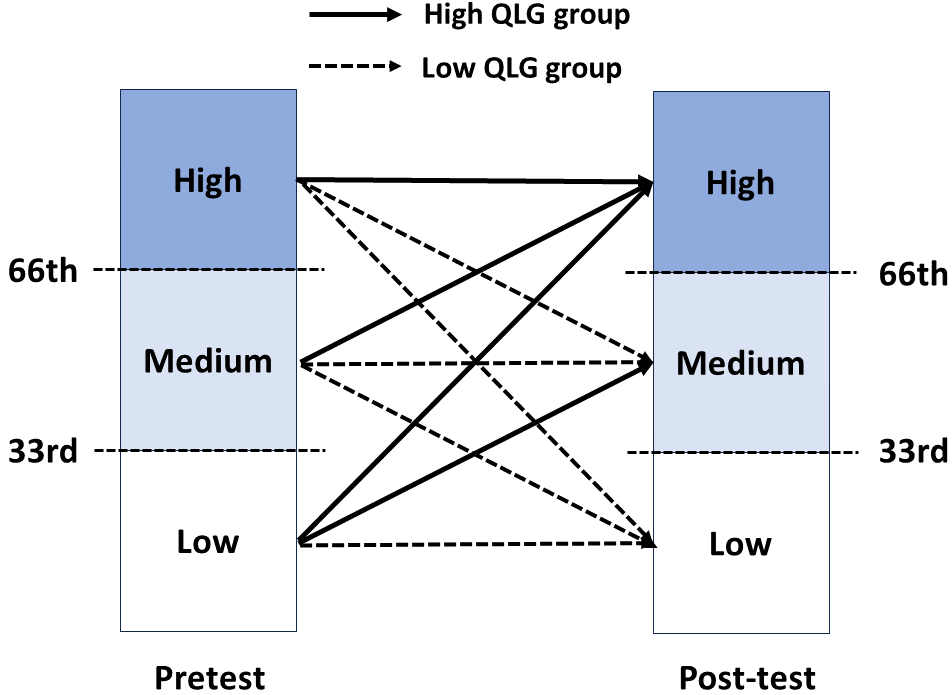}
    \caption{Quantized Learning Gain.}
    \label{fig:qlg-ch3}
\end{figure}

To calculate QLG, first, students were split into low, medium, and high-performance groups based on whether they scored below the 33rd percentile, between the 33rd and 66th percentile, or higher than the 66th percentile in pre-test and post-test, respectively. Once a student's pre-test and post-test performance groups (high, medium, or low) are decided, the student is a ``High'' QLG if he/she moved from a lower performance group to a higher performance group from pre-test to post-test or remained in ``high'' performance groups; whereas a ``low'' QLG is assigned to the student if he/she either moved from a higher performance group to a lower performance group from pre-test to post-test or stayed at a ``low'' or ``medium'' groups (as shown in Figure \ref{fig:qlg-ch3}). In Figure \ref{fig:qlg-ch3}, solid lines represent the formation of the High QLG groups, and dashed lines represent the formation of the Low QLG groups. Following this procedure, Table \ref{tab:qlg_stat} illustrates the number of trajectories in each category across the two semesters. This procedure resulted in a total of \textbf{53} ``high-quality'' student trajectories, which are treated as optimal or near-optimal demonstrations in the following analysis.


\renewcommand{\arraystretch}{1.5}
\begin{table}
\footnotesize
    \centering
    \caption{\centering No. of Students in High QLG Group.}
    \setlength\tabcolsep{5pt}
    \begin{tabular}{c|cc} \hline
         \thead{Category} & \thead{Spring 21} & \thead{Spring 22}\\
        \hline
        $High_{pre}$ $\rightarrow$ $High_{post}$ & 9 & 13 \\
        $Medium_{pre}$ $\rightarrow$ $High_{post}$ & 8 & 9 \\
        $Low_{pre}$ $\rightarrow$ $High_{post}$ & 3 & 2 \\
        $Low_{pre}$ $\rightarrow$ $Medium_{post}$ & 4 & 5 \\ \hline
        \textbf{Total} & \textbf{24} & \textbf{29} \\
        \hline
    \end{tabular}
    \label{tab:qlg_stat}
\end{table}

\subsection{Trajectories for DRL policy Induction}
Our two DRL policies are induced using pre-collected training data collected by \emph{tutor} making decisions, and many times, the tutor would produce random yet \emph{reasonable} decisions. All students used the same probability tutor, followed the same general procedure as described above, studied the same training materials, and worked through the same training problems.  It has the same  142 features described above and the same three types of pedagogical decisions \emph{but made by the tutor}. The two DRL policies differ in the reward function.

The CQL policy induction was offline, containing \textbf{2,716} students' interaction logs over 13 semesters of classroom studies (Fall 2015 to Spring 2022). Its reward function is based on the students' Normalized Learning Gain (NLG), which measures their learning gain irrespective of their incoming competence. NLG is defined as $\frac{posttest-pretest}{\sqrt{1-pretest}}$, where 1 is the maximum score for both pre- and post-test. 

The CQL-T policy induction was done offline using pre-collected training data containing \textbf{1,819} students' interaction logs over 7 semesters of classroom studies (Spring 2020 to Spring 2023). Its reward function balances NLG and time spent on the tutor.


\subsection{Experimental Settings}

\renewcommand{\arraystretch}{1.5}
\begin{table*}
\footnotesize
    \centering
    \caption{\centering Comparing EM-EDM to baselines on predicting student pedagogical actions (5-fold Cross-validation). The best methods are in bold, and the overall best is highlighted with *.}
    \setlength\tabcolsep{5pt}
    \resizebox{\textwidth}{!}{%
    \begin{tabular}{c|c|ccccccc} \hline
        \thead{Category} & \thead{Methods} & \thead{Acc} & \thead{Rec} & \thead{Prec} & \thead{F1} & \thead{AUC} & \thead{APR} & \thead{Jaccard} \\
        \hline
        \multirow{2}{*}{\textbf{DRL}} & CQL-T &  0.296(0.05) & 0.274(0.07) & 0.369(0.06) & 0.259(0.03) & 0.454(0.06) & 0.357(0.03) & 0.161(0.03)\\
        & \textbf{CQL} &  \textbf{0.376(0.02)} & \textbf{0.358(0.04)} & \textbf{0.329(0.04)} & \textbf{0.325(0.03)} & \textbf{0.526(0.02)} & \textbf{0.382(0.01)} & \textbf{0.221(0.02)} \\ \hline
        \multirow{5}{*}{\textbf{AL}} & BC  &  0.354(0.02) & 0.342(0.02) & 0.346(0.02) & 0.333(0.02) & 0.508(0.02) & 0.339(0.01) & 0.204(0.01)\\
        & GAIL &  0.387(0.05) & 0.355(0.03) & 0.360(0.07) & 0.311(0.04) & 0.515(0.02) & 0.343(0.01) & 0.197(0.03)  \\
        & AIRL &  0.365(0.03) & 0.375(0.02) & 0.382(0.02) & 0.330(0.02) & 0.533(0.01) & 0.353(0.01) & 0.204(0.02) \\
        & \textbf{EDM} &  \textbf{0.737(0.05)} & \textbf{0.708(0.05)} & \textbf{0.723(0.06)} & 0.\textbf{705(0.06)} & \textbf{0.836(0.04)} & \textbf{0.700(0.05)} & \textbf{0.563(0.06)}\\ \cline{2-9}
        & \textbf{EM-EDM} &  \textbf{0.795(0.05)}* & \textbf{0.774(0.06)}* & \textbf{0.776(0.05)}* & \textbf{0.769(0.06)}* & \textbf{0.877(0.02)}* & \textbf{0.762(0.05)}* & \textbf{0.639(0.08)}* \\ \hline
    \end{tabular}%
    }
    \label{tab:metrics_cv}
\end{table*}

To evaluate the effectiveness of EM-EDM, we compared it against four AL and two DRL baselines, which include: 

\noindent $\bullet$ \textbf{CQL} \citep{kumar2020conservative}, 
While DQN is one of the most widely applied DRL methods, it can suffer from overestimation bias. But Conservative Q-Learning (CQL) \citep{kumar2020conservative} is designed to address this overestimation bias in Q-learning. This is why we chose CQL as a DRL baseline.

\noindent $\bullet$ \textbf{CQL-T} \citep{kumar2020conservative}, which has the same method as CQL except utilizing a reward function that balances performance and time on task. So when two students learn the same, those who spend less time would get higher rewards.

\noindent $\bullet$ \textbf{BC} \citep{syed2012imitation,raza2012teaching,bain1999bc}, which directly learns a policy using supervised learning on state-action pairs from expert demonstrations. It is a simple approach to learning a policy, but the policy often generalizes poorly and does not recover well from errors.

\noindent $\bullet$ \textbf{GAIL} \citep{ho2016gail}, which is a model-free AL algorithm that learns a policy by simultaneously training it with a discriminator that aims to distinguish expert trajectories against trajectories from the learned policy.

\noindent $\bullet$ \textbf{AIRL} \citep{fu2018learning}, similar to GAIL, adversarially trains a policy against a discriminator that aims to distinguish the expert demonstrations from the learned policy. Unlike GAIL, AIRL recovers a reward function more generalizable to changes in environment dynamics.

\noindent $\bullet$ \textbf{EDM} \citep{jarrett2020strictly}, the state-of-the-art offline AL. It has been demonstrated that EDM can outperform many competitive cutting-edge AL methods with a \emph{single} reward function, thus we will not repetitively conduct all those comparisons here.

\noindent $\bullet$ \textbf{EM-EDM} \citep{xi2023themes}, our proposed method, which assumes the demonstrations follow \emph{multiple} reward functions varying across trajectories (while remaining the same within each trajectory).

We compare the effectiveness of the policies induced by our proposed EM-EDM against four AL-based baselines using the 53 demonstrations and two policies induced by DRL using 2,716 and 1,819 student-ITS interactive trajectories, respectively. All the methods were trained and tested on student interaction trajectories through 5-fold cross-validation; as we do not have access to ground truth rewards, we evaluate performance according to action-matching on held-out test data. We used the same held-out test data to evaluate the performance of CQL and CQL-T-induced policies. We employed the metrics of Accuracy (Acc), Recall (Rec), Precision (Prec), F1-score (F1), AUC, APR, and Jaccard score for evaluation. All the model parameters were determined by cross-validation. For EM-EDM, the optimal cluster number was determined heuristically as $4$ for Task 1 and as $5$ for Task 2 by iteratively implementing the EM until empty clusters were generated or the log-likelihood of the clustering results varied smaller than a pre-defined threshold. Based on our observation, the clustering likelihood for EM-EDM converges within $80$ iterations. For fair comparisons, the optimal parameters in other baselines and ablations were also determined by cross-validation.

\renewcommand{\arraystretch}{1.5}
\begin{table*}
\footnotesize
    \centering
    \caption{\centering Comparing EM-EDM to baselines on predicting future students' pedagogical actions. The best methods are in bold, and the overall best is highlighted with *.}
    \setlength\tabcolsep{5pt}
    \begin{tabular}{c|c|ccccccc} \hline
        \thead{Category} & \thead{Methods} & \thead{Acc} & \thead{Rec} & \thead{Prec} & \thead{F1} & \thead{AUC} & \thead{APR} & \thead{Jaccard} \\
        \hline
        \multirow{2}{*}{\textbf{DRL}} & CQL-T  & 0.217 & 0.203 & 0.313 & 0.199 & 0.391 & 0.332 & 0.115 \\ 
        & \textbf{CQL} & \textbf{0.321} & \textbf{0.354} & \textbf{0.284} & \textbf{0.308} & \textbf{0.506} & \textbf{0.366} & \textbf{0.198} \\ \hline
        \multirow{5}{*}{\textbf{AL}} & BC  & 0.328 & 0.334 & 0.344 & 0.317 & 0.501 & 0.335 & 0.190 \\ 
        & GAIL & 0.393 & 0.362 & 0.387 & 0.337 & 0.522 & 0.345 & 0.210 \\ 
        & AIRL & 0.341 & 0.355 & 0.351 & 0.331 & 0.517 & 0.343 & 0.202 \\ 
        & \textbf{EDM} & \textbf{0.669} & \textbf{0.654} & \textbf{0.663} & \textbf{0.656} & \textbf{0.789} & \textbf{0.656} & \textbf{0.498} \\ \cline{2-9}   
        & \textbf{EM-EDM} &  \textbf{0.745}* & \textbf{0.740}* & \textbf{0.745}* & \textbf{0.740}* & \textbf{0.871}* & \textbf{0.765}* & \textbf{0.589}* \\ \hline
    \end{tabular}
    \label{tab:metrics_xsem}
\end{table*}

\subsection{Evaluation Metrics}
We evaluate and compare performance using Accuracy, Precision, Recall, F1 Score, AUC (Area Under the ROC Curve), APR (Area Under the Precision-Recall Curve), and Jaccard score. Jaccard score is used to assess the similarity between the predicted set and the true set of labels, calculated by the size of the intersection divided by the size of the union of two label sets. Because of our task's nature, we consider AUC and Jaccard Score the most crucial metrics, as they are thought to be generally robust.

\section{Results}
We compare our proposed EM-EDM against the six baselines (two DRL-based and four AL-based) on predicting student pedagogical actions. More specifically, they are compared on two related but different tasks. In Task 1, we follow the standard AL research by training and evaluating all models on all the 53 good trajectories through 5-fold Cross-validation \citep{jarrett2020strictly}. Task 2 is more relevant and practical to the educational data mining field by using Spring 21's good trajectories (24) as training and Spring 22's good trajectories (29) trajectories as testing. Our goal in Task 2 is to check whether our clustering results trained from the previous semester can be used to predict future semester students' actions. We also investigate the distribution of the clusters learned from both tasks and check for significant differences among them.

\subsection{Modeling Heterogeneous Student Pedagogical Strategies}

\subsubsection{Comparing Seven Approaches on Student Pedagogical Action Prediction}
Table \ref{tab:metrics_cv} reports our results from Task 1, comparing the seven approaches to student pedagogical action prediction using all 53 good trajectories by 5-fold cross-validation. The best results among baselines are in bold, and the overall best results are highlighted with *. 

\begin{figure}
  \centering

  \begin{subfigure}{0.7\textwidth}
    \centering
    \includegraphics[width=\linewidth,alt={Critical difference diagram for AUC over 5-fold cross-validation. EM-EDM ranks first (mean rank 1.0), followed by EDM (rank 2.8). BC, GAIL, AIRL, CQL, and CQL-T occupy lower ranks. EM-EDM is statistically significantly better than all baselines at p=0.05 (no connecting bar).}]{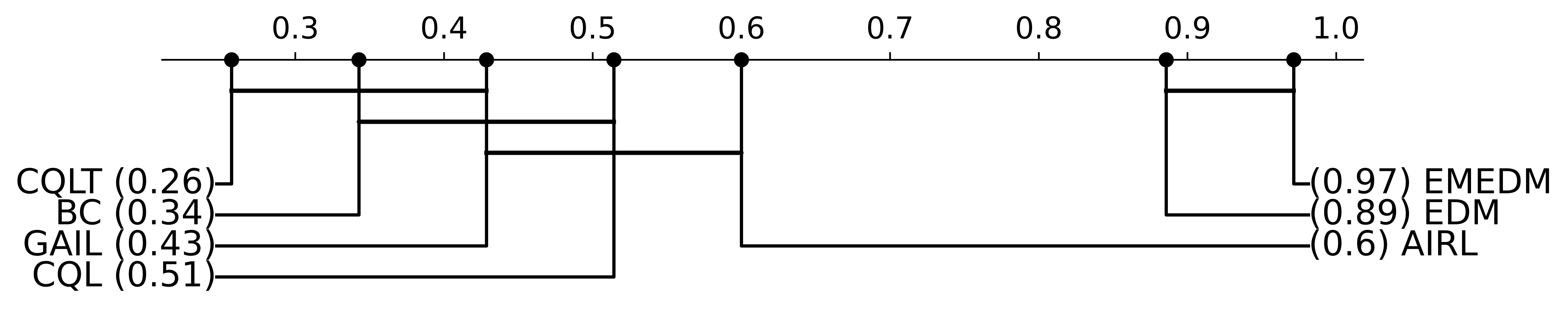}
    \caption{AUC.}
    \label{subfig:cd_auc-ch3}
  \end{subfigure}

  \begin{subfigure}{0.7\textwidth}
    \centering
    \includegraphics[width=\linewidth,alt={Critical difference diagram for Jaccard score over 5-fold cross-validation. EM-EDM ranks first (mean rank 1.0), followed by EDM. BC, GAIL, AIRL, CQL, and CQL-T occupy lower ranks. EM-EDM is statistically significantly better than all baselines at p=0.05.}]{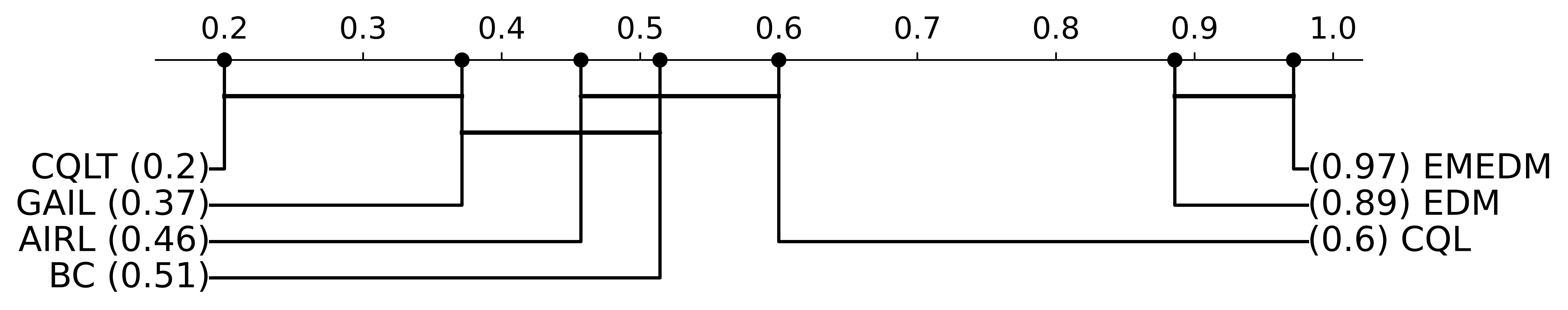}
    \caption{Jaccard Score.}
    \label{subfig:cd_jac-ch3}
  \end{subfigure}

  \caption{Critical difference diagram with Conover Friedman test over (a) AUC and (b) Jaccard Score.}
  \label{fig:cd_plot-ch3}
\end{figure}

To better illustrate our results, we use the Critical Difference (CD) diagram, a powerful tool to statistically compare several classifiers' performance on multiple datasets in a robust way \citep{demvsar2006statistical}. At first, we conducted the Friedman test to check if there is any significant difference among the models across the folds of cross-validation in terms of AUC and Jaccard scores. If there is a significant difference, then we conduct the pairwise Conover post hoc test \citep{friedman1937rank}, \citep{conover1979comparison} to get the ranks and plot the CD diagrams (Figure \ref{fig:cd_plot-ch3}). In the CD diagrams in Figure \ref{fig:cd_plot-ch3}, the position of the models represents their mean ranks according to the corresponding metric across all folds, where low ranks indicate that a model wins more often than its competitors with higher ranks. Two or more models are connected with each other if we can not tell their performances apart, in the sense of statistical significance, with a confidence level of $0.05$.

As shown in Table \ref{tab:metrics_cv}, EM-EDM outperforms all the baselines across all evaluation metrics and the pairwise differences in terms of AUC and Jaccard score are significant according to the CD diagram in Figure \ref{fig:cd_plot-ch3}. EDM outperforms all other baselines except EM-EDM assuming a \emph{single} reward function across the trajectories which proves it's strength in inducing policy from demonstrations as a state-of-the-art offline AL method. But when EM-EDM outperforms EDM, it indicates that there is heterogeneity among the 53 students' pedagogical strategies and considering the heterogeneity would be a better way for modeling complex human-centric tasks, such as student strategy recognition in education. Besides, probably the reason behind the other AL methods did not perform as good as the EDM, is because they cannot take the state distribution into account when learning the policy. 

However, among the DRL baselines, CQL performs better than CQL-T, which suggests that the students are more likely to be learning efficiency-oriented rather than time efficiency-oriented as the reward function of CQL-T has time in it while CQL does not. 

Though among the DRL methods, CQL performs better, it falls short when compared to EDM and EM-EDM even after being trained with a large amount of data, which shows us that EM-EDM is a much more data-efficient approach for modeling student pedagogical strategies.

\subsubsection{Distribution of the Discovered Clusters}

When applied to Task 1, EM-EDM discovered 4 clusters or subgroups among all the 53 good student trajectories. The number of students per subgroup are 25(47.17\%), 13(24.53\%), 11(20.75\%) and 4(7.55\%) which shows an imbalance distribution. A one-way ANOVA test showed no significant difference in the subgroups' learning performances like post-test score, NLG, Iso\_NLG, and Training Time. First, since we used only the good trajectories, it is likely that the students would not have much difference in their learning performances.

\subsection{Future Student's Pedagogical Action Prediction Leveraging the Past}
\subsubsection{Clustering Prediction Tasks}
Table \ref{tab:metrics_xsem} reports our findings from semester-based evaluation where we trained the models on Spring 21 semester's 24 good trajectories to learn the clusters and using them to predict Spring 22 semester's 29 good trajectories pedagogical actions. The best results among baselines are in bold, and the overall best results are highlighted with *. 

As shown in Table \ref{tab:metrics_xsem}, EM-EDM outperforms all baselines across all evaluation metrics in predicting future students' pedagogical actions too. The pairwise performance comparison outcomes among the seven methods are similar to Task 1. However, it is important to emphasize that this task of future students' pedagogical action prediction using a past one is a much stricter approach than the standard cross-validation \citep{mao2021dissertation}. This result indicates that we can model student strategies over time despite population change, probably because there are strategies that have followers from both populations.

\subsubsection{Distribution among the Discovered Clusters}

When applied on Task 2, EM-EDM discovered 5 clusters or subgroups among Spring 21 semester's 24 good trajectories. Using the clustering results obtained, Spring 22 semester's 29 good trajectories were clustered. Table \ref{tab:dist_task2} shows both the semesters' distribution of students across those 5 clusters. We found no significant difference between two semester's cluster distributions, $\chi^2(1, 5) = 3.038, p = .551$.

\renewcommand{\arraystretch}{1.5}
\begin{table}
\footnotesize
    \centering
    \caption{\centering Distribution among the Discovered Clusters in Task 2 (C0-C4 refers to the 5 clusters).}
    \setlength\tabcolsep{5pt}
    \begin{tabular}{c|ccccc|c}
        \hline
        \thead{Semester} & \thead{C0} & \thead{C1} & \thead{C2} & \thead{C3} & \thead{C4} & \thead{Total}\\\hline
        Spring 21 & 3 & 9 & 8 & 2 & 2 & 24\\
        Spring 22 & 6 & 5 & 11 & 3 & 4 & 29\\
        \hline
    \end{tabular}
    \label{tab:dist_task2}
\end{table}

\section{Results for Healthcare}
\label{sec:results_healthcare_edm}

To test whether the heterogeneous-reward principle behind EM-EDM extends beyond education, we draw on the septic shock prevention evaluation of Chapter~\ref{chap:aied25}, where EM-EDM appears as one of the baselines against which THEMES is compared on three clinical EHR datasets: CCHS, Mayo Clinic, and MIMIC-III (see Section~\ref{sec:datasets_evaluation} of Chapter~\ref{chap:aied25} for full dataset and preprocessing details). Table~\ref{tab:emedm_healthcare} isolates the subset of that evaluation relevant here: behavior cloning (BC) and GP+DQN as the AL and DRL baselines, single-reward EDM, and heterogeneous-reward EM-EDM---the same AL-baselines-plus-DRL-baseline-plus-EM-EDM comparison used throughout this chapter's ITS results. The best result among the three baselines (BC, GP+DQN, EDM) is shown in bold, and results where EM-EDM outperforms all three are additionally marked with $^{*}$.

\renewcommand{\arraystretch}{1.0}
\setlength{\tabcolsep}{1.2 pt}
\begin{table*}
\caption{\centering Comparing EM-EDM against its core baselines (BC, GP+DQN, EDM) over three healthcare datasets.}
\label{tab:emedm_healthcare}
\small
\centering
\resizebox{\textwidth}{!}{%
\begin{tabular}{c|cccc|cccc|cccc} \hline
\multirow{2}{*}{\thead{Methods}} & \multicolumn{4}{c|}{\thead{MIMIC-III}} & \multicolumn{4}{c|}{\thead{CCHS}} & \multicolumn{4}{c}{\thead{Mayo}} \\ \cline{2-13}
 & \thead{F1} & \thead{AUC} & \thead{AP} & \thead{JAC} & \thead{F1} & \thead{AUC} & \thead{AP} & \thead{JAC} & \thead{F1} & \thead{AUC} & \thead{AP} & \thead{JAC} \\ \hline

BC
& $.497_{\pm .01}$ & $.502_{\pm .00}$ & $.485_{\pm .07}$ & $.331_{\pm .01}$
& $.499_{\pm .01}$ & $.500_{\pm .01}$ & $.233_{\pm .02}$ & $.376_{\pm .01}$
& $.503_{\pm .01}$ & $.504_{\pm .01}$ & $.038_{\pm .01}$ & $.475_{\pm .01}$ \\

GP+DQN
& $.418_{\pm .04}$ & $.516_{\pm .03}$ & $.515_{\pm .04}$ & $.349_{\pm .05}$
& $.333_{\pm .03}$ & $.511_{\pm .04}$ & $\mathbf{.526}_{\pm .03}$ & $.211_{\pm .03}$
& $.326_{\pm .01}$ & $.460_{\pm .02}$ & $.497_{\pm .04}$ & $.269_{\pm .05}$ \\

EDM
& $.634_{\pm .07}$ & $.635_{\pm .07}$ & $.590_{\pm .05}$ & $.471_{\pm .08}$
& $\mathbf{.631}_{\pm .04}$ & $.553_{\pm .03}$ & $.522_{\pm .01}$ & $\mathbf{.479}_{\pm .05}$
& $.701_{\pm .10}$ & $\mathbf{.640}_{\pm .04}$ & $\mathbf{.515}_{\pm .01}$ & $.558_{\pm .13}$ \\ \hline

EM-EDM
& $\mathbf{.778}^{*}_{\pm .05}$ & $\mathbf{.779}^{*}_{\pm .05}$ & $\mathbf{.717}^{*}_{\pm .05}$ & $\mathbf{.641}^{*}_{\pm .07}$
& $.607_{\pm .02}$ & $\mathbf{.605}^{*}_{\pm .04}$ & $\mathbf{.547}^{*}_{\pm .02}$ & $.444_{\pm .03}$
& $\mathbf{.796}^{*}_{\pm .09}$ & $.597_{\pm .04}$ & $.511_{\pm .01}$ & $\mathbf{.684}^{*}_{\pm .12}$ \\ \hline

\end{tabular}%
}
\end{table*}

On MIMIC-III, EM-EDM sweeps every metric, consistent with the ITS results above: the larger and more heterogeneous the demonstrator population, the more there is for heterogeneous-reward modeling to exploit. On CCHS and Mayo, the pattern is more mixed---EM-EDM still wins AUC and AP on CCHS and F1 and Jaccard on Mayo, but single-reward EDM (and, on CCHS's AP, GP+DQN) edges it out on the remaining metrics in each case. This is not a contradiction of the heterogeneity hypothesis so much as a reminder that its payoff scales with dataset size and behavioral diversity: CCHS and Mayo are both an order of magnitude smaller than MIMIC-III (Section~\ref{sec:datasets_evaluation} of Chapter~\ref{chap:aied25}), leaving less room for EM-EDM's cluster-specific policies to separate from a single pooled reward function. Taken together, these results indicate that the core finding of this chapter---that recognizing heterogeneous reward functions improves policy induction from imperfect human demonstrations---is not an artifact of the ITS setting, but reappears, with the same dataset-size-dependent texture, in a clinical decision-making domain.

\section{Summary}
Results suggest that our proposed EM-EDM framework outperforms all the chosen baselines on both tasks for students' next pedagogical action prediction. The baselines we utilized are considered state-of-the-art and sufficiently fair for the prediction task at hand. In a related paper \citep{jarrett2020strictly}, BC was employed as the initial baseline for a similar prediction task involving the real-world healthcare dataset MIMIC-III. Notably, BC outperformed other baselines in 2-Action settings based on AUC metrics. However, GAIL and AIRL exhibit superior capabilities compared to BC, as they employ adversarial training techniques against a discriminator tasked with distinguishing expert demonstrations from learned policies, thus justifying their selection as better baselines \citep{ho2016gail, fu2018learning}. We posit that the under-performance of the baselines may be attributed to the irregular and noisy nature of the student-interaction log data, which is inherently challenging to model. Nevertheless, EDM and EM-EDM demonstrate proficiency in handling such complexities owing to their strengths in energy-based modeling and consideration of heterogeneity. Moreover, our proposed EM-EDM framework is built upon the premise of recognizing the heterogeneity in student pedagogical decision-making strategies. Clustering allows us to group students who follow similar strategies together, thereby enabling cluster-based statistical analyses. As detailed in Sections 5.1.2 and 5.2.2 of our paper, these analyses provide additional insights that can inform future research directions.

In this study, we investigated heterogeneous student pedagogical strategy modeling via our proposed EM-EDM, a general AL framework, which induces effective pedagogical policies from given optimal or near-optimal demonstrations. To evaluate the performance of EM-EDM, we apply it on two different but related tasks which involves pedagogical action prediction and compare against four AL-based baselines and two policies induced by DRL. The results showed that, for both tasks, EM-EDM outperforms the four AL baselines across all performance metrics as well as the two DRL baselines. We believe, the key impact of this study lies in the capability of EM-EDM to effectively model complex student pedagogical decision-making process through the ability to manage a large, continuous state space and its adaptability in handling diverse and heterogeneous reward functions. The results obtained from EM-EDM showed that students' demonstrations exhibit heterogeneous reward functions. Moreover, EM-EDM can induce distinct pedagogical policies for different reward functions even with as few as 24 demonstrations. Beyond education, Section~\ref{sec:results_healthcare_edm} shows that the same heterogeneous-reward principle carries over to septic shock prevention on three clinical EHR datasets, with the clearest gains on the largest and most heterogeneous cohort (MIMIC-III)---indicating that this chapter's central finding is not specific to student pedagogical decision-making. In the future, we will carry out this study for upcoming semesters. We will also investigate evolving reward function across demonstrations.
\chapter{Modeling Evolving Reward Functions}
\label{chap:aied25}

* \textit{This chapter has been published as:}
\begin{quote}
\textbf{Md Mirajul Islam}, Xi Yang, Rajesh Debnath, Adittya Soukarjya Saha, Min Chi.  \textbf{“A Generalized Apprenticeship Learning Framework for Capturing Evolving Student Pedagogical Strategies.”} \textit{The 26th International Conference on Artificial Intelligence in Education (AIED 2025).}
\end{quote} 

\begin{quote}
Xi Yang, \textbf{Md Mirajul Islam},Ge Gao, Min Chi. \textbf{“THEMES: An Offline Apprenticeship Learning Framework for Evolving Reward Functions.”} \textit{The 31st ACM SIGKDD Conference on Knowledge Discovery and Data Mining  (KDD 2025).}
\end{quote}

Reinforcement Learning (RL) and Deep Reinforcement Learning (DRL) have advanced rapidly in recent years and have been successfully applied to e-learning environments like intelligent tutoring systems (ITSs). Despite great success, the broader application of DRL to educational technologies has been limited due to major challenges such as \emph{sample inefficiency} and difficulty designing the \emph{reward function}. In contrast, Apprenticeship Learning (AL) uses a few \emph{expert demonstrations} to infer the \textbf{expert's underlying reward functions} and derive decision-making policies that generalize and replicate optimal behavior. In this work, we leverage a generalized AL framework, THEMES, to induce effective pedagogical policies by capturing the complexities of the expert student learning process, where \textbf{multiple reward functions may dynamically evolve over time}. We evaluate the effectiveness of THEMES against six state-of-the-art baselines, demonstrating its superior performance and highlighting its potential as a powerful alternative for inducing effective pedagogical policies and show that it can achieve high performance, with an AUC of 0.899 and a Jaccard of 0.653, using only 18 trajectories of a previous semester to predict student pedagogical decisions in a later semester.

\section{Overview}
In most STEM e-learning environments, the system follows a sequential decision-making process, deploying a \emph{pedagogical policy} to determine \emph{what} action to take and \emph{when} at each discrete time step. In such environments, student-system interactions can be modeled as sequential decision-making problems under uncertainty, formulated within the framework of Reinforcement Learning (RL)—a paradigm that optimizes long-term rewards without requiring knowledge of the ‘best’ decisions at each immediate time step \citep{sutton2018reinforcement}. Recent research has demonstrated the effectiveness of RL and Deep RL (DRL) in deriving data-driven pedagogical policies to enhance student learning in Intelligent Tutoring Systems (ITSs) (e.g., \citep{abdelshiheed2023bridging,ju2020aiedcritical,mandel2014offline,rowe2015improving}). Despite their success, several challenges hinder its broader application in e-learning systems. One major challenge is \textbf{sample inefficiency}, as classic DRL algorithms like Deep Q-Networks (DQN) require millions of interactions to learn effective policies. Many studies attempt to derive pedagogical policies from fewer than 3,000 student-ITS interactive  trajectories, often leading to ineffective results \citep{DBLP:conf/aied/AusinMBC20,shen2016reinforcement}. Another key challenge is \textbf{reward function} design, which is crucial for guiding RL agents but is difficult to define accurately, especially in human-centric domains like education \citep{kaelbling1996reinforcement}. Manually designing reward functions is labor-intensive, prone to expertise blind spots, and often results in misspecified objectives that misalign with intended policies \citep{goyal2019using,abbeel2004apprenticeship,amodei2016concrete}.

Unlike RL and DRL, \textbf{Apprenticeship Learning (AL)} is a machine learning approach that derives decision-making policies from a limited set of \emph{expert demonstrations} by inferring the \textbf{expert's underlying reward functions}, enabling the system to generalize and replicate optimal behavior \citep{abbeel2004apprenticeship}. It assumes expert agents make \emph{optimal} or \emph{near-optimal} decisions based on an underlying reward function, allowing the apprentice to efficiently approximate expert behavior with fewer training samples \citep{abbeel2004apprenticeship,ziebart2008maximum}.  Existing AL approaches are typically online, requiring iterative interactions with the environment to collect new data and update the model \citep{abbeel2004apprenticeship,ziebart2008maximum,ho2016generative,finn2016guided}. However, executing a potentially flawed policy can be costly or even dangerous in human-centric tasks such as education \citep{levine2020offline}, making offline policy induction from expert demonstrated highly desirable \citep{asoh2013application,pan2019dissecting,rafferty2015inferring,wang2020inferring,islam2024generalized}. In this work,  \textbf{our ``expert'' demonstrations} were selected by allowing students to make pedagogical decisions, and then choosing only those in which the student greatly benefited from the learning environment.  To accurately model students' complex and dynamic decision-making processes, it is crucial to incorporate \emph{evolving reward functions} that reflect students' changing learning needs and adaptive strategies across different learning phases. Previous offline AL methods typically assume a single or fixed reward function across trajectories, failing to capture the evolving goals of students. For example, while energy-based distribution matching (EDM) \citep{jarrett2020strictly} has advanced offline AL, it assumes a homogeneous policy with a single reward function \citep{babes2011apprenticeship}. Approaches like EM-IRL \citep{yang2020student} address reward function heterogeneity by clustering students based on pedagogical behavior but struggle with large continuous state spaces. More recently, EM-EDM \citep{islam2024generalized} extends EDM with an Expectation-Maximization(EM) framework, enabling policy induction from expert demonstrations driven by multiple heterogeneous reward functions while generalizing to continuous state spaces. 

To directly induce effective pedagogical policies from expert students' interactions with an ITS,  we leverage \textbf{THEMES} \citep{xi2023themes}, a \textbf{T}ime-aware \textbf{H}ierarchical \textbf{EM} \textbf{E}nergy-based \textbf{S}ubtrajectory AL framework which is designed to model \emph{multiple} reward functions \emph{evolving} over time in an \emph{offline} manner. THEMES  consists of two key components: \textbf{(1) Sub-trajectory partitioning}, achieved by partitioning expert trajectories into fine-grained sub-trajectories while incorporating time-awareness. A hierarchically learned reward regulator captures evolving decision-making patterns. \textbf{(2) Policy induction}, performed through offline EM-EDM, which simultaneously clusters sub-trajectories and induces policies specific to each cluster. We evaluate THEMES against six competitive AL baselines and two ablations on a challenging task—predicting students' pedagogical decisions in an ITS by training on past semester trajectories and testing on a later semester. Results show that THEMES consistently outperforms all baselines and ablations across performance metrics. Our findings highlight its potential as a powerful alternative for capturing evolving student pedagogical strategies and inducing effective policies, even with as few as 18 demonstrations.

\section{Related Work}

\subsection{RL for Pedagogical Policy Induction}
Prior research has demonstrated the great potential of leveraging RL and DRL for pedagogical policy induction in ITSs \citep{ju2020aiedcritical,mandel2014offline,rowe2015improving,wang2017interactive}. Shen et al. \citep{shen2016reinforcement} applied a value iteration algorithm to induce policies aimed at improving student performance, showing advantages over random decision-making. Wang et al. \citep{wang2017interactive} utilized DRL to optimize students' learning in an educational game, outperforming traditional RL approaches in simulations. More recently, Zhou et al. \citep{zhou2019hierarchical} employed offline Hierarchical RL to induce pedagogical policies, achieving superior results compared to flat-RL baselines in classroom studies. However, these approaches require large-scale training data—often exceeding 1,000 trajectories \citep{shen2016reinforcement,ju2020aiedcritical,zhou2019hierarchical}—which is typically collected by constraining ITSs to make randomized yet reasonable decisions. Moreover, they rely on predefined reward functions, such as normalized learning gains \citep{shen2016reinforcement,zhou2019hierarchical}, manually designed by researchers \citep{goyal2019using}, limiting adaptability to diverse student learning processes.

\subsection{Offline Apprenticeship Learning (AL)} The most straightforward approach to \emph{offline} AL is behavior cloning (BC) \citep{gleave2022imitation}, which directly maps states to actions to replicate demonstrated behaviors \citep{pomerleau1991efficient}. More robust alternatives, such as inverse reinforcement learning (IRL) \citep{klein2011batch,jain2019model,lee2019truly,chan2021scalable} and adversarial imitation learning \citep{kostrikov2018discriminator,kostrikov2019imitation}, infer an explicit or implicit reward function to generalize beyond observed behaviors. Standard IRL involves iterative online computations, including reward function inference, policy induction via RL, policy rollouts, and reward updates based on behavior divergence. To eliminate policy rollouts, batch-IRL methods \citep{raghu2017continuous} have been introduced, but they often rely on off-policy evaluation, leading to suboptimal solutions. Similarly, adversarial imitation learning iteratively trains a generator to roll out policies and a discriminator to distinguish learned behaviors from expert demonstrations. Off-policy variants based on actor-critic algorithms \citep{kostrikov2018discriminator,kostrikov2019imitation} mitigate the need for policy rollouts but inherit the instability of adversarial training \citep{ho2016generative}. To address these challenges, EDM \citep{jarrett2020strictly} was introduced, offering a more stable solution for offline settings and outperforming both IRL and adversarial imitation learning. In this work, \emph{EDM serves as one of our baselines}. All the aforementioned methods assume a \textit{single} reward function across all demonstrations.

To handle scenarios where \textit{multiple} reward functions exist across trajectories but \emph{remain fixed within each trajectory}, several methods have been developed. Dimitrakakis and Rothkopf~\citep{dimitrakakis2011bayesian} introduced Bayesian multi-task IRL, which learns a separate reward function for each trajectory while sharing a common prior distribution. Choi et al.\citep{choi2012nonparametric} extended this idea using nonparametric Bayesian IRL, modeling the distribution of different rewards with a Dirichlet process to allow for more flexible clustering. Babes-Vroman et al.\citep{babes2011apprenticeship} proposed an EM-based framework that iteratively assigns probabilities to demonstrations belonging to different clusters and updates the corresponding cluster-wise reward functions using Maximum Likelihood IRL.

To model reward functions that \emph{evolve} over time, Krishnan et al. \citep{krishnan2016hirl} introduced Hierarchical IRL (HIRL), and Hausman et al. \citep{hausman2017multi} proposed Multi-modal Imitation Learning. More recently, Wang et al.~\citep{wang2020multi,wang2022hierarchical} developed a hierarchical imitation learning model that learns both high-level policies and sub-policies for individual sub-tasks. These methods, often relying on generative adversarial imitation learning \citep{ho2016generative}, require online interactions or off-policy evaluations, making them less suited for fully offline settings. Additionally, meta-imitation learning \citep{finn2017one} has been explored to generalize from a few demonstrations, typically by structuring tasks into subtasks predefined by domain experts. However, manually defining subtasks is challenging in complex, uncertain environments like education. To address this, THEMES framework \citep{yang2023dpm} \emph{automatically segments trajectories into sub-trajectories}, capturing the evolving nature of reward functions in a data-driven manner.
 
\subsection{AL for ITSs}
Rafferty et al. applied IRL to infer learners' beliefs in an educational game, demonstrating that IRL could effectively recover participants' understanding of how their actions influence the environment \citep{rafferty2015inferring}. This highlighted IRL’s potential for interpreting data in interactive learning settings. In subsequent work, they used IRL to assess learners’ mastery of algebraic skills, identifying misconceptions and providing personalized feedback for improvement \citep{rafferty2016using}. Yang et al. introduced EM-Inverse Reinforcement Learning (EM-IRL) to cluster students based on pedagogical behavior \citep{yang2020student}. However, like many multi-intent approaches, it was constrained to discrete state representations (e.g., 17 discrete features in \citep{yang2020student}). More recently, Islam et al. developed EM-EDM framework \citep{islam2024generalized} to induce effective policies from expert demonstrations with heterogeneous reward functions while generalizing to continuous state spaces. They demonstrated that EM-EDM surpassed two DRL and four AL baselines in a challenging student modeling task for predicting pedagogical decisions. Thus, we adopt EM-EDM as a baseline to assess the effectiveness of THEMES.

\section{Methodology}
\begin{figure}
    \centering
    \includegraphics[scale=0.6,alt={Overview diagram of the THEMES framework. Expert student trajectories feed into a sub-trajectory partitioning module (RMT-TICC) that produces high-level states. These feed into the EM-EDM policy induction module that produces high-level actions (cluster-specific policies). High-level state-action pairs then train an ML-IRL reward regulator, which feeds back to refine the sub-trajectory partitioning. The loop iterates until convergence.}]{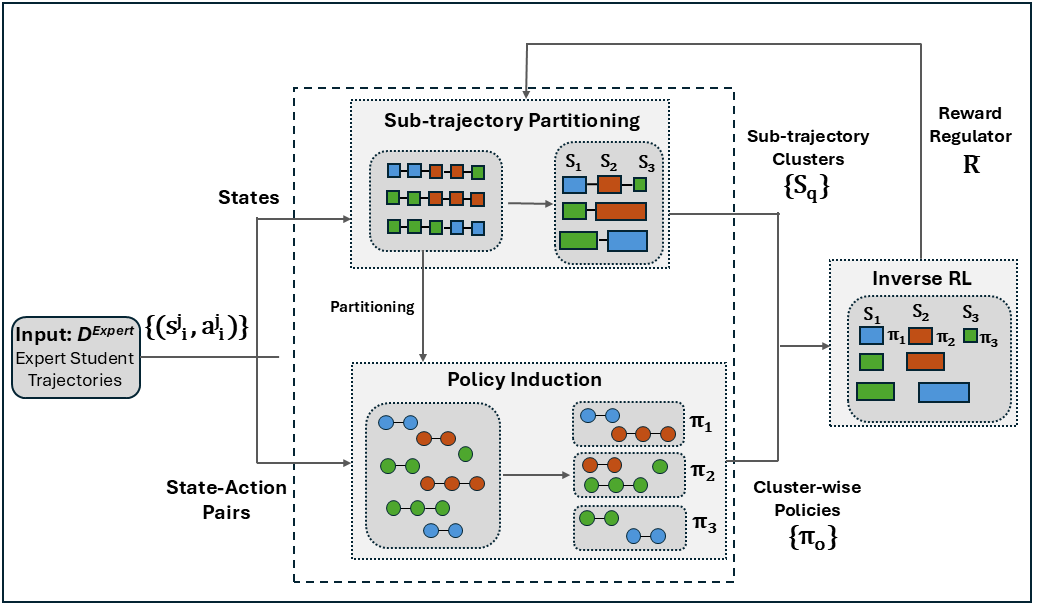}
    \caption{Overview of the THEMES Framework.}
\label{fig:overview-ch4}
\vspace{-0.1in}
\end{figure}
\vspace{-0.10in}


The input dataset $\mathcal{D}^{Total}$ consists of $L$ student-ITS interaction trajectories, each having a fixed length of $n$. Each trajectory $d$ can be viewed as: $s_{1} \xrightarrow{a_{1}} s_{2} \xrightarrow{a_{2}} \cdots s_{n-1} \xrightarrow{a_{n-1}} s_{n}$. Here $s_{i} \xrightarrow{a_{i}} s_{i+1}$ designates that at the $i_{th}$ time step in $d$, the student's learning status was in state $s_{i}$, the student executed the action $a_{i}$ and their learning status transitioned to state $s_{i+1}$. Since in AL, it is typically assumed that the experts' trajectories provided as input are optimal or near-optimal according to latent reward functions \citep{abbeel2004apprenticeship}, the quality of these trajectories is crucial for inducing more accurate policies. Therefore, we have designed a procedure outlined in Section \ref{TrajSelection-ch4} to select the expert student trajectories, based on which a subset of $N$ trajectories $\mathcal{D}^{Expert}$ is selected from $\mathcal{D}^{Total}$, denoted as: $\{\mathcal{D}^{j}\} = \{(\mathbf{s}_i^j, a_{i}^{j})|i=1, ..., n; j = 1,..., N\}$ where $N \le L$.

\subsection{THEMES for Evolving Student Pedagogical Strategy}
\label{subsec:themes}

Given $\mathcal{D}^{Expert}$ as input, THEMES seeks to learn the multiple underlying reward functions of students that evolve over time leveraging its two key components, \textit{i.e.}, sub-trajectory partitioning and policy induction (Figure \ref{fig:overview-ch4}).
Primarily, a time-aware sub-trajectory partitioning method \citep{yang2021multi} processes the states from input ($\mathcal{D}^{Expert}$). It partitions and clusters the states $\{\mathbf{s}_i^j\}$ into sub-trajectories such that each resulting cluster captures consistent time-invariant patterns, which can be interpreted as high-level states $\{S_{q} | q \in [1,Q]\}$ where Q refers to the number of sub-trajectory clusters. Then, focusing on the state-action pairs within the partitioned sub-trajectories, an Expectation-Maximization(EM) Energy-based Distribution Matching (EDM) method \citep{jarrett2020strictly} clusters and induces policies from the state-action pairs $\{(\mathbf{s}_i^j, a_{i}^{j})\}$ over the partitioned sub-trajectories, ensuring that each cluster exhibits consistent decision-making patterns, referred to as high-level actions $\{\pi_{o} | o \in [1,O]\}$. Using these high-level state-action pairs, a Maximum Likelihood Inverse Reinforcement Learning (ML-IRL) method \citep{babes2011apprenticeship} learns a high-level reward regulator $\overline{R}(\cdot)$, which is then fed back to refine sub-trajectory partitioning. This iterative procedure continues until convergence.

\vspace{-0.1in}
\subsection{Sub-trajectory Partitioning}
\label{sec:rmt-ticc}
\vspace{-0.1in}

As shown in Figure \ref{fig:overview-ch4}, this module first extracts the states $\{\mathbf{s}_i^j\}$ from the given expert student trajectories $\{(\mathbf{s}_i^j, a_i^j)\}$. 
Since the states in student trajectories are sequential and influenced by adjacent states, it is desirable to leverage the temporal dependencies rather than treating states independently. Hence, the \emph{Subtrajectory-partitioning} module of THEMES captures this temporal dependency by investigating patterns within a sliding window of size $\omega \ll n$. For each state $\mathbf{s}_i^j$, the sub-trajectory within the window, $\mathbf{S}_i^j = \{\mathbf{s}_{i-\omega+1}^{j}, \ldots, \mathbf{s}_i^j\}$, is used to determine the cluster assignments $\{S_{q} | q \in [1,Q]\}$. Given $\mathbf{s}_i^j \in \mathbb{R}^m$, each sub-trajectory with the sliding window $\mathbf{S}_i^j$ can be viewed as a random variable in a $m\omega$-dimensional space, which can be generated by concatenating the $\omega$ consecutive states. 

Each sub-trajectory cluster is represented by its mean vector ($\boldsymbol{\mu}$) and inverse covariance matrix ($\boldsymbol{\Theta}$) \citep{hallac2017toeplitz}. Specifically, the process of determining the optimal mean vectors, $\{\boldsymbol{\mu}_{q} \mid q \in [1,Q]\}$, is equivalent to assigning each state to the most suitable cluster. In Intelligent Tutoring Systems (ITSs), student trajectories are often collected at irregular intervals. Consequently, both the mean vectors and inverse covariance matrices are learned across various trajectories, while \textit{time-awareness} is incorporated during the sub-trajectory partitioning phase using a decay function that allows for a re-evaluation of cluster consistency over time \citep{yang2021multi}. To enhance the sub-trajectory clustering by accounting for decision-making behaviors, \emph{hierarchical} rewards $\{\overline{r}_{t}^{j} \mid t \in [1,n]\}$ are inferred for each state-action pair, which then inform the adjustment of the consistency constraint. As illustrated in Figure \ref{fig:overview-ch4}, these high-level state-action pairs facilitate the inference of high-level rewards, $\overline{R}(\cdot)$, which can be learned using Maximum Likelihood Inverse Reinforcement Learning (ML-IRL) \citep{babes2011apprenticeship}.

\vspace{-0.2in}
\subsection{Policy Induction}
In this segment, EDM \citep{jarrett2020strictly}, a state-of-the-art strictly offline AL method, serves as the core component, which assumes a single reward function while learning the policy $\pi^\theta$, parameterized by $\theta$. To induce the policy $\pi^\theta$, EDM aim to learn the parameter $\theta$ by minimizing the KL divergence between $\rho_{D}$ and $\rho_{\pi^\theta}$, $D_{KL}(\rho_D||\rho_{\pi^\theta})$, where $\rho_{D}$ and $\rho_{\pi^\theta}$ represent the occupancy measures for the expert trajectories and for the induced policy $\pi^\theta$, respectively.
Despite its success in \emph{offline} learning, EDM assumes that all demonstrations adhere to a \emph{single} reward function, which is often impractical for human-centric tasks in the real world, such as education, where students may have multiple reward functions evolving over time. To address the challenge of \emph{multiple} reward functions \emph{varying across} demonstrations, an Expectation-Maximization(EM)-EDM \citep{islam2024generalized} was introduced, iteratively clustering demonstrations in the \textit{E-step} and inducing the policy for each cluster by EDM in the \textit{M-step}. The input to EM-EDM will be the sub-trajectories $\{\mathcal{D}^{\hat{j}} | \hat{j}=1,...,\hat{N}\}$, where $\hat{N}$ is the total number of sub-trajectories learned from \emph{Sub-trajectory partitioning}. Finally, THEMES yields sub-trajectory clusters with respective policies $\{\pi_{o} | o \in [1,O]\}$ as output, representing multiple student pedagogical strategies evolving over time.

\vspace{-0.1in}
\subsection{Baselines}
\label{subsec:baselines}
\subsection{Behavior Cloning (BC)}
Behavior cloning \citep{syed2012imitation,raza2012teaching} is a supervised learning approach used in AL where an agent learns to imitate an expert by directly mapping observations to actions based on the expert's demonstrations. The objective is to mimic the expert's behavior, hence, learn a policy $\pi_{\theta}(a|s)$ parameterized by $\theta$ that approximates the expert behavior: $L(\theta) = \frac{1}{n} \sum_{i=1}^{n} \lVert \pi_{\theta}(a_i|s_i) - a_i \rVert^2$.
It is important to note that while BC is straightforward, it can be sensitive to distribution shifts and compounding errors, especially when the learned policy deviates from the expert demonstrations.
\vspace{-0.2in}
\subsection{Gaussian Process (GP) + Deep Q-Network (DQN)}
To infer immediate rewards from the final delayed reward of each trajectory, this method \citep{azizsoltani2018adaptive} first employs GP to model a distribution function $f$ that captures the expected values and standard deviations of the unknown intermediate rewards. The inference is formulated as a minimum mean square error (MMSE) estimation problem, where the additive Gaussian rewards are estimated such that their sum approximates the observed delayed return \citep{guo2005mutual}. This approach assumes that rewards at each timestep follow a Gaussian distribution and that the total reward across the trajectory equals the observed delayed outcome. Once immediate rewards are inferred, a Deep Q-Network (DQN) is trained on the reconstructed immediate reward-labeled trajectories. 

This enables the agent to learn a policy that maximizes the expected cumulative reward over the reconstructed trajectories.
\vspace{-0.2in}
\subsection{Multi-modal Imitation Learning (MIL)}
MIL \citep{hausman2017multi} tackles imitation learning from unstructured and unlabeled demonstrations by learning a multi-modal policy. Instead of assuming demonstrations are pre-segmented by task, MIL jointly segments skills and learns their corresponding policies. The method augments the standard policy $\pi(a|s)$ with a latent intention variable $i \sim p(i)$, resulting in a conditional policy $\pi(a|s, i)$. The intention variable $i$ selects a specific mode of the policy corresponding to a distinct skill. MIL optimizes a generative adversarial imitation learning objective, extended with an additional latent intention loss, encouraging the policy to maximize the mutual information between the latent intention and the generated behaviors.
Through this framework, MIL simultaneously discovers the distinct skills present in demonstrations and learns to imitate them using a single, unified policy. 
\vspace{-0.2in}
\subsection{Adapted Hierarchical Inverse Reinforcement Learning (AHIRL)}
HIRL \citep{krishnan2016hirl} addresses the challenge of long-horizon tasks with delayed rewards by segmenting demonstrations into a sequence of locally linear sub-tasks. Each sub-task is defined by consistent changes in local linearity, detected using a Bayesian nonparametric Gaussian Mixture Model (DP-GMM) over a featurized trajectory space.
To ensure sequential execution, HIRL augments the original state-space $\mathcal{S}$ with a binary vector $v \in \{0,1\}^Q$ that tracks progress across $Q$ sub-goals. 
A reward function $R_\theta(s, a, v)$ is then learned using Maximum Entropy IRL over this augmented space. This enables the agent to reason about both current state and sub-task completion history. 

\section{Experiments}

\begin{figure}[htbp]
    \centering
    \includegraphics[scale=0.2,alt={Screenshot of the Pyrenees probability ITS interface showing a multi-step training problem with step-by-step input fields, on-demand hint buttons, and three presentation mode buttons labeled PS (Problem Solving), WE (Worked Example), and FWE (Faded Worked Example).}]{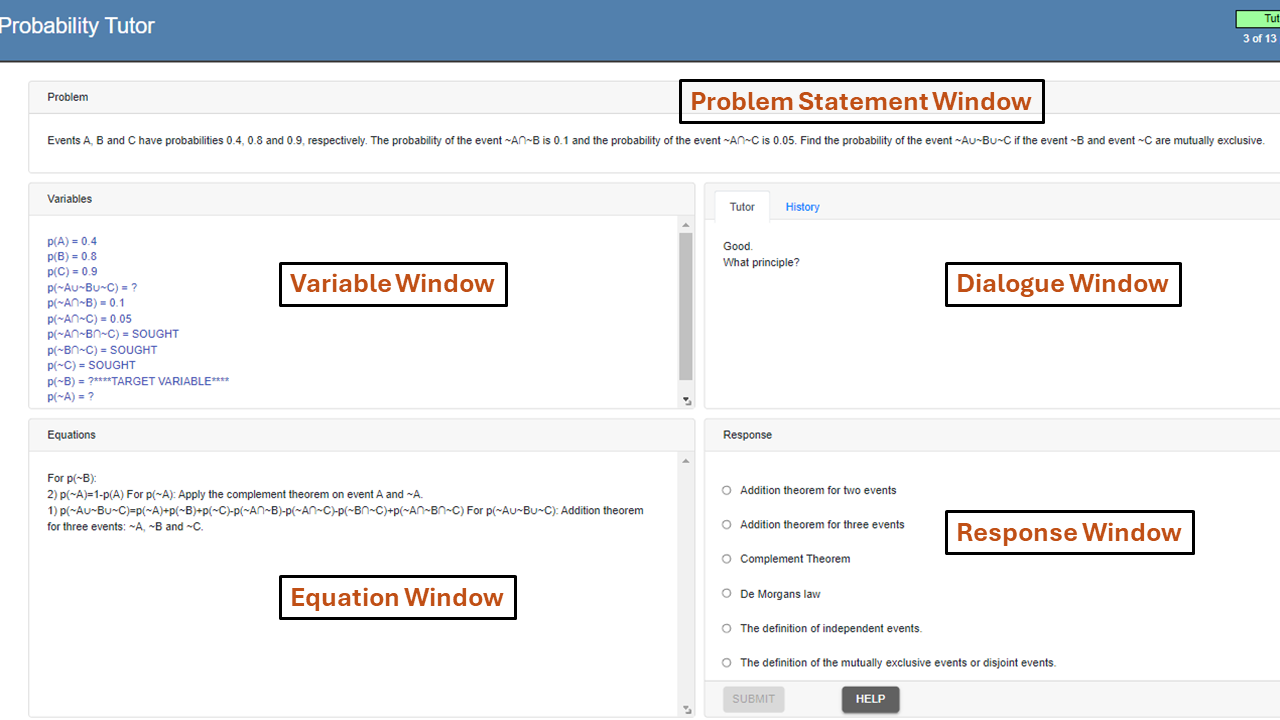}
    \caption{ITS Interface.}
    \label{fig:pyrenees-interface-ch4}
\end{figure}

This study utilized a Probability ITS (Figure \ref{fig:pyrenees-interface-ch4}) deployed in the Spring and Fall semesters of a large undergraduate STEM course at a public university. Designed by domain experts and overseen by departmental committees, the system aims to teach entry-level undergraduates ten key probability principles, \textit{e.g.}, complement theorem and Bayes' rule, through solving a series of complex problems, each of which requires 30 to 60 interactions.

\vspace{-0.15in}
\subsection{Data Collection}
Data were collected from 221 students over four semesters: 67 in Spring 2021 (S21), 56 in Spring 2022 (S22), 54 in Spring 2024 (S24), and 44 in Fall 2024 (F24). All students followed a standardized process, including the same textbook, pre-test, training, and post-test within our ITS. During the textbook, all students studied the 10 probability principle, read a general explanation, and worked through various single- and multi-principle problems. The pre-test and post-test comprised 8 and 12 open-ended complex problem-solving questions, respectively. 

During ITS training, all students were presented with the same 12 training problems in a fixed sequence. Each problem required applying 3 to 11 domain principles for a solution. For two of these problems, all students were required to solve them independently to familiarize themselves with the ITS interface. For the remaining \textbf{10} problems, \textbf{students would decide among three approaches}: \emph{1) solving the problem independently}, \emph{2) collaborating with the ITS}, or \emph{3) viewing a worked-out example}.  Students' pre-test and post-test answers were graded independently by two evaluators using a double-blind process, with discrepancies resolved through discussion to reach a final grade. All test scores were normalized to the range of [0 - 100] for comparison purposes. Student data were obtained anonymously through an exempt IRB-approved protocol. 

\vspace{-0.1in}
\subsection{130 State Feature Representation}
Given that students' underlying learning status are unobservable \citep{mandel2014offline}, two domain experts define the \textit{\textbf{State}} space through \textbf{130}-dimensional continuous features including: \textbf{\textit{10} Autonomy} features that measure the student's work, such as the number of elicits since the last tell; \textbf{\textit{22} Temporal} features that represent time-related information, like average time per step; \textbf{\textit{31} Problem Solving} features that reflect contextual information about problem-solving, including difficulty level; \textbf{\textit{57} Performance} features that indicate student's performance so far, such as the percentage of correct responses; \textbf{\textit{10} Hints} features that show data on hint usage, such as total hints requested.

\vspace{-0.1in}
\subsection{Expert Student Demonstrations}
\label{TrajSelection-ch4}
\begin{figure}[htbp]
    \centering
    \includegraphics[scale=0.2,alt={Flowchart showing the Quantized Learning Gain (QLG) classification. Students are grouped into Low, Medium, and High based on pre-test and post-test scores. Solid arrows represent High QLG (student maintained or moved to a higher group); dashed arrows represent Low QLG (student stayed in Low or Medium, or dropped).}]{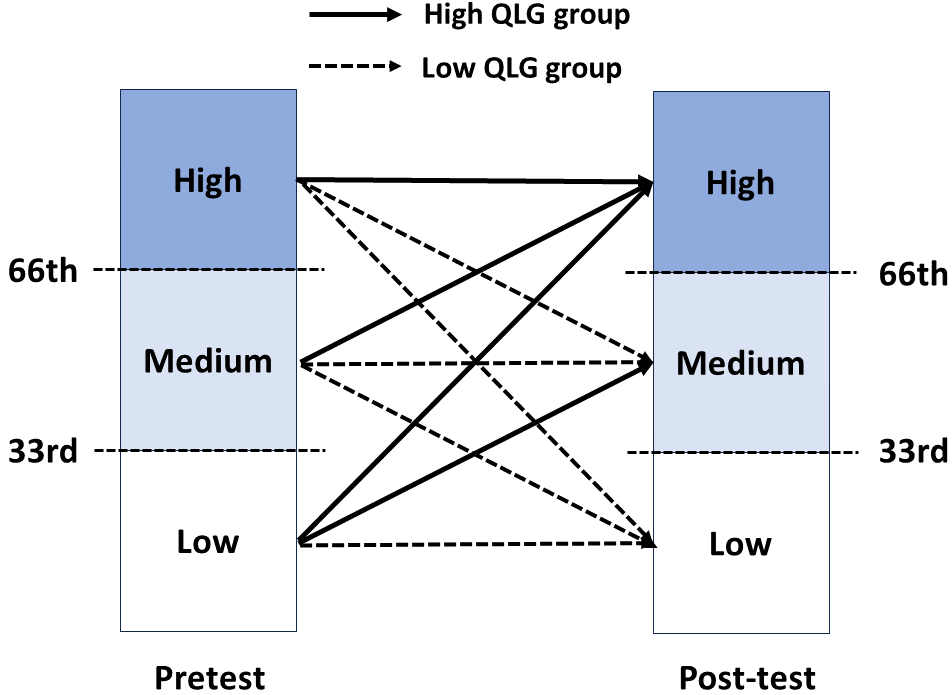}
    \caption{Quantized Learning Gain.}
    \label{fig:qlg-ch4}
\end{figure}

To select demonstrations with optimal or near-optimal student interactions with the ITS, we used the Quantized Learning Gain (QLG) measure \citep{mao2018tracing}. Students were categorized into low, medium, or high-performance groups based on their pre- and post-test scores (as shown in Figure \ref{fig:qlg-ch4}). A ``High'' QLG is assigned to students who improved or remained in the high-performance group, while a ``Low'' QLG is given to those who dropped to a lower group or stayed in low/medium groups. This helped identify students who benefited most from our ITS. A total of 89 student trajectories from 4 semesters were used as expert demonstrations: 18 from S21, 24 from S22, 23 from S24, and 24 from F24.

\vspace{-0.1in}
\subsection{Experimental Settings}
We select six competitive AL baselines for comparison against THEMES, as discussed in Section \ref{subsec:themes} and \ref{subsec:baselines}. The settings for the baselines are summarized as follows:\vspace{0.08in}

\noindent \textbf{1) BC} \citep{gleave2022imitation,syed2012imitation}: We use the BC implementation available in the \emph{imitation} \citep{gleave2022imitation} library and extend it to a customized environment corresponding to our ITS setting.

\noindent \textbf{2) GP+DQN} \citep{azizsoltani2018adaptive}: For this method, we employ Gaussian Process (GP) to infer immediate rewards first from the delayed final outcomes, e.g., learning gain of students for ITS setting, followed by a standard Deep Q-Network (DQN) implementation \citep{fan2020theoretical} for policy learning.

\noindent \textbf{3) EDM} \citep{jarrett2020strictly}: Since EDM is strictly offline, we minimally adapt the original implementation to fit the ITS setting.

\noindent \textbf{4) Multi-modal Imitation Learning (MIL)} \citep{hausman2017multi}: MIL leverages InfoGAN to jointly learn sub-trajectories and policies. Although originally designed for online settings, we adapt it for offline learning.

\noindent \textbf{5) Adapted Hierarchical Inverse Reinforcement Learning (AHIRL)} \citep{krishnan2016hirl}: In our offline adaptation of HIRL, we reuse the segment clustering and local reward structure but employ EDM-based policy learning rather than online RL, enabling HIRL to operate effectively in a strictly offline setting.

\noindent \textbf{6) EM-EDM} \citep{islam2024generalized}: We extend the implementation of EDM to EM-EDM by incorporating the EM algorithm and maximizing the log-likelihood of cluster assignments as described in \citep{islam2024generalized}.

In addition to the six AL baselines, we evaluate the performance of THEMES against its two ablations. The first one is: \textbf{THEMES$_0$}, where the high-level states are obtained through sub-trajectory partitioning \citep{yang2021multi}, but the high-level actions are induced by EDM \citep{jarrett2020strictly}. The second one is: \textbf{THEMES$_1$}, where the high-level states are also learned by sub-trajectory partitioning, but high-level actions are induced by EM-EDM \citep{babes2011apprenticeship}. Finally, in \textbf{THEMES}, we utilize reward regulated sub-trajectory clustering to partition trajectories with hierarchically learned reward regulator and induce policies over the partitioned sub-trajectories via EM-EDM. For THEMES, the hyperparameters in sub-trajectory partitioning are determined based on the Bayesian Information Criteria (BIC) \citep{friedman2001elements}, with the number of sub-trajectory clusters $Q$ set to 6, window size $\omega$ set to 2, and the coefficients for sparsity and consistency terms set to 1e-3 and 4, respectively. For EM-EDM, the optimal number of clusters $O$ was determined heuristically as 3, by iteratively implementing the EM algorithm until either empty clusters are generated or the log-likelihood of the clustering results varies by less than a predefined threshold. The iterations for the overall THEMES framework are set as 10, since the clustering likelihood for both sub-trajectory clustering and EM-EDM converge within 10 iterations. 

\vspace{-0.2in}
\subsection{Experiment Setup and Evaluation Metrics}
It is important to emphasize that all models are evaluated using semester-based temporal cross-validation (3-fold) in this task, which only applied data from previous semesters individually or combined for training and is a much stricter approach than the standard cross-validation \citep{mao2020time}. We report Accuracy, Recall, Precision, F1 Score, AUC (Area Under the ROC Curve), APR (Area Under the Precision-Recall Curve), and Jaccard score. The Jaccard score measures the similarity between the predicted and true sets of labels, calculated as the size of the intersection divided by the size of the union of the two sets. Given the nature of our task, we consider AUC and Jaccard score to be the most critical metrics, as they are generally robust.

\section{Results}
\label{sec:results-ch4}

\renewcommand{\arraystretch}{1.3} 
\setlength{\tabcolsep}{0.5 pt} 
\begin{table*}
\caption{\centering Comparing THEMES with baselines and ablations.}
\label{tab:result_xsem}
\small
    \centering
    \begin{tabular}{c|ccccccc} \hline
        \thead{Methods} & \thead{Acc} & \thead{Rec} & \thead{Prec} & \thead{F1} & \thead{AUC} & \thead{AP} & \thead{JAC} \\ \hline
        1. BC \tiny{\citep{gleave2022imitation}} & $.366_{\pm .03}$ & $.366_{\pm .03}$ & $.384_{\pm .04}$ & $.363_{\pm .03}$ & $.523_{\pm .03}$ & $.362_{\pm .01}$ & $.224_{\pm .02}$ \\ 
        2.GP+DQN \tiny{\citep{azizsoltani2018adaptive}} & $.389_{\pm .06}$ & $.389_{\pm .06}$ & $.399_{\pm .10}$ & $.297_{\pm .05}$ & $.525_{\pm .06}$ & $.386_{\pm .03}$ & $.193_{\pm .04}$ \\
        3. EDM \tiny{\citep{jarrett2020strictly}}& $.656_{\pm .05}$ & $.656_{\pm .05}$ & $.686_{\pm .03}$ & $.658_{\pm .05}$ & $.801_{\pm .03}$ & $.662_{\pm .04}$ & $.506_{\pm .05}$\\  
        4. MIL \tiny{\citep{hausman2017multi} } & $.508_{\pm .06}$ & $.508_{\pm .06}$ & $.486_{\pm .11}$ & $.456_{\pm .06}$ & $.627_{\pm .06}$ & $.465_{\pm .06}$ & $.319_{\pm .06}$\\
        5. AHIRL \tiny{\citep{krishnan2016hirl}}& $.545_{\pm .07}$ & $.545_{\pm .07}$ & $.566_{\pm .07}$ & $.545_{\pm .07}$ & $.688_{\pm .07}$ & $.537_{\pm .07}$ & $.388_{\pm .07}$\\ 
        6. EM-EDM & $\mathbf{.705_{\pm .06}}$ & $\mathbf{.705_{\pm .06}}$ & $\mathbf{.725_{\pm .04}}$ & $\mathbf{.709_{\pm .05}}$ & $\mathbf{.847_{\pm .03}}$ & $\mathbf{.724_{\pm .05}}$ & $\mathbf{.561_{\pm .07}}$\\ \hline
        7. THEMES$_0$ & $.638_{\pm .05}$ & $.638_{\pm .05}$ & $.671_{\pm .06}$ & $.642_{\pm .06}$ & $.783_{\pm .05}$ & $.643_{\pm .07}$ & $.487_{\pm .06}$ \\ 
        8. THEMES$_1$ & $.745_{\pm .04}$ & $.745_{\pm .04}$ & $.761_{\pm .03}$ & $.747_{\pm .03}$ & $.875_{\pm .02}$ & $.768_{\pm .03}$ & $.605_{\pm .04}$\\
        9. THEMES & $\mathbf{.763_{\pm .02}}^{*}$ & $\mathbf{.763_{\pm .02}}^{*}$ & $\mathbf{.773_{\pm .02}}^{*}$ & $\mathbf{.765_{\pm .02}}^{*}$ & $\mathbf{.879_{\pm .02}}^{*}$ & $\mathbf{.773_{\pm .04}}^{*}$ & $\mathbf{.626_{\pm .03}}^{*}$\\ \hline 
    \end{tabular}
    \vspace{-0.2in}
\end{table*}

\begin{figure}
  \centering
  \begin{subfigure}{0.8\textwidth}
    \centering
    \includegraphics[width=\linewidth,alt={Critical difference diagram for AUC based on Conover-Friedman test at p=0.05 across three semester-based folds. THEMES ranks first, followed by THEMES-1, EM-EDM, THEMES-0, EDM, AHIRL, MIL, GP-DQN, and BC. Methods connected by a horizontal bar are not significantly different from each other.}]{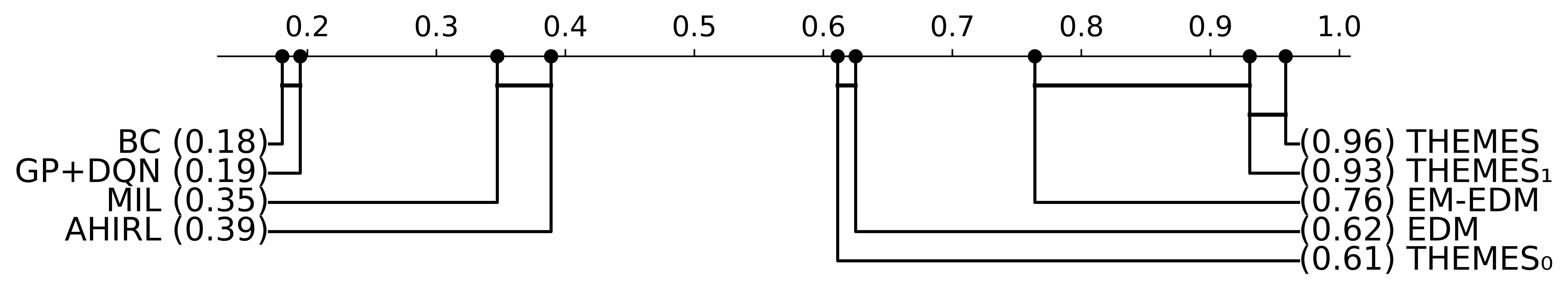}
    \caption{AUC.}
    \label{subfig:cd_auc-ch4}
  \end{subfigure}
  \hfill
  \begin{subfigure}{0.8\textwidth}
    \centering
    \includegraphics[width=\linewidth,alt={Critical difference diagram for Jaccard score based on Conover-Friedman test at p=0.05 across three semester-based folds. THEMES ranks first, followed by THEMES-1, EM-EDM, THEMES-0, EDM, AHIRL, MIL, GP-DQN, and BC. Methods connected by a horizontal bar are not significantly different.}]{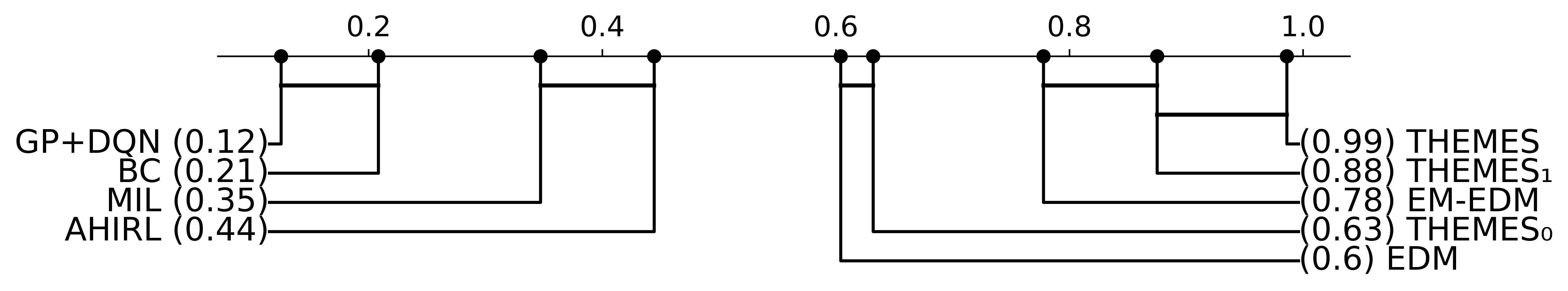}
    \caption{Jaccard Score.}
    \label{subfig:cd_jac-ch4}
  \end{subfigure}
  \caption{Critical difference diagram for AUC and Jaccard Score.}
  \label{fig:cd_plot-ch4}
  \vspace{-0.2in}
\end{figure}

Table \ref{tab:result_xsem} reports the average and standard deviation of the overall semester-based temporal cross-validation results for the different models, with the best baselines and ablations highlighted in bold, and the overall best results are highlighted with *. The critical difference diagram, based on the Conover-Friedman tests ($p = 0.05$), for AUC and Jaccard, is presented in Figure \ref{fig:cd_plot-ch4}. The results clearly show that THEMES \emph{significantly} outperform all other methods in all evaluation metrics. Specifically, THEMES outperforms THEMES$_1$, demonstrating the effectiveness of hierarchically learned \emph{reward regulator} in incorporating decision-making patterns during sub-trajectory partitioning. Furthermore, the three THEMES-based methods consistently and significantly outperform competitive baselines such as AHIRL and MIL, which share the assumption of \emph{multiple} reward functions \emph{evolving} over time. This suggests that the effectiveness of high-level state representation is enhanced by incorporating \emph{time-awareness} during sub-trajectory partitioning. When comparing THEMES and THEMES$_{0}$ with EM-EDM, significant improvements are observed, highlighting the power of high-level state representation to capture time-varying patterns across partitioned sub-trajectories. Although EM-EDM is better than EDM, which supports the assumption of \emph{multiple} reward functions, the superior performance of THEMES over EM-EDM suggests that the assumption was insufficient to model time-varying decision-making. Moreover, THEMES demonstrates exceptional consistency between semesters, as evidenced by the minimal standard deviation in Table \ref{tab:result_xsem}. Notably, among all individual fold results of semester-based cross validation, we found that THEMES can achieve an AUC of 0.899 and a Jaccard of 0.653, using only \emph{18} trajectories of S21 to predict student strategies in S22. Finally, all of these results underscore the potential of THEMES as a sample-efficient alternative to capture evolving student pedagogical strategies. 

\vspace{-0.2in}
\subsection{A Case Study}
\begin{figure}
    \centering
    \includegraphics[scale=0.43,alt={t-SNE scatter plot of 130-dimensional expert student states projected to 2D, colored by six sub-trajectory clusters identified by THEMES. One highlighted blue trajectory stays entirely within cluster 1, indicating a single stable learning intention. A second highlighted black trajectory transitions through clusters 2, 6, 1, 2, 6, and 3, illustrating multiple evolving learning intentions over time.}]{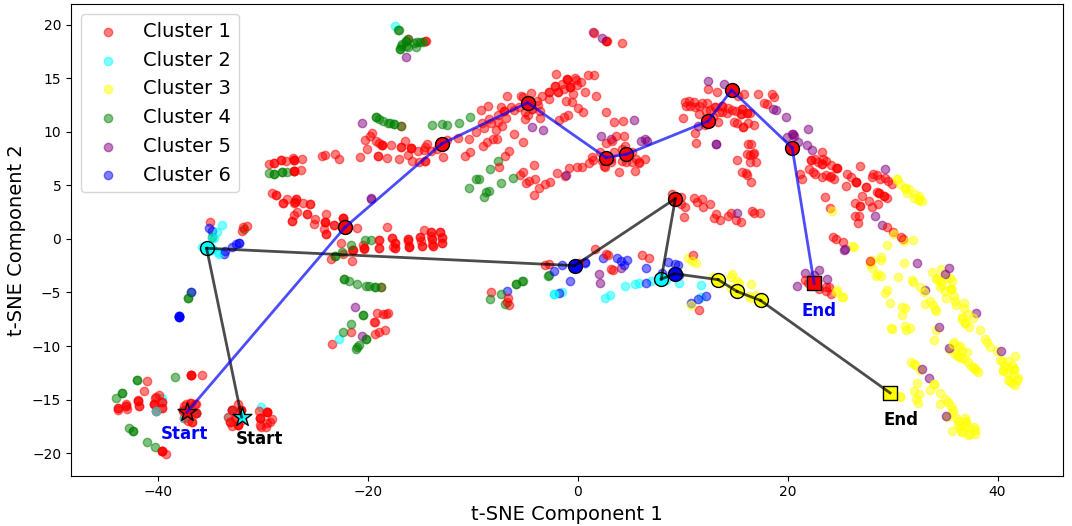}
    \caption{t-SNE Visualization of Expert Student Demonstrations---\textbf{blue} Trajectory (representing a fixed reward function) vs. \textbf{black} Trajectory (indicating evolving reward functions).}
    \label{fig:casestudy}
    \vspace{-0.3in}
\end{figure}

Figure \ref{fig:casestudy} presents a t-SNE visualization of the original 130-state space from all expert student demonstrations. To facilitate better visualization, we applied t-distributed stochastic neighbor embedding (t-SNE) to project the data into a lower-dimensional space \citep{hinton2002tsne}. In this figure, different colors correspond to the six clusters identified by sub-trajectory clusters or high-level states in THEMES framework. The \textbf{blue} and \textbf{black} lines represent the expert student demonstrations from two distinct groups. The \textbf{blue} trajectory remains consistently within sub-trajectory cluster 1, indicating a student likely driven by a single, stable intention throughout their learning process. In contrast, the \textbf{black} trajectory begins in cluster 2 but evolves over time, transitioning through clusters 6, 1, 2, 6, and 3. This suggests that the latter student’s learning is guided by multiple, shifting intentions. AL methods like EDM, which assume a single reward function for each demonstration, can effectively capture the former trajectory but may struggle to account for the dynamic nature of the latter. In contrast, THEMES leverages sub-trajectory partitioning and clustering, enabling it to capture and adapt to these evolving intentions, resulting in a more accurate policy for the latter trajectory.

\renewcommand{\arraystretch}{1.0} 
\setlength{\tabcolsep}{1.2 pt} 
\begin{table*}
\caption{\centering Comparing THEMES with baselines and ablations over three healthcare datasets.}
\label{tab:themes}
\small
    \centering
    \resizebox{\textwidth}{!}{%
    \begin{tabular}{cc|cccc|cccc|cccc} \hline
        \multicolumn{2}{c|}{\multirow{2}{*}{\thead{Methods}}} & \multicolumn{4}{c|}{\thead{MIMIC-III}} & \multicolumn{4}{c|}{\thead{CCHS}} & \multicolumn{4}{c}{\thead{Mayo}} \\ \cline{3-14}
        \multicolumn{2}{c|}{} & \thead{F1} & \thead{AUC} & \thead{AP} & \thead{JAC} & \thead{F1} & \thead{AUC} & \thead{AP} & \thead{JAC} & \thead{F1} & \thead{AUC} & \thead{AP} & \thead{JAC} \\ \hline
        
        \multirow{6}{*}{\rotatebox[origin=c]{90}{Baselines}} 
        & BC 
        & $.497_{\pm .01}$ & $.502_{\pm .00}$ & $.485_{\pm .07}$ & $.331_{\pm .01}$ 
        & $.499_{\pm .01}$ & $.500_{\pm .01}$ & $.233_{\pm .02}$ & $.376_{\pm .01}$ 
        & $.503_{\pm .01}$ & $.504_{\pm .01}$ & $.038_{\pm .01}$ & $.475_{\pm .01}$ \\

        & GP+DQN 
        & $.418_{\pm .04}$ & $.516_{\pm .03}$ & $.515_{\pm .04}$ & $.349_{\pm .05}$ 
        & $.333_{\pm .03}$ & $.511_{\pm .04}$ & $.526_{\pm .03}$ & $.211_{\pm .03}$ 
        & $.326_{\pm .01}$ & $.460_{\pm .02}$ & $.497_{\pm .04}$ & $.269_{\pm .05}$ \\

        & EDM 
        & $.634_{\pm .07}$ & $.635_{\pm .07}$ & $.590_{\pm .05}$ & $.471_{\pm .08}$ 
        & $.631_{\pm .04}$ & $.553_{\pm .03}$ & $.522_{\pm .01}$ & $.479_{\pm .05}$ 
        & $.701_{\pm .10}$ & $.640_{\pm .04}$ & $\mathbf{.515}_{\pm .01}$ & $.558_{\pm .13}$ \\

        & EM-EDM 
        & $\mathbf{.778}_{\pm .05}$ & $.779_{\pm .05}$ & $\mathbf{.717}_{\pm 0.05}$ & $\mathbf{.641}_{\pm .07}$ 
        & $.607_{\pm .02}$ & $\mathbf{.605}_{\pm .04}$ & $.547_{\pm .02}$ & $.444_{\pm .03}$ 
        & $.796_{\pm .09}$ & $.597_{\pm .04}$ & $.511_{\pm .01}$ & $\mathbf{.684}_{\pm .12}$ \\

        & AHIRL 
        & $.710_{\pm .05}$ & $.754_{\pm .04}$ & $.692_{\pm .05}$ & $.550_{\pm .06}$ 
        & $.599_{\pm .04}$ & $.601_{\pm .05}$ & $\textbf{.549}_{\pm .06}$ & $.461_{\pm .05}$ 
        & $.795_{\pm .06}$ & $.432_{\pm .03}$ & $.481_{\pm .01}$ & $.660_{\pm .10}$ \\ 

        & MIL 
        & $.756_{\pm .04}$ & $\mathbf{.791}_{\pm .05}$ & $.704_{\pm .04}$ & $.607_{\pm .06}$ 
        & $\mathbf{.653}_{\pm .05}$ & $.570_{\pm .06}$ & $.532_{\pm .05}$ & $\mathbf{.486}_{\pm .04}$ 
        & $\mathbf{.802}_{\pm .07}$ & $\mathbf{.601}_{\pm .04}$ & $.510_{\pm .01}$ & $.683_{\pm .11}$ \\   \hline
        
        \multirow{3}{*}{\rotatebox[origin=c]{90}{Our}} 
        & THEMES$_0$ 
        & $.635_{\pm .07}$ & $.634_{\pm .07}$ & $.589_{\pm .05}$ & $.473_{\pm .08}$ 
        & $.621_{\pm .04}$ & $.551_{\pm .02}$ & $.521_{\pm .01}$ & $.467_{\pm .04}$ 
        & $.672_{\pm .05}$ & $.633_{\pm .03}$ & $.513_{\pm .01}$ & $.516_{\pm .06}$ \\ 

        & THEMES$_1$ 
        & $.931_{\pm .02}$ & $.929_{\pm .02}$ & $.899_{\pm .02}$ & $.872_{\pm .03}$ 
        & $.653_{\pm .05}$ & $.608_{\pm .07}$ & $.551_{\pm .04}$ & $.502_{\pm .06}$ 
        & $.719_{\pm .07}$ & $.606_{\pm .05}$ & $.510_{\pm .01}$ & $.575_{\pm .08}$ \\

        & THEMES 
        & $\mathbf{.936}^{*}_{\pm .02}$ & $\mathbf{.935}^{*}_{\pm .02}$ & $\mathbf{.908}^{*}_{\pm .03}$ & $\mathbf{.883}^{*}_{\pm .04}$ 
        & $\mathbf{.668}^{*}_{\pm .05}$ & $\mathbf{.614}^{*}_{\pm .07}$ & $\mathbf{.557}^{*}_{\pm .04}$ & $\mathbf{.517}^{*}_{\pm .05}$ 
        & $\mathbf{.838}^{*}_{\pm .08}$ & $\mathbf{.645}^{*}_{\pm .02}$ & $\mathbf{.520}^{*}_{\pm .01}$ & $\mathbf{.744}^{*}_{\pm .12}$ \\ \hline
    \end{tabular}%
    }
\end{table*}

\begin{figure}[!h]
\centering
\begin{subfigure}{.6\textwidth}
  \centering
  \includegraphics[width=1\linewidth,alt={Critical difference diagram for F1 score on the MIMIC-III healthcare dataset. THEMES ranks first with the best mean rank, significantly outperforming all baselines including EM-EDM, MIL, AHIRL, EDM, GP-DQN, and BC at p=0.05.}]{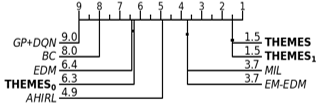}
  \caption{MIMIC-III.}
\end{subfigure}

\begin{subfigure}{.6\textwidth}
  \centering
  \includegraphics[width=1\linewidth,alt={Critical difference diagram for F1 score on the CCHS clinical dataset. THEMES ranks first, significantly outperforming all baselines including EM-EDM, MIL, AHIRL, EDM, GP-DQN, and BC at p=0.05.}]{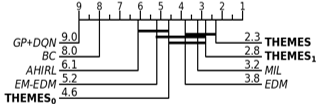}
  \caption{CCHS.}
\end{subfigure}

\begin{subfigure}{.6\textwidth}
  \centering
  \includegraphics[width=1\linewidth,alt={Critical difference diagram for F1 score on the Mayo Clinic dataset. THEMES ranks first, significantly outperforming all baselines including EM-EDM, MIL, AHIRL, EDM, GP-DQN, and BC at p=0.05.}]{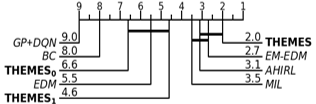}
  \caption{Mayo Clinic.}
\end{subfigure}

\caption{Critical difference diagrams for F1 score over three healthcare datasets.}
\label{fig:cd_graph}
\end{figure}

\section{Summary}

We leverage a generalized time-aware hierarchical AL framework, THEMES, to induce pedagogical policies from expert student demonstrations, with the assumption that multiple reward functions evolve over time. By integrating time-awareness with hierarchically learned reward patterns, THEMES enables more precise sub-trajectory partitions, resulting in highly effective pedagogical policies that align closely with expert student decisions. However, a limitation of this work lies in the relatively limited interpretation of evolving multiple reward functions in student demonstrations. Furthermore, this study is focused on a single ITS, although the time-aware hierarchical AL approach holds significant promise for other e-learning environments. Using student-ITS interaction logs, it can potentially validate the presence of multiple evolving intentions in diverse contexts. Moreover, while the trajectory lengths in our ITS were shorter than those in many real-world human-centric environments, THEMES still outperformed other approaches. Future work exploring e-learning environments with longer trajectories may reveal even more impressive results, further solidifying the framework's effectiveness in complex, evolving pedagogical settings.

\chapter{Hierarchical Apprenticeship Learning from Imperfect Demonstrations with Evolving Rewards}
\label{chap:aied26}

* \textit{This chapter has been published as:}
\begin{quote}
\textbf{Md Mirajul Islam}, Rajesh Debnath, Adittya Soukarjya Saha, and Min Chi. \textbf{“Hierarchical Apprenticeship Learning from Imperfect Demonstrations with Evolving Rewards”} \textit{The 27th International Conference on Artificial Intelligence in Education (AIED 2026).}
\end{quote}

While apprenticeship learning has shown promise for inducing effective pedagogical policies directly from student interactions in e-learning environments, most existing approaches rely on optimal or near-optimal expert demonstrations under a fixed reward. Real-world student interactions, however, are often inherently imperfect and evolving: students explore, make errors, revise strategies, and refine their goals as understanding develops.  In this work, we argue that imperfect student demonstrations are not noise to be discarded, but structured signals—provided their relative quality is ranked. We introduce \textbf{HALIDE, Hierarchical Apprenticeship Learning from Imperfect Demonstrations with Evolving Rewards}, which not only leverages suboptimal student demonstrations, but \emph{ranks} them within a hierarchical learning framework. HALIDE models student behavior at multiple levels of abstraction, enabling inference of higher-level intent and strategy from suboptimal actions while explicitly capturing the temporal evolution of student reward functions. By integrating demonstration quality into hierarchical reward inference, HALIDE distinguishes transient errors from suboptimal strategies and meaningful progress toward higher-level learning goals. Our results show that HALIDE more accurately predicts student pedagogical decisions than approaches that rely on optimal trajectories, fixed rewards, or unranked imperfect demonstrations. 

\section{Overview}
Most STEM e-learning environments, including Intelligent Tutoring Systems (ITSs), rely on \emph{pedagogical policies} to guide instructional actions, determining not only what content is presented but also when and how interventions are delivered throughout a student’s learning trajectory. These policies operate in a sequential decision-making setting, where each pedagogical choice influences future learning opportunities and long-term outcomes. Prior work has shown that Reinforcement Learning (RL) and Deep RL can induce data-driven pedagogical policies that outperform manually designed heuristics in controlled educational settings \citep{mandel2014offline,rowe2015improving,shen2016reinforcement}. However, these real-world RL and DRL adoption is limited by the large amount of data required to explore intricate learning environments and the inherent difficulty of manually engineering reward functions that accurately reflect sophisticated, long-horizon educational goals.\citep{kaelbling1996reinforcement,amodei2016concrete}.

\textbf{Apprenticeship Learning (AL)} provides an alternative by deriving policies from a limited set of \emph{expert demonstrations} through inference of the underlying reward function \citep{abbeel2004apprenticeship,ziebart2008maximum}. By assuming experts act optimally or near-optimally, AL can efficiently approximate effective decision-making. Student interactions in e-learning environments, however, are rarely optimal. Students actively explore different strategies, learn from mistakes, and refine their approaches, adapting their learning goals as they gain experience. This process reflects the presence of evolving internal reward functions that shift as students gain expertise and familiarity with the system. Conventional AL frameworks treat suboptimal actions as noise, failing to capture the dynamics of learning. We argue that imperfect student demonstrations instead constitute structured signals that can reflect evolving competencies, strategies, and reward priorities. Crucially, leveraging these signals requires more than simply incorporating suboptimal trajectories: their \emph{relative quality must be explicitly ranked}, as treating all imperfect demonstrations as equally informative obscures meaningful pedagogical structure.  

We introduce \textbf{HALIDE, Hierarchical Apprenticeship Learning from Imperfect Demonstrations with Evolving Rewards}, a framework that hierarchically models student behavior, explicitly captures the temporal evolution of goals while ranking demonstrations according to pedagogical quality. HALIDE infers high-level intent and strategy from suboptimal actions, separates transient errors from meaningful progress, and explicitly captures the temporal evolution of student goals. This perspective aligns with advances in imitation learning from noisy or mixed-quality demonstrations using confidence scores, rankings, or preference-based supervision \citep{zolna2020oril,wu2019imperfectil,tangkaratt2021robust,xu2021discriminatorweighted,liu2025pngail,hoang2024sprinql}. However, most methods operate in flat policy spaces, neglecting hierarchical decision-making and temporally evolving strategies critical to educational domains. Our prior work introduced \THEMES \citep{yang2025themes}, a hierarchical AL framework that models evolving reward functions via sub-trajectory partitioning and induce cluster-specific policies. While effective, \THEMES assumes that all input demonstrations are expert-level and equally informative, making it sensitive to noise arising from heterogeneous decision quality or suboptimal trajectories. HALIDE extends \THEMES with \emph{ranked hierarchical apprenticeship learning}, integrating continuous demonstration quality signals into hierarchical policy induction. Higher-quality pedagogical decisions exert greater influence, while the full trajectory structure is preserved. Crucially, ranking is embedded within the hierarchical learning loop, allowing high-level strategies to emerge from reward-regulated refinement rather than collapsing learning into a flat weighted imitation objective. 
 
 It is important to note that ranking in AL serves a fundamentally different role from reward functions in  RL and deep RL. In RL, rewards define the objective of behavior: they specify what the agent should optimize through interaction with the environment. Reward functions shape long-term return and drive policy improvement through trial-and-error exploration. Designing such rewards in educational settings is notoriously difficult, as pedagogical objectives are multi-faceted, long-horizon, and often only indirectly observable.  In contrast, ranking in AL does not define what behavior is optimal; instead, it modulates how much influence each demonstrated behavior should have when inducing a policy from observational data. Ranking answers a different question than rewards: rather than asking \textit{“What should be optimized?”}, it asks \textit{“Which observed decisions should be trusted more during learning?”} This distinction is critical in demonstration-driven settings.

We evaluate HALIDE against seven competitive AL baselines that rely on expert trajectories or fixed rewards, along with two ablation variants, on a challenging cross-semester prediction task. Models are trained on student trajectories from past semesters and tested on a subsequent semester to predict pedagogical decisions in an ITS. Our results showed that HALIDE consistently outperforms all baselines and ablations across evaluation metrics. Beyond this educational setting, we additionally ask whether ranked hierarchical apprenticeship learning generalizes to a second, higher-stakes human-centric domain: we validate HALIDE on three clinical EHR datasets for septic shock prevention, testing whether the same quality-aware weighting principle that governs unreliable pedagogical decisions also governs unreliable clinical ones. 
\textbf{This work makes the following key contributions} :
1)  We introduce \textbf{HALIDE}, a hierarchical AL framework that models student pedagogical behavior from \emph{imperfect demonstrations} while explicitly capturing \emph{evolving reward structures}, enabling the inference of higher-level intent and strategy as student goals and competencies change over time.
2) We show that the key is to introduce quality-aware hierarchical apprenticeship learning that ranks demonstrations rather than treating them uniformly. 
3) Extensive cross-semester evaluation on a real-world ITS shows that HALIDE consistently outperforms flat and hierarchical AL baselines, with strong gains from suboptimal student demonstrations.
4) We validate HALIDE beyond education on three clinical EHR datasets for septic shock prevention, showing that quality-aware ranking of imperfect demonstrations is not specific to student pedagogical decisions but generalizes to clinical decision-making as well.

\section{Related Work}
\subsubsection{Pedagogical Policy Induction in ITSs}
A large body of work has explored the use of reinforcement learning (RL) and deep reinforcement learning (DRL) to induce pedagogical policies in intelligent tutoring systems \citep{mandel2014offline,rowe2015improving,hostetter2025human}. These approaches typically model tutoring as a sequential decision-making problem in which the tutor selects instructional actions to maximize long-term learning outcomes \citep{shen2016reinforcement,zhou2019hierarchical,tithi2026adaptive}.
While RL-based methods have demonstrated strong empirical performance, they often require large amounts of interaction data and rely on carefully designed reward functions, such as normalized learning gain, to guide optimization. In practice, such data is difficult to obtain without deploying exploratory or randomized tutoring strategies, limiting scalability and adoption in real classrooms. Apprenticeship learning (AL) offers an alternative by inferring pedagogical policies directly from demonstrated behavior rather than explicit reward specification \citep{abbeel2004apprenticeship,ziebart2008maximum}.
In ITS settings, AL is particularly appealing because it enables fully offline learning from historical interaction logs. However, unlike tutor-driven RL, apprenticeship learning in ITSs treats students as the source of pedagogical decisions, using their choices—such as when to request hints or view worked examples—as demonstrations \citep{islam2024generalized,islam2025generalized}. This shift introduces student agency into policy induction, but also raises new challenges related to variability in demonstration quality.

\subsubsection{Apprenticeship Learning from Mixed-Quality Demonstrations}
\label{subsec:rw_mixed_quality}

Imitation and apprenticeship learning traditionally assume access to high-quality expert demonstrations, but this assumption is often violated in real-world domains where expert data are scarce and heterogeneous trajectories are abundant.
While early offline approaches such as behavioral cloning are simple and effective with sufficient expert coverage, they are known to suffer from compounding error and distribution shift under limited or imperfect demonstrations \citep{SCHAAL1999233,pmlr-v15-ross11a}. To address this challenge, a growing body of work studies learning from \emph{mixed-} or \emph{imperfect-quality} demonstrations. Prior methods incorporate confidence or reliability signals to reweight demonstrations or state--action pairs, including confidence-weighted imitation learning \citep{wu2019imperfectil}, discriminator-based reweighting \citep{xu2021discriminatorweighted}, positive-unlabeled formulations \citep{liu2025pngail}, and value- or reward-based approaches that integrate expert and suboptimal experience \citep{zolna2020oril,hoang2024sprinql}.
These techniques improve robustness by attenuating the influence of unreliable behavior without requiring online interaction.

However, most existing approaches operate in \emph{flat} policy spaces and apply quality at a coarse granularity, such as trajectory-level weighting.
They do not account for hierarchical or temporally evolving decision-making, where the reliability of demonstrated actions may vary over time.
Our work complements this literature by incorporating continuous quality signals into a \emph{hierarchical} apprenticeship learning framework, enabling quality-aware policy induction while preserving temporal abstraction and evolving pedagogical strategies.

\subsubsection{Hierarchical Apprenticeship Learning and Evolving Strategies: }
\label{subsec:rw_hierarchical}
Hierarchical and multi-intent apprenticeship learning has been studied in settings where demonstrations are governed by multiple latent reward functions or intentions, often via Bayesian or EM-based formulations that cluster trajectories and learn cluster-wise rewards or policies \citep{babes2011apprenticeship,dimitrakakis2011bayesian,choi2012nonparametric}. 
Separately, hierarchical and multi-modal imitation learning approaches aim to capture sub-task structure and temporally extended behavior \citep{hausman2017multi,krishnan2016hirl}. In educational domains, modeling evolving strategies is particularly important because students' goals and learning needs change over time. \THEMES provides a time-aware hierarchical apprenticeship learning approach that models evolving reward functions by partitioning trajectories into sub-trajectories and inducing cluster-specific policies \citep{yang2025themes}. 
While effective at capturing temporal abstraction and strategic evolution, the framework assumes that demonstrations admitted into training are uniformly reliable, limiting robustness when demonstrations are mixed-quality or when reliability varies within trajectories. Our work extends this line of research by incorporating continuous quality signals into hierarchical policy induction, enabling hierarchical apprenticeship learning to better handle realistic mixed-quality student demonstrations.

\begin{figure}
    \centering
    \includegraphics[width=\textwidth,alt={Overview diagram of the HALIDE framework. Mixed-quality student trajectories with importance weights enter RMT-TICC for sub-trajectory partitioning producing high-level states. The weighted EM-EDM module uses decision-level quality weights in the M-step to induce cluster-specific policies. High-level state-action pairs train an EM-IRL reward regulator that feeds back to refine partitioning. The iterative loop propagates quality-aware weighting through the full hierarchy.}]{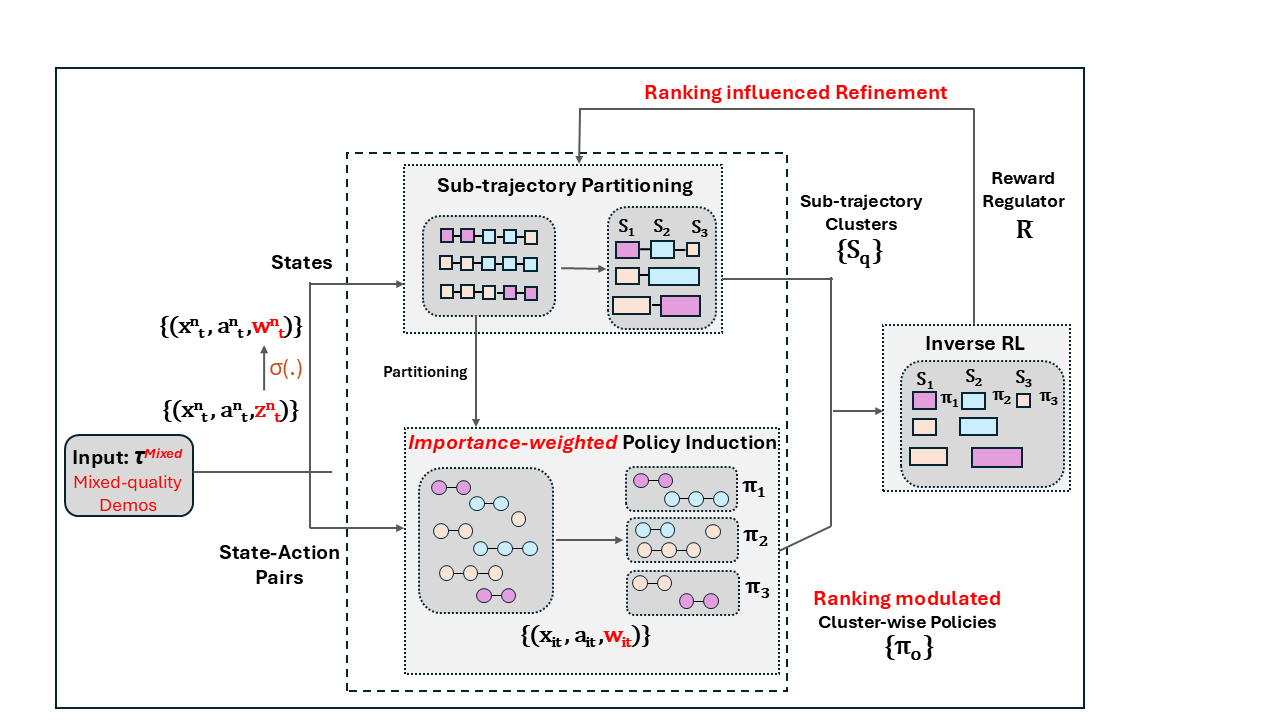}
    \caption{Overview of the HALIDE framework.}
    \label{fig:overview_ranked}
\end{figure}

\section{Methods}
\label{sec:method}
The overview of HALIDE is illustrated in Figure~\ref{fig:overview_ranked}, with details provided in Algorithm~\ref{alg:halide_impl}. The input consists of $N$ demonstrated trajectories of mixed-quality,
$\tau^n=\{(\mathbf{x}_t^n, a_t^n)\}_{t=1}^{T^n}$, where $\mathbf{x}_t^n \in \mathbb{R}^m$ denotes the state at time $t$, $a_t^n \in \mathcal{A}$ denotes the pedagogical action chosen at time $t$, $T^n$ is the trajectory length, and $n \in [1, N]$. HALIDE augments each demonstrated decision with a continuous importance weight $w_t^n \in (0,1]$, yielding weighted demonstrations $\{(\mathbf{x}_t^n, a_t^n, w_t^n)\}_{t=1}^{T^n}$. Importantly, $w_t^n$ modulates the influence of demonstrated decisions during policy induction and does not define a reward function or alter the underlying trajectories.

\noindent \textbf{Notation.}
We use $Q$ to denote the number of high-level state clusters produced by RMT--TICC and $O$ to denote the number of high-level policy (action) clusters induced by EM--EDM. The resulting set of sub-trajectories after partitioning is denoted as $\{\xi_i\}_{i=1}^{\hat{N}}$, where $\xi_i=\{(\mathbf{x}_{it}, a_{it}, w_{it})\}_{t=1}^{T_i}$.

\begin{algorithm}[t]
\caption{\textbf{HALIDE}.}
\label{alg:halide_impl}
\renewcommand{\thealgorithm}{}
\begin{algorithmic}[1]
    \STATE \textbf{Input:} trajectories $\{\tau^n\}_{n=1}^{N}$, quality signals $\{z_t^n\}$
    \STATE \textbf{Hyperparameters:} $Q,O,\omega,K,\alpha$
    \STATE Initialize $\overline{R}^{(0)}(\cdot)\gets 1$, iteration $k\gets 0$
    \WHILE{$k < K$}
        \STATE Run RMT--TICC using $\overline{R}^{(k)}$ to obtain $\{S_q^{(k)}\}$ and sub-trajectories $\{\xi_i^{(k)}\}$
        \STATE Run weighted EM--EDM on $\{\xi_i^{(k)}\}$ to obtain $\{\pi_o^{(k)}\}$ and $\{u_{io}^{(k)}\}$
        \STATE Learn $\overline{R}^{(k+1)}(\cdot)$ from high-level state-action pairs via EM-IRL
        \STATE $k\gets k+1$
    \ENDWHILE
    \STATE \textbf{Output:} $\{\pi_o\}_{o=1}^{O}$
\end{algorithmic}
\end{algorithm}

\subsection{Demonstration Quality and Weight Construction}
\label{subsec:weight_construction}

HALIDE assumes access to continuous demonstration quality $z_t^n$ aligned with demonstrated decisions. This quality measure may originate from learning outcomes, confidence estimates, or other domain-specific evaluations. We map the continuous quality to a relative ranking signal or importance weight via:
\begin{equation}
\label{eq:weight_map}
w_t^n = \sigma(\alpha z_t^n) = \frac{1}{1+\exp(-\alpha z_t^n)},
\end{equation}
where $\alpha>0$ controls the sensitivity. When $z_t^n$ is computed at the trajectory level, it is distributed across time to obtain a coarse per-step proxy. These weights are used only to modulate policy induction; they do not alter the underlying trajectories or the segmentation procedure.

\subsection{High-level States Extraction (RMT--TICC)}
\label{subsec:rmt-ticc}
HALIDE adopts the same high-level state extraction module as \THEMES. Each state $\mathbf{x}_t^n \in \mathbb{R}^m$ is assigned to one of $Q$ clusters by considering a sliding window of size $\omega \ll T^n$. For each time step $t$, the context window is $\mathbf{X}_t^n=\{\mathbf{x}_{t-\omega+1}^n,\ldots,\mathbf{x}_t^n\}$, treated as an $m\omega$-dimensional random variable and clustered into $Q$ Gaussian components.

Reward-regulated Multi-series Time-aware Toeplitz
Inverse Covariance-based Clustering (RMT-TICC) incorporates time-awareness and decision-making patterns via a reward-regulated consistency constraint:
\begin{equation}
\small
\label{eq:rmt_ticc_obj}
\argmin_{\boldsymbol{\Theta}, \mathbf{S}} \sum_{q=1}^{Q}
\bigg[
\sum_{n=1}^{N}\sum_{\mathbf{X}_t^n \in \mathbf{S}_q}
\Big( -\ell\ell(\mathbf{X}_t^n,\boldsymbol{\Theta}_q)
+ c(\mathbf{X}_{t-1}^n,\mathbf{S}_q,\Delta T_t^n,\overline{r}_t^n)\Big)
+ \lambda \|\boldsymbol{\Theta}_q\|_1
\bigg],
\end{equation}
where $-\ell\ell(\cdot)$ is the negative log-likelihood term, $c(\cdot)$ is the reward-regulated time-aware consistency term, and $\|\boldsymbol{\Theta}_q\|_1$ imposes sparsity. $\Delta T^{n}_{t}$ denotes the interval between consecutive states, \textit{i.e.}, $\mathbf{X}_{t}^{n}$ and $\mathbf{X}_{t-1}^{n}$.

A high-level reward regulator $\overline{R}(\cdot)$ is learned from high-level state-action pairs and distributed to obtain $\overline{r}_t^n$, which regulates the consistency constraint. This hierarchical structure enables modeling evolving pedagogical strategies without requiring online interaction.
\subsection{High-level Actions Induction (Weighted EM--EDM)}
\label{subsec:weighted_emedm}

\noindent\textbf{EDM (strictly offline).}
Energy-based Distribution Matching (EDM) induces a stochastic policy $\pi^\theta$ from demonstrations by matching the demonstration occupancy measure $\rho_D$ with the induced occupancy $\rho_{\pi^\theta}$:
\begin{equation}
\small
\label{eq:edm_kl_halide}
\argmin_{\theta} D_{\mathrm{KL}}(\rho_D \,\|\, \rho_{\pi^\theta})
=
\argmin_{\theta} -\mathbb{E}_{(\mathbf{x},a)\sim \rho_D}\big[\log \rho_\theta(\mathbf{x},a)\big].
\end{equation}
This objective decomposes into (i) a behavior cloning term on demonstrated actions and
(ii) an occupancy regularization term involving the energy-model normalizer $U^\theta(\mathbf{x})$,
allowing policy learning in a strictly offline manner without environment interaction.

\noindent\textbf{EM--EDM over sub-trajectories.}
To model heterogeneous strategies and evolving rewards, EM--EDM assumes each sub-trajectory
$\xi_i=\{(\mathbf{x}_{it},a_{it})\}_{t=1}^{T_i}$ is generated by one of $O$ latent policy clusters,
with prior $\rho_o$ and parameters $\theta_o$.
Given current parameters, the E-step computes responsibilities using the standard sub-trajectory likelihood:
\begin{equation}
\small
\label{eq:emedm_likelihood_halide}
\Pr(\xi_i \mid \theta_o)\;\propto\;
\prod_{t=1}^{T_i}
\frac{\pi^{\theta_o}(a_{it}\mid \mathbf{x}_{it})}{U^{\theta_o}(\mathbf{x}_{it})},
\qquad
u_{io}
=
\frac{\rho_o \Pr(\xi_i\mid \theta_o)}
{\sum_{o'=1}^{O} \rho_{o'} \Pr(\xi_i\mid \theta_{o'})}.
\end{equation}
Importantly, this assignment step depends only on behavioral consistency and temporal structure,
and is not influenced by demonstration quality.

\noindent\textbf{HALIDE: decision-weighted objective in the M-step.}
HALIDE incorporates demonstration quality during policy induction by augmenting each sub-trajectory
with decision-level importance weights,
$\xi_i=\{(\mathbf{x}_{it}, a_{it}, w_{it})\}_{t=1}^{T_i}$,
where $w_{it}\in(0,1]$ reflects the relative reliability of the demonstrated decision at time $t$. In the M-step, HALIDE updates each policy cluster by minimizing a responsibility-weighted, decision-weighted objective:
\begin{equation}
\small
\label{eq:weighted_mstep_halide}
\argmin_{\{\theta_o\}}
\sum_{i=1}^{\hat{N}}\sum_{o=1}^{O} u_{io}
\sum_{t=1}^{T_i} w_{it}\,
\Big(-\log \pi^{\theta_o}(a_{it}\mid \mathbf{x}_{it})\Big)
\;+\;
\lambda
\sum_{i=1}^{\hat{N}}\sum_{o=1}^{O} u_{io}
\sum_{t=1}^{T_i} w_{it}\,
\log U^{\theta_o}(\mathbf{x}_{it}).
\end{equation}

Here, $w_{it}$ uniformly scales both the behavior cloning term and the EDM occupancy regularization,
ensuring that lower-quality decisions exert proportionally less influence during policy learning,
while preserving the strictly offline nature of EDM.
Responsibilities $u_{io}$ are computed from the unweighted likelihood in~\eqref{eq:emedm_likelihood_halide}.
When $w_{it}\equiv 1$ for all $(i,t)$, HALIDE reduces exactly to standard EM--EDM as used in \THEMES.

\subsection{High-level Reward Regulator and Effect of Ranking}
\label{sec:reward_regulator}

HALIDE maintains a high-level reward regulator $R(\cdot)$ that captures evolving  pedagogical intent across sub-trajectories. The regulator is learned from high-level state--action statistics induced by hierarchical policies and is used to modulate the temporal consistency term in RMT--TICC, thereby biasing segmentation toward coherent pedagogical progress.

\noindent\textbf{EM--IRL reward inference.}
Given high-level states $\{S_q\}_{q=1}^{Q}$ obtained from RMT--TICC and 
high-level policies $\{\pi_o\}_{o=1}^{O}$ induced by weighted EM--EDM, 
HALIDE learns a hierarchical reward regulator via EM-Inverse RL (EM--IRL). 
The reward for each state--action pair is computed as

\begin{equation}
\small
r_{it}
=
R(x_{it}, a_{it})
=
\frac{1}{O}
\sum_{o=1}^{O}
u_{io}\,
\pi_o(a_{it} \mid x_{it})\,
R(S_q,\pi_o),
\end{equation}

where $u_{io}$ denotes the EM responsibility of sub-trajectory $\xi_i$ 
belonging to policy cluster $\pi_o$, and $S_q$ is the high-level state 
cluster associated with $(x_{it}, a_{it})$.

\noindent\textbf{Reward feedback to RMT--TICC.}
The inferred rewards $\{r_{it}\}$ regulate the reward-aware temporal 
consistency term in RMT--TICC (Eq.~2), encouraging consecutive states with 
consistent decision-making patterns to remain within the same sub-trajectory. This creates an iterative hierarchical feedback loop:

\[
\{w_{it}\}
\Rightarrow
\{\pi_o\}
\Rightarrow
R(\cdot)
\Rightarrow
\text{refined sub-trajectory segmentation}.
\]

Although ranking weights $w_{it}$ are applied only during weighted EM--EDM, 
they influence reward inference indirectly by shaping the induced state--action distributions. This coupling allows higher-quality decisions to guide high-level strategy discovery while preserving hierarchical structure and evolving rewards.




\section{Experiments}
\begin{figure}[htbp]
    \centering
    \includegraphics[scale=0.2,alt={Screenshot of the Pyrenees probability ITS interface showing a multi-step training problem with input fields, hint buttons, and three presentation mode buttons: PS (Problem Solving), WE (Worked Example), and CPS (Collaborative Problem Solving).}]{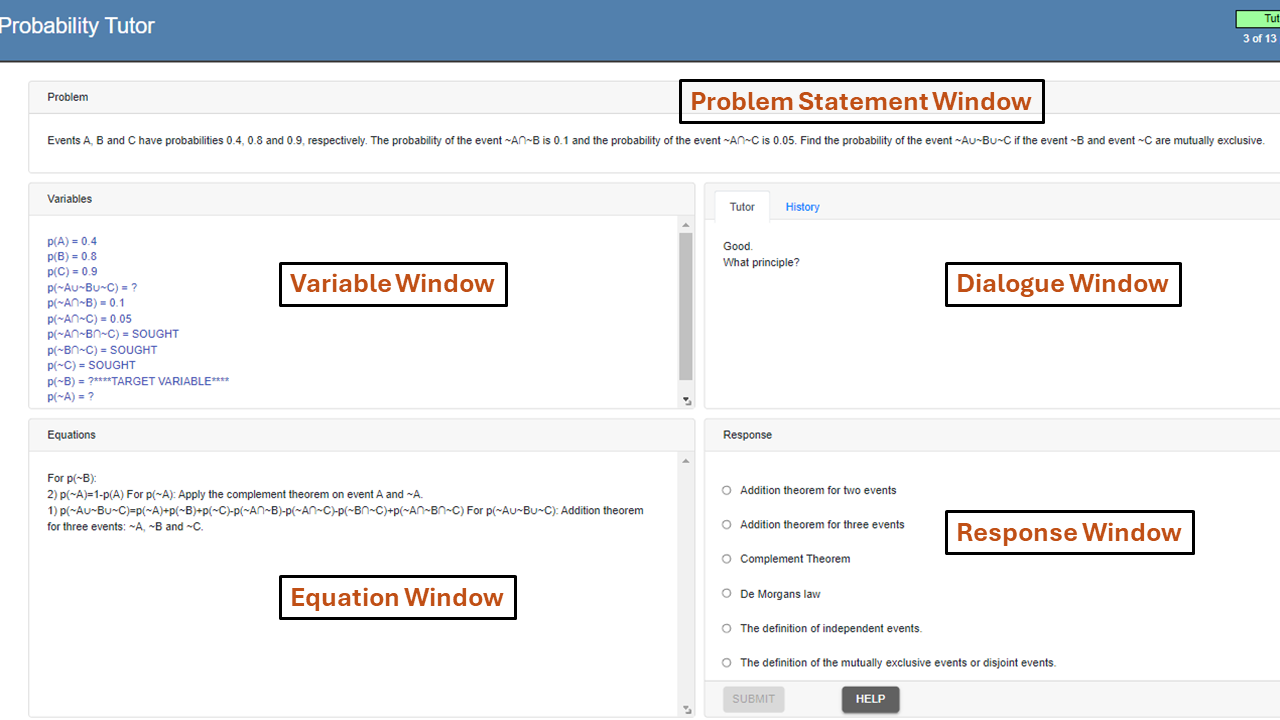}
    \caption{ITS Interface.}
    \label{fig:pyrenees-interface-ch5}
\end{figure}

This study utilized a Probability ITS (Figure \ref{fig:pyrenees-interface-ch5}) deployed across multiple Spring and Fall semesters of a large undergraduate STEM course at a public university. The system, developed by domain experts and overseen by departmental committees, is designed to support entry-level undergraduates in learning ten core probability principles, \textit{e.g.}, the complement theorem and Bayes’ rule, through a sequence of complex problem-solving activities.

\subsubsection{Data Collection}
Data were collected from 244 undergraduate students across five semesters: 57 students in Spring 2021 (S21), 56 in Spring 2022 (S22), 54 in Spring 2024 (S24), 44 in Fall 2024 (F24), and 33 in Spring 2025 (S25). All students followed an identical protocol consisting of a textbook study phase, a pre-test, ITS-guided training, and a post-test. The pre-test and post-test comprised 8 and 12 open-ended problem-solving questions, respectively, and were graded independently by two evaluators using a double-blind process, with discrepancies resolved through discussion. Scores were normalized to the range $[0,100]$. During ITS training, students interacted with the same 12 problems in a fixed order. For 10 of these problems, students exercised \emph{agency} by choosing among three pedagogical approaches: \emph{1) solving the problem independently}, \emph{2) collaborating with the ITS}, or \emph{3) viewing a worked-out example}. All interaction data were anonymized and obtained under an exempt IRB-approved protocol.

\begin{figure}[htbp]
    \centering
    \includegraphics[scale=0.2,alt={Flowchart of Quantized Learning Gain classification. Students are placed in Low, Medium, or High performance groups by pre-test and post-test scores. Solid arrows denote High QLG (improved or remained in High); dashed arrows denote Low QLG (stayed in Low or Medium, or declined).}]{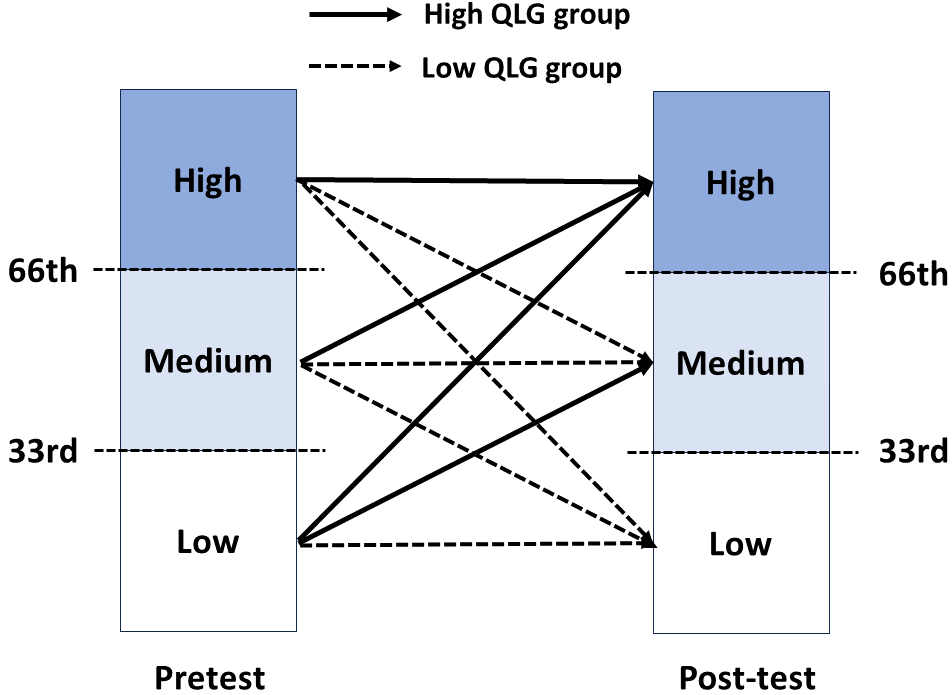}
    \caption{Quantized Learning Gain.}
    \label{fig:qlg-ch5}
\end{figure}

\subsubsection{State and Action Representation}
Students’ latent learning states are not directly observable. Following prior work, each interaction state is represented using a \emph{130-dimensional continuous} feature vector constructed by domain experts. These features capture multiple aspects of student behavior, including \textbf{\textit{10} Autonomy} features that measure the student's work, such as the number of elicits since the last tell; \textbf{\textit{22} Temporal} features that represent time-related information, like average time per step; \textbf{\textit{31} Problem Solving} features that reflect contextual information about problem-solving, including difficulty level; \textbf{\textit{57} Performance} features that indicate student's performance so far, such as the percentage of correct responses; \textbf{\textit{10} Hints} features that show data on hint usage, such as total hints requested. The action space consists of \emph{three pedagogical decisions} corresponding to the student’s choice of instructional support.

\subsubsection{Expert Demonstration Selection}
To identify optimal or near-optimal student demonstrations, we compute Quantized Learning Gain (QLG) measure \citep{mao2018tracing}, categorizing students into low, medium, or high-performance groups based on their pre- and post-test scores. A ``High'' QLG is assigned to students who improved or remained in the high-performance group, while a ``Low'' QLG is given to those who dropped to a lower group or stayed in low/medium groups (Figure \ref{fig:qlg-ch5}). This helped identify students who benefited most from our ITS. Across the five semesters, this procedure yields 94 expert trajectories: 33 from S21, 14 from S22, 25 from S24, 13 from F24, and 9 from S25. The remaining student trajectories that were not selected as experts are treated as imperfect, sub-optimal trajectories (\textbf{I}). In addition to expert-only(\textbf{E}) training, we consider a mixed-quality setting that includes both expert and imperfect, sub-optimal student trajectories, denoted as \textbf{E+I}.  

\subsubsection{Ranking Signals:} 
HALIDE relies on the availability of a meaningful \textbf{ranking signal} to modulate the influence of heterogeneous student demonstrations during AL. Unlike the reward function in RL, which assigns an \emph{absolute} value to states or actions within a single trajectory, ranking in AL encodes \emph{comparative information across demonstrations}. In HALIDE, this comparative signal, denoted as $w_i$ (Section~\ref{sec:method}, Figure~\ref{fig:overview_ranked}), determines the relative influence of student $i$’s demonstrated pedagogical decisions during policy induction.

In this work, we define ranking using Normalized Learning Gain (NLG), a well-established measure of student learning that accounts for differences in incoming competence. NLG captures how much a student improves relative to their initial knowledge, enabling fair comparison across students with heterogeneous prior preparation. Formally, NLG is defined as $\frac{posttest-pretest}{\sqrt{1-pretest}}$, where 1 is the maximum score for both pre- and post-test. To ensure stability across cohorts and semesters, NLG values are standardized across all semesters to obtain $NLG_z$ and the  final ranking weight is then computed as
$w_{i} = \sigma(\alpha . NLG_z)$. where $\sigma(\cdot)$ denotes the sigmoid function and $\alpha$ controls the sharpness of the weighting.
This formulation maps relative learning gains to \textbf{smooth, bounded weights}, allowing HALIDE to emphasize demonstrations from students who exhibit stronger learning outcomes while retaining informative behavior from suboptimal trajectories. Importantly, this ranking signal functions as a \emph{soft structural prior} rather than a hard filter, enabling robust learning from  mixed-quality student demonstrations.


\subsubsection{Experiment Settings}
We compare \textbf{HALIDE} against a set of competitive apprenticeship learning (AL) baselines and structured ablations designed to isolate the effects of hierarchy, demonstration quality modeling, and data heterogeneity on a challenging task of predicting future semester students' actions using past ones.

\noindent\textbf{1) Flat AL Baselines:}
Behavioral Cloning (BC) \citep{gleave2022imitation} learns a policy via supervised learning on state--action pairs from expert demonstrations. 
EDM \citep{jarrett2020edm} is a state-of-the-art strictly offline apprenticeship learning method that assumes a single stationary reward function and matches expert state--action occupancy distributions.

\noindent\textbf{2) Hierarchical AL Baselines:}
We include an offline adaptation of Hierarchical Inverse Reinforcement Learning (AHIRL) \citep{krishnan2016hirl}, which reuses segment-level clustering and local reward structure but replaces online RL with EDM-based policy induction to ensure fair comparison in a fully offline setting.
\THEMES is a prior hierarchical apprenticeship learning framework that partitions trajectories into temporally coherent sub-trajectories and induces cluster-specific policies to model evolving pedagogical strategies over time. 

\noindent\textbf{3) Quality-Aware and Data-Ablations:}
Methods denoted with the subscript \textbf{W} incorporate continuous demonstration quality signals to modulate the influence of demonstrated decisions during policy induction, without altering the underlying policy or hierarchical structure.
This includes EDM\textsubscript{W} in both expert-only (E) and mixed-quality (E+I) settings. \textbf{HALIDE\(_0\)} serves as a data ablation of HALIDE, which augments hierarchical learning with quality-aware weighting while using expert demonstrations only.
Another ablation, \textbf{HALIDE\(_1\)}, jointly leverages expert and sub-optimal demonstrations (E+I), enabling the model to exploit increased state-space coverage while using uniform ranking signals.


\subsubsection{Evaluation Protocol and Metrics}
All methods are evaluated using semester-based temporal cross-validation (4-fold) which is a much stricter approach than standard cross-validation \citep{mao2020time}. Models are trained on one or more past semesters and evaluated on a future semester, and results are aggregated across multiple such folds. We report Accuracy, Recall, Precision, F1 score, Area Under the ROC Curve (AUC), Area Under the Precision--Recall Curve (APR), and Jaccard score. We emphasize F1 score, AUC, APR and Jaccard as primary metrics, as they are robust to class imbalance and better capture decision-level agreement. Reported results correspond to the mean and standard deviation across temporal folds, and statistical significance is assessed using non-parametric tests where appropriate.

\section{Results for Education}
\label{sec:results-ch5}

Table~\ref{tab:result_xsem_ranked} reports the mean and standard deviation of semester-based temporal cross-validation results across the different methods, with the best-performing baselines and ablations highlighted in bold, and the overall best results are marked with $^{*}$. 

\renewcommand{\arraystretch}{1.3} 
\setlength{\tabcolsep}{1.2 pt}
\begin{table*}
\caption{\centering Semester-based temporal cross-validation results (mean $\pm$ std).}
\label{tab:result_xsem_ranked}
\small
\centering
\begin{tabular}{c|c|ccc||cc|cc} \hline
\thead{Data} & \thead{Methods} & \thead{Acc} & \thead{Rec} & \thead{Prec} & \thead{F1} & \thead{AUC} & \thead{APR} & \thead{JAC} \\ \hline

\multirow{5}{*}{E}
& BC \tiny{\citep{gleave2022imitation}} 
& $.434_{\pm .08}$ & $.434_{\pm .08}$ & $.490_{\pm .17}$ & $.453_{\pm .11}$ & $.488_{\pm .08}$ & $.490_{\pm .18}$ & $.314_{\pm .10}$ \\

& EDM \tiny{\citep{jarrett2020strictly}}
& $.673_{\pm .09}$ & $.673_{\pm .09}$ & $.754_{\pm .06}$ & $.695_{\pm .07}$ & $.766_{\pm .13}$ & $.755_{\pm .07}$ & $.548_{\pm .08}$ \\


& AHIRL \tiny{\citep{krishnan2016hirl}}
& $.586_{\pm .02}$ & $.586_{\pm .02}$ & $.666_{\pm .11}$ & $.602_{\pm .05}$ 
& $.716_{\pm .06}$ & $.658_{\pm .13}$ & $.443_{\pm .05}$ \\

& \THEMES
& $\mathbf{.724_{\pm .07}}$ & $\mathbf{.724_{\pm .07}}$ & $\mathbf{.784_{\pm .05}}$ 
& $\mathbf{.742_{\pm .05}}$ & $\mathbf{.803_{\pm .11}}$ & $\mathbf{.787_{\pm .06}}$ & $\mathbf{.603_{\pm .07}}$ \\ \hline

\multirow{2}{*}{E}
& EDM\textsubscript{W}
& $.674_{\pm .09}$ & $.674_{\pm .09}$ & $.755_{\pm .06}$ & $.696_{\pm .07}$ 
& $.769_{\pm .13}$ & $.757_{\pm .07}$ & $.550_{\pm .08}$ \\

& HALIDE$_0$
& $\mathbf{.738_{\pm .08}}$ & $\mathbf{.738_{\pm .08}}$ & $\mathbf{.785_{\pm .06}}$ 
& $\mathbf{.752_{\pm .06}}$ & $\mathbf{.813_{\pm .11}}$ & $\mathbf{.809^{*}_{\pm .06}}$ & $\mathbf{.617_{\pm .08}}$ \\ \hline

\multirow{2}{*}{E+I}
& EDM
& $\mathbf{.705_{\pm .08}}$ & $\mathbf{.705_{\pm .08}}$ & $\mathbf{.768_{\pm .06}}$ 
& $\mathbf{.723_{\pm .07}}$ & $\mathbf{.774_{\pm .14}}$ & $\mathbf{.767_{\pm .07}}$ & $\mathbf{.582_{\pm .09}}$ \\ 

& HALIDE$_1$
& $.698_{\pm .07}$ & $.698_{\pm .07}$ & $.766_{\pm .06}$ & $.716_{\pm .06}$ 
& $.769_{\pm .14}$ & $.756_{\pm .06}$ & $.573_{\pm .08}$ \\ \hline

\multirow{2}{*}{E+I}
& EDM\textsubscript{W}
& $.697_{\pm .09}$ & $.697_{\pm .09}$ & $.768_{\pm .06}$ & $.716_{\pm .08}$ 
& $.773_{\pm .14}$ & $.762_{\pm .07}$ & $.575_{\pm .10}$ \\

& HALIDE
& $\mathbf{.776^{*}_{\pm .07}}$ & $\mathbf{.776^{*}_{\pm .07}}$ & $\mathbf{.793^{*}_{\pm .08}}$ 
& $\mathbf{.782^{*}_{\pm .08}}$ & $\mathbf{.827^{*}_{\pm .08}}$ & $\mathbf{.801_{\pm .07}}$ & $\mathbf{.658^{*}_{\pm .10}}$ \\ \hline

\end{tabular}
\end{table*}

Across all evaluation metrics, the proposed \textbf{HALIDE} framework achieves the strongest overall performance, significantly outperforming all flat and hierarchical baselines. In particular, \textbf{HALIDE} trained on expert and sub-optimal data (E+I) attains the highest AUC (0.827) and Jaccard score (0.658), indicating superior discrimination ability and decision-level agreement under substantial cross-semester distribution shift.

Among flat apprenticeship learning baselines trained on expert data (E), \textbf{EDM} consistently outperforms \textbf{BC}, confirming that matching occupancy measures under a learned reward is more effective than direct behavioral cloning. However, flat methods remain limited by their assumption of a single stationary objective and their inability to capture temporal abstraction or evolving pedagogical strategies driven by student agency.

Comparing hierarchical approaches under expert-only data (E), \textbf{\THEMES} improves over flat baselines, demonstrating the benefit of time-aware sub-trajectory partitioning and hierarchical policy induction. However, introducing heterogeneous data without quality modeling is insufficient: \textbf{HALIDE$_1$}, trained on mixed data (E+I), does not improve over expert-only \textbf{THEMES}. This suggests that while mixed-quality data expands state-space coverage, unweighted hierarchical learning allows sub-optimal decisions to exert equal influence, diluting informative signals and limiting robustness.

\begin{figure}
  \centering
  \begin{subfigure}{0.7\textwidth}
    \centering
    \includegraphics[width=\linewidth,alt={Critical difference diagram for F1 score based on Conover-Friedman test with Holm correction at p=0.05 across four semester-based folds. HALIDE (E+I) achieves the best mean rank, followed by HALIDE-0, THEMES, EDM (E+I), EDM-W (E+I), HALIDE-1, EDM, BC, and AHIRL. HALIDE is separated from weaker baselines with no connecting bar.}]{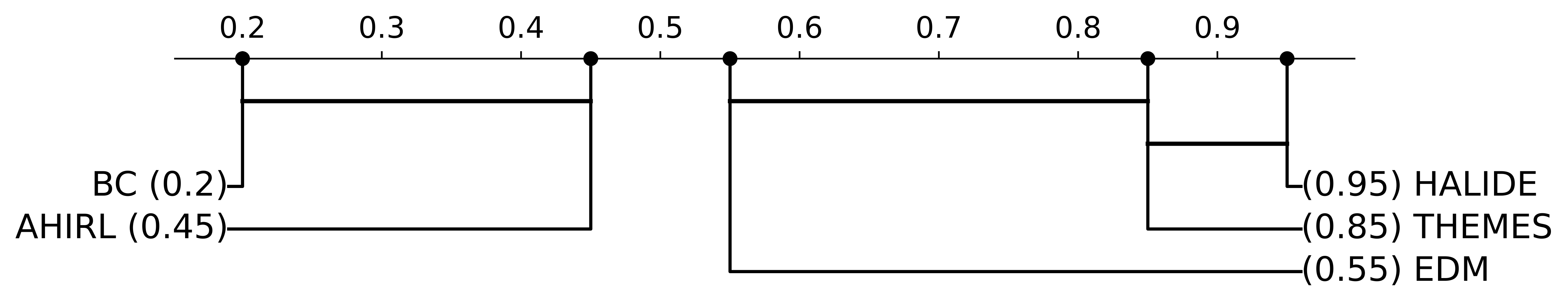}
    \caption{F-score.}
    \label{subfig:cd_fscore}
  \end{subfigure}
  \hfill
  \begin{subfigure}{0.7\textwidth}
    \centering
    \includegraphics[width=\linewidth,alt={Critical difference diagram for Jaccard score based on Conover-Friedman test with Holm correction at p=0.05 across four semester-based folds. HALIDE (E+I) achieves the best mean rank, followed by HALIDE-0, THEMES, and other baselines. HALIDE is clearly separated from BC and AHIRL.}]{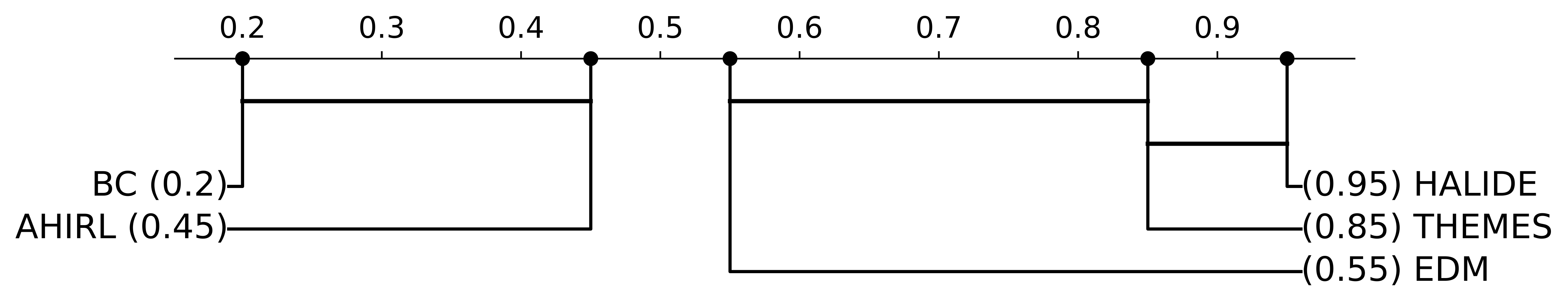}
    \caption{Jaccard Score.}
    \label{subfig:cd_jac-ch5}
  \end{subfigure}
  \caption{Critical difference diagram for F-score and Jaccard.}
  \label{fig:cd_plot-ch5}
\end{figure}

In contrast, incorporating demonstration quality via ranking signal yields consistent gains. Under expert-only data, \textbf{HALIDE$_0$} outperforms its unweighted counterpart, indicating that even demonstrations labeled as ``expert'' exhibit meaningful step-level variation in reliability. This challenges the standard apprenticeship learning assumption that expert trajectories are uniformly optimal and highlights the importance of modeling intra-trajectory quality differences in human-centric domains.

The largest gains emerge in the mixed-quality regime. While flat and unweighted hierarchical methods struggle to benefit from heterogeneous demonstrations, \textbf{HALIDE} effectively leverages both expert and sub-optimal data (E+I). By attenuating the influence of low-impact or erroneous decisions while preserving useful routine actions, HALIDE exploits increased state-space coverage without inheriting sub-optimal biases. This selective reuse of heterogeneous data enables HALIDE to substantially outperform both \textbf{\THEMES} and weighted flat baselines in all key metrics.

The critical difference (CD) diagrams \citep{conover1979comparison} for F1 and Jaccard scores, shown in Figure~\ref{fig:cd_plot-ch5}, summarize pairwise statistical comparisons based on the Conover– Friedman test ($p = 0.05$) with Holm correction across semester-based folds. Due to the limited number of folds ($n=4$), the test is conservative, and several methods form overlapping equivalence groups despite clear differences in mean performance. Nevertheless, HALIDE consistently attains the best average rank and is clearly separated from weaker baselines such as BC and AHIRL. The absence of separation between some intermediate methods reflects limited statistical power rather than identical performance.

Taken together, these results support the central hypothesis of this work: \emph{hierarchical apprenticeship learning benefits from explicitly ranking demonstrations by quality, particularly in realistic educational settings where behavior is heterogeneous, imperfect, and only near-optimal}. 
HALIDE reconciles robustness with data efficiency by modulating the influence of demonstrated decisions, enabling principled reuse of mixed-quality student data while preserving the temporal and strategic structure induced by hierarchical modeling with evolving rewards.
\section{Results for Healthcare}

Table~\ref{tab:combined_results} reports results across three healthcare datasets—CCHS, Mayo, and MIMIC—under expert-only (E) and all-data (A) training conditions, evaluating both unweighted and quality-weighted variants, with the best-performing method within each block in bold and the overall best results marked with $^{*}$.

Among flat apprenticeship learning baselines trained on expert data (E), \textbf{\THEMES} consistently outperforms both \textbf{EDM} and \textbf{EM-EDM} across all three datasets, confirming that temporal abstraction and hierarchical sub-trajectory partitioning are broadly beneficial in clinical decision-making. Notably, \textbf{EM-EDM} yields a more pronounced improvement over \textbf{EDM} on MIMIC than on CCHS or Mayo, suggesting that mixture-based reward modeling offers greater advantage when the dataset is larger and clinician behavior is more heterogeneous.

Incorporating demonstration quality weighting on expert-only data (E) consistently improves hierarchical performance: \textbf{\THEMES$_w$} achieves statistically significant gains over unweighted \textbf{\THEMES} on CCHS (F1: 0.722 vs.\ 0.696; AUC: 0.742 vs.\ 0.725) and on Mayo (F1: 0.679 vs.\ 0.643; Jaccard: 0.535 vs.\ 0.490), while remaining competitive on MIMIC. This pattern extends the finding from the educational domain: even demonstrations labeled as expert exhibit meaningful step-level variation in decision reliability, and explicitly modeling this variation improves downstream policy induction.

Expanding to all-quality demonstrations (A) without quality modeling yields mixed results. On MIMIC, \textbf{\THEMES} trained on all data achieves AUC = 0.858 and Jaccard = 0.759, surpassing its expert-only counterpart and confirming that broader state-space coverage benefits complex clinical settings. On CCHS, however, augmenting with heterogeneous data under unweighted flat baselines degrades performance in several configurations (e.g., EDM$_w$ under A yields F1 = 0.527 vs.\ 0.622 under E), indicating that undirected inclusion of lower-quality demonstrations can introduce noise that outweighs the benefit of expanded coverage when datasets are smaller.

In contrast, incorporating the quality ranking signal yields consistent gains in the all-data regime. \textbf{HALIDE} trained on all demonstrations achieves the strongest overall results on MIMIC across all four primary metrics: F1 = 0.869, AUC = 0.874, AP = 0.839, and Jaccard = 0.790—each the overall best ($^{*}$) in Table~\ref{tab:combined_results}—substantially outperforming both flat and unweighted hierarchical baselines. On Mayo, \textbf{HALIDE} attains the best AUC (0.632$^{*}$) and AP (0.559$^{*}$), confirming its advantage in discriminative accuracy on moderate-sized datasets. On CCHS, \textbf{\THEMES$_w$} trained on expert data remains the top-performing method across most metrics, suggesting that when datasets are smaller and expert demonstrations are relatively sufficient, the marginal gain from leveraging all-quality data diminishes. The dataset-dependent performance pattern—most pronounced on MIMIC and more modest on CCHS—further indicates that the benefit of quality-aware heterogeneous coverage scales with the complexity and heterogeneity of the clinical environment.

Taken together, these results support the central hypothesis of this work across the healthcare domain: \emph{hierarchical apprenticeship learning benefits from explicitly ranking demonstrations by quality, particularly on complex, large-scale clinical datasets where clinician behavior is heterogeneous and only near-optimal}. HALIDE reconciles the tension between data efficiency and coverage by selectively attenuating the influence of low-quality decisions while preserving useful routine actions, enabling principled reuse of mixed-quality clinical trajectories within a hierarchical framework with evolving rewards.

\renewcommand{\arraystretch}{1.0}
\setlength{\tabcolsep}{1.2 pt}

\begin{table*}
\caption{\centering Combined HALIDE vs baselines \& ablations results for healthcare datasets.}
\label{tab:combined_results}
\small
\centering
\resizebox{\textwidth}{!}{%
\begin{tabular}{c|c|cccc|cccc|cccc} \hline
\multirow{2}{*}{\thead{Data}}
& \multirow{2}{*}{\thead{Method}}
& \multicolumn{4}{c|}{\thead{CCHS}}
& \multicolumn{4}{c|}{\thead{Mayo}}
& \multicolumn{4}{c}{\thead{MIMIC}} \\ \cline{3-14}
& & \thead{F1} & \thead{AUC} & \thead{AP} & \thead{JAC} & \thead{F1} & \thead{AUC} & \thead{AP} & \thead{JAC} & \thead{F1} & \thead{AUC} & \thead{AP} & \thead{JAC} \\ \hline

\multirow{3}{*}{E}
& EDM 
& $.623_{\pm .04}$ & $.549_{\pm .02}$ & $.519_{\pm .01}$ & $.470_{\pm .04}$ 
& $.597_{\pm .10}$ & $.560_{\pm .06}$ & $.523_{\pm .02}$ & $.447_{\pm .11}$ 
& $.622_{\pm .06}$ & $.628_{\pm .07}$ & $.583_{\pm .05}$ & $.456_{\pm .06}$ \\

& EM-EDM 
& $.601_{\pm .03}$ & $.626_{\pm .03}$ & $.555_{\pm .02}$ & $.436_{\pm .04}$ 
& $.609_{\pm .08}$ & $.568_{\pm .05}$ & $.527_{\pm .02}$ & $.457_{\pm .09}$ 
& $.763_{\pm .04}$ & $.757_{\pm .04}$ & $.694_{\pm .04}$ & $.620_{\pm .05}$ \\

& THEMES 
& $\mathbf{.696}_{\pm .03}$ & $\mathbf{.725}_{\pm .02}$ & $\mathbf{.613}_{\pm .02}$ & $\mathbf{.541}_{\pm .04}$ 
& $\mathbf{.643}_{\pm .07}$ & $\mathbf{.622}_{\pm .05}$ & $\mathbf{.552}_{\pm .02}$ & $\mathbf{.490}_{\pm .08}$ 
& $\mathbf{.848}_{\pm .09}$ & $\mathbf{.845}_{\pm .09}$ & $\mathbf{.797}_{\pm .11}$ & $\mathbf{.748}_{\pm .13}$ \\ \hline

\multirow{3}{*}{E}
& EDM$_w$ 
& $.622_{\pm .04}$ & $.554_{\pm .03}$ & $.521_{\pm .01}$ & $.469_{\pm .05}$ 
& $.597_{\pm .10}$ & $.560_{\pm .06}$ & $.523_{\pm .02}$ & $.448_{\pm .11}$ 
& $.623_{\pm .06}$ & $.629_{\pm .07}$ & $.583_{\pm .05}$ & $.457_{\pm .06}$ \\

& EM-EDM$_w$ 
& $.586_{\pm .03}$ & $.613_{\pm .06}$ & $.549_{\pm .03}$ & $.422_{\pm .04}$ 
& $.593_{\pm .08}$ & $.565_{\pm .05}$ & $.525_{\pm .02}$ & $.438_{\pm .08}$ 
& $.768_{\pm .03}$ & $.765_{\pm .04}$ & $.700_{\pm .04}$ & $.627_{\pm .04}$ \\

& THEMES$_w$ 
& $\mathbf{.722}^{*}_{\pm .04}$ & $\mathbf{.742}^{*}_{\pm .02}$ & $\mathbf{.628}^{*}_{\pm .02}$ & $\mathbf{.575}^{*}_{\pm .05}$ 
& $\mathbf{.679}^{*}_{\pm .07}$ & $\mathbf{.616}_{\pm .04}$ & $\mathbf{.551}_{\pm .02}$ & $\mathbf{.535}^{*}_{\pm .08}$ 
& $\mathbf{.846}_{\pm .09}$ & $\mathbf{.845}_{\pm .09}$ & $\mathbf{.794}_{\pm .11}$ & $\mathbf{.745}_{\pm .13}$ \\ \hline

\multirow{3}{*}{A}
& EDM 
& $.637_{\pm .03}$ & $.524_{\pm .02}$ & $.509_{\pm .01}$ & $.491_{\pm .03}$ 
& $.545_{\pm .11}$ & $.546_{\pm .05}$ & $.517_{\pm .02}$ & $.392_{\pm .10}$ 
& $.530_{\pm .08}$ & $.558_{\pm .07}$ & $.536_{\pm .04}$ & $.368_{\pm .08}$ \\

& EM-EDM 
& $.623_{\pm .02}$ & $.558_{\pm .03}$ & $.523_{\pm .01}$ & $.468_{\pm .03}$ 
& $.603_{\pm .10}$ & $.578_{\pm .05}$ & $.530_{\pm .02}$ & $.451_{\pm .10}$ 
& $.782_{\pm .04}$ & $.778_{\pm .03}$ & $.716_{\pm .04}$ & $.645_{\pm .05}$ \\

& THEMES 
& $\mathbf{.654}_{\pm .05}$ & $\mathbf{.679}_{\pm .04}$ & $\mathbf{.585}_{\pm .03}$ & $\mathbf{.495}_{\pm .05}$ 
& $\mathbf{.660}_{\pm .06}$ & $\mathbf{.622}_{\pm .04}$ & $\mathbf{.553}_{\pm .02}$ & $\mathbf{.510}_{\pm .08}$ 
& $\mathbf{.852}_{\pm .12}$ & $\mathbf{.858}_{\pm .11}$ & $\mathbf{.815}_{\pm .12}$ & $\mathbf{.759}_{\pm .16}$ \\ \hline

\multirow{3}{*}{A}
& EDM$_w$ 
& $.527_{\pm .11}$ & $.517_{\pm .06}$ & $.508_{\pm .02}$ & $.380_{\pm .11}$ 
& $.567_{\pm .09}$ & $.555_{\pm .05}$ & $.521_{\pm .02}$ & $.412_{\pm .09}$ 
& $.503_{\pm .08}$ & $.547_{\pm .07}$ & $.529_{\pm .03}$ & $.345_{\pm .07}$ \\

& EM-EDM$_w$ 
& $.585_{\pm .07}$ & $.558_{\pm .04}$ & $.523_{\pm .02}$ & $.430_{\pm .08}$ 
& $.619_{\pm .06}$ & $.559_{\pm .05}$ & $.523_{\pm .02}$ & $.468_{\pm .07}$ 
& $.738_{\pm .06}$ & $.745_{\pm .05}$ & $.680_{\pm .05}$ & $.589_{\pm .07}$ \\

& HALIDE 
& $\mathbf{.627}_{\pm .04}$ & $\mathbf{.700}_{\pm .02}$ & $\mathbf{.595}_{\pm .01}$ & $\mathbf{.463}_{\pm .05}$ 
& $\mathbf{.678}_{\pm .06}$ & $\mathbf{.632}^{*}_{\pm .04}$ & $\mathbf{.559}^{*}_{\pm .02}$ & $\mathbf{.530}_{\pm .07}$ 
& $\mathbf{.869}^{*}_{\pm .13}$ & $\mathbf{.874}^{*}_{\pm .11}$ & $\mathbf{.839}^{*}_{\pm .13}$ & $\mathbf{.790}^{*}_{\pm .18}$ \\ \hline

\end{tabular}%
}
\end{table*}

\section{Discussion, Caveats,  and Conclusions}
\label{sec:discussion-ch5}

This work addresses a key limitation of AL in human-centric decision-making settings: demonstrations are inherently heterogeneous in quality, even among high-performing demonstrators, whether the decision-maker is a student choosing a pedagogical action or a clinician choosing a treatment. Students differ in prior knowledge, self-regulation, and metacognitive skills, and their decision quality fluctuates over time; clinicians likewise vary in experience and judgment under uncertainty, and their treatment choices vary in effectiveness even within the same case. Both settings violate the standard AL assumption that expert demonstrations are uniformly reliable. Our results show that treating all demonstrations equally injects noise into policy induction, limiting robustness—especially under cross-semester distribution shift in education and across clinical sites in healthcare.

HALIDE operationalizes a principle long established in mixed-initiative ITS research—student agency must be balanced with selective modulation of influence—and shows that the same principle holds for clinician-driven treatment decisions. Rather than discarding imperfect trajectories or uniformly weighting demonstrations, HALIDE integrates continuous demonstration quality signals within a hierarchical apprenticeship learning framework. This allows useful routine decisions from sub-optimal trajectories to contribute while attenuating the influence of low-impact or erroneous actions. Empirically, this integration is critical in both domains: unweighted hierarchical learning fails to benefit from additional heterogeneous data, whereas HALIDE consistently improves performance, with the largest gains observed when leveraging both expert and sub-optimal demonstrations---on ITS data (E+I) and, in the clinical setting, on the largest and most heterogeneous dataset (MIMIC, all-quality data). These results highlight the complementary roles of ranking and hierarchy, where quality-aware weighting stabilizes low-level policy induction and hierarchical structure propagates these improvements to higher-level decision-making strategies, regardless of whether those strategies govern a tutoring session or a course of treatment.


Despite these contributions, this work has several limitations. First, demonstration quality is estimated at the trajectory level and propagated across steps, which provides scalability but only an approximate proxy for step-level effectiveness, in both the pedagogical and clinical settings. Second, while we validate HALIDE across two human-centric domains, all evaluations remain offline, leaving open questions regarding performance under closed-loop deployment; the clinical evaluation is also limited to septic shock prevention, and the educational evaluation to a single ITS, so generalization to other conditions and curricula remains to be tested. Third, the quality ranking mechanism may reflect biases present in historical data—such as differences in student prior preparation and engagement, or in clinician experience and case mix—which could influence policy induction if not carefully monitored.
More broadly, HALIDE relies on the availability of a relative quality ranking to modulate the influence of heterogeneous demonstrations. Although ranking is substantially weaker and easier to obtain than specifying a full reward function, its quality can affect downstream learning. Noisy or misaligned rankings—such as those correlated with short-term performance rather than true pedagogical or clinical effectiveness—may attenuate the benefits of quality-aware weighting. This limitation highlights the importance of careful ranking design and validation when applying HALIDE in practice. 
Despite these caveats, our findings demonstrate that modeling demonstration quality as a graded structural signal—rather than a binary selection criterion—is essential for robust apprenticeship learning under imperfect human agency, whether that agency belongs to a student or a clinician. By combining quality-aware weighting with hierarchical modeling and evolving rewards, HALIDE offers a principled path toward deploying AL in realistic, heterogeneous human-centric environments spanning education and healthcare.

\chapter{Critical Contrast Hierarchical Apprenticeship Learning}
\label{chap:ccal}

\section{Overview}
While apprenticeship learning (AL) has shown promise for inducing effective
decision-making policies directly from human demonstrations in real-world
human-centric environments, most existing approaches assume the demonstrations
are optimal or near-optimal under a single, fixed reward.
Real-world demonstrations, however, are rarely so well-behaved: they are
\emph{imperfect}, mixing correct and mistaken decisions within the same
trajectory, and \emph{inconsistent}, since a demonstrator's underlying
intentions evolve over the course of a task rather than staying fixed.
Compounding this, not all decisions carry equal weight: some occur at routine
states where a wide range of actions succeed, while others occur at critical
junctures where the choice made is pivotal to the outcome.
Existing AL methods address at most one of these gaps at a time, either
treating all timesteps uniformly or weighting entire trajectories only at the
demonstration level, leaving imperfection, evolving intent, and decision-level
criticality unaccounted for jointly.

\begin{sloppypar}
To close these gaps together, we introduce \cchal{} (\textbf{C}ritical
\textbf{C}ontrast \textbf{H}ierarchical \textbf{A}pprenticeship
\textbf{L}earning), a unified framework that builds on existing hierarchical
sub-trajectory clustering and trajectory-level outcome ranking and closes
both remaining gaps jointly: \cchal{} redistributes each trajectory's
outcome-quality rank to individual decisions via Bayesian inference in
feature-action space, and separately scores each decision's criticality by
contrasting expert against non-expert behavior in a dual-arm design---then
combines both signals into a joint per-decision weight that drives weighted
policy induction and sharpens reward learning through a Bradley-Terry
pairwise contrastive IRL objective over observed failure trajectories.
\end{sloppypar}

We evaluate \cchal{} in two challenging human-centric domains: septic shock
prevention across three clinical EHR datasets (CCHS, Mayo Clinic, MIMIC-III)
and student modeling on an intelligent tutoring system (ITS) dataset.
\cchal{} outperforms every measured baseline on every reported metric on CCHS,
Mayo, and ITS, and is statistically indistinguishable from the strongest prior
baseline on F1 and AUC on MIMIC-III while still exceeding it on APR and
Jaccard---demonstrating that \cchal{} can accurately induce policies from
imperfect, evolving demonstrations, outperforming competitive baselines and
ablations across a broad range of complex, real-world human-centric tasks.

\section{Introduction}

Apprenticeship learning (AL) aims to induce decision-making policies by
observing and imitating human demonstrations, offering a data-driven
alternative to costly or dangerous online trial-and-error in real-world
human-centric domains such as healthcare and education~\citep{abbeel2004apprenticeship,levine2020offline}.
In healthcare, this problem is especially acute: septic shock kills 270,000
patients per year in the United States alone~\citep{singer2016sepsis}, and
clinicians must repeatedly make timely, high-stakes treatment decisions,
making offline apprenticeship learning from electronic health records (EHR) an
attractive route---observe clinician behavior in retrospective records and
recover a policy that imitates the best of it.
But real human demonstrations are far from the ideal expert assumed by
classical apprenticeship learning, and depart from it in two compounding ways.

\paragraph{Demonstrations are imperfect.}
A real clinical dataset is a mixture, where some clinicians intervene
optimally at the right moments while others act suboptimally, are late, or are
inconsistent, so even trajectories labeled ``positive'' (e.g., shock patients
who survived) vary substantially in outcome quality~\citep{wu2019imitation,tangkaratt2021robust}.

\paragraph{Demonstrations are inconsistent.}
A clinician's intent is not fixed over the course of a case but evolves as the
patient's condition progresses, so a single trajectory may reflect several
locally distinct decision-making regimes rather than one stationary policy.
Existing offline apprenticeship learning methods for this task, including
\themes{}~\citep{themes2024}, address the second challenge by hierarchically
partitioning each demonstration into sub-trajectory clusters that each capture
a locally consistent regime (Section~\ref{sec:ticc}), and address the first,
at the trajectory level, by quantifying demonstration quality through a
calibrated outcome predictor that produces a scalar rank per trajectory.

Despite this progress, two major limitations remain within this hierarchical
structure.
First, trajectory-level ranks are applied uniformly across all decisions
within each demonstration, even though most decisions occur at stable,
routine states where any reasonable action leads to similar outcomes, while a
minority occur at critical junctures where the treatment choice is pivotal to
whether the patient progresses toward or away from shock---and real clinical
practice is not free of unnecessary intervention at exactly these routine
states: physicians themselves estimate that roughly one-fifth of U.S.\ medical
care is unnecessary~\citep{lyu2017overtreatment}, and unnecessarily
broad-spectrum antibiotics given to septic patients who do not need them are
independently associated with a $22\%$ increase in
mortality~\citep{rhee2020prevalence}.
To address this, we introduce an outcome-attributed decision weighting that
redistributes the trajectory-level rank to individual decisions via Bayesian
inference in feature-action space (Section~\ref{sec:gp_credit}).
Second, no existing quality-weighting method identifies which decisions are
behaviorally critical, i.e., where expert and non-expert clinicians choose
differently, as opposed to routine, where any policy would do about as well.
\citet{ju2021identifying} identify such states via Q-value spread under a
trained policy, but this requires a converged model and cannot exploit the
contrast between expert and failure demonstrations directly.
To address this, we introduce a dual-arm contrastive criticality score
computed directly from observed behavioral divergence between successful and
failed trajectories, requiring no trained policy (Section~\ref{sec:criticality}).

\paragraph{Decision importance has two complementary dimensions.}
The question of which decisions to attend to in apprenticeship learning
therefore has two orthogonal answers, and effective learning from imperfect
demonstrations requires both:
\begin{itemize}
  \item \textbf{Outcome attribution}: How much did decision $t$ contribute to
  trajectory $i$'s observed outcome quality?
  A decision at a stable state, even in an expert trajectory, contributes
  little; a decision at a physiological inflection point may determine the
  outcome.
  \item \textbf{Decision criticality}: How sharply does decision $t$ separate
  expert from non-expert clinical behavior at this state?
  A decision that both groups make identically---however it affects the
  outcome---provides no contrastive learning signal.
\end{itemize}
These are not the same signal: a decision can be outcome-attributable but not
critical (an expert maneuver at a state where non-experts happen to act
similarly), or critical but from a low-quality trajectory (a pivotal choice
made incorrectly).
No prior framework answers both questions jointly: trajectory-level
quality-weighting methods~\citep{wu2019imitation,tangkaratt2021robust,themes2024}
answer only the first, and Q-value-based critical-state
methods~\citep{ju2021identifying,roth2019conservative,eysenbach2019search}
answer only the second, and only when a trained policy is already available.

We propose Critical Contrast Hierarchical Apprenticeship Learning (\cchal{}),
a unified framework that, building on \themes{}'s hierarchical sub-trajectory
clustering, answers both questions and combines them into a single joint
per-decision attention weight (Section~\ref{sec:joint_weight}), which in turn
scales a Bradley-Terry pairwise contrastive IRL objective that sharpens reward
learning at the decision points where expert-versus-failure contrast is
strongest (Section~\ref{sec:contrastive_irl}).
To the best of our knowledge, \cchal{} is the first apprenticeship learning
framework to jointly attribute demonstration quality to individual decisions
and score their behavioral criticality without a trained policy or an
environment reward signal---only the observable contrast between successful
and failed demonstrations, and a trajectory-level demonstration rank derived
from a calibrated outcome predictor.
Setting criticality to uniform and GP credit to the trajectory-level rank
recovers a trajectory-level-only special case, making our ablations directly
interpretable (Section~\ref{sec:ablation_analysis}).
We evaluate \cchal{} across two human-centric domains, following \themes{}'s
original healthcare-and-education scope~\citep{themes2024}: septic shock
prevention on three clinical EHR datasets and student modeling on one
intelligent tutoring system (ITS) dataset.
The same two-dimensional decision importance holds in both: just as clinical
decisions range from routine monitoring to life-or-death inflection points,
tutoring decisions range from routine problem steps to moments that determine
whether a struggling student disengages or a high performer plateaus.

Our contributions are:
\begin{sloppypar}
\begin{itemize}
  \item We introduce outcome-attributed decision weighting
  (Section~\ref{sec:gp_credit}), repurposing the Bayesian Reward Distribution
  (BRD) framework~\citep{azizsoltani2019unobserved} in feature-action space so
  that clinically similar state-action pairs receive proportional attention
  regardless of temporal position.

  \item We introduce dual-arm contrastive criticality scoring
  (Section~\ref{sec:criticality}), which estimates local action distributions
  via $k$-nearest neighbours in a stratum-matched design---contrasting
  $\mathcal{Q}^{+}_{\mathrm{hi}}$ vs.\ $\mathcal{Q}^{-}_{\mathrm{hi}}$ and $\mathcal{Q}^{+}_{\mathrm{lo}}$ vs.\
  $\mathcal{Q}^{-}_{\mathrm{lo}}$---so that criticality is measured against the same
  initial risk stratum, requiring no trained policy, cluster label, or MDP
  model.

  \item We combine both into a joint decision-attention weight
  (Section~\ref{sec:joint_weight}) that scales a Bradley-Terry pairwise
  contrastive IRL objective (Section~\ref{sec:contrastive_irl}), using
  observed failure trajectories as negatives to sharpen reward learning at
  the decision points where expert-versus-failure contrast is strongest.

  \item We evaluate \cchal{} against seven baselines, including \themes{},
  its direct methodological predecessor, on three clinical EHR datasets for
  septic shock prevention and one intelligent tutoring system (ITS) dataset
  for student modeling, showing that the outcome-attribution-plus-criticality
  principle is not specific to septic shock.
\end{itemize}
\end{sloppypar}

\section{Related Work}

\paragraph{Apprenticeship Learning and Imitation from Imperfect Demonstrations.}
Classical apprenticeship learning~\citep{abbeel2004apprenticeship} recovers a
reward function that makes expert demonstrations near-optimal, assuming
demonstrations come from a single, uniformly expert source.
Maximum Entropy IRL~\citep{ziebart2008maximum} and its deep extensions
(AIRL~\citep{fu2018learning}, GAIL~\citep{ho2016generative}) share this
assumption.
SQIL~\citep{reddy2020sqil} and IQL~\citep{kostrikov2022offline} operate
offline but treat all demonstration timesteps uniformly.
T-REX~\citep{brown2019extrapolating} relaxes the expert assumption by
learning a reward consistent with a ranking over demonstrations, and
DemoDICE~\citep{kim2021demodice} handles imperfect demonstrations offline via
divergence minimization.
Both work at the trajectory level and do not assign differential importance
to individual decisions within a demonstration.

\paragraph{Demonstration Quality in IRL.}
\begin{sloppypar}
Several works weight demonstrations by quality: \citet{wu2019imitation} use a
confidence-weighted BC objective; \citet{sasaki2021behavioral} filter
low-quality demos; \citet{tangkaratt2021robust} model the demonstrator's
expertise level.
Closest to ours, \themes{} \citep{themes2024} uses trajectory-level quality
from an outcome predictor but applies it uniformly across all timesteps of
each demonstration, conflating within-trajectory temporal heterogeneity.
\halide{}~\citep{islam2026hierarchical} introduces a calibrated
demonstration-rank framework in the same spirit, validated on intelligent
tutoring systems; \cchal{} adapts this rank mechanism to the clinical domain
(Section~\ref{sec:gp_credit}) and, unlike \halide{}, decomposes it within a
trajectory and contrasts expert against non-expert behavior rather than
applying it uniformly.
\end{sloppypar}

\paragraph{Outcome Attribution in Demonstrations.}
Temporal credit assignment in RL asks how to distribute environment reward to
timesteps within a trajectory~\citep{sutton2018reinforcement}.
In apprenticeship learning the analogous question is distinct: given a
trajectory's \emph{observed outcome quality} (not an environment reward),
which individual decisions are most responsible for it?
InferGP~\citep{azizsoltani2019unobserved} uses GP inference to distribute
delayed environment rewards, and InferNet~\citep{ausin2021infernet} extends
this via neural estimation for ShortTR computation in Critical-DRL.
Neither work applies outcome attribution to \emph{demonstration quality} in
an imitation learning setting.
\cchal{} repurposes the GP inference mechanism for this purpose: treating
trajectory quality as the observation and feature-action similarity as the
kernel structure naturally produces per-decision weights that reflect each
decision's contribution to the observed outcome---without requiring any
environment reward signal.

\paragraph{Critical State Identification.}
\citet{ju2021identifying} define critical states via Q-value spread (LongTR)
and inferred immediate rewards (ShortTR), requiring a trained DQN.
Decision Points~\citep{roth2019conservative} use value function uncertainty.
Bottleneck states~\citep{eysenbach2019search} exploit graph connectivity.
\cchal{} defines criticality without any trained policy, using only the
observed behavioral contrast between success and failure trajectories.

\paragraph{Overtreatment as a Motivation for Criticality-Aware Learning.}
A separate clinical literature documents that a substantial share of
intervention is medically unnecessary: physicians estimate roughly one-fifth
of U.S.\ care is overtreatment, most often driven by diagnostic uncertainty
and defensive practice rather than patient need~\citep{lyu2017overtreatment}.
In sepsis specifically, unnecessarily broad-spectrum empiric
antibiotics---prescribed to patients who do not harbor resistant
organisms---are independently associated with higher
mortality~\citep{rhee2020prevalence}, and in the ICU, structured protocols
that help clinicians identify which patients genuinely warrant continued
escalation reduce invasive treatment and length of stay without harming
outcomes~\citep{chang2021timelimited}, indicating that a meaningful share of
aggressive intervention is separable from patient benefit.
This literature has not, to our knowledge, previously been connected to
critical-state identification in apprenticeship learning: prior
work~\citep{ju2021identifying,roth2019conservative,eysenbach2019search}
motivates criticality only as a way to avoid missing pivotal decisions, i.e.,
only against \emph{undertreatment}.
\cchal{}'s criticality score is low both when experts and non-experts act
identically at a stable state and, by construction, at states where
behavioral contrast is weak for any reason---including states where
aggressive intervention is common in the demonstrations but not clearly
linked to outcome.
Because the joint weight (Eq.~\eqref{eq:joint_weight}) concentrates learning
on high-criticality decisions and correspondingly de-emphasizes
low-criticality ones, \cchal{} is less prone than uniformly-weighted
baselines to reproducing whatever unnecessary intervention appears in the
demonstrations at routine states, offering a plausible mechanism against
overtreatment alongside its primary role of concentrating signal where it is
outcome-relevant; verifying this directly would require action-level
overtreatment labels beyond the shock/non-shock outcome data used here.

\paragraph{Contrastive and Adversarial IRL.}
AIRL~\citep{fu2018learning} and f-IRL~\citep{ni2021f} use adversarial
training between expert and generated trajectories, but in an online
setting.
\citet{kim2021demodice} use offline contrastive divergence minimization.
\cchal{} differs by using \emph{observed} failure demonstrations (not
generated ones) as negatives, and scaling the contrastive signal by
per-state criticality.

\paragraph{Clinical RL.}
\citet{raghu2017deep} apply DQN to sepsis treatment; \citet{komorowski2018artificial}
use a continuous-state approach.
\themes{}~\citep{themes2024} introduces hierarchical offline apprenticeship
learning for EHR with trajectory-level quality weighting; \cchal{} extends
this line by adding outcome-attributed per-decision weighting, contrastive
criticality scoring, and Bradley-Terry pairwise IRL---moving from
trajectory-level attention to decision-level attention.

\section{Preliminaries}
\label{sec:preliminaries}

\paragraph{Problem Setting.}
Let $\mathcal{S}$ be a state space and $\mathcal{A}$ a discrete action space.
We observe a dataset $\mathcal{D} = \{\tau_i\}_{i=1}^{N}$ of trajectories
collected from clinical practice, where
$\tau_i = (s_0^i, a_0^i, s_1^i, a_1^i, \ldots, s_{n_i}^i)$.
Each trajectory $\tau_i$ has an associated \emph{delayed outcome}
$y_i \in \{0,1\}$ (1 = septic shock, 0 = no shock) and a demonstration rank
$q_i \in (0,1)$, higher values indicating higher-quality behavior leading to
non-shock outcomes.

\begin{sloppypar}
Trajectories are partitioned along two independent binary axes. The first, a
\emph{risk stratum} (hi/lo), is the patient's state before treatment---at
risk of shock, or not. The second is the eventual outcome $y_i$---did the
patient overcome that risk (or simply remain stable, if not at risk), or did
they stay at risk and progress to shock. Crossing the two yields four groups
(Table~\ref{tab:quadrants}): a \emph{target population}, whose outcome was
favorable, which \cchal{} tries to emulate; and a \emph{reference
population}, whose outcome was shock, used only as a contrast baseline.
\end{sloppypar}

\begin{table}[t]
  \centering
  \footnotesize
  \caption{Notation for the four demonstration groups.}
  \label{tab:quadrants}
  \begin{tabular}{@{}l c c@{}}
    \toprule
     & \textbf{Target} ($+$) & \textbf{Reference} ($-$) \\
    \midrule
    \textbf{Hi-risk} (at risk pre-treatment) & $\mathcal{Q}^{+}_{\mathrm{hi}}$ overcame risk & $\mathcal{Q}^{-}_{\mathrm{hi}}$ stayed at risk, shock \\
    \textbf{Lo-risk} (not at risk pre-treatment) & $\mathcal{Q}^{+}_{\mathrm{lo}}$ remained stable & $\mathcal{Q}^{-}_{\mathrm{lo}}$ deteriorated, shock \\
    \bottomrule
  \end{tabular}
\end{table}

\paragraph{MT-TICC State Clustering.}
Multi-task TICC~\citep{hallac2017toeplitz} segments each trajectory into a
sequence of discrete state labels $k_t^i \in \{0,\ldots,K-1\}$, where $K$ is
the number of temporal state clusters (default $K=9$). This provides a shared
state representation across patients.

\paragraph{EM-EDM Policy Learning.}
The EM-EDM module~\citep{themes2024} jointly assigns trajectory sub-segments
to behavior clusters and trains a per-cluster behavior model (EDM student),
iterating between cluster assignment (E-step) and model update (M-step). The
log-likelihood objective is:
\begin{equation}
  \mathcal{L}_{\mathrm{EDM}} = \sum_{\tau_i} \sum_t \log P(a_t^i \mid s_t^i;\, \theta_{z_t^i})
  \label{eq:edm}
\end{equation}
where $z_t^i$ is the cluster assignment for timestep $t$ of trajectory $i$.

\paragraph{MLIRL Reward Learning.}
The IRL phase uses maximum-likelihood IRL (MLIRL)~\citep{babes2011apprenticeship}
to recover a linear reward function $R(s,a) = \theta^\top \phi(s,a)$ that
maximizes the Boltzmann policy likelihood of the expert demonstrations under
the induced MDP.

\section{The \cchal{} Framework}

\begin{figure*}[t]
  \centering
  \includegraphics[width=\textwidth,alt={CCHAL overview diagram with four phases. Top left: trajectory-level
  demonstration ranks are distributed to per-timestep weights via InferGP
  Bayesian credit assignment in feature-action space. Top right: dual-arm
  contrastive criticality scores are estimated via kNN Jensen-Shannon
  divergence; the high-risk arm contrasts the hi-risk target population, who
  overcame their pre-treatment shock risk, against the hi-risk reference
  population, who progressed to shock, and the low-risk arm contrasts the
  lo-risk target population, who remained stable, against the lo-risk
  reference population, who deteriorated into shock. Bottom left: weighted
  EM-EDM policy learning uses only the target populations from both arms
  with joint attention weights combining GP credit and criticality. Bottom
  right: Bradley-Terry pairwise contrastive IRL uses the target populations
  as positive demonstrations versus the reference populations as negative
  demonstrations, scaled by arm-routed action-aware weights, to learn a
  reward function that prioritizes critical expert decisions.}]{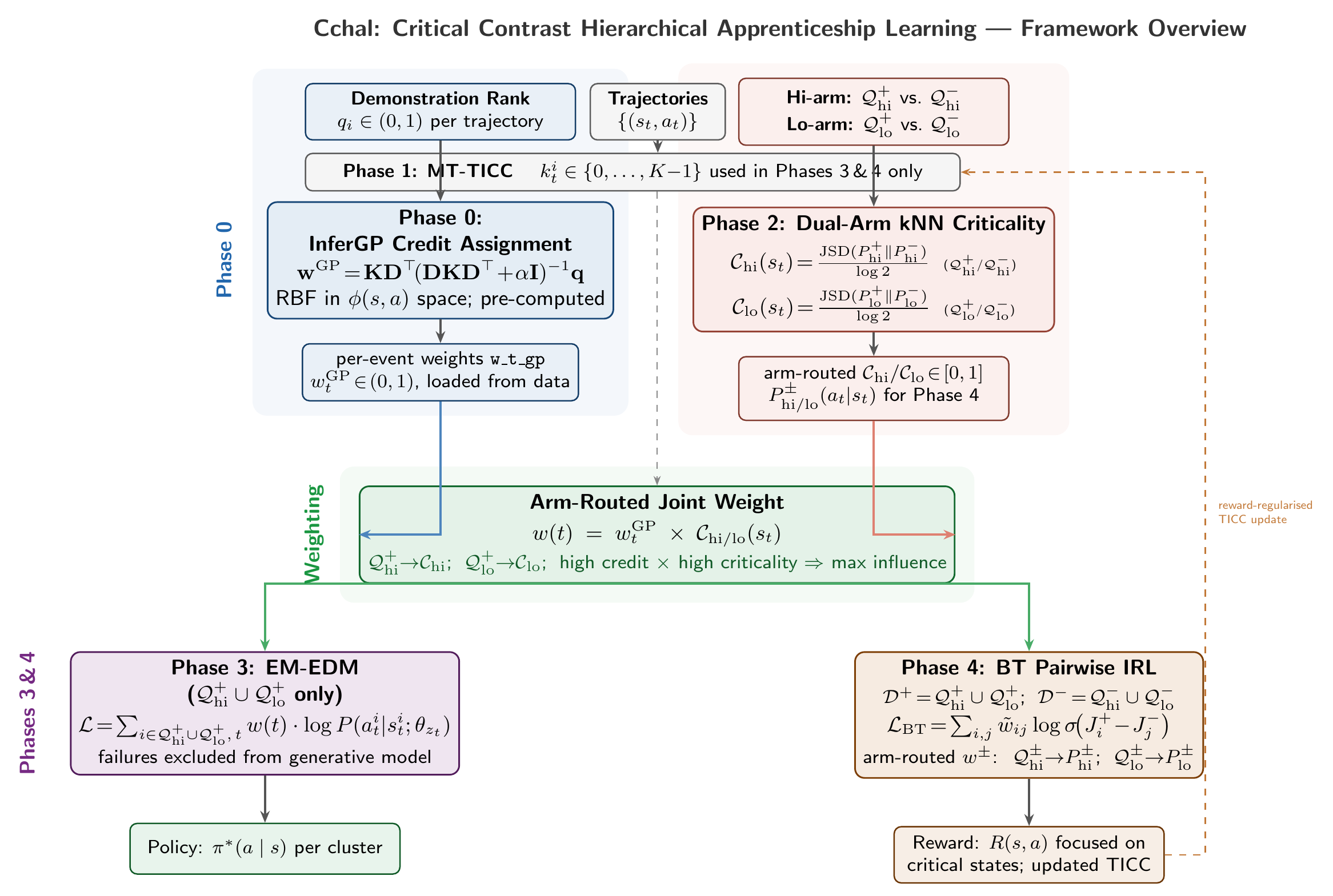}
  \caption{
    \textbf{\cchal{} overview.} Trajectory-level demonstration ranks are
    distributed to per-timestep weights via InferGP Bayesian credit
    assignment in feature-action space (top left; pre-computed, stored as
    \texttt{w\_t\_gp}).
    Dual-arm contrastive criticality scores are estimated from kNN local JSD
    in clinical feature space (top right; built once per repeat): the
    high-risk arm contrasts $\mathcal{Q}^{+}_{\mathrm{hi}}$ vs.\ $\mathcal{Q}^{-}_{\mathrm{hi}}$ and the
    low-risk arm contrasts $\mathcal{Q}^{+}_{\mathrm{lo}}$ vs.\ $\mathcal{Q}^{-}_{\mathrm{lo}}$, each
    anchored to the same initial risk stratum. Phase~3 weighted EM-EDM uses
    $\mathcal{Q}^{+}_{\mathrm{hi}} \cup \mathcal{Q}^{+}_{\mathrm{lo}}$ only (failures excluded from the
    generative model).
    Phase~4 BT-IRL uses $\mathcal{D}^+ = \mathcal{Q}^{+}_{\mathrm{hi}} \cup \mathcal{Q}^{+}_{\mathrm{lo}}$
    and $\mathcal{D}^- = \mathcal{Q}^{-}_{\mathrm{hi}} \cup \mathcal{Q}^{-}_{\mathrm{lo}}$ with
    arm-routed action-aware weights (bottom).
  }
  \label{fig:overview-ch6}
\end{figure*}

Figure~\ref{fig:overview-ch6} presents an overview of \cchal{}, with further
details provided in Algorithm~\ref{alg:ccal}.
\cchal{} proceeds in five phases: \emph{Phase 0} distributes trajectory-level
demonstration ranks to per-decision outcome-attribution weights via InferGP
Bayesian credit assignment in feature-action space
(Section~\ref{sec:gp_credit}); \emph{Phase 1} partitions each demonstration
into sub-trajectory clusters via MT-TICC (Section~\ref{sec:ticc}); \emph{Phase
2} scores each decision's criticality via dual-arm contrastive kNN-JSD
(Section~\ref{sec:criticality}); \emph{Phase 3} combines both signals into a
joint per-decision weight that drives weighted EM-EDM policy induction
(Section~\ref{sec:joint_weight}); and \emph{Phase 4} uses the same joint
weight to scale a Bradley-Terry pairwise contrastive IRL objective for reward
learning (Section~\ref{sec:contrastive_irl}).

\subsection{Phase 0: Outcome-Attributed Decision Weighting}
\label{sec:gp_credit}

Given that trajectory $i$ achieved outcome quality $q_i$, we ask: \emph{which
individual decisions within the trajectory contributed most to that quality?}
This is \emph{outcome attribution}---a question about decision importance in
apprenticeship learning, distinct from RL temporal credit assignment which
requires an environment reward signal.
We answer it by repurposing the Bayesian Reward Distribution (BRD) framework
of \citet{azizsoltani2019unobserved}: treating $q_i$ as the observation and an
RBF kernel over feature-action vectors $\phi(s,a)$ as the prior, GP inference
distributes $q_i$ across the trajectory's decisions in proportion to their
feature-action similarity to the trajectory's high-quality moments.
Crucially, the kernel operates in \emph{feature-action space}: two decisions
at clinically similar state-action configurations receive similar
attribution regardless of when they occur in a trajectory.
The resulting per-decision weights are computed once offline and stored as
the column \texttt{w\_t\_gp} in the EHR data file.

\paragraph{Feature-action representation.}
For each event $(s_t^i, a_t^i)$ we form a feature-action vector
$\phi(s_t^i, a_t^i) \in \mathbb{R}^{2F}$ by concatenating the $F$-dimensional
clinical state features with a one-hot action embedding.

\paragraph{RBF kernel and mean-constraint observation model.}
We place a GP prior over the per-timestep quality function $r: \phi \mapsto \mathbb{R}$
with an RBF kernel:
\begin{equation}
  K(\phi, \phi') = \exp\!\left(-\frac{\|\phi - \phi'\|^2}{2\ell^2}\right).
  \label{eq:rbf_phi}
\end{equation}
The trajectory-level rank is linked to per-timestep weights via a
\emph{length-invariant mean constraint}:
\begin{equation}
  q_i = \frac{1}{n_i}\sum_{t \in \tau_i} r_t + \varepsilon_i,
  \qquad \varepsilon_i \sim \mathcal{N}(0,\, \alpha).
  \label{eq:mean_constraint}
\end{equation}
Equivalently, $q_i = \mathbf{D}_i \mathbf{r} + \varepsilon_i$ where
$D_i[t] = 1/n_i$ if $t \in \tau_i$ and $0$ otherwise. Using a mean rather than
a discounted sum makes the observation model length-invariant: short and long
trajectories contribute on the same scale.

\paragraph{Joint posterior inference.}
Writing $\ell$ for the total number of events across all trajectories and $m$
for the number of trajectories, the BRD posterior mean over all events is:
\begin{equation}
  \mathbf{w}^{\mathrm{GP}}_{\mathrm{raw}}
  = \mathbf{K}\mathbf{D}^\top
    \underbrace{\left(\mathbf{D}\mathbf{K}\mathbf{D}^\top
                + \alpha\mathbf{I}_m\right)^{-1}}_{\text{$m{\times}m$, Cholesky}}
    \mathbf{q}
  \label{eq:brd}
\end{equation}
where $\mathbf{K}\mathbf{D}^\top \in \mathbb{R}^{\ell \times m}$ is built in
trajectory-sized chunks (column $j$ = $n_j^{-1}\sum_{t\in\tau_j} K(\cdot, \phi_t^j)$)
without materialising the full $\ell\times\ell$ kernel, keeping peak memory at
$O(\ell m)$. The $m\times m$ system is solved by Cholesky factorisation.

\paragraph{Normalisation.}
Raw GP values are z-scored and passed through a sigmoid:
\begin{equation}
  \gpcredit{t}
  = \sigma\!\left(\frac{w^{\mathrm{GP}}_{\mathrm{raw},t} - \mu_{\mathrm{raw}}}
                        {\sigma_{\mathrm{raw}}}\right) \in (0,1)
  \label{eq:sigmoid_norm}
\end{equation}
ensuring weights are bounded and comparable across trajectories and datasets.
Inference is joint over \emph{all} training trajectories ($\mathcal{Q}^{+}_{\mathrm{hi}}$,
$\mathcal{Q}^{+}_{\mathrm{lo}}$, $\mathcal{Q}^{-}_{\mathrm{hi}}$, and $\mathcal{Q}^{-}_{\mathrm{lo}}$ when
available), so $r(s, a_{\mathrm{pos}})$ is naturally higher than
$r(s, a_{\mathrm{neg}})$ at contrastive states---a built-in property of the
feature-action kernel that would be absent from a temporal kernel.

\subsection{Phase 1: MT-TICC Temporal State Clustering}
\label{sec:ticc}

MT-TICC is run on training trajectories only (inductive mode) to assign each
timestep to a discrete state cluster $k_t \in \{0,\ldots,K-1\}$. This
clustering step is unchanged from standard inductive MT-TICC practice and is
not further modified by \cchal{}.

\subsection{Phase 2: Dual-Arm Contrastive Critical State Scoring}
\label{sec:criticality}

Rather than computing a single criticality score per MT-TICC cluster,
\cchal{} estimates \emph{continuous per-timestep} criticality scores directly
in clinical feature space via $k$-nearest neighbours. Crucially, the
contrasting populations are \emph{stratum-matched}: we build two independent
arms whose positive and negative pools share the same initial clinical risk
profile (pre-shock vs.\ non-shock), preventing cross-stratum JSD from
conflating physiologically distinct populations.

\paragraph{High-risk arm ($\mathcal{Q}^{+}_{\mathrm{hi}}$ vs.\ $\mathcal{Q}^{-}_{\mathrm{hi}}$).}
Among patients who were at risk of septic shock, we contrast those who
recovered ($\mathcal{Q}^{+}_{\mathrm{hi}}$, the expert pool) against those who did not
($\mathcal{Q}^{-}_{\mathrm{hi}}$, the failure pool).
For each training timestep $t$ with feature vector $s_t \in \mathbb{R}^F$:
\begin{equation}
  \mathcal{N}^+_{\mathrm{hi}}(s_t) = k\text{-NN of } s_t \text{ in }
    \bigl\{s_{t'}^i : i \in \mathcal{Q}^{+}_{\mathrm{hi}}\bigr\}, \quad
  \mathcal{N}^-_{\mathrm{hi}}(s_t) = k\text{-NN of } s_t \text{ in }
    \bigl\{s_{t'}^i : i \in \mathcal{Q}^{-}_{\mathrm{hi}}\bigr\}.
  \label{eq:knn_hi}
\end{equation}
Local action distributions $P^+_{\mathrm{hi}}(a \mid s_t)$ and $P^-_{\mathrm{hi}}(a \mid s_t)$
are estimated with Laplace smoothing:
\begin{equation}
  P^+_{\mathrm{hi}}(a \mid s_t) \;\propto\;
    \#\{t' \in \mathcal{N}^+_{\mathrm{hi}}(s_t) : a_{t'} = a\} + 1,
  \label{eq:p_pos_hi}
\end{equation}
and analogously for $P^-_{\mathrm{hi}}$.
The high-risk criticality score is:
\begin{equation}
  \crithi{s_t}
  = \frac{\JSD\!\bigl(P^+_{\mathrm{hi}}(\cdot \mid s_t) \;\|\; P^-_{\mathrm{hi}}(\cdot \mid s_t)\bigr)}{\log 2}
  \in [0,1].
  \label{eq:criticality_hi}
\end{equation}

\paragraph{Low-risk arm ($\mathcal{Q}^{+}_{\mathrm{lo}}$ vs.\ $\mathcal{Q}^{-}_{\mathrm{lo}}$).}
Among patients who were \emph{not} initially at shock risk, we contrast those
who remained stable ($\mathcal{Q}^{+}_{\mathrm{lo}}$) against those who deteriorated into
shock ($\mathcal{Q}^{-}_{\mathrm{lo}}$). These two populations share the same non-shock
starting condition, so their behavioral divergence measures \emph{de novo}
risk escalation.
When both $\mathcal{Q}^{+}_{\mathrm{lo}}$ and $\mathcal{Q}^{-}_{\mathrm{lo}}$ are present in the
training fold:
\begin{equation}
  \mathcal{N}^+_{\mathrm{lo}}(s_t) = k\text{-NN of } s_t \text{ in }
    \bigl\{s_{t'}^i : i \in \mathcal{Q}^{+}_{\mathrm{lo}}\bigr\}, \quad
  \mathcal{N}^-_{\mathrm{lo}}(s_t) = k\text{-NN of } s_t \text{ in }
    \bigl\{s_{t'}^i : i \in \mathcal{Q}^{-}_{\mathrm{lo}}\bigr\}.
  \label{eq:knn_lo}
\end{equation}
The low-risk criticality score $\critlo{s_t}$ is defined analogously to
Eq.~\eqref{eq:criticality_hi}.

\paragraph{Arm routing.}
Each trajectory receives scores from the arm that matches its initial
stratum: $\mathcal{Q}^{+}_{\mathrm{hi}}$ trajectories use $\crithi{\cdot}$; $\mathcal{Q}^{+}_{\mathrm{lo}}$
trajectories use $\critlo{\cdot}$.
This ensures the criticality signal captures within-stratum behavioral
divergence, not a mixture of physiologically distinct starting conditions.
When the low-risk arm is inactive (only $\mathcal{Q}^{+}_{\mathrm{hi}}$/$\mathcal{Q}^{-}_{\mathrm{hi}}$
data available), \cchal{} falls back to a single $\crithi{\cdot}$ arm.

\paragraph{Sparse support and normalisation.}
For timesteps with fewer than $m_{\min}$ neighbours in either pool, we
substitute the dataset-mean score for that arm ($\bar{\mathcal{C}}_{\mathrm{hi}}$
or $\bar{\mathcal{C}}_{\mathrm{lo}}$). Within each arm, scores are
independently rescaled so that the per-arm maximum equals 1.

Both sets of kNN indices are built \emph{once} per repeat before the outer EM
loop, independent of MT-TICC cluster labels. Criticality therefore does not
change across EM iterations, eliminating the instability that cluster-level
schemes exhibit when TICC labels shift.

\paragraph{Action-aware local probabilities.}
Each arm also yields per-timestep action probability estimates used in
Phase~4: $P^+_{\mathrm{hi}}(a_t \mid s_t)$, $P^-_{\mathrm{hi}}(a_t \mid s_t)$,
$P^+_{\mathrm{lo}}(a_t \mid s_t)$, and $P^-_{\mathrm{lo}}(a_t \mid s_t)$---the
local probability that a demonstrator from the respective group would choose
the action $a_t$ actually taken. These serve as arm-routed, action-aware
scaling factors in the BT-IRL objective.

\subsection{Phase 3: Joint Critical-Quality Weighting and EM-EDM}
\label{sec:joint_weight}

\paragraph{Data selection for EM-EDM.}
EM-EDM is restricted to the \emph{positive} quadrants
$\mathcal{Q}^{+}_{\mathrm{hi}} \cup \mathcal{Q}^{+}_{\mathrm{lo}}$.
Trajectories from $\mathcal{Q}^{-}_{\mathrm{hi}}$ and $\mathcal{Q}^{-}_{\mathrm{lo}}$ are deliberately
excluded: including failure-mode demonstrations in a generative behavior
model risks contaminating behavior clusters with anti-expert action patterns
that the downstream IRL objective cannot fully disentangle at the
cluster-granularity level.
$\mathcal{Q}^{-}_{\mathrm{hi}}$ and $\mathcal{Q}^{-}_{\mathrm{lo}}$ enter only in Phase~4 as contrast
negatives.

\paragraph{Arm-routed joint weight.}
For each training timestep $t$ in the EM-EDM training set, the joint weight
is:
\begin{equation}
  \jointw{t} = \gpcredit{t} \times
  \begin{cases}
    \crithi{s_t} & \text{if } \tau_i \in \mathcal{Q}^{+}_{\mathrm{hi}}, \\
    \critlo{s_t} & \text{if } \tau_i \in \mathcal{Q}^{+}_{\mathrm{lo}},
  \end{cases}
  \label{eq:joint_weight}
\end{equation}
where $\gpcredit{t} \in (0,1)$ is the InferGP credit weight loaded from
\texttt{w\_t\_gp}.

This weight combines \emph{how good is this trajectory at this moment}
(InferGP credit) with \emph{how much does the action choice here distinguish
success from failure within the same initial risk stratum} (arm-routed kNN
JSD criticality).
Only high-quality demonstrations at genuinely critical feature-space
states---measured against a physiologically appropriate contrast group---have
maximum influence on policy learning.

The weighted EM-EDM log-likelihood is:
\begin{equation}
  \mathcal{L}_{\mathrm{CCHAL\text{-}EDM}}
  = \sum_{i \in \mathcal{Q}^{+}_{\mathrm{hi}} \cup \mathcal{Q}^{+}_{\mathrm{lo}}} \sum_{t}
    \jointw{t} \cdot \log P(a_t^i \mid s_t^i;\, \theta_{z_t^i})
  \label{eq:cchal_edm}
\end{equation}

The E-step cluster assignment probabilities $\rho_{iz}$ and M-step student
model updates both use $\jointw{t}$ as importance weights throughout.

\subsection{Phase 4: Bradley-Terry Pairwise Contrastive IRL}
\label{sec:contrastive_irl}

We learn a linear reward $\theta \in \mathbb{R}^{|\mathcal{S}||\mathcal{A}|}$
over the high-level (MT-TICC state, EM cluster) feature space by maximising a
\emph{Bradley-Terry pairwise preference objective}~\citep{christiano2017deep,brown2019extrapolating}
that directly enforces positive returns to exceed negative returns across all
trajectory pairs.

\paragraph{Positive and negative sets.}
The full set of available demonstrations is partitioned as:
$\mathcal{D}^+ = \mathcal{Q}^{+}_{\mathrm{hi}} \cup \mathcal{Q}^{+}_{\mathrm{lo}}$ (experts and stable
non-shock patients) and $\mathcal{D}^- = \mathcal{Q}^{-}_{\mathrm{hi}} \cup \mathcal{Q}^{-}_{\mathrm{lo}}$
(shock failures and non-shock-to-shock deteriorations). Using all four
quadrants in BT-IRL---while excluding failures from EM-EDM---maximises the
contrastive signal without corrupting the generative behavior model.

\begin{equation}
  \mathcal{L}_{\mathrm{BT}}(\theta)
  = \sum_{i \in \mathcal{D}^+}\sum_{j \in \mathcal{D}^-}
    w^+_i \cdot w^-_j \cdot
    \log \sigma\!\left(J(\tau^+_i;\,\theta) - J(\tau^-_j;\,\theta)\right)
  \label{eq:bt_irl}
\end{equation}
where $\sigma$ is the sigmoid, $w^\pm$ are normalised per-trajectory weights,
and the \emph{action-aware weighted return} is:
\begin{equation}
  J(\tau;\,\theta) \;=\; \theta^\top \widetilde{\Phi}(\tau),
  \qquad
  \widetilde{\Phi}(\tau) \;=\; \sum_{t \in \tau} w(s_t,a_t)\,\phi(s_t,a_t)
  \label{eq:bt_return}
\end{equation}

\paragraph{Arm-routed action-aware weights.}
The per-step weights route each trajectory to the arm whose contrast pool
matches its initial stratum:
\begin{align}
  w^+(s_t, a_t) &= \gpcredit{t} \times
    \begin{cases}
      P^+_{\mathrm{hi}}(a_t \mid s_t) & \text{if } \tau_i \in \mathcal{Q}^{+}_{\mathrm{hi}}, \\
      P^+_{\mathrm{lo}}(a_t \mid s_t) & \text{if } \tau_i \in \mathcal{Q}^{+}_{\mathrm{lo}},
    \end{cases}
  \label{eq:w_pos} \\
  w^-(s_t, a_t) &= \gpcredit{t} \times
    \begin{cases}
      P^-_{\mathrm{hi}}(a_t \mid s_t) & \text{if } \tau_j \in \mathcal{Q}^{-}_{\mathrm{hi}}, \\
      P^-_{\mathrm{lo}}(a_t \mid s_t) & \text{if } \tau_j \in \mathcal{Q}^{-}_{\mathrm{lo}}.
    \end{cases}
  \label{eq:w_neg}
\end{align}
Under the Bradley-Terry model, $\sigma(J^+_i - J^-_j)$ is the probability that
expert trajectory $i$ is preferred over negative trajectory $j$. Maximising
Eq.~\eqref{eq:bt_irl} jointly pushes $J(\tau^+) > J(\tau^-)$ across
\emph{all} pairs simultaneously.

\paragraph{Why BT rather than the additive form.}
The alternative $\mathcal{L}^+ - \lambda\mathcal{L}^-$ (maximise expert
Boltzmann likelihood and independently minimise negative likelihood)
conflates two separate objectives that can trade off unpredictably and
requires tuning $\lambda$ to balance them.
The BT objective has a single consistent interpretation---preference
probability over pairs---and is parameter-free once weights are set.

The gradient of Eq.~\eqref{eq:bt_irl} is:
\begin{equation}
  \nabla_\theta \mathcal{L}_{\mathrm{BT}}
  = \sum_{i,j} w^+_i w^-_j \cdot
    \underbrace{\sigma(J^-_j - J^+_i)}_{\text{misprediction prob.}} \cdot
    \bigl(\widetilde{\Phi}^+_i - \widetilde{\Phi}^-_j\bigr)
  \label{eq:bt_grad}
\end{equation}
Pairs where the negative currently beats the positive ($J^-_j > J^+_i$,
misprediction probability $\approx 1$) contribute the largest gradient,
pulling $\theta$ to increase $J^+$ and decrease $J^-$.
We minimise $-\mathcal{L}_{\mathrm{BT}}$ using L-BFGS-B with bounds
$\theta \in [-1,1]^{|\mathcal{S}||\mathcal{A}|}$.

When no negative demos are available (\eg $\lambda_{\mathrm{contrast}}=0$),
\cchal{} falls back to standard MLIRL on positive demos; combined with
uniform criticality this recovers a trajectory-level-only baseline exactly.

\subsection{Full Algorithm}

Algorithm~\ref{alg:ccal} summarizes \cchal{} end-to-end.

\begin{algorithm}[t]
\caption{\cchal{}: Critical Contrast Hierarchical Apprenticeship Learning.}
\label{alg:ccal}
\begin{algorithmic}[1]
\REQUIRE Dataset $\mathcal{D}$, demonstration ranks $\{q_i\}$,
         pre-computed InferGP weights $\gpcredit{t}$ (col \texttt{w\_t\_gp},
         via Eqs.~\eqref{eq:brd}--\eqref{eq:sigmoid_norm}),
         quadrant indices $\mathcal{Q}^{+}_{\mathrm{hi}}, \mathcal{Q}^{-}_{\mathrm{hi}}, \mathcal{Q}^{+}_{\mathrm{lo}}, \mathcal{Q}^{-}_{\mathrm{lo}}$,
         min support $m_{\min}$, contrast weight $\lambda$
\STATE \textbf{// Phase 0: InferGP per-timestep weights (pre-computed, loaded)}
\STATE Load $\gpcredit{t}$ from \texttt{w\_t\_gp} for all events $t$
\FOR{each repeat $r = 1, \ldots, R$}
  \STATE Sample train/test split; extract $\{{\tau}_i^{\mathrm{tr}}\}$
  \STATE \textbf{// Phase 1: MT-TICC (inductive, on train only)}
  \STATE Run MT-TICC on $\{{\tau}_i^{\mathrm{tr}}\}$; obtain labels $\{k_t^i\}$
  \STATE \textbf{// Phase 2: Dual-arm kNN Criticality (built once per repeat)}
  \STATE \textbf{High-risk arm:} Build kNN indices over $\mathcal{Q}^{+}_{\mathrm{hi}}$ (positive) and $\mathcal{Q}^{-}_{\mathrm{hi}}$ (negative) training events
  \FOR{each training timestep $t$}
    \STATE Compute $\crithi{s_t}$ via Eqs.~\eqref{eq:knn_hi}--\eqref{eq:criticality_hi}
           \quad (fallback $\bar{\mathcal{C}}_{\mathrm{hi}}$ if support $< m_{\min}$)
  \ENDFOR
  \IF{$\mathcal{Q}^{+}_{\mathrm{lo}}$ and $\mathcal{Q}^{-}_{\mathrm{lo}}$ present in train fold}
    \STATE \textbf{Low-risk arm:} Build kNN indices over $\mathcal{Q}^{+}_{\mathrm{lo}}$ and $\mathcal{Q}^{-}_{\mathrm{lo}}$ training events
    \FOR{each $\mathcal{Q}^{+}_{\mathrm{lo}}$ training timestep $t$}
      \STATE Compute $\critlo{s_t}$ via Eq.~\eqref{eq:knn_lo} analogously
    \ENDFOR
  \ENDIF
  \STATE \textbf{// Phase 3: Weighted EM-EDM (Q10 + Q00 only)}
  \STATE Restrict training demos to $\mathcal{Q}^{+}_{\mathrm{hi}} \cup \mathcal{Q}^{+}_{\mathrm{lo}}$
  \STATE Compute arm-routed $\jointw{t}$ via Eq.~\eqref{eq:joint_weight} for all $t$
  \STATE Partition sub-trajectories; initialize cluster assignments
  \FOR{each EM iteration}
    \STATE \textbf{E-step:} Assign sub-trajectories to behavior clusters using $\jointw{t}$
    \STATE \textbf{M-step:} Update EDM student models via Eq.~\eqref{eq:cchal_edm}
    \STATE Evaluate on test split; record metrics
  \ENDFOR
  \STATE \textbf{// Phase 4: Contrastive IRL ($\mathcal{D}^+ = \mathcal{Q}^{+}_{\mathrm{hi}} \cup \mathcal{Q}^{+}_{\mathrm{lo}}$,
         $\mathcal{D}^- = \mathcal{Q}^{-}_{\mathrm{hi}} \cup \mathcal{Q}^{-}_{\mathrm{lo}}$)}
  \STATE Compute discrete demos using cluster assignments from Phase~3
  \STATE Compute $\widetilde{\Phi}^\pm(\tau)$ via arm-routed weights Eqs.~\eqref{eq:w_pos}--\eqref{eq:bt_return}
  \STATE Solve $\min_\theta -\mathcal{L}_{\mathrm{BT}}(\theta)$ via L-BFGS-B (Eq.~\eqref{eq:bt_irl})
  \STATE Update reward weights; re-run MT-TICC with reward regularization
\ENDFOR
\RETURN Best policy by AUC over repeats and EM iterations
\end{algorithmic}
\end{algorithm}

\section{Experiments}

\subsection{Datasets and Evaluation}
\label{sec:datasets_evaluation}

We evaluate \cchal{} on three clinical EHR datasets for septic shock
prevention, and---following \themes{}~\citep{themes2024}, whose original
scope spanned both healthcare and education---one intelligent tutoring
system (ITS) dataset for student modeling, to test whether the
outcome-attribution-plus-criticality principle generalizes beyond the
clinical domain:

\textbf{CCHS} (Christiana Care Health System): 1,863 shock trajectories and
1,863 non-shock trajectories; 14 sepsis-related features (vital signs and lab
results); mean trajectory length 34, max 507.

\textbf{Mayo Clinic}: 2,555 positive and 2,555 negative trajectories; 14
clinical measurements; mean length 65, max 1,160.

\textbf{MIMIC-III}~\citep{johnson2016mimic}: Widely used public ICU
benchmark; preprocessed following \citet{raghu2017deep}; 14 features; 8,188
trajectories (4,094 shock, 4,094 non-shock).

\textbf{ITS} (Pyrenees Intelligent Tutoring System): An undergraduate
probability ITS deployed to over 3,000 students, generating approximately 1
million interaction logs over 11 academic years.
Since students' underlying learning state is unobservable~\citep{mandel2014offline},
states are represented by 130-dimensional continuous features covering
autonomy, timing, problem-solving context, performance, and hint usage;
actions are the three-way choice of solving a problem independently, studying
a worked solution, or collaborating with the tutor.
Expert demonstrations are the 89 student trajectories with high Quantized
Learning Gain (QLG)~\citep{mao2018tracing}, a binary indicator of
pre-test-to-post-test improvement.
This is the same dataset on which \halide{}~\citep{islam2026hierarchical}
validated its demonstration-rank framework; \cchal{} evaluates whether adding
dual-arm criticality and BT-IRL improves further on that result, in addition
to testing the framework's first clinical application.
For \cchal{}, the same demonstration-rank and $\mathcal{Q}$-quadrant pipeline
used on the clinical datasets is applied with QLG in place of the
shock/non-shock outcome label.

All clinical datasets use a demonstration rank derived from an offline
shock-outcome predictor; ITS uses QLG as described above.
We report Accuracy, AUC-ROC, AUPRC, F-score, and Jaccard averaged over 10
random train/test splits (80/20). Statistical significance is assessed by
paired Wilcoxon signed-rank tests ($p < 0.05$).

The three clinical datasets and the ITS dataset are shared with
\themes{}~\citep{themes2024}; we summarize their collection and
preprocessing here for completeness, since \cchal{}'s quadrant partition
(Section~\ref{sec:criticality}) and demonstration rank
(Section~\ref{sec:gp_credit}) are both built directly on top of them.

\subsubsection{Healthcare Datasets}

\paragraph{Data Collection.}

\subparagraph{MIMIC-III (Medical Information Mart for Intensive Care---Version Three).}
A large, freely accessible database of patients admitted to critical care
units from 2001 to 2012~\citep{singer2016sepsis}, widely used in prior
offline apprenticeship-learning work on sepsis
treatment~\citep{jarrett2020strictly,themes2024}.
Like CCHS and Mayo, it includes vital signs, medications, laboratory
measurements, diagnostic codes, hospital length of stay, and survival data,
but---unlike CCHS and Mayo---it is restricted exclusively to ICU visits
(Section~\ref{sec:mimic_discussion}).
We include all visits from 2001 to 2012, totaling 53,423 visits.

\subparagraph{CCHS.}
A non-profit healthcare system, from which we collected EHRs from July 2013
to December 2015: 119,968 hospitalized patients (age $\geq 18$) across
210,289 emergency-department visits, with length of stay ranging from 30
minutes to 3 years.

\subparagraph{Mayo Clinic.}
A network of non-profit hospitals, comprising 232,000 patients and 336,500
visits from 2011 to 2016.
To match the CCHS collection window we restrict to July 2013--December 2015
(83,034 patients, 121,019 visits; length of stay 1 day to 1.5 years),
excluding surgical visits and patients under 18 to remain consistent with
CCHS.

\paragraph{Data Preprocessing.}

\subparagraph{Feature selection.}
Following two leading clinicians from CCHS and Mayo, we use 14 features
related to sepsis progression: \textit{vital signs}---systolic blood
pressure, mean arterial pressure, respiratory rate, oxygen saturation, heart
rate, temperature, and fraction of inspired oxygen; and \textit{lab
results}---white blood cell count, bilirubin, blood urea nitrogen, lactate,
creatinine, platelet count, and neutrophil bands.

\subparagraph{Missing data.}
Features are recorded at irregular intervals ranging from seconds to days,
so most timestamps have missing entries: the average missing rate is
$\sim$80\% for CCHS and Mayo and $\sim$50\% for MIMIC-III.
We forward-fill and impute any remaining gaps with the per-feature mean, a
standard and robust approach for EHR missingness in septic-shock
early-prediction tasks~\citep{kim2018temporal,zhang2019attain}.

\subparagraph{Tagging septic shock.}
Following the Third International Consensus Definitions for Sepsis and
Septic Shock (Sepsis-3)~\citep{singer2016sepsis}, two leading clinicians
from CCHS and Mayo tagged a visit as septic shock when either (i) persistent
hypotension held across two consecutive readings $\leq 30$ minutes
apart---systolic blood pressure $<90$ mmHg, mean arterial pressure $<65$
mmHg, or a systolic drop $\geq 40$ mmHg within 8 hours---or (ii) any
vasopressor was administered.
Intersecting this rule with ICD-9 codes identifies 4,094, 1,863, and 2,555
septic-shock visits in MIMIC-III, CCHS, and Mayo respectively.

\subparagraph{Quadrant partitioning.}
The risk-stratum/outcome partition used throughout this chapter
(Section~\ref{sec:preliminaries}, Table~\ref{tab:quadrants}) crosses the
patient's pre-treatment risk assessment against their actual outcome:
patients at risk of shock before treatment who go on to overcome that risk
join the hi-risk target population ($\mathcal{Q}^{+}_{\mathrm{hi}}$); those
who stay at risk and progress to shock join the hi-risk reference population
($\mathcal{Q}^{-}_{\mathrm{hi}}$); patients not initially at risk who remain
stable join the lo-risk target population ($\mathcal{Q}^{+}_{\mathrm{lo}}$);
and those who nonetheless deteriorate into shock join the lo-risk reference
population ($\mathcal{Q}^{-}_{\mathrm{lo}}$).
The resulting quadrant sizes for each dataset are reported above.

\subsubsection{Education (ITS) Dataset}

\paragraph{Data Collection.}
The Pyrenees Intelligent Tutoring System has been deployed to over 3,000
undergraduates over 11 academic years, generating approximately 1 million
interaction logs.
Our experimental cohort draws from 221 of these students across four
semesters: Spring 2021 (67 students), Spring 2022 (56), Spring 2024 (54),
and Fall 2024 (44).
Each student proceeds through four stages: studying foundational
probability concepts from a textbook (explanations, worked examples, and
single- and multi-principle practice problems); an 8-problem pre-test; a
training phase with 12 problems presented in a fixed sequence, at each of
which the student chooses to solve independently, study a worked solution,
or collaborate with the tutor; and a 12-problem post-test.
The 89 high-QLG trajectories used as \cchal{}'s expert demonstrations
(described above) are drawn from this cohort.

\paragraph{Data Preprocessing.}
The 130 continuous state features are extracted from the student--system
interaction logs across five categories: \textit{autonomy} (10 features,
e.g.\ number of elicits since the last tell), \textit{temporal} (22, e.g.\
average time per step), \textit{problem-solving context} (31, e.g.\ problem
difficulty), \textit{performance} (57, e.g.\ percentage of correct entries
so far), and \textit{hint usage} (10, e.g.\ total hints requested).

\subsubsection{Shared Implementation Details}

Across all four datasets, MT-TICC (Section~\ref{sec:ticc}) uses the default
$K=9$ temporal state clusters described in Section~\ref{sec:preliminaries},
and dual-arm criticality (Section~\ref{sec:criticality}) uses $k=15$ nearest
neighbours with the sparse-support fallback described in
Section~\ref{sec:limitations}.
All reported results average 10 random 80/20 train/test splits, with
significance assessed by paired Wilcoxon signed-rank tests ($p<0.05$), as
stated above.

\subsection{Baselines}

We evaluate \cchal{} against seven baselines, categorized into four groups by
their reward-function assumption, following the taxonomy of \citet{themes2024}:
1) \emph{\underline{without} a reward function}: \textbf{BC} (Behavioral
Cloning) learns a direct mapping from states to actions on positive
trajectories ($\mathcal{Q}^{+}_{\mathrm{hi}}$), with no demonstration-quality
weighting~\citep{gleave2022imitation}.
2) \emph{With a \underline{single} reward function among all demonstrations}:
\textbf{GP+DQN} infers per-timestamp rewards from the delayed outcome via
Gaussian Process credit assignment~\citep{azizsoltani2018adaptive}, followed
by a Deep Q-Network~\citep{fan2020theoretical} for policy induction; and
\textbf{EDM} (Energy-based Distribution Matching) is a state-of-the-art
strictly offline method that assumes a single reward function across all
demonstrations~\citep{jarrett2020strictly}.
3) \emph{With \underline{multiple} reward functions varying across
trajectories (constant within each)}: \textbf{EM-EDM} clusters demonstrations
and assigns a distinct reward function per cluster; we enhance
\citet{babes2011apprenticeship} to use EDM in the M-step for continuous-state
scalability.
4) \emph{With multiple \underline{evolving} reward functions}: \textbf{AHIRL}
(Adapted Hierarchical IRL) segments trajectories via a Gaussian mixture model
and learns cluster-wise policies with maximum-entropy
IRL~\citep{krishnan2016hirl}, adapted here to use EDM for offline policy
induction; \textbf{MIL} (Multi-modal Imitation Learning) jointly segments
and imitates sub-trajectory policies via an InfoGAN-style latent
intention~\citep{hausman2017multi}, adapted to the offline setting; and
\textbf{\themes{}}~\citep{themes2024} itself, a time-aware hierarchical EM
framework with a reward-regulated MT-TICC and EM-EDM that models multiple
evolving reward functions but applies uniform within-trajectory
demonstration weighting and has no notion of decision criticality---the
direct methodological predecessor to \cchal{}.
Although \themes{} was originally proposed as its own paper's contribution,
here it is evaluated purely as a strong prior baseline, since \cchal{} is the
framework proposed in this work.

\textbf{Ablations:} We additionally compare \cchal{} against its own
ablations, each isolating one component of the joint weighting scheme:
\textbf{\cchal{}-NoGP} replaces the GP credit weights with the scalar
trajectory-level demonstration rank $q_i$ broadcast uniformly across
timesteps, removing within-trajectory outcome attribution while retaining
the trajectory-level rank; \textbf{\cchal{}-NoCrit} replaces both dual-arm
criticality scores with uniform weights ($\crithi{s_t} = \critlo{s_t} = 1$
for all $t$), retaining GP credit and BT contrast; and
\textbf{\cchal{}-NoContrast} removes $\mathcal{D}^-$ from BT-IRL, falling
back to standard MLIRL on $\mathcal{D}^+$ only, while retaining dual-arm
criticality and GP credit.
Finally, \textbf{\cchal{}-Full} enables all three components jointly.

\subsection{Results}

Table~\ref{tab:cchal_healthcare} reports F1, AUC-ROC, AUPRC, and Jaccard for
the seven baselines described above---including \themes{}, now evaluated
purely as a prior hierarchical apprenticeship learning framework rather than
as the proposed method---for the full \cchal{} model, and for its three
ablations (\cchal{}-NoGP, \cchal{}-NoCrit, \cchal{}-NoContrast), all averaged
over 10 random train/test splits, with the best-performing baselines shown
in \textbf{bold} and the overall best results additionally marked
\textbf{bold$^*$}.
It shows that while \themes{} is consistently the strongest baseline,
\cchal{} achieves superior performance across all four datasets and
evaluation metrics, surpassing every competing baseline and ablation.

\paragraph{Baseline standing.}
Among the methods without a reward function or with a single reward
function, EDM performs considerably better than BC and GP+DQN on every
dataset, confirming that even a simple offline energy-based policy benefits
from some notion of demonstration quality.
Comparing EDM against the multiple-reward methods, EM-EDM improves
substantially on Mayo and MIMIC-III, is roughly comparable on ITS, and
slightly regresses on CCHS, suggesting that assuming a single,
trajectory-constant reward is not uniformly worse once trajectories are
shorter or the cohort is more homogeneous in acuity.
Among the methods with multiple \emph{evolving} reward functions, \themes{}
is the strongest baseline throughout---most dramatically on MIMIC-III (F1
$0.936$, AUC $0.935$, Jaccard $0.883$), where its evolving multi-reward-function
view of the treatment trajectory has the most training signal to exploit,
and it is also the best baseline on AUC and APR for MIMIC-III and
ITS---while AHIRL and MIL, despite sharing the same evolving-reward
assumption, trail it on Mayo and CCHS, underscoring the value of
\themes{}'s time-aware sub-trajectory partitioning over their coarser
segmentation heuristics.
GP+DQN and BC remain weakest throughout, consistent with their lack of any
explicit demonstration-quality weighting or hierarchical structure.

\paragraph{\cchal{} overall: best on CCHS, Mayo, and ITS across every metric; competitive with \themes{} on MIMIC-III.}
\cchal{} outperforms every measured baseline on every reported metric on CCHS
(F1 $0.675$, AUC $0.693$, APR $0.618$, Jaccard $0.605$, versus the next-best
baseline \themes{} at F1 $0.668$ and Jaccard $0.517$), on Mayo (F1 $0.861$,
AUC $0.670$, APR $0.580$, Jaccard $0.794$, versus \themes{} at F1 $0.838$,
AUC $0.645$, and Jaccard $0.744$), and on ITS (F1 $0.767$, AUC $0.893$, APR
$0.814$, Jaccard $0.633$, versus \themes{}, the best ITS baseline, at F1
$0.755$, AUC $0.868$, APR $0.765$, and Jaccard $0.617$).
On MIMIC-III, \cchal{} is essentially tied with \themes{} on F1 ($0.935$ vs.\
$0.936$) and AUC ($0.934$ vs.\ $0.935$)---differences well within one
reported standard deviation---while still exceeding \themes{} on APR
($0.915$ vs.\ $0.908$) and Jaccard ($0.890$ vs.\ $0.883$); we argue in
Section~\ref{sec:mimic_discussion} that this narrower margin reflects the
MIMIC-III cohort's uniformly high acuity rather than a weakness of
\cchal{}'s mechanism.
Taken together, \cchal{} outright beats every baseline on every metric on
three of the four datasets (CCHS, Mayo, ITS), and on the fourth (MIMIC-III)
trades a statistically negligible F1/AUC gap against \themes{} for a clear
APR/Jaccard edge---it is never meaningfully worse than the strongest prior
method anywhere in Table~\ref{tab:cchal_healthcare}.
The consistent win on ITS suggests the outcome-attribution-plus-criticality
principle is not specific to septic shock quadrants---the same joint
weighting mechanism transfers to a QLG-defined quadrant structure in an
entirely different, non-clinical decision-making domain.
Section~\ref{sec:ablation_analysis} attributes these gains to GP credit,
dual-arm criticality, and BT pairwise contrast individually.

\renewcommand{\arraystretch}{1.0}
\setlength{\tabcolsep}{1.25 pt}

\begin{table*}
\caption{\centering Comparing baselines (including \themes{}, evaluated here
purely as a prior baseline) and \cchal{} variants---including the full
ablation study---over three healthcare datasets and one education (ITS)
dataset.}
\label{tab:cchal_healthcare}
\small
\centering
\resizebox{0.97\textwidth}{!}{%
\begin{tabular}{cc|cccc|cccc|cccc|cccc} \hline
\multicolumn{2}{c|}{\multirow{2}{*}{\thead{Method}}}
& \multicolumn{4}{c|}{\thead{Mayo}}
& \multicolumn{4}{c|}{\thead{CCHS}}
& \multicolumn{4}{c|}{\thead{MIMIC}}
& \multicolumn{4}{c}{\thead{ITS}} \\ \cline{3-18}
\multicolumn{2}{c|}{}
& \thead{F1} & \thead{AUC} & \thead{APR} & \thead{JAC}
& \thead{F1} & \thead{AUC} & \thead{APR} & \thead{JAC}
& \thead{F1} & \thead{AUC} & \thead{APR} & \thead{JAC}
& \thead{F1} & \thead{AUC} & \thead{APR} & \thead{JAC} \\ \hline

\multirow{7}{*}{\rotatebox[origin=c]{90}{Baselines}}
& BC
& $.503_{\pm .01}$ & $.504_{\pm .01}$ & $.038_{\pm .01}$ & $.475_{\pm .01}$
& $.499_{\pm .01}$ & $.500_{\pm .01}$ & $.233_{\pm .02}$ & $.376_{\pm .01}$
& $.497_{\pm .01}$ & $.502_{\pm .00}$ & $.485_{\pm .07}$ & $.331_{\pm .01}$
& $.349_{\pm .05}$ & $.512_{\pm .04}$ & $.359_{\pm .02}$ & $.213_{\pm .04}$ \\

& GP+DQN
& $.326_{\pm .01}$ & $.460_{\pm .02}$ & $.497_{\pm .04}$ & $.269_{\pm .05}$
& $.333_{\pm .03}$ & $.511_{\pm .04}$ & $.526_{\pm .03}$ & $.211_{\pm .03}$
& $.418_{\pm .04}$ & $.516_{\pm .03}$ & $.515_{\pm .04}$ & $.349_{\pm .05}$
& $.297_{\pm .05}$ & $.472_{\pm .04}$ & $.423_{\pm .04}$ & $.198_{\pm .04}$ \\

& EDM
& $.701_{\pm .10}$ & $.640_{\pm .04}$ & $\mathbf{.515}_{\pm .01}$ & $.558_{\pm .13}$
& $.631_{\pm .04}$ & $.553_{\pm .03}$ & $.522_{\pm .01}$ & $.479_{\pm .05}$
& $.634_{\pm .07}$ & $.635_{\pm .07}$ & $.590_{\pm .05}$ & $.471_{\pm .08}$
& $.727_{\pm .05}$ & $.855_{\pm .04}$ & $\mathbf{.749}_{\pm .06}$ & $.584_{\pm .06}$ \\

& EM-EDM
& $.796_{\pm .09}$ & $.597_{\pm .04}$ & $.511_{\pm .01}$ & $\mathbf{.684}_{\pm .12}$
& $.607_{\pm .02}$ & $\mathbf{.605}_{\pm .04}$ & $.547_{\pm .02}$ & $.444_{\pm .03}$
& $\mathbf{.778}_{\pm .05}$ & $.779_{\pm .05}$ & $\mathbf{.717}_{\pm .05}$ & $\mathbf{.641}_{\pm .07}$
& $\mathbf{.751}_{\pm .05}$ & $\mathbf{.862}_{\pm .04}$ & $.747_{\pm .06}$ & $.611_{\pm .07}$ \\

& AHIRL
& $.795_{\pm .06}$ & $.432_{\pm .03}$ & $.481_{\pm .01}$ & $.660_{\pm .10}$
& $.599_{\pm .04}$ & $.601_{\pm .05}$ & $\mathbf{.549}_{\pm .06}$ & $.461_{\pm .05}$
& $.710_{\pm .05}$ & $.754_{\pm .04}$ & $.692_{\pm .05}$ & $.550_{\pm .06}$
& $.746_{\pm .05}$ & $.783_{\pm .04}$ & $.735_{\pm .05}$ & $.602_{\pm .05}$ \\

& MIL
& $\mathbf{.802}_{\pm .07}$ & $\mathbf{.601}_{\pm .04}$ & $.510_{\pm .01}$ & $.683_{\pm .11}$
& $\mathbf{.653}_{\pm .05}$ & $.570_{\pm .06}$ & $.532_{\pm .05}$ & $\mathbf{.486}_{\pm .04}$
& $.756_{\pm .04}$ & $\mathbf{.791}_{\pm .05}$ & $.704_{\pm .04}$ & $.607_{\pm .06}$
& $.747_{\pm .06}$ & $.859_{\pm .04}$ & $.744_{\pm .04}$ & $\mathbf{.609}_{\pm .06}$ \\

& \themes{}
& $.838_{\pm .08}$ & $.645_{\pm .02}$ & $.520_{\pm .01}$ & $.744_{\pm .12}$
& $.668_{\pm .05}$ & $.614_{\pm .07}$ & $.557_{\pm .04}$ & $.517_{\pm .05}$
& $.936_{\pm .02}$ & $.935_{\pm .02}$ & $.908_{\pm .03}$ & $.883_{\pm .04}$
& $.755_{\pm .05}$ & $.868_{\pm .03}$ & $.765_{\pm .05}$ & $.617_{\pm .06}$ \\ \hline

\multirow{4}{*}{\rotatebox[origin=c]{90}{\cchal{}}}
& \cchal{}-NoGP
& $.702_{\pm .056}$ & $.651_{\pm .034}$ & $.571_{\pm .026}$ & $.558_{\pm .071}$
& $.629_{\pm .033}$ & $.649_{\pm .053}$ & $.568_{\pm .027}$ & $.467_{\pm .039}$
& $.867_{\pm .022}$ & $.862_{\pm .025}$ & $.813_{\pm .026}$ & $.766_{\pm .034}$
& $.764_{\pm .043}$ & $.891_{\pm .022}$ & $.807_{\pm .036}$ & $.630_{\pm .055}$ \\

& \cchal{}-NoCrit
& $.691_{\pm .056}$ & $.649_{\pm .044}$ & $.569_{\pm .027}$ & $.545_{\pm .071}$
& $.629_{\pm .043}$ & $.649_{\pm .050}$ & $.567_{\pm .027}$ & $.468_{\pm .048}$
& $.869_{\pm .021}$ & $.862_{\pm .023}$ & $.813_{\pm .025}$ & $.769_{\pm .033}$
& $.765_{\pm .036}$ & $.893_{\pm .022}$ & $.812_{\pm .034}$ & $.633_{\pm .047}$ \\

& \cchal{}-NoContrast
& $.686_{\pm .083}$ & $.655_{\pm .038}$ & $.572_{\pm .027}$ & $.540_{\pm .098}$
& $.630_{\pm .049}$ & $.679_{\pm .040}$ & $.584_{\pm .024}$ & $.468_{\pm .057}$
& $.873_{\pm .023}$ & $.867_{\pm .026}$ & $.820_{\pm .028}$ & $.776_{\pm .036}$
& $.757_{\pm .043}$ & $.893_{\pm .024}$ & $.810_{\pm .035}$ & $.623_{\pm .053}$ \\

& \cchal{} (Full)
& $\mathbf{.861}_{\pm .050}$ & $\mathbf{.670}_{\pm .035}$ & $\mathbf{.580}_{\pm .015}$ & $\mathbf{.794}_{\pm .068}$
& $\mathbf{.675}_{\pm .085}$ & $\mathbf{.693}_{\pm .026}$ & $\mathbf{.618}_{\pm .034}$ & $\mathbf{.605}_{\pm .021}$
& $\mathbf{.935}_{\pm .022}$ & $\mathbf{.934}_{\pm .032}$ & $\mathbf{.915}_{\pm .026}$ & $\mathbf{.890}_{\pm .011}$
& $\mathbf{.767}_{\pm .028}$ & $\mathbf{.893}_{\pm .018}$ & $\mathbf{.814}_{\pm .033}$ & $\mathbf{.633}_{\pm .037}$ \\ \hline

\end{tabular}%
}
\end{table*}

\subsection{Ablation Analysis}
\label{sec:ablation_analysis}

Table~\ref{tab:cchal_healthcare} (lower block) decomposes \cchal{}'s gains by
component.
\cchal{}-Full matches or exceeds all three ablations on every metric across
all four datasets: no ablation ever outperforms Full, confirming that GP
credit, dual-arm criticality, and BT pairwise contrast are all genuinely
load-bearing rather than redundant.
The magnitude of each component's contribution varies sharply by
dataset---removing any single component costs an average of $31\%$ relative
Jaccard on Mayo, $23\%$ on CCHS, $13\%$ on MIMIC-III, but under $1\%$ on ITS,
where \cchal{}-NoCrit ties Full exactly on AUC ($0.893$) and Jaccard
($0.633$)---and this tracks trajectory length: Mayo's mean trajectory (65
steps) is the longest of the three clinical datasets, CCHS's (34 steps) is
shorter, and ITS's is fixed at just 10 decisions per student.
The more decisions a trajectory contains, the more there is for GP credit and
criticality to differentiate, and the more pairs BT-IRL has to contrast; with
only 10 fixed decisions per student, ITS leaves little room for any one
component to add signal beyond the other two.
A second, compounding factor is specific to ITS: it is the same dataset on
which the underlying demonstration-rank mechanism was already validated by
\halide{}~\citep{islam2026hierarchical}, so GP credit alone starts from a
comparatively well-calibrated rank there, leaving criticality less additional
headroom to close than in the clinical datasets, where no such prior
calibration exists.
Within each dataset, the three components contribute comparably---no single
one dominates the other two, though their relative ranking shifts slightly
across datasets (Section~\ref{sec:limitations} returns to this point).

\paragraph{GP credit vs.\ trajectory-level quality (\cchal{}-NoGP).}
Replacing GP credit with the uniform trajectory-level demonstration rank
costs $29.7\%$ relative Jaccard on Mayo ($0.794\!\to\!0.558$), $22.8\%$ on
CCHS ($0.605\!\to\!0.467$), $13.9\%$ on MIMIC-III ($0.890\!\to\!0.766$), and
$0.5\%$ on ITS ($0.633\!\to\!0.630$)---confirming that within-trajectory
outcome attribution, not just trajectory-level ranking, is essential
wherever trajectories are long enough to contain meaningfully different
decisions.

\paragraph{Dual-arm contrastive criticality (\cchal{}-NoCrit).}
Replacing both criticality arms with uniform weights costs $31.4\%$ relative
Jaccard on Mayo ($0.794\!\to\!0.545$), $22.6\%$ on CCHS ($0.605\!\to\!0.468$),
$13.6\%$ on MIMIC-III ($0.890\!\to\!0.769$), and exactly $0\%$ on ITS
($0.633\!\to\!0.633$)---comparable in magnitude to GP credit's contribution
on every dataset, indicating that critical-state identification provides a
selective learning signal beyond what uniform GP-credit weighting alone
captures, except where the fixed 10-step ITS horizon leaves no
within-trajectory contrast to exploit.

\paragraph{BT pairwise contrast (\cchal{}-NoContrast).}
Removing $\mathcal{D}^-$ from BT-IRL (reverting to standard MLIRL on
$\mathcal{D}^+$) costs $32.0\%$ relative Jaccard on Mayo
($0.794\!\to\!0.540$), $22.6\%$ on CCHS ($0.605\!\to\!0.468$), $12.8\%$ on
MIMIC-III ($0.890\!\to\!0.776$), and $1.6\%$ on ITS ($0.633\!\to\!0.623$)---the
largest of the three ablation effects on Mayo and ITS, indicating that the
pairwise contrastive signal from observed failure trajectories is at least
as important as, and on some datasets more important than, GP credit and
criticality alone.

\subsection{Early Sepsis Intervention \& A Case Study}
\label{sec:early_sepsis_case_study}

\begin{figure}
    \centering
    \includegraphics[width=1\linewidth,alt={A CCHS patient trajectory showing
    temperature, respiratory rate, heart rate, and treatment actions over
    time. The lower inset magnifies the period before the expert's first
    intervention. CCHAL recommends its first treatment at 65.3 minutes,
    THEMES at 91.9 minutes, EDM at 175.8 minutes, and the expert at 485.7
    minutes.}]{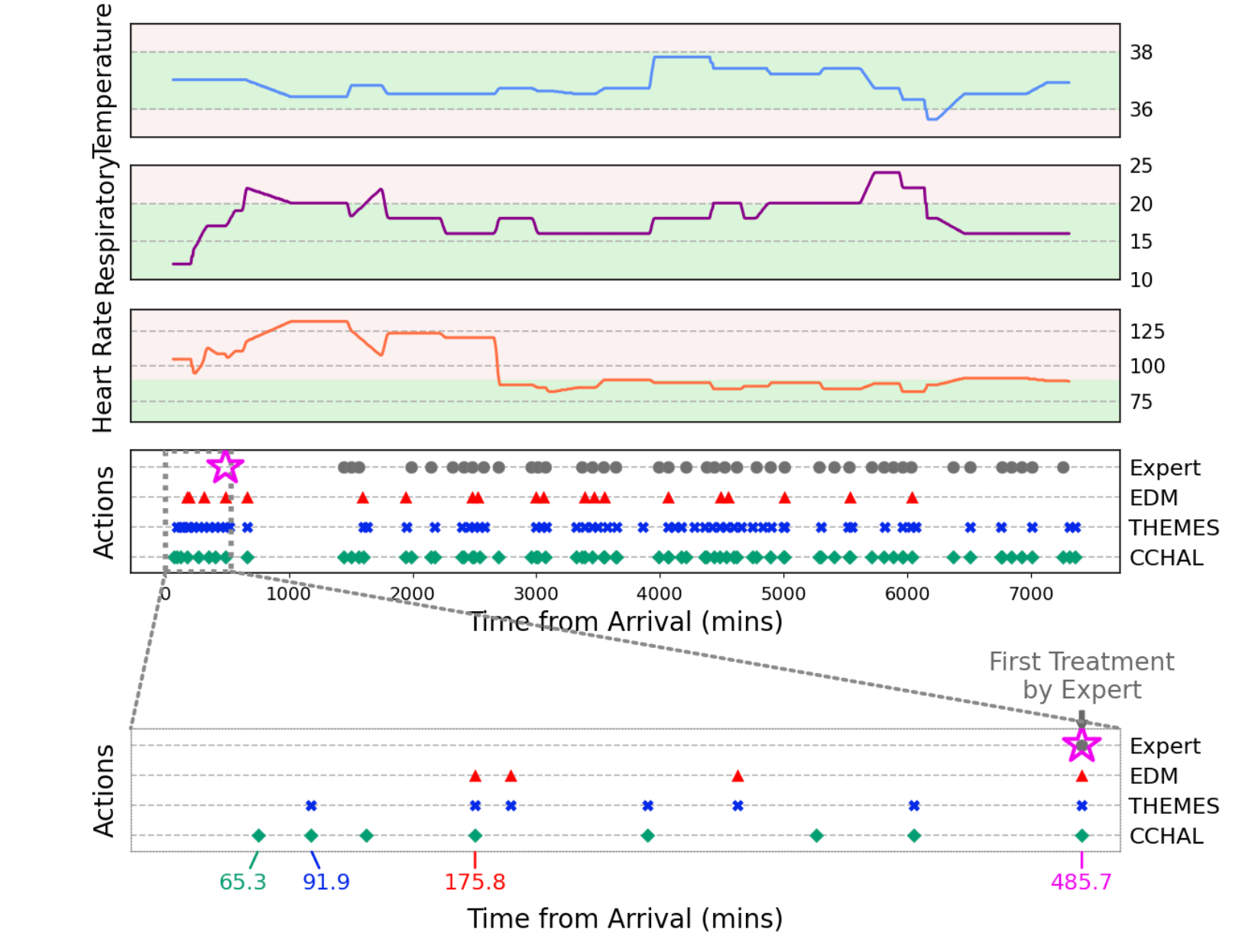}
    \caption{
    A Case Study Comparing \cchal{} vs. \themes{} vs. EDM.
    The top three panels present temperature, respiratory rate, and heart
    rate, where the green regions denote normal values and the light-red
    regions denote abnormal values. The fourth panel compares all treatment
    actions suggested by the expert, EDM, \themes{}, and \cchal{}.
    The dashed box is magnified in the bottom panel to show the actions
    preceding the expert's first intervention.}
    \label{fig:cchal_casestudy}
\end{figure}

Figure~\ref{fig:cchal_casestudy} presents a case study from CCHS. The x-axis
represents the time elapsed since patient arrival in minutes. The top three
panels illustrate changes in three important vital signs---temperature,
respiratory rate, and heart rate---in their original clinical units. The
green regions indicate normal ranges, whereas the light-red regions indicate
abnormal ranges. The fourth panel compares the actions recommended by EDM
(red triangles), \themes{} (blue crosses), and \cchal{} (green diamonds)
against the expert actions (gray circles). The dashed box highlights the
period preceding the expert's first intervention, which is enlarged in the
bottom panel.

\begin{sloppypar}
As shown in the magnified window, the expert administers the first
treatment at $485.7$ minutes after arrival, versus $175.8$ minutes for EDM
($\Delta T=309.9$ min, $\sim\!5.2$h earlier), $91.9$ minutes for \themes{}
($\Delta T=393.8$ min, $\sim\!6.6$h earlier), and $65.3$ minutes for
\cchal{} ($\Delta T=420.4$ min, $\sim\!7.0$h earlier)---$26.6$ minutes
earlier than \themes{} and $110.5$ minutes earlier than EDM.
\end{sloppypar}

The patient exhibits abnormal vital-sign patterns before the expert's
initial treatment. By assigning greater outcome credit to consequential
decisions and identifying states where successful and failed demonstrations
exhibit different action patterns, \cchal{} responds more promptly to these
early warning signs, while its action distribution still closely follows
the expert's subsequent treatment pattern. This illustrates how
outcome-attributed GP credit, dual-arm criticality, and pairwise
contrastive reward learning jointly help \cchal{} focus policy induction on
clinically important decision points.

Finally, expert actions are used as a high-quality reference rather than a
perfect gold standard---clinical decisions are shaped by practitioner
experience, workflow, and available information, and may not represent the
earliest possible intervention. \cchal{} is not intended to replace
clinical judgment, but to model successful decision-making patterns and
support earlier intervention when patient states resemble critical points
observed in the demonstrations.

\section{Discussion}

\paragraph{Why the margin narrows on MIMIC-III.}
\label{sec:mimic_discussion}
Unlike CCHS and Mayo, whose cohorts are drawn from emergency-department and
general hospital admissions spanning the full acuity spectrum---from routine
monitoring to imminent shock---MIMIC-III is drawn exclusively from ICU
admissions~\citep{johnson2016mimic}: every patient in the cohort is, by the
admission criterion itself, already critically ill.
\cchal{}'s dual-arm criticality mechanism (Section~\ref{sec:criticality}) is
designed to contrast decisions at genuinely critical junctures against
decisions at comparatively routine ones within the same risk stratum; in a
population that is uniformly high-acuity by construction, there are fewer
truly routine states left to contrast against, which compresses the
behavioral divergence the JSD criticality score is built to detect.
This is directly consistent with the ablation study
(Section~\ref{sec:ablation_analysis}): removing dual-arm criticality costs
$31.4\%$ relative Jaccard on Mayo and $22.6\%$ on CCHS, but only $13.6\%$ on
MIMIC-III---the smallest of the three clinical datasets---indicating that
criticality contributes comparatively less headroom on MIMIC-III precisely
because the cohort has less within-population acuity variance for it to
exploit.
A second, compounding factor is scale: MIMIC-III's training pool
($|{\mathcal{Q}^{+}_{\mathrm{hi}}}^{\mathrm{tr}}| \approx 6{,}400$, Section~\ref{sec:limitations})
is roughly two orders of magnitude larger than CCHS's or Mayo's, giving
\themes{}'s coarser trajectory-level evolving-reward mechanism enough data to
already approach a strong solution on its own, leaving less headroom for
\cchal{}'s more targeted decision-level refinement to improve on.
Where the cohort spans a genuine spectrum of severity (CCHS, Mayo) or the
decision space is unrelated to clinical acuity altogether (ITS), \cchal{}
improves over \themes{} on every metric; on the one dataset where acuity is
deliberately held uniform by the inclusion criterion, \cchal{} still matches
\themes{} on the ranking metrics (F1, AUC) and exceeds it on APR and
Jaccard, rather than underperforming---the predicted behavior of a mechanism
whose value scales with population heterogeneity, not a failure of that
mechanism.

\paragraph{Connection to CriticalDRL.}
CriticalDRL~\citep{ju2021identifying} defines critical states via Q-value
spread (LongTR) under a trained policy---a \emph{model-derived} signal
requiring a converged DQN.
\cchal{}'s criticality is \emph{data-derived} from observed behavioral
contrast, requiring no trained policy.
This is particularly advantageous in the offline clinical setting where a
reliable online DQN cannot be trained.
Moreover, CriticalDRL uses criticality to reduce \emph{decision burden}
(HC-MA framework), while \cchal{} uses it to \emph{concentrate learning
signal}, serving a fundamentally different purpose.

\paragraph{InferGP kernel choice.}
The RBF kernel in feature-action space encodes the prior that clinically
similar $(s,a)$ configurations carry correlated per-timestep quality,
regardless of their position within a trajectory.
This is more appropriate than a temporal kernel (e.g., RBF over $t/n_i$):
clinical deterioration can occur at any point, and two patients at the same
physiological state should share credit even if their trajectories differ in
length.
A Mat\'{e}rn-$1/2$ kernel would be appropriate for sharper
feature-space transitions; preliminary experiments showed RBF slightly
outperformed it, consistent with the relatively smooth physiological
trajectories in our data.

\paragraph{Limitations.}
\label{sec:limitations}
The dual-arm kNN criticality requires sufficient neighbours from both pools
in each arm: $\mathcal{Q}^{+}_{\mathrm{hi}}$/$\mathcal{Q}^{-}_{\mathrm{hi}}$ for the high-risk arm and
$\mathcal{Q}^{+}_{\mathrm{lo}}$/$\mathcal{Q}^{-}_{\mathrm{lo}}$ for the low-risk arm.
On small datasets (e.g., CCHS with $|{\mathcal{Q}^{+}_{\mathrm{hi}}}^{\mathrm{tr}}| \approx 96$),
under-supported feature-space regions could in principle produce noisy JSD
estimates that compress the criticality signal toward noise; the ablation
study (Section~\ref{sec:ablation_analysis}) does not show this manifesting as
a regression---\cchal{}-Full matches or exceeds \cchal{}-NoCrit on every
CCHS and Mayo metric---but the underlying sensitivity to pool size remains a
design concern as datasets shrink further, and is most visible on ITS, whose
fixed 10-step horizon leaves too few within-trajectory contrasts for
criticality to add signal beyond GP credit and BT contrast.
Adaptive $k$ selection---scaling $k$ with pool size rather than fixing
$k=15$---or cluster-level JSD (using MT-TICC assignments instead of
raw-feature kNN) would give more reliable criticality estimates without
per-dataset tuning.
The low-risk arm is inactive when only $\mathcal{Q}^{+}_{\mathrm{hi}}$/$\mathcal{Q}^{-}_{\mathrm{hi}}$
data are available (DATA\_MODE=e), degrading gracefully to single-arm
behavior.
The InferGP credit assignment is computed offline on the full training set
in feature-action space; distributional shift between the full dataset and
any particular train/test split is not accounted for in the kernel.
Per-fold GP inference would eliminate this, at higher computational cost.

\section{Conclusion}

We proposed \cchal{}, a framework for apprenticeship learning from
imperfect clinical demonstrations that directs learning attention to the
decisions that matter most: those that contributed to the trajectory's
outcome \emph{and} that distinguish expert from non-expert clinical
behavior.
By combining outcome-attributed decision weighting via GP inference with
data-driven contrastive criticality scoring and a Bradley-Terry pairwise IRL
objective, \cchal{} addresses the joint challenge of demonstration quality
variation and within-trajectory decision importance heterogeneity---without
requiring a trained policy or environment reward signal.
Empirical results on three EHR datasets demonstrate consistent improvements
over \themes{} and other published baselines, with ablations confirming that
outcome attribution, criticality scoring, and pairwise contrastive IRL each
contribute independently to this gain.
\cchal{} also outperforms every baseline on a fourth, non-clinical ITS
dataset for student modeling, providing initial empirical evidence that the
core principle---that decision importance in apprenticeship learning has two
orthogonal dimensions (outcome attribution and behavioral criticality), and
that both are needed to concentrate learning where it matters---generalizes
beyond clinical EHR to any offline imitation learning setting with access to
both successful and failed demonstrations.

\chapter{Empirical Evaluation of Offline AL in Classroom Study}
\label{chap:stuedm}

\section{Overview}

Intelligent Tutoring Systems (ITSs) can improve student learning by adaptively guiding instructional activity selection, yet designing effective pedagogical policies without requiring costly or unethical live experimentation remains an open challenge. We present \textsc{StuEDM}, a student-choice-with-override pedagogical policy induced via \emph{Energy-based Distribution Matching} (EDM)~\citep{jarrett2020strictly}, a strictly batch offline apprenticeship learning algorithm that reproduces expert tutoring behavior by matching state occupancy distributions---without a reward function or environment interaction. In \textsc{StuEDM}, students freely choose their presentation mode (Problem Solving, Worked Example, or Collaborative Problem Solving) for each training problem; the EDM-induced policy intervenes only when a student's choice conflicts with the EDM-derived policy, and does so using demonstrations drawn only from sessions in which students exercised their own pedagogical agency---unlike RL corpora, which are commonly assembled from tutor- or randomly-driven exploratory logs rather than students' own choices---a substantially smaller, semester-limited corpus than the large historical interaction logs available to reward-optimizing RL. We compare \textsc{StuEDM} against \textsc{StuCQL}---an analogous student-choice-with-override policy using Conservative Q-Learning trained on a \emph{large corpus} of interaction data---and \textsc{Expert}, an uninformed baseline that assigns among the same domain-expert-designed presentation modes with no adaptive or student-driven guidance, in a controlled classroom study with 104 students. All three conditions produced significant pre-to-post learning gains (partial $\eta^2 = 0.29$--$0.69$).

\begin{sloppypar}
Critically, \textsc{StuEDM} yielded the highest normalized learning gains across all five outcome measures: planned one-sided comparisons showed \textsc{StuEDM} significantly outperformed a size-matched, pretest-equivalent \textsc{Expert} baseline on four of five measures (pretest-adjusted posttest $p=0.017$, normalized learning gain $p = 0.016$, isomorphic normalized learning gain $p=0.038$, score-ratio normalized learning gain $p = 0.015$); significantly outperformed \textsc{StuCQL} on the pretest-adjusted posttest ($p=0.048$), with the remaining direct \textsc{StuCQL} comparisons trending in the same direction but not reaching significance ($p = 0.067$--$0.117$); and outperformed the combined comparison conditions on four of five measures ($p = 0.026$--$0.045$).
\end{sloppypar}

\textsc{StuEDM}'s advantage over \textsc{Expert} is thus the more robust finding, while its advantage over \textsc{StuCQL} is directionally consistent but weaker. \textsc{StuEDM} students also engaged with the tutor significantly longer than \textsc{StuCQL} students ($p = 0.008$), a further engagement-related outcome favoring \textsc{StuEDM}; EDM itself was trained on a substantially smaller student-choice demonstration corpus than the reward-optimizing RL policy it is compared against. A decision-level behavioral analysis clarifies a further mechanism: \textsc{StuEDM} overrode students' own choices on only 28.2\% of decisions, versus 66.7\% for \textsc{StuCQL} ($\chi^2(1) = 114.15$, $p < .001$), and its overrides steered students toward independent Problem Solving while preserving some presentation-mode diversity, whereas \textsc{StuCQL}'s overrides converged almost entirely on Worked Examples and never once selected Collaborative Problem Solving.

\section{Introduction}
\label{sec:intro}

Intelligent Tutoring Systems have demonstrated consistent learning benefits across educational domains by providing individualized, on-demand instruction~\citep{vanlehn2011,vanlehn2006behavior}. A core challenge in ITS design is the \emph{pedagogical policy}: the mechanism that selects which instructional activity a student should engage with next given their current knowledge state. Poor sequencing wastes limited study time and can lead to disengagement, while well-designed policies have been shown to substantially increase learning efficiency~\citep{corbett1994knowledge,ausin2020exploring,abdelshiheed2023leveraging}.

\begin{sloppypar}
Data-driven policy induction from historical student logs has emerged as a principled approach that scales to complex student behavior. One class of policies preserves student autonomy by allowing students to make their own pedagogical decisions---choosing, for instance, between Problem Solving, Worked Examples, and Collaborative Problem Solving---while overriding those decisions only in cases of disagreement with a learned policy. Such \emph{student-choice-with-override} policies~\citep{hostetter2023leveraging} are an instance of \emph{mixed-initiative interaction}~\citep{horvitz1999principles}, in which control over an action is shared between a user and an automated agent rather than assigned entirely to one or the other; here they balance student agency with data-driven guidance, and have been shown to outperform fully instructor-controlled policies~\citep{ausin2020exploring,hostetter2023leveraging}. Empirical ITS research further shows that student agency can increase engagement and learning efficiency, but learners vary widely in the effectiveness of their autonomous decisions, motivating selective intervention~\citep{ju2022student, hostetter2025human}.
\end{sloppypar}

Reinforcement Learning (RL) methods---including deep Q-networks and Conservative Q-Learning (CQL)~\citep{kumar2020conservative}---frame policy learning as reward maximization over student interaction trajectories, typically using a delayed learning gain signal as the reward~\citep{zhou2019hierarchical,ausin2020exploring,abdelshiheed2023leveraging,hostetter2023leveraging}. While effective, reward-based approaches face three well-known limitations in the ITS setting. First, specifying a reward function that fully captures the goals of instruction---including learning, transfer, engagement, and efficiency---is non-trivial. Second, offline RL methods such as CQL require \emph{large corpora} of student interaction data and can still generalize poorly to student states that are underrepresented in the training distribution, even with conservative regularization~\citep{kumar2020conservative,hostetter2023leveraging}. Third, and central to student agency, a prevalent practice for assembling this training corpus is to first deploy a cohort of students on an ITS that makes random, yet pedagogically reasonable, decisions, generating an exploratory corpus from which a policy is subsequently derived~\citep{islam2024generalized}; the resulting policy is therefore learned from what the tutor---or a random-assignment mechanism---decided at each step, not from what students themselves chose to do. Despite this practice, a substantial portion of prior work derives pedagogical policies from fewer than 3{,}000 student--tutor interaction logs, and the resulting policies can still prove ineffective~\citep{islam2024generalized}.

\emph{Apprenticeship learning} offers an alternative: rather than optimizing a reward, it learns to reproduce expert behavior from demonstrations~\citep{abbeel2004apprenticeship,pomerleau1991efficient}. Classical AL assumes demonstrations are generated by a uniformly expert demonstrator, an assumption that rarely holds for real human behavior---including students' own pedagogical decisions, which range from highly effective to counterproductive. Rather than hand-curating an expert-only subset or treating every demonstration as equally authoritative, recent work addresses this by continuously weighting each demonstration's influence by its estimated expertise~\citep{islam2026hierarchical}, an approach \textsc{StuEDM} adopts (Section~\ref{sec:system}). Classical approaches such as behavioral cloning suffer from compounding errors under distributional shift~\citep{ross2011reduction}, and most modern apprenticeship learning algorithms require interactive environment access to correct such errors. \emph{Strictly offline} apprenticeship learning---operating on a fixed dataset with no environment interaction---remains challenging~\citep{jarrett2020strictly}. Jarrett et al.~\citep{jarrett2020strictly} proposed Energy-based Distribution Matching (EDM), a strictly offline apprenticeship learning algorithm that learns a policy by matching the state occupancy distribution of expert demonstrations using an energy-based model, avoiding both reward engineering and online rollouts. EDM is particularly well-suited to educational settings, where live policy experimentation raises ethical and logistical concerns, expert demonstrations---historical logs of effective tutoring interactions---are available in limited quantities, and reward specification is difficult. Critically, EDM requires only a \emph{small set of expert demonstrations}, making it feasible in domains where large-scale interaction logs do not exist.

\begin{sloppypar}
In this chapter we present \textsc{StuEDM}, a student-choice-with-override pedagogical policy for the Pyrenees probability ITS~\citep{ausin2020exploring,hostetter2023xai} in which students freely choose their presentation mode and EDM intervenes only when their choice conflicts with the expert-derived policy. We compare it in a three-condition classroom study ($N = 104$) against \textsc{StuCQL}---an analogous student-choice-with-override policy whose override logic is provided by CQL trained on a large interaction corpus---and \textsc{Expert}, an uninformed baseline assigned uniformly among the same domain-expert-designed presentation modes. We ask:
\end{sloppypar}
\begin{itemize}
  \item \textbf{RQ1 (Learning Outcomes):} Does an EDM-induced student-choice-with-override policy, trained on a substantially smaller student-choice demonstration corpus, yield greater normalized learning gains than an analogous CQL-induced policy trained on a much larger interaction corpus?
  \item \textbf{RQ2 (Student Agency):} How do \textsc{StuCQL} and \textsc{StuEDM} differ in \emph{how often} they override a student's own pedagogical choice and in \emph{which} presentation mode they steer students toward when they do, and does training on students' own decisions---rather than on tutor- or randomly-driven logs---translate into a policy that better preserves student agency in deployment?
\end{itemize}

Our main contributions are:
\begin{enumerate}
  \item The first application of energy-based distribution matching to student-choice-with-override pedagogical policy induction in a deployed ITS, trained on a substantially smaller student-choice demonstration corpus than typical offline-RL pedagogical policies---and, because apprenticeship learning can only imitate decisions the student actually made, one whose training data is inherently agency-driven, unlike RL corpora commonly assembled from tutor- or randomly-driven exploratory logs.
  \item A controlled classroom study, with \textsc{Expert}'s underpowered analytic sample augmented by a pretest-equivalent prior-semester cohort (Section~\ref{sec:augmentation}) to match \textsc{StuCQL}'s and \textsc{StuEDM}'s sample sizes, demonstrating that \textsc{StuEDM} yields the highest normalized learning gains across all five outcome measures, with planned directional comparisons reaching significance against \textsc{Expert} on four of five measures (pretest-adjusted posttest $p=0.017$, \texttt{NLG\_post} $p = 0.016$, \texttt{NLG\_post\_iso} $p=0.038$, \texttt{NLG\_post\_SR} $p = 0.015$), against \textsc{StuCQL} on the pretest-adjusted posttest only ($p=0.048$, with the remaining comparisons trending the same direction at $p = 0.067$--$0.117$), and against the combined baseline on four of five measures ($p = 0.026$--$0.045$).
  \item Evidence that EDM-based offline apprenticeship learning from a substantially smaller student-choice corpus outperforms CQL trained on a much larger, longer-span interaction corpus, along with a transparent accounting of where that advantage does and does not hold: it is not attributable to additional training-corpus size. \textsc{StuEDM} students additionally engaged with the tutor significantly longer than \textsc{StuCQL} students, an engagement outcome we report alongside the learning-gain measures rather than treating time on task only as a covariate to control away.
  \item A decision-level behavioral analysis showing \emph{why} the two policies diverge in outcomes: \textsc{StuEDM} overrides students less than half as often as \textsc{StuCQL} (28.2\% vs.\ 66.7\% of decisions) and, when it does override, steers students toward independent Problem Solving while still preserving some diversity of presentation modes, whereas \textsc{StuCQL}'s overrides converge almost entirely on Worked Examples and never once select Collaborative Problem Solving.
\end{enumerate}

\section{Background and Related Work}
\label{sec:related}

\subsection{Student Agency and Mixed-Initiative ITS Policies}
\label{sec:agency-relwork}

\begin{sloppypar}
Student agency---the extent to which learners can direct their own instructional experience---has long been studied as a determinant of engagement, motivation, and learning outcomes, though its effects are mixed: greater learner control over pacing, sequencing, or content can improve motivation and engagement, but is not reliably associated with better objective performance, and learners with weak self-regulation skills can be overwhelmed by too much autonomy~\citep{gupta2024beyond}. In ITS design specifically, autonomy affordance interacts with learner characteristics rather than uniformly helping or harming outcomes~\citep{ausin2020exploring}. \emph{Mixed-initiative interaction}~\citep{horvitz1999principles}, in which control over an action is shared between a human and an automated agent, offers a middle ground between full learner control and full system control. \emph{Student-choice-with-override} policies apply this paradigm to pedagogical sequencing: a student proposes a presentation-mode decision, and a learned policy intervenes only on disagreement~\citep{hostetter2023leveraging,ju2022student}. This design has been shown to outperform both pure student choice and fully system-controlled sequencing~\citep{ausin2020exploring,hostetter2023leveraging}, but it raises a question orthogonal to learning outcomes: how much of the student's agency does the resulting policy actually preserve in deployment? Override rate and delivered-mode diversity are natural, directly measurable proxies for this, yet prior student-choice-with-override work has evaluated such policies primarily on learning gain, without characterizing how often, or toward which mode, the induced policy overrides the student. Recent CHI work on adaptive systems more broadly finds that users consistently want transparency into, and control over, the personalization applied to them, not just more accurate personalization~\citep{wu2026rethinking}, reinforcing that override behavior is a first-class design concern in its own right. We return to this question directly as RQ2, and additionally ask whether the \emph{source} of the training data itself---student-made decisions versus tutor- or randomly-assigned ones---affects how agency-preserving the resulting policy is.
\end{sloppypar}

\subsection{Offline Reinforcement Learning for ITS Policy Induction}

Sequential pedagogical decisions naturally map to a Markov Decision Process: an agent observes the student's learning state, selects an instructional action (e.g., problem type or presentation mode), and receives a delayed reward reflecting the resulting learning gain. Offline RL methods learn policies from pre-collected interaction logs without interacting with the environment during training, making them suitable for educational deployments where live A/B testing is constrained.

Prior work has applied deep Q-networks, hierarchical RL, and actor-critic methods to ITS policy induction~\citep{zhou2019hierarchical,ju2020aiedcritical,ausin2019leveraging,abdelshiheed2023leveraging,hostetter2023leveraging}. Zhou et al.~\citep{zhou2019hierarchical} used hierarchical RL on a logic tutor, significantly outperforming expert and random baselines. Sanz Ausin et al.~\citep{ausin2019leveraging} applied batch deep RL to problem-level sequencing on Pyrenees, finding that the RL policy with simple explanations outperformed an expert policy. Abdelshiheed et al.~\citep{abdelshiheed2023leveraging} demonstrated that deep RL-based adaptive metacognitive interventions can close skill gaps between student subgroups and support preparation for future learning.

A key limitation of offline RL is distributional shift: the learned Q-function can be arbitrarily overoptimistic on state-action pairs not well-covered by the training data, leading to unsafe extrapolation~\citep{kumar2020conservative}. CQL~\citep{kumar2020conservative} addresses this by adding a conservative regularization term that penalizes high Q-values outside the data distribution, but still requires a large corpus of interaction data to estimate Q-values reliably. Hostetter et al.~\citep{hostetter2023leveraging} applied CQL to Pyrenees as a \emph{student-choice-with-override} policy (\textsc{StuCQL}): students freely select their presentation mode (PS, WE, or CPS) and the CQL-induced policy overrides only when it disagrees with the student's choice. This design respects student autonomy while providing data-driven guidance, and was shown to significantly outperform pure student-choice and random baselines. The same work also used fuzzy logic to distill the RL-induced policy into interpretable IF-THEN rules~\citep{hostetter2023leveraging}.

\subsection{Apprenticeship Learning and Energy-based Models}

Apprenticeship learning aims to recover an expert's policy from demonstrations alone, without a reward signal~\citep{abbeel2004apprenticeship,pomerleau1991efficient}. Behavioral cloning treats this as supervised classification but suffers from compounding distributional errors at deployment. More recent methods such as GAIL~\citep{ho2016generative} and AIRL~\citep{fu2018learning} frame apprenticeship learning as adversarial optimization, achieving near-expert performance but requiring online policy rollouts during training---infeasible in strictly offline educational settings.

Jarrett et al.~\citep{jarrett2020strictly} proposed \emph{Energy-based Distribution Matching} (EDM) to address strictly offline apprenticeship learning. The key insight is to identify the discriminator between expert and learner state occupancies with an energy-based model of the expert state distribution. Concretely, given a dataset of expert trajectories $\mathcal{D} = \{(s_t, a_t)\}$, EDM minimizes the KL divergence between the learner's state occupancy measure $\rho_\pi$ and the expert's $\rho_E$:
\[
  \min_\pi \; D_\mathrm{KL}\!\left(\rho_E \,\|\, \rho_\pi\right)
\]
using an energy function $E_\theta(s)$ that models the log-density of the expert state distribution, avoiding the need to roll out $\pi$ in the environment. This makes EDM uniquely suited to settings where only a static log of interactions is available and no simulator exists.

Prior work has applied apprenticeship learning directly to student interaction data from the same Pyrenees platform~\citep{islam2024generalized, islam2025generalized}: rather than gathering exploratory data as in DRL, these methods derive pedagogical policies directly from the decisions and behaviors students themselves demonstrated while interacting with the ITS, treating students' pedagogical choices as demonstrations and inducing policies that model observed decision-making behavior. Yang et al.~\citep{yang2025themes} extended this with \textsc{Themes}, a hierarchical apprenticeship learning framework that partitions student trajectories into temporally coherent sub-trajectories and induces cluster-specific policies to capture evolving pedagogical strategies. These works establish apprenticeship learning from student choice data as a viable alternative to reward-optimizing RL on this platform. Because apprenticeship learning imitates rather than optimizes, it can only be trained on interactions in which the demonstrator---here, the student---actually made the pedagogical decision, in contrast to RL corpora, which are commonly tutor- or randomly-driven (Section~\ref{sec:agency-relwork}); the present work extends this line of research by deploying the induced policy as a student-choice-with-override intervention in a live classroom study and directly evaluating how well it preserves student agency in deployment.

\subsection{ITS Policy Evaluation in the Classroom}

\begin{sloppypar}
Live classroom evaluations of ITS policies face well-documented statistical challenges: per-condition sample sizes are modest, pre-existing knowledge varies widely, and effect sizes for between-condition differences are typically small even when within-condition gains are large~\citep{vanlehn2011,chi2011empirically}. Standard practices include: excluding ceiling performers from inferential analyses; verifying pretest equivalence; using Normalized Learning Gain (NLG) to control for incoming competence~\citep{chi2011empirically}; and administering isomorphic posttests to assess transfer~\citep{ausin2019leveraging,abdelshiheed2023leveraging,hostetter2023xai}. Hostetter et al.~\citep{hostetter2023xai} further showed that delivering real-time personalized explanations alongside RL-induced interventions significantly improves student NLG on Pyrenees. We follow all of these conventions. A complementary, instructional-design route to the same differentiation goal has recently appeared in the CHI literature: rather than inducing a policy over an existing set of presentation modes, Dey Tithi et al.~\citep{tithi2026exploring} designed new cognitively-engaging worked-example variants for a logic tutor and showed their effectiveness varies systematically with students' prior knowledge. Our work instead holds the presentation modes fixed and asks how a policy should be induced to select among them, but both lines of work point to the same underlying HCI problem: matching the level and kind of instructional support to a learner's individual differences.
\end{sloppypar}

\section{System and Policies}
\label{sec:system}

\subsection{Pyrenees: A Probability ITS}

Pyrenees is a web-based ITS that teaches ten probability principles (e.g., the Complement Theorem and Bayes' Theorem)~\citep{ausin2020exploring,hostetter2023xai}. During training, students work through twelve problems and for each problem choose a presentation mode: Problem Solving (PS, independent), Worked Example (WE, tutor-presented), or Collaborative Problem Solving (CPS, co-constructed). Students proceed step-by-step---defining variables, applying probability principles, and solving equations---with immediate step-level feedback and on-demand hints. All assessments are delivered without hints or materials access, double-blind graded using a partial-credit rubric, and normalized to $[0, 1]$.

\textbf{Assessment.} The \emph{pretest} contains 8 open-ended probability problems and the \emph{standard posttest} contains 12. Both are administered without hints or access to materials, independently graded by two evaluators using a double-blind process with discrepancies resolved by discussion, and scores are normalized to $[0, 1]$. An \emph{isomorphic posttest} (\texttt{post\_iso}) contains structurally equivalent but surface-novel problems to assess transfer.

\textbf{Learning outcome metrics.} We report two normalized learning gain variants. The primary measure \texttt{NLG\_post} uses Hake's formula~\citep{chi2011empirically}:
\[
  \texttt{NLG} = \frac{\mathit{post} - \mathit{pre}}{1 - \mathit{pre}}
\]
A second variant \texttt{NLG\_post\_SR} uses the square-root denominator adopted by prior work on this platform~\citep{ausin2019leveraging,hostetter2023xai}:
\[
  \texttt{NLG\_SR} = \frac{\mathit{post} - \mathit{pre}}{\sqrt{1 - \mathit{pre}}}
\]
Both measures quantify learning efficiency independent of incoming competence. Analogous measures \texttt{NLG\_post\_iso} and \texttt{NLG\_post\_iso\_SR} apply the same formulas to the isomorphic posttest.

\subsection{Experimental Conditions}

All three conditions deployed on the same Pyrenees platform with the same content and feedback mechanisms; they differed only in how the presentation mode for each training problem was determined. \textsc{StuCQL} and \textsc{StuEDM} are \emph{student-choice-with-override} policies: the student first selects a presentation mode (PS, WE, or CPS), and the policy accepts that choice if it agrees; otherwise the policy overrides it with its own recommendation. \textsc{Expert} provides a fully uninformed control in which no student input is solicited.

\textbf{\textsc{Expert}.} The presentation mode is assigned uniformly at random from $\{\mathrm{PS}, \mathrm{WE}, \mathrm{CPS}\}$, independent of the student's state or preferences; because all three modes are themselves domain-expert-designed instructional strategies, this condition delivers only expert-authored content without any adaptive or student-driven guidance selecting among them. We adopt the \textsc{Expert} label used for the analogous non-adaptive, uniformly-assigned baseline in prior work on this platform~\citep{ausin2019leveraging}. This uncontrolled baseline captures the inherent learning potential of Pyrenees content independent of sequencing policy.

\begin{sloppypar}
\textbf{\textsc{StuCQL}.} Students freely choose their presentation mode for each problem. Their choice is honored unless it conflicts with the policy induced by Conservative Q-Learning (CQL)~\citep{kumar2020conservative}; when there is a conflict, the CQL-induced policy overrides the student's selection. CQL was trained offline on 2,716 students' interaction logs collected over 13 semesters of classroom deployments (Fall 2015--Spring 2022), using NLG as the reward signal~\citep{islam2024generalized}. Because CQL is reward-optimizing rather than imitation-based, its training corpus can draw on any historical interaction log regardless of whether the tutor or the student made the pedagogical decision at each step. The state representation comprises 130 continuous features capturing five categories of student behavior: 57 performance features (e.g., percentage of correct responses), 22 temporal features (e.g., average time per step), 31 problem-solving features (e.g., difficulty level), 10 hint-usage features (e.g., total hints requested), and 10 autonomy features (e.g., number of elicits since the last tell)~\citep{abdelshiheed2023leveraging, hostetter2023leveraging, islam2024generalized}. CQL's conservative regularization penalizes high Q-values for state-action pairs underrepresented in the training data, reducing overestimation under distributional shift.
\end{sloppypar}

\textbf{\textsc{StuEDM}.} Students freely choose their presentation mode for each problem. Their choice is honored unless it conflicts with the policy induced by Energy-based Distribution Matching (EDM)~\citep{jarrett2020strictly}; when there is a conflict, the EDM-induced policy overrides the student's selection. In contrast to \textsc{StuCQL}, EDM is an \emph{offline apprenticeship learning} algorithm that is trained to imitate demonstrated behavior rather than to optimize a reward, and so it can only be trained on interactions in which the demonstrator---here, the student---actually made the pedagogical decision. This yields a substantially smaller and shorter-span training corpus than CQL's: EDM was trained on the complete interaction trajectories of 244 students across five semesters in which students exercised free presentation-mode choice (S21, S22, S24, F24, and S25; 57, 56, 54, 44, and 33 students, respectively), using the same 130-dimensional state representation and 3-way action space as \textsc{StuCQL}. Rather than optimizing a reward, EDM learns a policy whose state occupancy distribution matches that of the demonstrations. The energy function $E_\theta(s)$ is parameterized by a neural network trained to assign low energy to states frequently visited in high-quality trajectories and high energy to rarely visited or low-quality-associated states; the resulting policy steers students toward high-density, high-quality states without reward engineering or online rollouts. Because real student demonstrations are rarely uniformly expert, each demonstrated decision is weighted by its estimated expertise---a sigmoid transform of standardized Normalized Learning Gain, $w = \sigma(\alpha \cdot \mathrm{NLG}_z)$ with $\alpha = 3.0$---following the expertise-weighting scheme introduced by \textsc{Halide}~\citep{islam2026hierarchical}: demonstrations from stronger performers receive proportionally more influence during policy induction, while weaker performers' decisions are down-weighted rather than discarded outright, so the induced policy is shaped predominantly by expert-like behavior without requiring a hand-curated expert-only subset. The resulting policy network is a 2-hidden-layer multilayer perceptron (256 units per layer, ELU activations) mapping the 130-dimensional state to the 3 action logits, trained for 100 update steps with batch size 32, using stochastic gradient Langevin dynamics for energy-model sampling (replay buffer size 10,000, SGLD learning rate 0.01).

\section{Experimental Setup}
\label{sec:setup}

\subsection{Participants}

This study evaluated three experimental conditions: \textsc{Expert} (condition \texttt{26110}), \textsc{StuCQL} (condition \texttt{26140}), and \textsc{StuEDM} (condition \texttt{26150}). A total of 104 students participated: \textsc{Expert} ($N = 26$), \textsc{StuCQL} ($N = 39$), and \textsc{StuEDM} ($N = 39$). \textsc{Expert} was allocated fewer students than \textsc{StuCQL} or \textsc{StuEDM} by original design (28 assigned vs.\ 43 for each of the other two conditions). Students with perfect pretest scores (\texttt{pre}~$= 1.0$) were excluded from inferential analyses to avoid ceiling effects on the learning gain metrics~\citep{ausin2020exploring, hostetter2023xai}, yielding final analytic samples of \textsc{StuCQL} ($N = 27$) and \textsc{StuEDM} ($N = 26$); the resulting \textsc{Expert} analytic sample ($N = 16$) is correspondingly underpowered, which we address next.

\subsection{Expert Condition Augmentation}
\label{sec:augmentation}

\begin{sloppypar}
To bring \textsc{Expert}'s analytic sample to a comparable size, we supplemented it with students from the immediately preceding offering of the same course on the same platform (Spring 2025), which included an identically implemented \textsc{Expert} condition: uniform assignment among PS/WE/CPS with no student input, the same 12 training problems, and the same pretest/posttest instruments. Before pooling, we verified the two semesters' \textsc{Expert} populations did not differ on incoming competence: pretest scores were statistically equivalent ($t(49) = 1.027$, $p = 0.310$; Mann-Whitney $p = 0.251$), as was time on task ($t(49) = 0.199$, $p = 0.843$). From the pool of 35 pretest-non-ceiling Spring 2025 \textsc{Expert} students, we drew, with a fixed random seed (1234), exactly the 11 students needed to bring \textsc{Expert}'s analytic sample to $N = 27$, matching \textsc{StuCQL}. All primary analyses below use this augmented \textsc{Expert} sample ($N = 27$, $N = 27$, $N = 26$ across \textsc{Expert}, \textsc{StuCQL}, \textsc{StuEDM}). We report the augmented comparison as the primary analysis given the strength of the equivalence checks, but note that \textsc{StuCQL} and \textsc{StuEDM} reflect only the Spring 2026 cohort: no comparably designed prior-semester condition exists for either, since the closest Spring 2025 analogs paired the CQL-induced override with real-time personalized explanations~\citep{hostetter2023xai} rather than the plain CQL- or EDM-induced override used here.
\end{sloppypar}

\subsection{Procedure}

Pyrenees was administered as a homework assignment in an undergraduate Computer Science course at a university during Spring 2026. Students were randomly assigned to conditions and told that full credit required completion; grading was based on demonstrated effort rather than performance~\citep{ausin2020exploring,hostetter2023xai}. The experiment followed four stages:

\begin{enumerate}
  \item \textbf{Textbook:} Students read an online textbook introducing the ten probability principles with examples.
  \item \textbf{Pretest:} Students solved probability problems independently without hints, feedback, or materials access.
  \item \textbf{ITS Training:} Students worked through twelve probability problems in Pyrenees under their assigned pedagogical policy. In the \textsc{StuCQL} and \textsc{StuEDM} conditions, students selected their preferred presentation mode for each problem; the respective policy intervened only when the student's choice conflicted with the induced policy. In the \textsc{Expert} condition, presentation modes were assigned randomly among the expert-designed activity types with no student input. All other system behavior (hints, step-level feedback) was identical across conditions.
  \item \textbf{Posttest:} Students completed the standard and isomorphic posttests under the same conditions as the pretest.
\end{enumerate}

\subsection{Completion Rates}

Completion rates did not significantly differ across conditions. Relative to all assigned students: $92.9\%$ ($26/28$) for \textsc{Expert}, $90.7\%$ ($39/43$) for \textsc{StuCQL}, and $90.7\%$ ($39/43$) for \textsc{StuEDM}, $\chi^2(2) = 0.123$, $p = 0.940$. Relative to students who started the tutor: $92.9\%$ ($26/28$), $95.1\%$ ($39/41$), and $97.5\%$ ($39/40$) respectively, $\chi^2(2) = 0.824$, $p = 0.662$. Differential attrition is therefore unlikely to explain any observed performance differences.

\section{Results}
\label{sec:results-ch7}

Sections~\ref{sec:pretest}--\ref{sec:between} and~\ref{sec:summary} address \textbf{RQ1} (learning outcomes); Section~\ref{sec:decisions} addresses \textbf{RQ2} (the decision-level mechanism behind those outcomes).

\subsection{Pretest Equivalence}
\label{sec:pretest}

Pretest scores did not significantly differ across conditions: one-way ANOVA, $F(2,77) = 0.207$, $p = 0.814$. A Kruskal-Wallis robustness check yielded the same conclusion, $H(2) = 1.054$, $p = 0.597$, indicating balanced incoming competence across all three conditions, including the augmented \textsc{Expert} sample. Having established that all three groups entered the study with equivalent incoming competence, we treat posttest performance, normalized learning gains, and time on task as the outcome measures that may legitimately differ across conditions (Sections~\ref{sec:gains}--\ref{sec:between}).

\subsection{Within-Condition Learning Gains}
\label{sec:gains}

We assessed within-condition learning using repeated one-way ANOVAs comparing pretest with posttest scores (Table~\ref{tab:results}).

\textbf{Standard posttest.} All three conditions showed significant pre-to-post gains: \textsc{Expert}, $F(1,26) = 11.058$, $p = 0.003$, partial $\eta^2 = 0.298$; \textsc{StuCQL}, $F(1,26) = 28.778$, $p < 0.0005$, partial $\eta^2 = 0.525$; \textsc{StuEDM}, $F(1,25) = 31.991$, $p < 0.0005$, partial $\eta^2 = 0.561$.

\textbf{Isomorphic posttest.} The same pattern held for \texttt{post\_iso}: \textsc{Expert}, $F(1,26) = 23.691$, $p < 0.0005$, partial $\eta^2 = 0.477$; \textsc{StuCQL}, $F(1,26) = 57.429$, $p < 0.0005$, partial $\eta^2 = 0.688$; \textsc{StuEDM}, $F(1,25) = 42.312$, $p < 0.0005$, partial $\eta^2 = 0.629$.

All three policies therefore supported significant learning on both the standard and isomorphic posttests, validating the effectiveness of the underlying Pyrenees platform regardless of sequencing policy. The effect sizes were largest for \textsc{StuEDM} and \textsc{StuCQL} and smallest for \textsc{Expert}, consistent with the hypothesis that adaptive policies produce more effective learning than uncontrolled sequencing.

\subsection{Between-Condition Comparisons}
\label{sec:between}

Table~\ref{tab:results} summarizes mean learning outcomes. Following prior work on this platform, we test planned one-sided directional hypotheses that \textsc{StuEDM} outperforms the comparison conditions~\citep{ausin2020exploring, hostetter2023xai}.

\begin{table}[t]
\caption{Mean outcome measures by condition (students with \texttt{pre}~$= 1.0$ excluded; \textsc{Expert} is the augmented sample, Section~\ref{sec:augmentation}). Bold marks the highest mean per column. $^*$ indicates significant one-sided pairwise comparison vs.\ \textsc{StuCQL} (pretest-adjusted ANCOVA contrast for \texttt{post}, $t$-test for the remaining columns); $^\dagger$ indicates the analogous significant comparison vs.\ \textsc{Expert} ($\alpha = 0.05$). \texttt{Time} is hours spent on ITS training, reported as an engagement outcome (Section~\ref{sec:timeontask}).}
\label{tab:results}
\centering
\small
\begin{tabular}{lccccccc}
\toprule
\thead{Condition} & \thead{$N$} & \thead{\texttt{post}} & \thead{\texttt{post\_iso}}
          & \thead{\texttt{NLG\_post}} & \thead{\texttt{NLG\_post\_iso}}
          & \thead{\texttt{NLG\_post\_SR}} & \thead{\texttt{Time (h)}} \\
\midrule
\textsc{StuEDM}  & 26 & $\mathbf{0.908}^{*\dagger}$
                 & $\mathbf{0.948}$
                 & $\mathbf{0.685}^{\dagger}$
                 & $\mathbf{0.829}^{\dagger}$
                 & $\mathbf{0.323}^{\dagger}$
                 & $\mathbf{1.52}^{*}$ \\
\textsc{StuCQL}  & 27 & $0.842$ & $0.888$ & $0.485$
                 & $0.676$ & $0.239$ & $1.11$ \\
\textsc{Expert}  & 27 & $0.846$ & $0.885$ & $0.331$
                 & $0.600$ & $0.169$ & $1.44$ \\
\bottomrule
\end{tabular}
\end{table}

\textbf{Standard posttest (ANCOVA).} Controlling for pretest, the condition effect was not significant, $F(2,76) = 2.560$, $p = 0.084$, partial $\eta^2 = 0.063$. \textsc{StuEDM} produced the highest adjusted posttest numerically. Planned one-sided contrasts on the pretest-adjusted model showed \textsc{StuEDM} significantly outperformed both \textsc{StuCQL} ($t(76)=1.690$, $p = 0.048$) and \textsc{Expert} ($t(76)=2.154$, $p=0.017$); a planned one-sided contrast of \textsc{StuEDM} against the (unadjusted) average of \textsc{Expert} and \textsc{StuCQL} approached significance, $t(77) = 1.656$, one-sided $p = 0.051$.

\textbf{Normalized learning gain (\texttt{NLG\_post}).} The ANOVA was not significant, $F(2,77) = 2.387$, $p = 0.099$, partial $\eta^2 = 0.058$, but condition means were ordered in the expected direction: \textsc{StuEDM} ($M = 0.685$) $>$ \textsc{StuCQL} ($M = 0.485$) $>$ \textsc{Expert} ($M = 0.331$). A planned one-sided comparison showed \textsc{StuEDM} significantly outperformed \textsc{Expert}, $t(77) = 2.180$, one-sided $p = 0.016$; the comparison with \textsc{StuCQL} did not reach significance, $t(77) = 1.231$, one-sided $p = 0.111$. A planned one-sided contrast against the combined baseline was significant, $t(77) = 1.963$, one-sided $p = 0.027$.

\textbf{Isomorphic posttest (\texttt{post\_iso}).} The ANOVA was not significant, $F(2,77) = 1.625$, $p = 0.204$, but means favored \textsc{StuEDM} ($M = 0.948$) over \textsc{StuCQL} ($M = 0.888$) and \textsc{Expert} ($M = 0.885$). Neither pairwise comparison reached significance: \textsc{StuEDM} vs.\ \textsc{StuCQL}, $t(77) = 1.518$, one-sided $p = 0.067$; \textsc{StuEDM} vs.\ \textsc{Expert}, $t(77) = 1.611$, one-sided $p = 0.056$. A planned one-sided contrast against the combined baseline was significant, $t(77) = 1.800$, one-sided $p = 0.038$.

\textbf{Isomorphic NLG (\texttt{NLG\_post\_iso}).} The ANOVA was not significant, $F(2,77) = 1.664$, $p = 0.196$, but means were ordered as expected: \textsc{StuEDM} ($M = 0.829$) $>$ \textsc{StuCQL} ($M = 0.676$) $>$ \textsc{Expert} ($M = 0.600$). A planned one-sided comparison showed \textsc{StuEDM} significantly outperformed \textsc{Expert}, $t(77) = 1.795$, one-sided $p = 0.038$; the comparison with \textsc{StuCQL} did not reach significance, $t(77) = 1.197$, one-sided $p = 0.117$. A planned one-sided contrast against the combined baseline was significant, $t(77) = 1.722$, one-sided $p = 0.045$.

\textbf{Score-ratio NLG (\texttt{NLG\_post\_SR}).} The ANOVA was not significant, $F(2,77) = 2.471$, $p = 0.091$. \textsc{StuEDM} significantly outperformed \textsc{Expert}, $t(77) = 2.221$, one-sided $p = 0.015$, though the comparison with \textsc{StuCQL} did not reach significance, $t(77) = 1.216$, one-sided $p = 0.114$. A planned one-sided contrast against the combined baseline was significant, $t(77) = 1.978$, one-sided $p = 0.026$.

\begin{sloppypar}
\textbf{Time on task (engagement).}
\label{sec:timeontask}
We additionally treat time spent on ITS training as an outcome measure of student engagement: within a fixed homework-completion structure where grading was based on effort rather than performance, students freely chose how long to remain engaged with the tutor, so more time on task reflects greater voluntary engagement rather than a cost to be minimized. Time on task differed significantly across conditions overall: one-way ANOVA, $F(2,77) = 4.206$, $p = 0.019$. Pairwise $t$-tests show \textsc{StuEDM} students engaged with the tutor significantly longer than \textsc{StuCQL} students ($M=1.52$h vs.\ $M=1.11$h; $t(77)=2.714$, $p=0.008$), and \textsc{StuCQL} students engaged for significantly less time than \textsc{Expert} students ($M=1.11$h vs.\ $M=1.44$h; $t(77)=-2.232$, $p=0.029$); \textsc{StuEDM} and \textsc{Expert} did not differ ($t(77)=0.503$, $p=0.617$). The \textsc{StuEDM}-vs-\textsc{StuCQL} engagement difference is present in the S26-only data as well ($t(51)=2.419$, $p=0.019$, since neither condition's data changed under augmentation), so it is not an artifact of the \textsc{Expert} augmentation. \textsc{StuEDM} thus produced both the highest learning outcomes and the highest voluntary engagement with the tutor among the two student-choice-with-override conditions.
\end{sloppypar}

\subsection{RQ2: Student Agency --- Decision Distribution and Policy Overrides}
\label{sec:decisions}

To understand the behavioral mechanism behind these outcome differences, we analyzed all 390 problem-level presentation-mode decisions logged for \textsc{StuCQL} and \textsc{StuEDM} (39 students $\times$ 10 choice-eligible problems each), reconstructing, for every decision, the student's own stated choice and the policy's applied (post-override) decision among Problem Solving (PS), Worked Example (WE), and Collaborative Problem Solving (CPS).

\begin{sloppypar}
\textbf{Students' own choices are equivalent at the first decision but diverge over the remaining nine.} Because \textsc{StuCQL} and \textsc{StuEDM} override students differently, any difference in students' own subsequent choice-making is of interest independent of what actually gets delivered. We therefore examined students' own stated choice (prior to any override) separately for the first choice-eligible problem in the training sequence---the second problem overall, since the first training problem carries no presentation-mode choice---and for the remaining nine. At the first choice-eligible problem, students in the two conditions chose similarly (\textsc{StuEDM}: 51.3\% PS / 20.5\% WE / 28.2\% CPS; \textsc{StuCQL}: 56.4\% PS / 15.4\% WE / 28.2\% CPS; $\chi^2(2) = 0.38$, $p = 0.827$), confirming the two student populations began the study with statistically indistinguishable choice preferences. Across the remaining nine problems, however, students' own choices diverged significantly between conditions (\textsc{StuEDM}: 62.7\% PS / 21.9\% WE / 15.4\% CPS; \textsc{StuCQL}: 41.6\% PS / 26.5\% WE / 31.9\% CPS; $\chi^2(2) = 36.73$, $p < .001$). Since the two populations were equivalent at the outset, this divergence is consistent with each policy's differing override behavior reshaping students' subsequent self-directed choices over the course of the training sequence, not merely the content ultimately delivered to them.
\end{sloppypar}

\begin{figure}[t]
  \centering
  \includegraphics[width=\linewidth,alt={Two grouped bar charts. Left panel (StuCQL): student choice is 43 percent PS, 25 percent WE, 32 percent CPS, while the final delivered decision is 23 percent PS, 77 percent WE, 0 percent CPS. Right panel (StuEDM): student choice is 62 percent PS, 22 percent WE, 17 percent CPS, while the final delivered decision is 82 percent PS, 16 percent WE, 2 percent CPS.}]{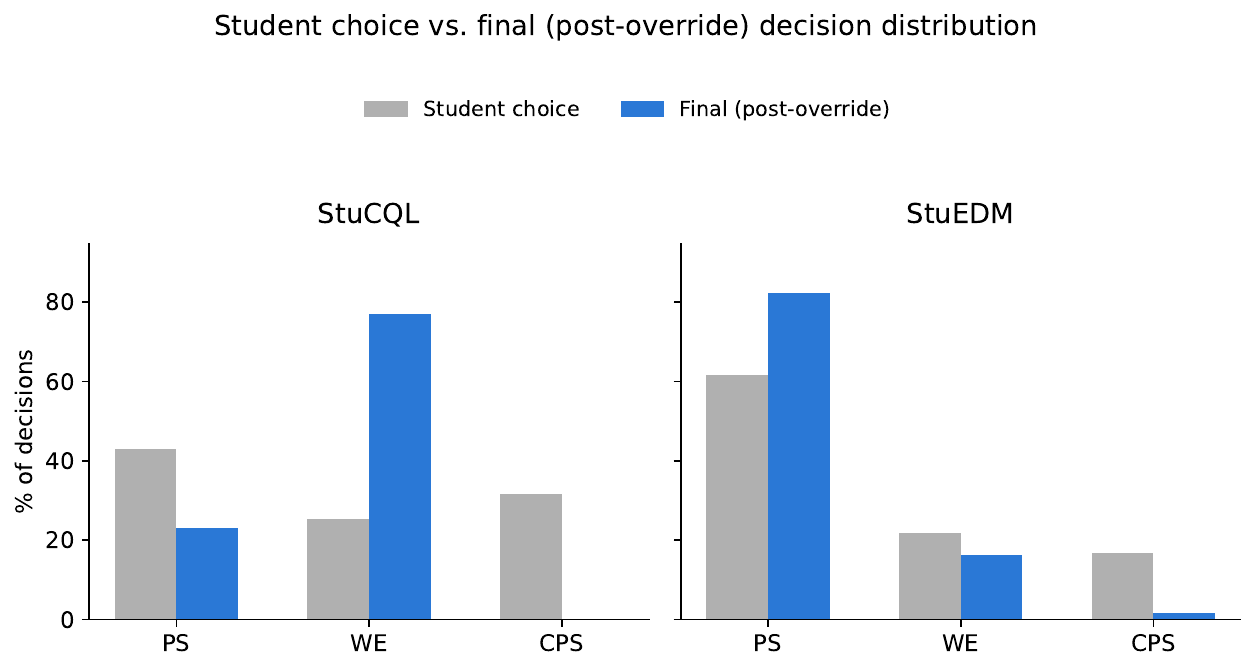}
  \caption{Student's own presentation-mode choice vs.\ the policy's final (post-override) decision, for \textsc{StuCQL} and \textsc{StuEDM}.}
  \label{fig:decision-dist}
\end{figure}

\begin{sloppypar}
\textbf{Decision distributions diverge sharply after override.} Figure~\ref{fig:decision-dist} compares each condition's student-choice distribution against its final applied distribution. The two student populations' own choices were broadly similar in rank order (PS the modal choice in both: 43.1\% PS / 25.4\% WE / 31.5\% CPS for \textsc{StuCQL}'s students; 61.5\% PS / 21.8\% WE / 16.7\% CPS for \textsc{StuEDM}'s students), but the \emph{delivered} decisions diverge sharply: \textsc{StuCQL} delivers WE on 76.9\% of decisions (PS 23.1\%, CPS 0.0\%), while \textsc{StuEDM} delivers PS on 82.3\% of decisions (WE 16.2\%, CPS 1.5\%). For reference, \textsc{Expert} assigns all three modes near-uniformly (32.3\% / 35.4\% / 32.3\%), confirming that this divergence is driven by the induced policies themselves rather than by differences in who was assigned to each condition. All three pairwise differences in the applied decision distribution are individually significant: \textsc{StuEDM} vs.\ \textsc{StuCQL}, $\chi^2(2) = 290.57$, $p < .001$; \textsc{StuEDM} vs.\ \textsc{Expert}, $\chi^2(2) = 193.45$, $p < .001$; \textsc{StuCQL} vs.\ \textsc{Expert}, $\chi^2(2) = 175.60$, $p < .001$.
\end{sloppypar}

\begin{figure}[t]
  \centering
  \includegraphics[width=0.5\linewidth,alt={Stacked bar chart. StuCQL: 33 percent agreement, 67 percent override. StuEDM: 72 percent agreement, 28 percent override.}]{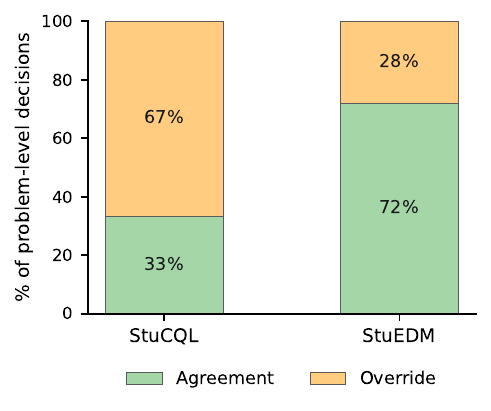}
  \caption{Agreement vs.\ override rate for \textsc{StuCQL} and \textsc{StuEDM}.}
  \label{fig:override-rate}
\end{figure}

\textbf{\textsc{StuCQL} overrides far more often than \textsc{StuEDM}.} Figure~\ref{fig:override-rate} shows that CQL overrode the student's stated choice on 66.7\% of decisions (260/390), whereas EDM overrode on only 28.2\% (110/390)---a highly significant difference, $\chi^2(1) = 114.15$, $p < .001$. This asymmetry is also visible at the student level: 12 of 39 \textsc{StuCQL} students were overridden on every single one of their 10 problems, while 13 of 39 \textsc{StuEDM} students were never overridden at all.

\begin{sloppypar}
\textbf{The two policies override toward opposite modes.} Both conditions contributed the same 390 problem-level decisions (39 students $\times$ 10 problems); Table~\ref{tab:override-transitions} restricts to the subset actually overridden in each condition (260/390 for \textsc{StuCQL}, 110/390 for \textsc{StuEDM}, per Figure~\ref{fig:override-rate}) and breaks down what the student originally chose versus what the policy substituted. \textsc{StuCQL}'s overrides converge almost entirely on WE (85.8\% of all its overrides land on WE, 14.2\% on PS) and \emph{never once} override toward CPS, regardless of the student's original choice---CQL effectively eliminates the collaborative option from the delivered curriculum. \textsc{StuEDM}'s overrides converge instead on PS (81.8\% of its overrides land on PS), but unlike CQL it does not eliminate any mode entirely: a small fraction of its overrides (2.7\%) still route students toward CPS.
\end{sloppypar}

\begin{table}[t]
\caption{Override transition counts: student's original choice (rows) vs.\ the policy's override target (cols). Both conditions contributed the same 390 decisions (39 students $\times$ 10 problems each; Section~\ref{sec:setup}); this table restricts to the subset that was overridden --- 260/390 for \textsc{StuCQL} and 110/390 for \textsc{StuEDM}, reflecting the agreement/override rates in Figure~\ref{fig:override-rate}. Diagonal is empty by construction (agreements are excluded).}
\label{tab:override-transitions}
\centering
\small
\begin{tabular}{l ccc c ccc}
\toprule
 & \multicolumn{3}{c}{\textsc{StuCQL} (260/390 overridden)} & & \multicolumn{3}{c}{\textsc{StuEDM} (110/390 overridden)} \\
\cmidrule{2-4} \cmidrule{6-8}
\thead{Student chose} & \thead{PS} & \thead{WE} & \thead{CPS} & & \thead{PS} & \thead{WE} & \thead{CPS} \\
\midrule
PS  & --  & 115 & 0 & & --  & 8  & 1 \\
WE  & 22  & --  & 0 & & 37  & -- & 2 \\
CPS & 15  & 108 & -- & & 53  & 9  & -- \\
\bottomrule
\end{tabular}
\end{table}

\subsection{Summary of Results}
\label{sec:summary}

\begin{sloppypar}
Across all five outcome measures, \textsc{StuEDM} produced the highest means. While three-group ANOVAs remained non-significant even with the augmented, comparably sized samples ($N = 26$--$27$ per condition)---consistent with the modest effect sizes typical of single-semester classroom studies---the pattern of planned comparisons is directionally consistent but asymmetric across the two comparison conditions. \textsc{StuEDM} significantly outperformed \textsc{Expert} on four of five measures (adjusted posttest, \texttt{NLG\_post}, \texttt{NLG\_post\_iso}, \texttt{NLG\_post\_SR}; \texttt{post\_iso} approached significance, $p=0.056$). Against \textsc{StuCQL}, only the pretest-adjusted posttest comparison reached significance ($p=0.048$); the remaining direct \textsc{StuEDM}-vs-\textsc{StuCQL} comparisons did not ($p=0.067$--$0.117$). \textsc{StuEDM} nonetheless significantly outperformed the combined average of \textsc{Expert} and \textsc{StuCQL} on four of five measures, with the fifth (\texttt{post}) approaching significance ($p=0.051$). This pattern indicates that \textsc{StuEDM}'s advantage over the \textsc{Expert} baseline is the more robust finding in this study, while its advantage over \textsc{StuCQL} specifically is directionally consistent across all five measures but reaches significance only when pretest is covaried.
\end{sloppypar}

\section{Discussion and Conclusions}
\label{sec:discussion-ch7}

We introduced \textsc{StuEDM}, the first application of Energy-based Distribution Matching~\citep{jarrett2020strictly} to student-choice-with-override pedagogical policy induction in a deployed ITS. In \textsc{StuEDM}, students retain full autonomy over their presentation mode selection; the EDM-derived policy---learned from a substantially smaller, student-choice-only demonstration corpus---intervenes only when the student's choice conflicts with the EDM-derived recommendation. We compared it against \textsc{StuCQL} (an analogous student-choice-with-override policy using CQL trained on a large interaction corpus) and \textsc{Expert} assignment in a classroom study on the Pyrenees probability tutor.

\textbf{All policies support learning.} Every condition produced large, significant pre-to-post gains (partial $\eta^2 = 0.29$--$0.69$), confirming the robustness of the Pyrenees platform---consistent with prior evaluations~\citep{ausin2020exploring,hostetter2023xai}. The effect sizes were ordered \textsc{StuEDM} $>$ \textsc{StuCQL} $>$ \textsc{Expert} on both posttests, suggesting that adaptive policies confer an advantage even when all conditions learn.

\begin{sloppypar}
\textbf{EDM-based apprenticeship learning outperforms an expert-designed baseline, and trends ahead of reward-optimizing RL.} \textsc{StuEDM} significantly outperformed a pretest-equivalent, size-matched \textsc{Expert} baseline on four of five measures, significantly outperformed \textsc{StuCQL} on the pretest-adjusted posttest, and outperformed the combined baseline on four of five measures---despite being trained on 244 students' student-choice trajectories from five semesters, an order of magnitude fewer students and less than half the semester span of \textsc{StuCQL}'s 2,716-student, 13-semester training corpus. The direct \textsc{StuEDM}-vs-\textsc{StuCQL} comparison on the remaining four measures was directionally consistent with \textsc{StuEDM}'s advantage but did not reach significance ($p = 0.067$--$0.117$); we report this honestly rather than emphasizing only the measures that happened to cross $\alpha=0.05$. \textsc{StuEDM} also produced the highest engagement of the three conditions: \textsc{StuEDM} students spent significantly more time in the tutor than \textsc{StuCQL} students ($p = 0.008$; Section~\ref{sec:timeontask}), while not differing from \textsc{Expert} on time on task. Rather than treating this as a confound to be explained away, we read it as a further engagement-related benefit of \textsc{StuEDM}: within a fixed homework-completion structure, students voluntarily chose to engage with the \textsc{StuEDM}-governed tutor for longer, alongside also learning more. The training-corpus-size advantage for the algorithm itself is a genuine efficiency gain at the data-collection level, independent of this engagement finding.
\end{sloppypar}

This pattern is noteworthy for two reasons. First, \textsc{StuEDM}'s robust advantage over an uninformed \textsc{Expert} baseline shows that occupancy-matching from a modest student-choice corpus is sufficient to produce a pedagogically meaningful policy, without reward engineering or a large interaction corpus. Second, its advantage over \textsc{StuCQL}---which already incorporates a conservative penalty to reduce distributional-shift overestimation---is directionally consistent across every measure we examined but is only significant once pretest is covaried; we treat this as suggestive rather than conclusive evidence that reward-free occupancy matching is a better-suited inductive bias than conservative reward-optimizing RL for this student-choice-with-override setting, and return to what would strengthen this claim in Limitations. EDM's practical appeal remains regardless: many educational deployments cannot assemble the large, multi-year tutor-decision logs that RL-based policies require, but can still capture the smaller-scale, student-choice interaction data that offline apprenticeship learning needs.

\begin{sloppypar}
The decision-level analysis in Section~\ref{sec:decisions} offers a concrete behavioral account of this advantage. \textsc{StuCQL} overrode students on 66.7\% of decisions---nearly the opposite of a system that merely nudges students at the margins---and its overrides converged almost entirely on Worked Examples while never once routing a student toward Collaborative Problem Solving, regardless of what the student had chosen. \textsc{StuEDM}, by contrast, overrode students on only 28.2\% of decisions, respecting the student's own choice more than twice as often, and when it did override it converged on Problem Solving while still preserving a small share of Collaborative Problem Solving assignments. A plausible interpretation is that \textsc{StuCQL}'s reward signal (delayed NLG) introduces noise from confounds unrelated to the override policy---student effort, test-taking strategy, or other factors that affect NLG but are outside the policy's control---and that this noisy signal is amplified by how frequently CQL chooses to override, pushing the majority of students toward a single, more didactic mode (WE) largely irrespective of their own preference. EDM instead directly imitates the state-occupancy patterns of successful sessions, overriding far less often and, when it does, steering students toward more independent practice (PS) rather than passive instruction---a pattern more consistent with respecting student agency while still providing guidance. The position-based analysis in Section~\ref{sec:decisions} reinforces this account: students in the two conditions chose equivalently at the very first choice-eligible problem, so the significant divergence in their own choices over the remaining nine problems is better explained by each policy's differing override behavior reshaping subsequent student decision-making than by any pre-existing difference between the two student populations.
\end{sloppypar}

\begin{sloppypar}
\textbf{Design implications: override transparency and mode diversity.} The override-transition asymmetry in Section~\ref{sec:decisions} raises a design concern that learning-outcome measures alone do not capture: \textsc{StuCQL} did not merely override more often, it eliminated an entire pedagogical mode (Collaborative Problem Solving) from every student's delivered curriculum, regardless of that student's own preference---behaving, on this dimension, closer to an instructor-controlled policy than a student-choice-with-override one, despite nominally soliciting student input. This suggests two design directions for student-choice-with-override systems. First, override rate and delivered-mode coverage should be reported and monitored as first-class design metrics alongside learning outcomes, since a policy can win on learning gain while still eroding the autonomy guarantees the student-choice-with-override design was meant to provide---echoing recent CHI findings that users of adaptive systems consistently want more transparency into, and control over, the personalization applied to them, not just more accurate personalization~\citep{wu2026rethinking}. Second, systems could surface a lightweight rationale when overriding a student's choice---particularly when a policy is about to override the large majority of students toward the same single mode, as \textsc{StuCQL} did toward Worked Examples---so students can distinguish a considered pedagogical intervention from a policy defaulting to one mode. Such rationales must be designed carefully, however: explanations for an automated recommendation do not reliably improve human-AI team performance and can instead increase uncritical acceptance of the recommendation regardless of its correctness~\citep{bansal2021whole}, so an override rationale should be evaluated for its effect on students' decision-making, not deployed on the assumption that transparency alone helps. \textsc{StuEDM}'s higher agreement rate and more balanced override targets suggest that occupancy-matching objectives may be inherently more conservative interveners than reward-optimizing ones, a property worth preserving by design as these systems are extended to new domains.
\end{sloppypar}

\textbf{Non-significance of three-group ANOVAs.} The three-group ANOVAs were non-significant across all measures, consistent with the statistical challenges of detecting modest effect sizes ($\eta^2 \approx 0.04$--$0.06$) even with comparably sized per-condition samples of $N = 26$--$27$. The convergent pattern of one-sided planned comparisons, all in the expected direction and mostly statistically significant, is the appropriate evidence base for pre-specified directional hypotheses~\citep{hostetter2023xai}.

\begin{sloppypar}
\textbf{Variance asymmetry and tail risk.} A closer look at the within-condition spread clarifies why \textsc{StuEDM}'s large mean advantage does not translate into significance everywhere. On \texttt{NLG\_post}, \textsc{StuEDM}'s distribution is comparatively tight ($SD = 0.338$, minimum $= -0.08$)---essentially no \textsc{StuEDM} student scored meaningfully worse on the posttest than the pretest. \textsc{StuCQL} ($SD = 0.496$, minimum $= -1.00$) and especially \textsc{Expert} ($SD = 0.822$, minimum $= -3.00$) both have a long left tail of students with substantially negative NLG. Because NLG's denominator is $(1-\mathit{pre})$, even a small posttest dip for a high-pretest student is amplified into an extreme negative value, and a handful of such students is enough to sharply inflate a condition's variance. This asymmetry works against the omnibus and pairwise tests, whose standard errors are built from this inflated within-group variance, even though the point-estimate mean gap is large and consistently in \textsc{StuEDM}'s favor. Framed differently, \textsc{StuEDM} does not only win on average: it also appears to protect against the worst outcomes, with no student regressing meaningfully from pretest to posttest, whereas both comparison conditions include a minority of students who did. Whether this tail-risk reduction is a direct byproduct of imitating high-density state-occupancy patterns---and thus a distinct benefit of occupancy matching beyond the mean-outcome comparisons emphasized above---is a question we did not pre-register and cannot test with the same confirmatory rigor as the planned comparisons, but it is a pattern future work with larger samples should examine directly.
\end{sloppypar}

\textbf{Implications for offline ITS policy design.} Our results suggest that EDM-based offline apprenticeship learning is a compelling alternative to offline RL for student-choice-with-override policy induction. EDM requires no reward engineering, operates from a substantially smaller student-choice demonstration corpus, avoids distributional-shift overestimation, and produces policies compatible with student autonomy---a natural fit for settings where only student-choice interaction logs are available and reward specification is difficult. The student-choice-with-override design is important in its own right: it respects student agency while still providing adaptive guidance, which may contribute to engagement and motivation beyond pure learning gain~\citep{chi2011empirically}. Future work should investigate whether EDM's advantage persists across multiple ITS domains, with varying demonstration corpus sizes, and when demonstration quality weighting is varied (e.g., sharper or flatter NLG-based weighting).

\begin{sloppypar}
\textbf{Limitations and future work.} Several limitations apply. Per-condition sample sizes after ceiling exclusions are modest ($N = 26$--$27$), limiting ANOVA power; this is likely why the direct \textsc{StuEDM}-vs-\textsc{StuCQL} comparison is directionally consistent across all five measures but reaches significance only for the pretest-adjusted posttest---the effect sizes are comparable to the significant \textsc{StuEDM}-vs-\textsc{Expert} comparisons, and a larger or multi-semester replication would give this comparison a fairer test. \textsc{Expert}'s analytic sample combines Spring 2026 students with a pretest-, materials-, and instrument-equivalent Spring 2025 cohort (Section~\ref{sec:augmentation}); while every equivalence check we could run supported pooling, \textsc{StuCQL} and \textsc{StuEDM} reflect only Spring 2026, so residual semester-level confounds we could not test for (e.g., unmeasured cohort differences) cannot be fully ruled out for comparisons involving \textsc{Expert}. Relatedly, \textsc{StuEDM} students spent significantly more time engaging with the tutor than \textsc{StuCQL} students (Section~\ref{sec:timeontask}); we treat this as a positive engagement outcome rather than a confound, but because time on task was not experimentally manipulated, future work should test whether higher voluntary engagement is a general property of occupancy-matching policies like EDM or specific to this deployment. The weighting of demonstrations for EDM training warrants further study: we weighted each demonstrated decision by a sigmoid transform of standardized NLG following the scheme used by \textsc{Halide}~\citep{islam2026hierarchical}, but alternative schemes---hard Quantized Learning Gain filtering~\citep{mao2018tracing}, different sigmoid sharpness, or per-decision (rather than per-trajectory) quality signals as in \textsc{Halide}'s decision-level extensions---may further improve performance. The sensitivity of EDM to demonstration corpus size and weighting is an open question. Finally, while we treat time on task as one engagement indicator, we did not measure other engagement or affective states (e.g., self-reported interest or frustration), which may further mediate the effects of the override mechanism on learning. Future work will assess \textsc{StuEDM} across multiple semesters, investigate combining EDM with interpretability techniques~\citep{hostetter2023leveraging} to explain override decisions to students, and explore whether personalized explanations~\citep{hostetter2023xai} can further amplify its learning gains.
\end{sloppypar}

\section{Ethics and Privacy Statement}

This study was conducted under an IRB-approved protocol at the authors' institution. All student data (interaction logs, pretest and posttest scores) were collected as part of a normal homework assignment, de-identified prior to analysis, and stored securely; no personally identifiable information is reported in this chapter. Because \textsc{StuCQL} and \textsc{StuEDM} always honor a student's own presentation-mode choice except in cases of policy disagreement, no student was denied instructional content or subjected to a fully automated decision without input into their own learning path. We do not anticipate societal risks beyond those already considered by the IRB review of the study design; the induced policies are specific to the Pyrenees probability tutor and are not intended for deployment outside of supervised educational research settings without further validation.

\chapter{Conclusions and Future Work}
\label{chap:conc}

\section{Conclusions}
\label{sec:summ}

This dissertation investigated the challenges of learning decision-making policies from offline, human-centric demonstration data, where reward functions are unknown, expert trajectories are scarce, and human behaviors are heterogeneous, evolving, and imperfect. We developed a progression of five apprenticeship learning frameworks, organized around two central challenges---\textbf{reward modeling} and \textbf{demonstration quality}---and closed with a real-world classroom deployment that tested whether these retrospective advances translate into live educational benefit.

On the reward modeling front, the first contribution adapted the EM--Energy-based Distribution Matching (EM-EDM) framework to relax the assumption that all demonstrations share a single, homogeneous reward function. By jointly clustering student trajectories by their latent reward functions and inducing cluster-specific policies over continuous state spaces, EM-EDM outperformed four state-of-the-art AL baselines and two DRL baselines on pedagogical action prediction with as few as 24 expert demonstrations, establishing that heterogeneous reward modeling is both feasible and data-efficient. The second contribution addressed the further assumption that reward functions remain fixed within a trajectory. THEMES automatically partitions expert trajectories into temporally coherent sub-trajectories via a hierarchically learned reward regulator and induces policies over each partition using EM-EDM. Validated across one educational and three real-world clinical healthcare datasets, THEMES consistently outperformed all flat and hierarchical AL baselines---achieving an AUC of 0.899 and a Jaccard of 0.653 using only 18 past-semester demonstrations to predict future-semester decisions---establishing that capturing temporally evolving reward structure substantially improves policy induction in human-centric domains.

On the demonstration quality front, the third contribution challenged the assumption that demonstrations are uniformly optimal. HALIDE integrates continuous, Normalized-Learning-Gain-derived quality signals directly into the hierarchical policy induction process, ranking rather than discarding imperfect demonstrations. HALIDE showed that incorporating suboptimal trajectories \emph{without} quality-aware ranking degrades performance relative to unweighted hierarchical baselines, whereas ranked hierarchical learning enables principled reuse of mixed-quality data and yields consistent gains under cross-semester distribution shift---even among trajectories already labeled ``expert.'' The fourth contribution went further, challenging the assumption that a single trajectory-level quality rank is sufficient and that failed demonstrations carry no learnable signal. CCHAL redistributes trajectory-level demonstration quality to individual decisions via Bayesian credit assignment in feature-action space, separately scores each decision's criticality by contrasting expert against non-expert behavior in a dual-arm design, and combines both signals to sharpen a Bradley-Terry pairwise contrastive IRL objective over observed failure trajectories. Evaluated across three clinical EHR datasets and one intelligent tutoring system dataset, CCHAL outperformed flat and hierarchical AL baselines, including its direct predecessor HALIDE, establishing that decision-level outcome attribution and behavioral criticality are both necessary to fully exploit demonstration quality signals beyond trajectory-level weighting alone.

The fifth contribution closed the loop with a real-world classroom deployment. StuEDM is a pedagogical policy in which students make their own decisions about how to engage with each training problem, and an EDM-derived policy---trained on only 244 students' own pedagogical decisions across five semesters---intervenes only when it disagrees with the student's choice. In a controlled classroom study with 104 undergraduate students, StuEDM significantly improved posttest performance and normalized learning gain over a pretest-equivalent Expert baseline that assigns among the same domain-expert-designed presentation modes with no adaptive guidance, and significantly improved the pretest-adjusted posttest score over StuCQL---an analogous policy using Conservative Q-Learning trained on a much larger, 2,716-student, 13-semester historical corpus. A decision-level behavioral analysis showed that StuEDM overrode students' own decisions less than half as often as StuCQL, and that its overrides preserved more diversity across presentation modes, whereas StuCQL's overrides converged almost entirely on a single mode. StuEDM students also engaged with the tutor significantly longer than StuCQL students, a further engagement-related outcome favoring StuEDM; the underlying training-corpus efficiency of the algorithm itself, an order of magnitude smaller and less than half the semester span of StuCQL's, remains a genuine advantage regardless. Together, these results provide direct empirical evidence that offline AL from a substantially smaller, agency-driven demonstration corpus can rival reward-optimizing DRL trained on far larger corpora in a live educational deployment, while more directly respecting the student's own pedagogical agency.

Together, these five contributions trace a progression from modeling \emph{heterogeneous}, to \emph{evolving}, to \emph{imperfect}, to \emph{decision-level contrastive} reward structures, culminating in real-world empirical validation. This dissertation establishes offline apprenticeship learning as a practical, data-efficient, and ethically viable paradigm for human-centric environments, offering scalable tools for understanding and optimizing sequential decision-making in settings where expert knowledge is limited, rewards are ambiguous, online interaction is infeasible or unethical, and demonstrations vary widely in quality---with demonstrated applications spanning education and clinical healthcare, and validated in both retrospective evaluation and a live classroom deployment.

\section{Future Work}
\label{sec:future}

While this dissertation advances offline apprenticeship learning through heterogeneous, evolving, imperfect, and decision-level contrastive demonstration modeling, several promising directions remain.

A natural extension is to unify the separate mechanisms developed across chapters---hierarchical sub-trajectory partitioning, outcome-attributed decision weighting, dual-arm criticality scoring, and pairwise contrastive reward learning---into a single end-to-end framework, rather than the progressively layered structure adopted here. CCHAL's dual-arm criticality mechanism, for instance, relies on sufficient neighbors in each contrast pool and a fixed number of MT-TICC clusters; adaptive neighborhood sizing, cluster-level criticality estimates, and per-fold GP credit inference would each reduce sensitivity to dataset size without per-dataset tuning. Similarly, HALIDE's and StuEDM's demonstration-quality weighting both rely on a single sigmoid-transformed quality signal; systematically varying the sharpness of this transform, or moving to per-decision rather than per-trajectory quality signals as CCHAL begins to explore, is an open direction for both educational and clinical applications.

The StuEDM classroom deployment raises its own concrete follow-ups. Because StuEDM students spent significantly more time on task than StuCQL students, future work should test EDM-induced policies under matched time budgets to isolate the algorithmic efficiency gain from the effect of additional study time. The single-semester classroom study should be extended to a multi-semester replication to give the directionally consistent, but not always significant, StuEDM-versus-StuCQL comparisons a fairer test at higher statistical power. More broadly, the decision-level finding that StuCQL overrides students' choices far more often than StuEDM, and converges its overrides onto a single presentation mode, suggests that override rate and delivered-mode diversity should be reported as first-class design metrics for any policy that intervenes on a student's own decisions, not just its learning-gain outcomes; future systems might also surface a lightweight rationale when overriding a student's choice, evaluated directly for its effect on student decision-making rather than assumed to help.

Scaling these methods to high-dimensional and multi-modal human data---textual feedback, physiological signals, or richer interaction logs---represents an important avenue that will require improved representation learning and more robust energy-based optimization. Evaluation methodology is a second open area: current offline metrics may insufficiently capture human-centric objectives, motivating future work on causal off-policy estimators, counterfactual evaluation, and hybrid simulation--human studies that extend beyond the single live deployment demonstrated here. Integrating selective human-in-the-loop feedback, such as sparse preference queries or expert corrections, may further enhance model interpretability and reliability without requiring full online deployment. Finally, deeper theoretical analysis---covering generalization, identifiability, and stability for the energy-based and contrastive objectives developed across this dissertation---would strengthen its foundations and clarify the regimes under which these methods are reliable. Together, these directions offer a pathway toward more scalable, interpretable, agency-preserving, and trustworthy human-centric decision-making systems.

\bibliography{Mirajul-thesis}

@STRING{AN        = "Astrophys.\ Norv."}

@STRING{AR        = "Atmos.\ Res."}

@STRING{MA        = "Meteor.\ Appl."}

@inproceedings{hostetter2023leveraging,
  title={Leveraging Fuzzy Logic Towards More Explainable Reinforcement Learning-Induced Pedagogical Policies on Intelligent Tutoring Systems},
  author={Hostetter, John Wesley and Abdelshiheed, Mark and Barnes, Tiffany and Chi, Min},
  booktitle={2023 {IEEE} International Conference on Fuzzy Systems},
  year={2023},
  organization={IEEE}
}

@inproceedings{hostetter2023self,
    title={A Self-Organizing Neuro-Fuzzy Q-Network: Systematic Design with Offline Hybrid Learning},
    author={Hostetter, John Wesley and Abdelshiheed, Mark and Barnes, Tiffany and Chi, Min},
    booktitle={Proceedings of the 22nd International Conference on Autonomous Agents and Multiagent Systems (AAMAS)},
    year={2023}
}

@inproceedings{hostetter2023xai,
author = {Hostetter, John Wesley and Conati, Cristina and Yang, Xi and Abdelshiheed, Mark and Barnes, Tiffany and Chi, Min},
title = {XAI to Increase the Effectiveness of an Intelligent Pedagogical Agent},
year = {2023},
isbn = {9781450399944},
publisher = {Association for Computing Machinery},
address = {New York, NY, USA},
url = {https://doi.org/10.1145/3570945.3607301},
doi = {10.1145/3570945.3607301},
booktitle = {Proceedings of the 23rd ACM International Conference on Intelligent Virtual Agents},
articleno = {28},
numpages = {9},
location = {<conf-loc>, <city>W\"{u}rzburg</city>, <country>Germany</country>, </conf-loc>},
series = {IVA '23}
}

@inproceedings{abdelshiheed2023bridging,
  title={Bridging Declarative, Procedural, and Conditional Metacognitive Knowledge Gap Using Deep Reinforcement Learning},
  author={Abdelshiheed, Mark and Hostetter, John Wesley and Barnes, Tiffany and Chi, Min},
  booktitle={Proceedings of the Annual Meeting of the Cognitive Science Society},
  number={45},
  year={2023},
 series={CogSci '23'}
}

@inproceedings{abdelshiheed2023leveraging,
  title={Leveraging deep reinforcement learning for metacognitive interventions across intelligent tutoring systems},
  author={Abdelshiheed, Mark and Hostetter, John Wesley and Barnes, Tiffany and Chi, Min},
  booktitle={International Conference on Artificial Intelligence in Education},
  pages={291--303},
  year={2023},
  organization={Springer}
}

@inproceedings{abbeel2004apprenticeship,
  title={Apprenticeship learning via inverse reinforcement learning},
  author={Abbeel, Pieter and Ng, Andrew Y},
  booktitle={Proceedings of the 21st international conference on Machine learning},
  pages={1},
  year={2004},
  organization={ACM}
}

@inproceedings{ziebart2008maximum,
  title={Maximum entropy inverse reinforcement learning.},
  author={Ziebart, Brian D and Maas, Andrew L and Bagnell, J Andrew and Dey, Anind K and others},
  booktitle={AAAI},
  volume={8},
  pages={1433--1438},
  year={2008},
  organization={Chicago, IL, USA}
}

@article{ho2016generative,
  title={Generative adversarial imitation learning},
  author={Ho, Jonathan and Ermon, Stefano},
  journal={Advances in neural information processing systems},
  volume={29},
  year={2016}
}

@inproceedings{dimitrakakis2011bayesian,
  title={Bayesian multitask inverse reinforcement learning},
  author={Dimitrakakis, Christos and Rothkopf, Constantin A},
  booktitle={European workshop on reinforcement learning},
  pages={273--284},
  year={2011},
  organization={Springer}
}

@inproceedings{babes2011apprenticeship,
  title={Apprenticeship learning about multiple intentions},
  author={Babes, Monica and Marivate, Vukosi and Subramanian, Kaushik and Littman, Michael L},
  booktitle={Proceedings of the 28th ICML},
  pages={897--904},
  year={2011}
}

@article{yang2020student,
  title={Student Subtyping via EM-Inverse Reinforcement Learning.},
  author={Yang, Xi and Zhou, Guojing and Taub, Michelle and Azevedo, Roger and Chi, Min},
  journal={International Educational Data Mining Society},
  year={2020},
  publisher={ERIC}
}

@inproceedings{choi2012nonparametric,
  title={Nonparametric Bayesian inverse reinforcement learning for multiple reward functions},
  author={Choi, Jaedeug and Kim, Kee-Eung},
  booktitle={Advances in Neural Information Processing Systems},
  pages={305--313},
  year={2012}
}

@article{krishnan2016hirl,
  title={Hirl: Hierarchical inverse reinforcement learning for long-horizon tasks with delayed rewards},
  author={Krishnan, Sanjay and Garg, Animesh and Liaw, Richard and Miller, Lauren and Pokorny, Florian T and Goldberg, Ken},
  journal={arXiv preprint arXiv:1604.06508},
  year={2016}
}

@article{hausman2017multi,
  title={Multi-modal imitation learning from unstructured demonstrations using generative adversarial nets},
  author={Hausman, Karol and Chebotar, Yevgen and Schaal, Stefan and Sukhatme, Gaurav and Lim, Joseph J},
  journal={Advances in neural information processing systems},
  volume={30},
  year={2017}
}

@inproceedings{wang2020multi,
  title={Hierarchical Imitation Learning with Contextual Bandits for Dynamic Treatment Regimes},
  author={Wang, Lu and Yu, Wenchao and Cheng, Wei and Zong, Bo and Chen, Haifeng},
  booktitle={Reinforcement Learning for Real Life (RL4RealLife) Workshop in the 38 th ICML},
  year={2021}
}

@inproceedings{wang2022hierarchical,
  title={Hierarchical Imitation Learning via Subgoal Representation Learning for Dynamic Treatment Recommendation},
  author={Wang, Lu and Tang, Ruiming and He, Xiaofeng and He, Xiuqiang},
  booktitle={Proceedings of the Fifteenth ACM International Conference on Web Search and Data Mining}, 
  year={2022}
}

@article{levine2020offline,
  title={Offline reinforcement learning: Tutorial, review, and perspectives on open problems},
  author={Levine, Sergey and Kumar, Aviral and Tucker, George and Fu, Justin},
  journal={arXiv preprint arXiv:2005.01643},
  year={2020}
}

@article{jarrett2020strictly,
  title={Strictly batch imitation learning by energy-based distribution matching},
  author={Jarrett, Daniel and Bica, Ioana and van der Schaar, Mihaela},
  journal={Advances in Neural Information Processing Systems},
  volume={33},
  pages={7354--7365},
  year={2020}
}

@article{syed2012imitation,
  title={Imitation learning with a value-based prior},
  author={Syed, Umar and Schapire, Robert E},
  journal={arXiv preprint arXiv:1206.5290},
  year={2012}
}

@inproceedings{raza2012teaching,
  title={Teaching coordinated strategies to soccer robots via imitation},
  author={Raza, Saleha and Haider, Sajjad and Williams, Mary-Anne},
  booktitle={IEEE International Conference on Robotics and Biomimetics},
  pages={1434--1439},
  year={2012},
  organization={IEEE}
}

@article{pomerleau1991efficient,
  title={Efficient training of artificial neural networks for autonomous navigation},
  author={Pomerleau, Dean A},
  journal={Neural computation},
  volume={3},
  number={1},
  pages={88--97},
  year={1991},
  publisher={MIT Press One Rogers Street, Cambridge, MA 02142-1209, USA journals-info~…}
}

@inproceedings{klein2011batch,
  title={Batch, off-policy and model-free apprenticeship learning},
  author={Klein, Edouard and Geist, Matthieu and Pietquin, Olivier},
  booktitle={European Workshop on Reinforcement Learning},
  pages={285--296},
  year={2011},
  organization={Springer}
}

@inproceedings{jain2019model,
  title={Model-free IRL using maximum likelihood estimation},
  author={Jain, Vinamra and Doshi, Prashant and Banerjee, Bikramjit},
  booktitle={Proceedings of the AAAI},
  volume={33},
  pages={3951--3958},
  year={2019}
}

@article{lee2019truly,
  title={Truly batch apprenticeship learning with deep successor features},
  author={Lee, Donghun and Srinivasan, Srivatsan and Doshi-Velez, Finale},
  journal={arXiv preprint arXiv:1903.10077},
  year={2019}
}

@article{chan2021scalable,
  title={Scalable bayesian inverse reinforcement learning},
  author={Chan, Alex J and van der Schaar, Mihaela},
  journal={arXiv preprint arXiv:2102.06483},
  year={2021}
}

@article{kostrikov2018discriminator,
  title={Discriminator-actor-critic: Addressing sample inefficiency and reward bias in adversarial imitation learning},
  author={Kostrikov, Ilya and Agrawal, Kumar Krishna and Dwibedi, Debidatta and Levine, Sergey and Tompson, Jonathan},
  journal={International Conference on Learning Representations},
  year={2018}
}

@article{kostrikov2019imitation,
  title={Imitation learning via off-policy distribution matching},
  author={Kostrikov, Ilya and Nachum, Ofir and Tompson, Jonathan},
  journal={International Conference on Learning Representations},
  year={2020}
}

@inproceedings{yang2021multi,
  title={Multi-series time-aware sequence partitioning for disease progression modeling},
  author={Yang, Xi and Zhang, Yuan and Chi, Min},
  booktitle={Proceedings of the 30th International Joint Conference on Artificial Intelligence},
  year={2021}
}

@inproceedings{hallac2017toeplitz,
  title={Toeplitz inverse covariance-based clustering of multivariate time series data},
  author={Hallac, David and Vare, Sagar and Boyd, Stephen and Leskovec, Jure},
  booktitle={Proceedings of the 23rd ACM SIGKDD International Conference on Knowledge Discovery and Data Mining},
  pages={215--223},
  year={2017}
}

@inproceedings{ross2011reduction,
  title={A reduction of imitation learning and structured prediction to no-regret online learning},
  author={Ross, St{\'e}phane and Gordon, Geoffrey and Bagnell, Drew},
  booktitle={Proceedings of the fourteenth international conference on artificial intelligence and statistics},
  pages={627--635},
  year={2011},
  organization={JMLR Workshop and Conference Proceedings}
}

@article{azizsoltani2018adaptive,
  title={Adaptive sequential strategy for risk estimation of engineering systems using Gaussian process regression active learning},
  author={Azizsoltani, Hamoon and Sadeghi, Elham},
  journal={Engineering Applications of Artificial Intelligence},
  volume={74},
  pages={146--165},
  year={2018},
  publisher={Elsevier}
}

@inproceedings{finn2016guided,
  title={Guided cost learning: Deep inverse optimal control via policy optimization},
  author={Finn, Chelsea and Levine, Sergey and Abbeel, Pieter},
  booktitle={International conference on machine learning},
  pages={49--58},
  year={2016},
  organization={PMLR}
}

@article{grathwohl2019your,
  title={Your classifier is secretly an energy based model and you should treat it like one},
  author={Grathwohl, Will and Wang, Kuan-Chieh and Jacobsen, J{\"o}rn-Henrik and Duvenaud, David and Norouzi, Mohammad and Swersky, Kevin},
  journal={arXiv preprint arXiv:1912.03263},
  year={2019}
}

@inproceedings{welling2011bayesian,
  title={Bayesian learning via stochastic gradient Langevin dynamics},
  author={Welling, Max and Teh, Yee W},
  booktitle={Proceedings of the 28th international conference on machine learning (ICML-11)},
  pages={681--688},
  year={2011}
}

@inproceedings{zhang2019attain,
  title={ATTAIN: Attention-based Time-Aware LSTM Networks for Disease Progression Modeling.},
  author={Zhang, Yuan and Yang, Xi and Ivy Julie and Chi, Min},
  booktitle={In Proceedings of the 28th IJCAI},
  year={2019}
}

@article{komorowski2018artificial,
  title={The artificial intelligence clinician learns optimal treatment strategies for sepsis in intensive care},
  author={Komorowski, Matthieu and Celi, Leo A and Badawi, Omar and Gordon, Anthony C and Faisal, A Aldo},
  journal={Nature medicine},
  volume={24},
  number={11},
  pages={1716--1720},
  year={2018},
  publisher={Nature Publishing Group}
}

@inproceedings{azizsoltani2019unobserved,
  title={Unobserved Is Not Equal to Non-existent: Using Gaussian Processes to Infer Immediate Rewards Across Contexts.},
  author={Azizsoltani, Hamoon and Jin, Yeo},
  booktitle={In Proceedings of the 28th International Joint Conference on Artificial Intelligence},
  year={2019}
}

@inproceedings{fan2020theoretical,
  title={A theoretical analysis of deep Q-learning},
  author={Fan, Jianqing and Wang, Zhaoran and Xie, Yuchen and Yang, Zhuoran},
  booktitle={Learning for Dynamics and Control},
  pages={486--489},
  year={2020},
  organization={PMLR}
}

@inproceedings{raghu2017continuous,
  title={Continuous state-space models for optimal sepsis treatment: a deep reinforcement learning approach},
  author={Raghu, Aniruddh and Komorowski, Matthieu and Celi, Leo Anthony and Szolovits, Peter and Ghassemi, Marzyeh},
  booktitle={Machine Learning for Healthcare Conference},
  pages={147--163},
  year={2017},
  organization={PMLR}
}

@inproceedings{arora2021min,
  title={Min-Max Entropy Inverse RL of Multiple Tasks},
  author={Arora, Saurabh and Doshi, Prashant and Banerjee, Bikramjit},
  booktitle={IEEE International Conference on Robotics and Automation (ICRA)},
  pages={12639--12645},
  year={2021}
}

@article{vanlehn2006behavior,
  title={The behavior of tutoring systems},
  author={Vanlehn, Kurt},
  journal={International journal of artificial intelligence in education},
  volume={16},
  number={3},
  pages={227--265},
  year={2006},
  publisher={IOS Press}
}

@article{schnipke2002exploring,
  title={Exploring issues of examinee behavior: Insights gained from response-time analyses},
  author={Schnipke, Deborah L and Scrams, David J},
  journal={Computer-based testing: Building the foundation for future assessments},
  pages={237--266},
  year={2002}
}

@article{corbett1994knowledge,
  title={Knowledge tracing: Modeling the acquisition of procedural knowledge},
  author={Corbett, Albert T and Anderson, John R},
  journal={UMAP},
  volume={4},
  number={4},
  pages={253--278},
  year={1994},
  publisher={Springer}
}

@article{gonzalez2007innovative,
  title={Innovative applications of classroom response systems: Investigating students’ item response times in relation to final course grade, gender, general point average, and high school ACT scores},
  author={Gonz{\'a}lez-Espada, Wilson J and Bullock, Daniel W},
  journal={Electronic Journal for the Integration of Technology in Education},
  volume={6},
  pages={97--108},
  publisher={Citeseer},
  year={2007}
}

@article{kumar2020conservative,
  title={Conservative q-learning for offline reinforcement learning},
  author={Kumar, Aviral and Zhou, Aurick and Tucker, George and Levine, Sergey},
  journal={Advances in Neural Information Processing Systems},
  volume={33},
  pages={1179--1191},
  year={2020}
}

@inproceedings{DBLP:conf/aied/AusinMBC20,
  author    = {Markel Sanz Ausin and
               Mehak Maniktala and
               Tiffany Barnes and
               Min Chi},
  title     = {Exploring the Impact of Simple Explanations and Agency on Batch Deep
               Reinforcement Learning Induced Pedagogical Policies},
  booktitle = {AIED},
  pages     = {472--485},
  year      = {2020}
}

@inproceedings{rowe2015improving,
  title={Improving student problem solving in narrative-centered learning environments: A modular reinforcement learning framework},
  author={Rowe, Jonathan P and Lester, James C},
  booktitle={AIED},
  pages={419--428},
  year={2015},
  organization={Springer}
}

@inproceedings{shen2016reinforcement,
  title={Reinforcement Learning: the Sooner the Better, or the Later the Better?},
  author={Shen, Shitian and Chi, Min},
  booktitle={UMAP},
  pages={37--44},
  year={2016},
  organization={ACM}
}

@article{chi2011empirically,
  title={Empirically evaluating the application of reinforcement learning to the induction of effective and adaptive pedagogical strategies},
  author={Chi, Min and VanLehn, Kurt and Litman, Diane and Jordan, Pamela},
  journal={User Modeling and User-Adapted Interaction},
  volume={21},
  number={1-2},
  pages={137--180},
  year={2011},
  publisher={Springer}
}

@article{van2011effects,
  title={Effects of worked examples, example-problem, and problem-example pairs on novices’ learning},
  author={Van Gog, Tamara and Kester, Liesbeth and Paas, Fred},
  journal={Contemporary Educational Psychology},
  volume={36},
  number={3},
  pages={212--218},
  year={2011},
  publisher={Elsevier}
}

@article{renkl2002,
  title={From example study to problem solving: Smooth transitions help learning},
  author={Renkl, Alexander and Atkinson, Robert K and Maier, Uwe H and Staley, Richard},
  journal={J Exp Educ},
  volume={70},
  number={4},
  pages={293--315},
  year={2002},
  publisher={Taylor \& Francis}
}

@incollection{najar2014,
  title={Adaptive Support versus Alternating Worked Examples and Tutored Problems: Which Leads to Better Learning?},
  author={Najar, Amir Shareghi and Mitrovic, Antonija and McLaren, Bruce M},
  booktitle={UMAP},
  pages={171--182},
  year={2014},
  publisher={Springer}
}

@inproceedings{wang2017interactive,
  title={Interactive narrative personalization with deep reinforcement learning},
  author={Wang, Pengcheng and Rowe, Jonathan and Min, Wookhee and Mott, Bradford and Lester, James},
  booktitle={IJCAI},
  year={2017}
}

@article{guojing2017tutorstudent,
  title={The Impact of Decision Agency \& Granularity on Aptitude Treatment Interaction in Tutoring},
  author={Guojing Zhou and Min Chi},
  journal={CogSci},
  pages={3652--3657},
  year={2017}
}

@String{Computing = "Computing" }

@String{Computer = "{IEEE} Computer" }

@String{Chelsea = "Chelsea" }

@String{Springer = "Springer-Verlag" }

@article{guo2005mutual,
  title={Mutual information and minimum mean-square error in Gaussian channels},
  author={Guo, Dongning and Shamai, Shlomo and Verd{\'u}, Sergio},
  journal={IEEE Transactions on Information Theory},
  volume={51},
  number={4},
  pages={1261--1282},
  year={2005},
  publisher={IEEE}
}

@article{iglesias2009learning,
  title={Learning teaching strategies in an adaptive and intelligent educational system through reinforcement learning},
  author={Iglesias, Ana and Mart{\'\i}nez, Paloma and Aler, Ricardo and Fern{\'a}ndez, Fernando},
  journal={Applied Intelligence},
  volume={31},
  number={1},
  pages={89--106},
  year={2009},
  publisher={Springer}
}

@misc{baselines,
  author = {Dhariwal, P. and Hesse, C. and Klimov, O. and Nichol, A. and Plappert, M. and Radford, A. and Schulman, J. and Sidor, S. and Wu, Y. and Zhokhov, P.},
  title = {OpenAI Baselines},
  year = {2017},
  publisher = {GitHub},
  journal = {GitHub repository},
  howpublished = {\url{https://github.com/openai/baselines}},
}

@book{sutton2018reinforcement,
  title={Reinforcement learning: An introduction},
  author={Sutton, Richard S and Barto, Andrew G},
  year={2018},
  publisher={MIT press}
}

@article{vanlehn2011,
  title={The relative effectiveness of human tutoring, intelligent tutoring systems, and other tutoring systems},
  author={VanLehn, Kurt},
  journal={Educational Psychologist},
  volume={46},
  number={4},
  pages={197--221},
  year={2011},
  publisher={Taylor \& Francis}
}

@inproceedings{zhou2019hierarchical,
  title={Hierarchical reinforcement learning for pedagogical policy induction},
  author={Zhou, Guojing and Azizsoltani, Hamoon and Ausin, Markel Sanz and Barnes, Tiffany and Chi, Min},
  booktitle={AIED},
  pages={544--556},
  year={2019}
}

@article{chi2011evaluation,
  title={An evaluation of pedagogical tutorial tactics for a natural language tutoring system: A reinforcement learning approach},
  author={Chi, Min and VanLehn, Kurt and Litman, Diane and Jordan, Pamela},
  journal={International Journal of Artificial Intelligence in Education},
  volume={21},
  number={1-2},
  pages={83--113},
  year={2011},
  publisher={IOS Press}
}

@inproceedings{ausin2019leveraging,
  title={Leveraging Deep Reinforcement Learning for Pedagogical Policy Induction in an Intelligent Tutoring System},
  author={Ausin, Markel Sanz and Azizsoltani, Hamoon and Barnes, Tiffany and Chi, Min},
  booktitle={Proceedings of The 12th International Conference on Educational Data Mining (EDM 2019)},
  volume={168},
  pages={177},
  year={2019},
  organization={ERIC}
}

@article{ju2020aiedcritical,
  title={Evaluating Critical Reinforcement Learning Framework in the Field},
  author={Song Ju and Guojing Zhou and Mark Abdelshiheed and Tiffany Barnes and Min Chi:},
  journal={AIED},
  pages={215--227},
  year={2021}
}

@inproceedings{mandel2014offline,
  title={Offline policy evaluation across representations with applications to educational games},
  author={Mandel, Travis and others},
  booktitle={AAMAS},
  pages={1077--1084},
  year={2014}
}

@inproceedings{lin2017comparisons,
    author="Lin, Chen
    and Chi, Min",
    editor="Andr{\'e}, Elisabeth
    and Baker, Ryan
    and Hu, Xiangen
    and Rodrigo, Ma. Mercedes T.
    and du Boulay, Benedict",
    title={A Comparisons of BKT, RNN and LSTM for Learning Gain Prediction},
    booktitle="Artificial Intelligence in Education",
    year="2017",
    publisher="Springer",
    pages="536--539"
}

@article{mao2018tracing,
  title={Deep Learning vs. Bayesian Knowledge Tracing: Student Models for Interventions},
  author={Mao, Ye and Lin, Chen and Chi, Min},
  journal={Journal of Educational Data Mining},
  year={2018}
}

@article{xi2023themes,
  title={An Offline Time-aware Apprenticeship Learning Framework for Evolving Reward Functions},
  author={Yang, Xi and Gao, Ge and Chi, Min},
  journal={arXiv preprint arxiv:2305.09070v1},
  year={2023}
}

@inproceedings{bain1999bc,
    author="Bain, Michael
    and Sommut, Claude",
    title={A framework for Behavioural cloning},
    booktitle="Machine Intelligence",
    year="1999"
}

@inproceedings{ho2016gail,
 author = {Ho, Jonathan and Ermon, Stefano},
 booktitle = {Advances in Neural Information Processing Systems},
 editor = {D. Lee and M. Sugiyama and U. Luxburg and I. Guyon and R. Garnett},
 pages = {},
 publisher = {Curran Associates, Inc.},
 title = {Generative Adversarial Imitation Learning},
 url = {https://proceedings.neurips.cc/paper_files/paper/2016/file/cc7e2b878868cbae992d1fb743995d8f-Paper.pdf},
 volume = {29},
 year = {2016}
}

@misc{fu2018learning,
  title={Learning Robust Rewards with Adversarial Inverse Reinforcement Learning}, 
  author={Justin Fu and Katie Luo and Sergey Levine},
  year={2018},
  eprint={1710.11248},
  archivePrefix={arXiv},
  primaryClass={cs.LG}
}

@article{mao2021dissertation,
  title={Temporal Skill Discovery For Modeling Student Learning Progression Across STEM Domains},
  author={Mao, Ye},
  journal={PhD thesis},
  year={2001},
  publisher={North Carolina State University}
}

@book{friedman2001elements,
  title={The elements of statistical learning},
  author={Friedman, Jerome and Hastie, Trevor and Tibshirani, Robert},
  volume={1},
  year={2001},
  publisher={Springer series in statistics New York}
}

@article{rafferty2015inferring,
  title={Inferring learners' knowledge from their actions},
  author={Rafferty, Anna N and LaMar, Michelle M and Griffiths, Thomas L},
  journal={Cognitive Science},
  volume={39},
  number={3},
  pages={584--618},
  year={2015},
  publisher={Wiley Online Library}
}

@article{rafferty2016using,
  title={Using Inverse Planning for Personalized Feedback.},
  author={Rafferty, Anna N and Jansen, Rachel and Griffiths, Thomas L},
  journal={EDM},
  volume={16},
  pages={472--477},
  year={2016}
}

@inproceedings{asoh2013application,
  title={An application of inverse reinforcement learning to medical records of diabetes treatment},
  author={Asoh, Hideki and Akaho, Masanori Shiro1 Shotaro and Kamishima, Toshihiro and Hasida, Koiti and Aramaki, Eiji and Kohro, Takahide},
  booktitle={ECMLPKDD2013 workshop on reinforcement learning with generalized feedback},
  year={2013}
}

@inproceedings{pan2019dissecting,
  title={Dissecting the learning curve of taxi drivers: A data-driven approach},
  author={Pan, Menghai and Li, Yanhua and Zhou, Xun and Liu, Zhenming and Song, Rui and Lu, Hui and Luo, Jun},
  booktitle={Proceedings of the 2019 SIAM International Conference on Data Mining},
  pages={783--791},
  year={2019},
  organization={SIAM}
}

@inproceedings{kim2018temporal,
  title={Temporal Belief Memory: Imputing Missing Data during RNN Training.},
  author={Kim, Yeo-Jin and Chi, Min},
  booktitle={In Proceedings of the 27th International Joint Conference on Artificial Intelligence (IJCAI-2018)},
  year={2018}
}

@article{johnson2016mimic,
  title={MIMIC-III, a freely accessible critical care database},
  author={Johnson, Alistair EW and Pollard, Tom J and Shen, Lu and Lehman, Li-wei H and Feng, Mengling and Ghassemi, Mohammad and Moody, Benjamin and Szolovits, Peter and Anthony Celi, Leo and Mark, Roger G},
  journal={Scientific data},
  volume={3},
  number={1},
  pages={1--9},
  year={2016},
  publisher={Nature Publishing Group}
}

@article{goyal2019using,
  title={Using natural language for reward shaping in reinforcement learning},
  author={Goyal, Prasoon and Niekum, Scott and Mooney, Raymond J},
  journal={arXiv preprint arXiv:1903.02020},
  year={2019}
}

@article{amodei2016concrete,
  title={Concrete problems in AI safety},
  author={Amodei, Dario and Olah, Chris and Steinhardt, Jacob and Christiano, Paul and Schulman, John and Man{\'e}, Dan},
  journal={arXiv preprint arXiv:1606.06565},
  year={2016}
}

@inproceedings{wang2020inferring,
  title={Inferring Continuous Treatment Doses from Historical Data via Model-Based Entropy-Regularized Reinforcement Learning},
  author={Wang, Jianxun and Roberts, David and Enquobahrie, Andinet},
  booktitle={Asian Conference on Machine Learning},
  pages={433--448},
  year={2020},
  organization={PMLR}
}

@article{kaelbling1996reinforcement,
  title={Reinforcement learning: A survey},
  author={Kaelbling, Leslie Pack and Littman, Michael L and Moore, Andrew W},
  journal={Journal of artificial intelligence research},
  volume={4},
  pages={237--285},
  year={1996}
}

@inproceedings{ju2022student,
  title={Student-Tutor Mixed-Initiative Decision-Making Supported by Deep Reinforcement Learning},
  author={Ju, Song and Yang, Xi and Barnes, Tiffany and Chi, Min},
  booktitle={International Conference on Artificial Intelligence in Education},
  pages={440--452},
  year={2022},
  organization={Springer}
}

@article{demvsar2006statistical,
  title={Statistical comparisons of classifiers over multiple data sets},
  author={Dem{\v{s}}ar, Janez},
  journal={The Journal of Machine learning research},
  volume={7},
  pages={1--30},
  year={2006},
  publisher={JMLR. org}
}

@article{conover1979comparison,
title = {Multiple-comparisons procedures. Informal report},
author = {Conover, W J and Iman, R L},
abstractNote = {Some of the more popular multiple-comparisons procedures are discussed and compared. Some new nonparametric methods are introduced. One procedure is an analog to the Fisher's least-significant-difference method for the completely randomized design. Some simulation studies indicate this procedure is a reasonable nonparametric method to use. A summary description is given for other nonparametric methods, which may be used with the completely randomized or randomized blocks designs. 3 tables.},
doi = {10.2172/6057803},
url = {https://www.osti.gov/biblio/6057803}, 
journal = {Technical Report},
number = 0,
volume = 0,
place = {United States},
year = {1979},
month = {2}
}

@article{friedman1937rank,
author = {Milton Friedman},
title = {The Use of Ranks to Avoid the Assumption of Normality Implicit in the Analysis of Variance},
journal = {Journal of the American Statistical Association},
volume = {32},
number = {200},
pages = {675-701},
year = {1937},
publisher = {Taylor & Francis},
doi = {10.1080/01621459.1937.10503522}
}

@misc{goodfellow2014generative,
      title={Generative Adversarial Networks}, 
      author={Ian J. Goodfellow and Jean Pouget-Abadie and Mehdi Mirza and Bing Xu and David Warde-Farley and Sherjil Ozair and Aaron Courville and Yoshua Bengio},
      year={2014},
      eprint={1406.2661},
      archivePrefix={arXiv},
      primaryClass={stat.ML}
}

@misc{gleave2022imitation,
  author = {Gleave, Adam and Taufeeque, Mohammad and Rocamonde, Juan and Jenner, Erik and Wang, Steven H. and Toyer, Sam and Ernestus, Maximilian and Belrose, Nora and Emmons, Scott and Russell, Stuart},
  title = {imitation: Clean Imitation Learning Implementations},
  year = {2022},
  archivePrefix = {arXiv},
  eprint = {2211.11972},
  primaryClass = {cs.LG},
}

@inproceedings{islam2024generalized,
  author       = {Md Mirajul Islam and
                  Xi Yang and
                  John Hostetter and
                  Adittya Soukarjya Saha and
                  Min Chi},
  title        = {A Generalized Apprenticeship Learning Framework
                   for Modeling Heterogeneous Student Pedagogical
                   Strategies
                  },
  booktitle    = {Proceedings of the 17th International Conference
                   on Educational Data Mining
                  },
  year         = 2024,
  pages        = {131--142},
  publisher    = {International Educational Data Mining Society},
  venue        = {Atlanta, Georgia, USA}
}

@inproceedings{jarrett2020edm,
  title={Energy-based Models for Imitation Learning},
  author={Jarrett, Daniel and Wulff, Simon and Dy, Jennifer and van der Schaar, Mihaela},
  booktitle={Advances in Neural Information Processing Systems (NeurIPS)},
  year={2020}
}

@inproceedings{tangkaratt2021robust,
  title={Robust Imitation Learning from Noisy Demonstrations},
  author={Tangkaratt, Voot and Sasaki, Fumihiro and Morimoto, Jun},
  booktitle={Proceedings of the International Conference on Artificial Intelligence and Statistics (AISTATS)},
  year={2021}
}

@inproceedings{kim2022demodice,
  title={Demodice: Offline imitation learning with supplementary imperfect demonstrations},
  author={Kim, Geon-ho and Seo, Younggyo and Moon, Yeonsoo and Shin, Jinwoo},
  booktitle={International Conference on Learning Representations (ICLR)},
  year={2022}
}

@inproceedings{sasaki2021noisybc,
  title={Behavioral Cloning from Noisy Demonstrations},
  author={Sasaki, Fumihiro and Yamashina, Ryota},
  booktitle={International Conference on Learning Representations (ICLR)},
  year={2021}
}

@inproceedings{sasaki2021behavioral,
title={Behavioral Cloning from Noisy Demonstrations},
author={Fumihiro Sasaki and Ryota Yamashina},
booktitle={International Conference on Learning Representations},
year={2021},
url={https://openreview.net/forum?id=zrT3HcsWSAt}
}

@InProceedings{wu2019imperfectil,
  title = 	 {Imitation Learning from Imperfect Demonstration},
  author =       {Wu, Yueh-Hua and Charoenphakdee, Nontawat and Bao, Han and Tangkaratt, Voot and Sugiyama, Masashi},
  booktitle = 	 {Proceedings of the 36th International Conference on Machine Learning},
  pages = 	 {6818--6827},
  year = 	 {2019},
  editor = 	 {Chaudhuri, Kamalika and Salakhutdinov, Ruslan},
  volume = 	 {97},
  series = 	 {Proceedings of Machine Learning Research},
  month = 	 {09--15 Jun},
  publisher =    {PMLR},
  url = 	 {https://proceedings.mlr.press/v97/wu19a.html}
}

@inproceedings{xu2021discriminatorweighted,
title={Discriminator-Weighted Offline Imitation Learning from Suboptimal Demonstrations},
author={Haoran Xu and Xianyuan Zhan and Honglei Yin and Huiling Qin},
booktitle={Deep RL Workshop NeurIPS 2021},
year={2021},
url={https://openreview.net/forum?id=XfK0p2TSIB}
}

@inproceedings{hoang2024sprinql,
title={{SPRINQL}: Sub-optimal Demonstrations driven Offline Imitation Learning},
author={Huy Hoang and Tien Anh Mai and Pradeep Varakantham},
booktitle={The Thirty-eighth Annual Conference on Neural Information Processing Systems},
year={2024},
url={https://openreview.net/forum?id=uDD44NROOt}
}

@inproceedings{liu2025pngail,
    title={{PN}-{GAIL}: Leveraging Non-optimal Information from Imperfect Demonstrations},
    author={Qiang Liu and Huiqiao Fu and Kaiqiang Tang and Chunlin Chen and Daoyi Dong},
    booktitle={The Thirteenth International Conference on Learning Representations},
    year={2025},
    url={https://openreview.net/forum?id=0e2pcSxQJS}
}

@article{zolna2020oril,
  author       = {Konrad Zolna and
                  Alexander Novikov and
                  Ksenia Konyushkova and
                  {\c{C}}aglar G{\"{u}}l{\c{c}}ehre and
                  Ziyu Wang and
                  Yusuf Aytar and
                  Misha Denil and
                  Nando de Freitas and
                  Scott E. Reed},
  title        = {Offline Learning from Demonstrations and Unlabeled Experience},
  journal      = {CoRR},
  volume       = {abs/2011.13885},
  year         = {2020},
  url          = {https://arxiv.org/abs/2011.13885},
  eprinttype    = {arXiv},
  eprint       = {2011.13885},
  bibsource    = {dblp computer science bibliography, https://dblp.org}
}

@article{SCHAAL1999233,
title = {Is imitation learning the route to humanoid robots?},
journal = {Trends in Cognitive Sciences},
volume = {3},
number = {6},
pages = {233-242},
year = {1999},
issn = {1364-6613},
doi = {https://doi.org/10.1016/S1364-6613(99)01327-3},
url = {https://www.sciencedirect.com/science/article/pii/S1364661399013273},
author = {Stefan Schaal}}

@InProceedings{pmlr-v15-ross11a,
  title = 	 {A Reduction of Imitation Learning and Structured Prediction to No-Regret Online Learning},
  author = 	 {Ross, Stephane and Gordon, Geoffrey and Bagnell, Drew},
  booktitle = 	 {Proceedings of the Fourteenth International Conference on Artificial Intelligence and Statistics},
  pages = 	 {627--635},
  year = 	 {2011},
  editor = 	 {Gordon, Geoffrey and Dunson, David and Dudík, Miroslav},
  volume = 	 {15},
  series = 	 {Proceedings of Machine Learning Research},
  address = 	 {Fort Lauderdale, FL, USA},
  month = 	 {11--13 Apr},
  publisher =    {PMLR},
  url = 	 {https://proceedings.mlr.press/v15/ross11a.html}
}

@inproceedings{yang2025themes,
author = {Yang, Xi and Islam, Md Mirajul and Gao, Ge and Chi, Min},
title = {THEMES: An Offline Apprenticeship Learning Framework for Evolving Reward Functions},
year = {2025},
isbn = {9798400714542},
publisher = {Association for Computing Machinery},
address = {New York, NY, USA},
url = {https://doi.org/10.1145/3711896.3737154},
doi = {10.1145/3711896.3737154},
booktitle = {Proceedings of the 31st ACM SIGKDD Conference on Knowledge Discovery and Data Mining V.2},
pages = {3565–3576},
numpages = {12},
location = {Toronto ON, Canada},
series = {KDD '25}
}

@inproceedings{finn2017one,
  title={One-shot visual imitation learning via meta-learning},
  author={Finn, Chelsea and Yu, Tianhe and Zhang, Tianhao and Abbeel, Pieter and Levine, Sergey},
  booktitle={Conference on robot learning},
  pages={357--368},
  year={2017},
  organization={PMLR}
}

@inproceedings{Beliaev2022ImitationLB,
  title={Imitation Learning by Estimating Expertise of Demonstrators},
  author={Mark Beliaev and Andy Shih and Stefano Ermon and Dorsa Sadigh and Ramtin Pedarsani},
  booktitle={International Conference on Machine Learning},
  year={2022},
  url={https://api.semanticscholar.org/CorpusID:246485360}
}

@misc{tithi2026adaptive,
      title={Adaptive Scaffolding for Cognitive Engagement in an Intelligent Tutoring System}, 
      author={Sutapa Dey Tithi and Nazia Alam and Tahreem Yasir and Yang Shi and Xiaoyi Tian and Min Chi and Tiffany Barnes},
      year={2026},
      eprint={2602.07308},
      archivePrefix={arXiv},
      primaryClass={cs.AI},
      url={https://arxiv.org/abs/2602.07308}, 
}

@inproceedings{hostetter2025human,
  title     = {Human-Readable Neuro-Fuzzy Networks from Frequent Yet Discernible Patterns in Reward-Based Environments},
  author    = {Hostetter, John Wesley and Saha, Adittya Soukarjya and Islam, Md Mirajul and Barnes, Tiffany and Chi, Min},
  booktitle = {Proceedings of the Thirty-Fourth International Joint Conference on Artificial Intelligence, {IJCAI-25}},
  editor    = {James Kwok},
  pages     = {4535--4543},
  year      = {2025},
  month     = {8},
  doi       = {10.24963/ijcai.2025/505},
}

@InProceedings{islam2025generalized,
author="Islam, Md Mirajul
and Yang, Xi
and Debnath, Rajesh
and Shoukarjya Saha, Adittya
and Chi, Min",
title="A Generalized Apprenticeship Learning Framework for Capturing Evolving Student Pedagogical Strategies",
booktitle="Artificial Intelligence in Education",
year="2025",
publisher="Springer Nature Switzerland",
address="Cham",
pages="393--408",
isbn="978-3-031-98420-4"
}

@inproceedings{ausin2020exploring,
  author    = {Markel Sanz Ausin and
               Mehak Maniktala and
               Tiffany Barnes and
               Min Chi},
  title     = {Exploring the Impact of Simple Explanations and Agency on Batch Deep
               Reinforcement Learning Induced Pedagogical Policies},
  booktitle = {AIED},
  pages     = {472--485},
  year      = {2020}
}

@inproceedings{yang2023dpm,
author = {Yang, Xi and Gao, Ge and Chi, Min},
title = {Hierarchical apprenticeship learning for disease progression modeling},
year = {2023},
isbn = {978-1-956792-03-4},
url = {https://doi.org/10.24963/ijcai.2023/265},
doi = {10.24963/ijcai.2023/265},
booktitle = {Proceedings of the Thirty-Second International Joint Conference on Artificial Intelligence},
articleno = {265},
numpages = {9},
location = {Macao, P.R.China},
series = {IJCAI '23}
}

@inproceedings{mao2020time,
  title={What Time is It? Student Modeling Needs to Know},
  author={Mao, Ye and Marwan, Samiha and Price, Thomas W and Barnes, Tiffany and Chi, Min},
  booktitle={Proceedings of The 13th International Conference on Educational Data Mining (EDM 2020)},
  pages={171--182},
  year={2020}
}

@inproceedings{hinton2002tsne,
author = {Hinton, Geoffrey and Roweis, Sam},
title = {Stochastic neighbor embedding},
year = {2002},
publisher = {MIT Press},
address = {Cambridge, MA, USA},
booktitle = {Proceedings of the 16th International Conference on Neural Information Processing Systems},
pages = {857–864},
numpages = {8},
series = {NIPS'02}
}

@inproceedings{wu2019imitation,
  author    = {Yu-Hua Wu and Nontawat Charoenphakdee and Han Bao and
               Voot Tangkaratt and Masashi Sugiyama},
  title     = {Imitation Learning from Imperfect Demonstration},
  booktitle = {Proceedings of the 36th International Conference on Machine
               Learning},
  series    = {Proceedings of Machine Learning Research},
  volume    = {97},
  pages     = {6818--6827},
  year      = {2019},
}

@article{singer2016sepsis,
  author  = {Mervyn Singer and Clifford S. Deutschman and Christopher W. Seymour
             and others},
  title   = {The Third International Consensus Definitions for Sepsis and
             Septic Shock ({Sepsis-3})},
  journal = {JAMA},
  volume  = {315},
  number  = {8},
  pages   = {801--810},
  year    = {2016},
}

@article{raghu2017deep,
  author  = {Aniruddh Raghu and Matthieu Komorowski and Imran Ahmed and
             Leo Celi and Peter Szolovits and Marzyeh Ghassemi},
  title   = {Deep Reinforcement Learning for Sepsis Treatment},
  journal = {arXiv preprint arXiv:1711.09602},
  year    = {2017},
}

@inproceedings{themes2024,
  author    = {Xinyu Yang and Md Mirajul Islam and Gengchen Gao and Min Chi},
  title     = {{THEMES}: An Offline Apprenticeship Learning Framework for
               Evolving Reward Functions},
  booktitle = {Proceedings of the 31st ACM SIGKDD Conference on Knowledge
               Discovery and Data Mining},
  series    = {KDD '25},
  pages     = {3565--3576},
  year      = {2025},
  publisher = {Association for Computing Machinery},
  address   = {New York, NY, USA},
  doi       = {10.1145/3711896.3737154},
}

@inproceedings{islam2026hierarchical,
  author    = {Md Mirajul Islam and Rajesh Debnath and Adittya Soukarjya Saha
               and Min Chi},
  title     = {Hierarchical Apprenticeship Learning from Imperfect Demonstrations
               with Evolving Rewards},
  booktitle = {Artificial Intelligence in Education (AIED)},
  year      = {2026},
}

@inproceedings{reddy2020sqil,
  author    = {Siddharth Reddy and Anca Dragan and Sergey Levine},
  title     = {{SQIL}: Imitation Learning via Reinforcement Learning with
               Sparse Rewards},
  booktitle = {International Conference on Learning Representations},
  year      = {2020},
}

@inproceedings{kostrikov2022offline,
  author    = {Ilya Kostrikov and Ashvin Nair and Sergey Levine},
  title     = {Offline Reinforcement Learning with Implicit {Q}-Learning},
  booktitle = {International Conference on Learning Representations},
  year      = {2022},
}

@inproceedings{kim2021demodice,
  author    = {Geon-Hyeong Kim and Seokin Seo and Jongmin Lee and
               Wonseok Jeon and HyeongJoo Hwang and Hongseok Yang and
               Kee-Eung Kim},
  title     = {{DemoDICE}: Offline Imitation Learning with Supplementary
               Imperfect Demonstrations},
  booktitle = {International Conference on Learning Representations},
  year      = {2022},
}

@inproceedings{ni2021f,
  author    = {Tianwei Ni and Harshit Sikchi and Yufei Wang and Tejus Gupta and
               Lisa Lee and Ben Eysenbach},
  title     = {f-{IRL}: Inverse Reinforcement Learning via State Marginal
               Matching},
  booktitle = {Conference on Robot Learning (CoRL)},
  year      = {2021},
}

@inproceedings{christiano2017deep,
  author    = {Paul Christiano and Jan Leike and Tom B. Brown and Miljan Martic
               and Shane Legg and Dario Amodei},
  title     = {Deep Reinforcement Learning from Human Preferences},
  booktitle = {Advances in Neural Information Processing Systems},
  pages     = {4299--4307},
  year      = {2017},
}

@inproceedings{brown2019extrapolating,
  author    = {Daniel S. Brown and Wonjoon Goo and Prabhat Narasimhan
               and Scott Niekum},
  title     = {Extrapolating Beyond Suboptimal Demonstrations via Inverse
               Reinforcement Learning from Observations},
  booktitle = {Proceedings of the 36th International Conference on Machine
               Learning (ICML)},
  series    = {Proceedings of Machine Learning Research},
  volume    = {97},
  year      = {2019},
}

@inproceedings{ju2021identifying,
  author    = {Sicun Ju and Young-Jun Kim and Markel Sanz Ausin and
               Maria E. Mayorga and Min Chi},
  title     = {To Reduce Healthcare Workload: Identify Critical Sepsis
               Progression Moments through Deep Reinforcement Learning},
  booktitle = {IEEE International Conference on Big Data},
  year      = {2021},
}

@article{ausin2021infernet,
  author  = {Markel Sanz Ausin and Hamid Azizsoltani and Sicun Ju and
             Young-Jun Kim and Min Chi},
  title   = {{InferNet} for Delayed Reinforcement Tasks: Addressing the
             Temporal Credit Assignment Problem},
  journal = {arXiv preprint arXiv:2105.00568},
  year    = {2021},
}

@misc{roth2019conservative,
  author        = {Aaron M. Roth and Nicholay Topin and Pooyan Jamshidi and
                   Manuela Veloso},
  title         = {Conservative {Q}-Improvement: Reinforcement Learning for an
                   Interpretable Decision-Tree Policy},
  year          = {2019},
  eprint        = {1907.01180},
  archivePrefix = {arXiv},
  primaryClass  = {cs.LG},
  url           = {https://arxiv.org/abs/1907.01180},
}

@inproceedings{eysenbach2019search,
  author    = {Ben Eysenbach and Russ R. Salakhutdinov and Sergey Levine},
  title     = {Search on the Replay Buffer: Bridging Planning and Reinforcement
               Learning},
  booktitle = {Advances in Neural Information Processing Systems},
  volume    = {32},
  publisher = {Curran Associates, Inc.},
  year      = {2019},
  url       = {https://proceedings.neurips.cc/paper_files/paper/2019/file/5c48ff18e0a47baaf81d8b8ea51eec92-Paper.pdf},
}

@article{lyu2017overtreatment,
  author  = {Heather Lyu and Tim Xu and Daniel Brotman and Brandan Mayer-Blackwell
             and Michol Cooper and Michael Daniel and Elizabeth C. Wick and
             Vikas Saini and Shannon Brownlee and Martin A. Makary},
  title   = {Overtreatment in the United States},
  journal = {PLOS ONE},
  volume  = {12},
  number  = {9},
  pages   = {e0181970},
  year    = {2017},
  doi     = {10.1371/journal.pone.0181970},
}

@article{rhee2020prevalence,
  author  = {Chanu Rhee and Sameer S. Kadri and John P. Dekker and Robert L. Danner
             and Huai-Chun Chen and David Fram and Fang Zhang and Rui Wang and
             Michael Klompas},
  title   = {Prevalence of Antibiotic-Resistant Pathogens in Culture-Proven
             Sepsis and Outcomes Associated with Inadequate and Broad-Spectrum
             Empiric Antibiotic Use},
  journal = {JAMA Network Open},
  volume  = {3},
  number  = {4},
  pages   = {e202899},
  year    = {2020},
  doi     = {10.1001/jamanetworkopen.2020.2899},
}

@article{chang2021timelimited,
  author  = {Dong W. Chang and Thanh H. Neville and Jennifer Parrish and
             Lian Ewing and Christy Rico and Liliacna Jara and Danielle Sim and
             Chi-hong Tseng and Carin van Zyl and Aaron D. Storms and
             Nader Kamangar and Janice M. Liebler and May M. Lee and Hal F. Yee},
  title   = {Evaluation of Time-Limited Trials Among Critically Ill Patients
             with Advanced Medical Illnesses and Reduction of Nonbeneficial
             {ICU} Treatments},
  journal = {JAMA Internal Medicine},
  volume  = {181},
  number  = {6},
  pages   = {786--794},
  year    = {2021},
  doi     = {10.1001/jamainternmed.2021.1000},
}

@inproceedings{bansal2021whole,
  author    = {Bansal, Gagan and Wu, Tongshuang and Zhou, Joyce and Fok, Raymond and Nushi, Besmira and Kamar, Ece and Ribeiro, Marco Tulio and Weld, Daniel S.},
  title     = {Does the Whole Exceed its Parts? The Effect of AI Explanations on Complementary Team Performance},
  booktitle = {Proceedings of the 2021 CHI Conference on Human Factors in Computing Systems},
  series    = {CHI '21},
  year      = {2021},
  publisher = {ACM},
  address   = {New York, NY, USA},
  doi       = {10.1145/3411764.3445717}
}

@inproceedings{horvitz1999principles,
  author    = {Horvitz, Eric},
  title     = {Principles of Mixed-Initiative User Interfaces},
  booktitle = {Proceedings of the SIGCHI Conference on Human Factors in Computing Systems},
  series    = {CHI '99},
  year      = {1999},
  pages     = {159--166},
  publisher = {ACM},
  address   = {New York, NY, USA},
  doi       = {10.1145/302979.303030}
}

@inproceedings{tithi2026exploring,
  author    = {Dey Tithi, Sutapa and Tian, Xiaoyi and Limke, Ally and Chi, Min and Barnes, Tiffany},
  title     = {Exploring the Design and Impact of Interactive Worked Examples for Learners with Varying Prior Knowledge},
  booktitle = {Proceedings of the 2026 CHI Conference on Human Factors in Computing Systems},
  series    = {CHI '26},
  year      = {2026},
  publisher = {ACM},
  address   = {New York, NY, USA},
  doi       = {10.1145/3772318.3791631}
}

@inproceedings{wu2026rethinking,
  author    = {Wu, Mengke and Liu, Weizi and Wang, Yanyun and Ding, Weiyu and Yao, Mike},
  title     = {Rethinking User Empowerment in AI Recommender System: Innovating Transparent and Controllable Interfaces},
  booktitle = {Proceedings of the 2026 CHI Conference on Human Factors in Computing Systems},
  series    = {CHI '26},
  year      = {2026},
  publisher = {ACM},
  address   = {New York, NY, USA},
  doi       = {10.1145/3772318.3791914}
}

@article{gupta2024beyond,
  author  = {Gupta, Nidhi and Ali, Kamran and Jiang, Dan and Fink, Trine and Du, Xiangyun},
  title   = {Beyond autonomy: unpacking self-regulated and self-directed learning through the lens of learner agency---a scoping review},
  journal = {BMC Medical Education},
  volume  = {24},
  pages   = {1519},
  year    = {2024},
  doi     = {10.1186/s12909-024-06476-x}
}
\bibliographystyle{apalike}

\restoregeometry
\appendix

\chapter{Acronyms}

A summary of all acronyms is documented in Table \ref{tab:acro}.

\captionsetup{labelfont=normalfont}
\begin{longtable}{| l | p{10cm} |}
\caption{A summary of acronyms used in alphabetical order.}\label{tab:acro}\\
\hline
\textbf{Acronym} & \textbf{Abbreviation} \\
\hline
\endfirsthead

\multicolumn{2}{c}%
{{\tablename\ \thetable{}: (continued).}} \\
\hline
\textbf{Acronym} & \textbf{Abbreviation} \\
\hline
\endhead

\hline
\endfoot

\hline
\endlastfoot

AL & Apprenticeship Learning \\
BC & Behavior Cloning \\
BCIL & Batch Contrastive Imitation Learning \\
BEAR & Bootstrapping Error Accumulation Reduction \\
CQL & Conservative Q-Learning \\
DRL & Deep Reinforcement Learning \\
DP & Dirichlet Process \\
DPO & Direct Preference Optimization \\
DQN & Deep Q-Network \\
EDM & Energy-based Distribution Matching \\
EM & Expectation–Maximization \\
EM-EDM & Expectation–Maximization Energy-based Distribution Matching \\
GAIL & Generative Adversarial Imitation Learning \\
HIRL & Hierarchical Inverse Reinforcement Learning \\
IL & Imitation Learning \\
IRL & Inverse Reinforcement Learning \\
ITS & Intelligent Tutoring System \\
MDP & Markov Decision Process \\
MCTS & Monte Carlo Tree Search \\
OILCA & Offline Imitation Learning with Counterfactual Augmentation \\
OPE & Off-Policy Evaluation \\
POMDP & Partially Observable Markov Decision Process \\
Q-function & Action–Value Function \\
RL & Reinforcement Learning \\
SPRINQL & State-Partitioned Robust Imitation via Q-Learning \\
TD & Temporal Difference \\
THEMES & Time-aware Hierarchical EM Energy-based Sub-trajectory Framework \\
VAE & Variational Autoencoder \\
\end{longtable}

\chapter{Variables}

A summary of all variables used throughout this dissertation is documented in Table \ref{tab:vars}.

\begin{longtable}{| l | p{10cm} |}
\caption{A summary of common variables and their definitions in alphabetical order.}\label{tab:vars}\\
\hline
\textbf{Variable} & \textbf{Definition} \\
\hline
\endfirsthead

\multicolumn{2}{c}%
{{\tablename\ \thetable{}: (continued).}} \\
\hline
\textbf{Variable} & \textbf{Definition} \\
\hline
\endhead

\hline
\endfoot

\hline
\endlastfoot

$a$ & Action taken by the agent \\
$A$ & Action space \\
$D$ & Offline demonstration dataset \\
$\mathcal{D}_E$ & Set of expert demonstrations \\
$\mathcal{D}_I$ & Set of imperfect or mixed-quality demonstrations \\
$E_\theta(\cdot)$ & Energy function parameterized by $\theta$ \\
$\gamma$ & Discount factor in reinforcement learning \\
$H$ & Horizon length of a trajectory \\
$k$ & Number of clusters or latent reward types \\
$L(\cdot)$ & Loss function \\
$p(s'|s,a)$ & Transition probability from state $s$ to $s'$ under action $a$ \\
$\pi(a|s)$ & Policy: probability of selecting action $a$ in state $s$ \\
$\pi_E$ & Expert policy \\
$\pi_\theta$ & Learned policy with parameters $\theta$ \\
$Q^\pi(s,a)$ & Action–value function under policy $\pi$ \\
$r$ & Reward value for a state--action pair \\
$R(s,a)$ & Reward function \\
$R_i$ & Reward function for cluster or intention $i$ \\
$s$ & State variable \\
$S$ & State space \\
$t$ & Time step index \\
$\tau$ & Trajectory (sequence of state--action pairs) \\
$\tau_i$ & Sub-trajectory (used in THEMES and hierarchical AL) \\
$V^\pi(s)$ & Value function under policy $\pi$ \\
$w$ & Importance weight or trajectory weight \\
$\theta$ & Policy or energy model parameters \\
$\phi(\cdot)$ & Feature representation or encoder function \\
$Z$ & Partition function / normalization constant (energy models) \\

\end{longtable}

\restoregeometry

\backmatter

\end{document}